\let\mypdfximage\pdfximage
\def\pdfximage{\immediate\mypdfximage}

\documentclass[10pt,journal,compsoc]{IEEEtran}
\usepackage{amssymb}
\usepackage{amsthm}

\usepackage{multirow}
\usepackage{color}
\usepackage[table]{xcolor}

\ifCLASSOPTIONcompsoc
  \usepackage[nocompress]{cite}
\else
  \usepackage{cite}
\fi
\ifCLASSINFOpdf
 \usepackage[pdftex]{graphicx}
\else
\fi
\usepackage{amsmath}
\usepackage{tabularx}
\ifCLASSOPTIONcompsoc
  \usepackage[caption=false,font=footnotesize,labelfont=sf,textfont=sf]{subfig}
\else
  \usepackage[caption=false,font=footnotesize]{subfig}
\fi
\usepackage{url}

\begin{document}
%
\title{\huge NPRportrait 1.0: A Three-Level Benchmark for Non-Photorealistic Rendering of Portraits}
%
%
%
%

\author{Paul L. Rosin,
        Yu-Kun Lai,
        David Mould,
        Ran Yi,
        Itamar Berger,
        Lars Doyle,
        Seungyong Lee,
        Chuan Li,
        Yong-Jin Liu,
		Amir Semmo,
        Ariel Shamir,
        Minjung Son,
        Holger Winnem{\"oller}
    \IEEEcompsocitemizethanks{%
        \IEEEcompsocthanksitem P.L. Rosin is with the School of Computer Science and Informatics, Cardiff University, UK
        E-mail: PaulRosin@cs.cf.ac.uk.
        \IEEEcompsocthanksitem Y.L. Lai is with the School of Computer Science and Informatics, Cardiff University, UK
        E-mail: Yukun.Lai@cs.cardiff.ac.uk.
        \IEEEcompsocthanksitem D. Mould is with the School of Computer Science, Carleton University, Canada
        E-mail: mould@scs.carleton.ca.
        \IEEEcompsocthanksitem R. Yi is with the Department of Computer Science and Technology, Tsinghua University, Beijing, China
        E-mail: yr16@mails.tsinghua.edu.cn.
        \IEEEcompsocthanksitem L. Doyle is with the School of Computer Science, Carleton University, Canada
        E-mail: larsdoyle@cmail.carleton.ca.
        \IEEEcompsocthanksitem I. Berger is with The Interdisciplinary Center Herzliya, Israel
        E-mail: berger.itamar@gmail.com.
        \IEEEcompsocthanksitem S. Lee is with the Department of Computer Science and Engineering, Pohang University of Science and Technology, South Korea
        E-mail: leesy@postech.ac.kr.
        \IEEEcompsocthanksitem C. Li is with the Lambda Labs, Inc., USA 
        E-mail: c@lambdal.com.
        \IEEEcompsocthanksitem Y.J. Liu is with the Department of Computer Science and Technology, Tsinghua University, Beijing, China
        E-mail: liuyongjin@tsinghua.edu.cn.
		\IEEEcompsocthanksitem A. Semmo is with the Hasso Plattner Institute, University of Potsdam, Germany
		E-mail: Amir.Semmo@hpi.de.
        \IEEEcompsocthanksitem A. Shamir is with The Interdisciplinary Center Herzliya, Israel
        E-mail: arik@idc.ac.il.
        \IEEEcompsocthanksitem M. Son is with the Multimedia Processing Laboratory, Samsung Advanced Institute of Technology, South Korea
        E-mail: minjungs.son@samsung.com.
        \IEEEcompsocthanksitem H. Winnem{\"oller} is with Adobe Systems, Inc. USA 
        E-mail: hwinnemo@adobe.com.
}
}

%
%

\markboth{}%
{}
%



\IEEEtitleabstractindextext{%
\begin{abstract}
Despite the recent upsurge of activity in image-based non-photorealistic rendering (NPR),
and in particular portrait image stylisation,
due to the advent of neural style transfer, the state of performance evaluation in this field is limited, especially compared to the norms in the computer vision and machine learning communities. 
Unfortunately, the task of evaluating image stylisation is 
thus far not well defined,
since it involves subjective, perceptual and aesthetic aspects.
To make progress towards a solution, this paper proposes a new structured, three level, benchmark dataset for the evaluation of stylised portrait images.
Rigorous criteria were used for its construction, and its consistency was validated by user studies.
Moreover, a new methodology has been developed for evaluating portrait stylisation algorithms, which makes use of the different benchmark levels
as well as annotations provided by user studies regarding the characteristics of the faces.  
We perform evaluation for a wide variety of image stylisation methods (both portrait-specific and general purpose, and also both traditional NPR approaches and neural style transfer) using the new benchmark dataset.
\end{abstract}

}

\maketitle

\IEEEdisplaynontitleabstractindextext

%
\IEEEpeerreviewmaketitle

\IEEEraisesectionheading{\section{Introduction}\label{sec:introduction}}

%
%
%
%

Image-based non-photorealistic rendering (NPR) is at the intersection
of computer graphics and computer vision, and has the aim of
synthesising new images based on the analysis of existing
images\footnote{ Note that NPR is normally assumed to refer
to \emph{artistic} rendering as opposed to e.g. simple intensity or
colour mapping.  }.  NPR can be applied in many different ways, such
as: the rendering of CAD models of furniture and interior designs as
watercolour style illustrations to provide more appealing renderings
for sales brochures~\cite{luft2008}, stylising images to different
degrees to provide stimuli for perceptual studies investigating the
theory of mind~\cite{krumhuber2019facial}, stylising images to reduce
patients' aversion to otherwise unpleasant pictures of surgical
procedures~\cite{besanccon2020reducing}, and image enhancement prior
to generating 3D bas-reliefs in order to emphasise salient structures
and reduce noise and visual clutter~\cite{wu}.  NPR can be applied to
images, video, and 3D models, but in this paper we will focus on
image-based NPR, and in particular the rendering of portrait images,
which is also known as portrait image stylisation.  A comprehensive
historical overview of 30 years of image-based NPR is provided by
Kyprianidis \emph{et al.}~\cite{KyprianidisCWI13}, while an overview
of the state of the art in 2013 is given by Rosin and
Collomosse~\cite{rosinBook}.  Shortly after this date the course of
NPR was dramatically changed with the advent of deep learning and the
huge popularity of neural style transfer that was initiated by
Gatys \emph{et al.}'s landmark paper~\cite{gatys2016image2}.

Despite the substantial amount of research activity in NPR/image stylisation,
the degree and level of evaluation of results reported in the literature is limited,
and falls far below the norms in the computer vision and machine learning communities.
We noted that one of the roots of NPR lies in computer graphics, and it is this aspect of image generation which is
very challenging,
in that evaluation of NPR results is less straightforward than for computer vision or machine learning for the following reasons:\footnote{
Some of these issues were identified by David Salesin in his NPAR 2002 keynote speech on seven Grand Challenges for NPR,
where amongst other things he talked about (1) How can we quantify success, and (5) the Artistic Turing Test: can NPR achieve products indistinguishable from an artist's works?
}
\begin{itemize}
\item
First, for a typical computer vision or machine learning task such as
classification, regression or detection, there is normally assumed to
be a correct solution, often referred to as ``ground
truth''.  However, for stylisation, ground truth generally does not exist; for
instance, if a particular NPR task is to produce stylisations in the
manner of the artist Monet it is not possible to acquire ideal images
before and after stylisation\footnote{Recently, in an attempt to address this,
Kerdreux \emph{et al.}~\cite{kerdreux2020interactive} show some preliminary work in
which they construct a set of photograph-painting pairs for two
buildings: the Notre Dame de Paris Cathedral and the Notre Dame de
Rouen Cathedral, which includes Monet, who made about 40 paintings of
Notre Dame de Rouen Cathedral.}.
Moreover, NPR algorithms are often
designed to produce novel styles, for which no existing examples
exist.
\item
Second, in addition to novel styles,
there are many aspects of stylisations that can differ, independently of the quality of the rendering.
Possibilities include the medium (oil paint, watercolour, pen, crayon),
level of control (tight versus loose), or
artistic school (Renaissance, impressionist, cubist).
Thus, there is a vast range of possible stylisations for any given image,
and it would be impossible to generate in advance all the possible ground truths, even for one image.
\item
Even if the above problems were somehow overcome,
there still remains the issue of how to quantify the similarity of a rendered image with ground truth.
Some partial solutions appear in the literature; their shortcomings will be described later in the paper.
\item
Finally, it is not clear how to compare the results of two methods that produce different styles.
A detailed rendering created with a large number of carefully placed colourful strokes
will look very different from one consisting of a few sparse and abstracted black lines.
Therefore, unlike in tasks such as classification where all algorithms aim to return the same (i.e. correct) solution,
here there is no unique ground truth.
\end{itemize}

The above problems have led to the situation where there is little in the way of standard benchmark data sets available
(although some progress exists: see the recent work by Mould and Rosin~\cite{mould2017developing} and Rosin \emph{et al.}~\cite{rosin2017benchmarking})
for researchers and practitioners in the area of NPR, and also a lack of evaluation methodology.
Naturally, this is an undesirable situation, since rigorous evaluation and comparison of methods
will help identify strengths and weaknesses in the field,
will make it easier to identify real improvements from amongst the large body of incremental work, and will advance the overall field.

Amongst the many applications of NPR, this paper concentrates on portrait stylisation.
There is a huge amount of portrait photography, from formal portrait to selfies,
and particularly with online applications there has been increased demand for personalised portraits.
With the consequent surge in portrait stylisation methods there is thus a need for more resources for portrait benchmarking.
This paper makes a step in improving the evaluation methodology for
portrait stylisation,
and makes the following three specific contributions:
\begin{itemize}
\item
A new structured, three level, benchmark dataset has been created for the evaluation of stylised portrait images.
Rigorous criteria were used for its construction, and several user studies have been used to generate annotations.
\item
A new methodology has been developed for evaluating portrait stylisation algorithms,
and makes use of the different benchmark levels and annotations.
\item
We perform evaluation for a wide variety of NPR methods
(both portrait-specific and general) using the new benchmark dataset.
\end{itemize}

This paper follows on from the previous conference version by Rosin \emph{et al.}~\cite{rosin2017benchmarking},
and makes substantial changes to that prior benchmark, namely \emph{NPRportrait0.1}.\footnote{
The benchmark released in \cite{rosin2017benchmarking} was presented at the time as a basic ``version 0.1'', with
the intention of performing user studies and extending the number of levels.
}
The major differences are that:
\begin{itemize}
\item
More rigorous criteria for image selection were used compared to Rosin \emph{et al.}~\cite{rosin2017benchmarking}.
This particularly affected level~1 of \emph{NPRportrait0.1}, which was therefore totally replaced by a better controlled image set.
\item
Images are now more rigorously checked against the design matrix requirements by running user studies for validation.
\item
A third and new level has been added to the benchmark
to provide more challenging test images for state-of-the-art methods.
Overall, under a third of \emph{NPRportrait1.0} consists of images from \emph{NPRportrait0.1}.
\item
The set of NPR algorithms that have been systematically evaluated has been expanded to include another six styles from the literature,
ensuring that they cover:
(1) both portrait-specific and general purpose methods,
(2) both traditional NPR and neural style transfer methods,
(3) stylisation of both texture and geometry,
(4) colour as well as black and white stylisations.
\item
A new set of experimental procedures is defined, and the NPR algorithms are quantitatively evaluated according to them.
Specifically, (1) the correctness of perceived facial characteristics are tested for stylisations (making use of the benchmark annotations),
and (2) the quality of the NPR algorithms' outputs are checked for trends across the benchmark levels.
\end{itemize}
The benchmark data (images and annotations) are made available to the research community, and
provide a framework for others to use and to extend.

\section{Related work}

Two critical elements in benchmarking are the datasets and the evaluation of the results.

\subsection{Benchmark Datasets}

In computer vision there is a huge range of benchmark datasets.\footnote{
CVonline~\cite{CVonline} lists 1170 unique computer vision datasets.
}
They incorporate
(1) both data and annotations
(e.g. ground truth class labels, bounding boxes, segmentations),
(2) cover many areas
(e.g.  medical, remote sensing, surveillance, agriculture),
and
(3) range from specific high level applications
(e.g. detection of various medical conditions),
to specific low level tasks such as segmentation, image registration, or feature localisation.
Further, websites such as
the Middlebury Vision Pages~\cite{Middlebury}
and the MIT/Tuebingen Saliency Benchmark~\cite{mit-tuebingen-saliency-benchmark}
allow users to submit results, and the benchmark organisers will perform evaluation and add the scores to published leaderboards.
Over the years these benchmark datasets have
become increasingly large, especially in recent years so as to facilitate machine learning.

The situation in NPR is very different. Until recently there were no benchmark datasets.\footnote{
To the best of our knowledge no websites equivalent to the Middlebury Vision Pages or the MIT/Tuebingen Saliency Benchmark
exist for NPR.
}
Mould and Rosin~\cite{mould2017developing} created the first one, \emph{NPRgeneral}, which as its name indicates,
was designed to provide images for the general task of NPR.
It contains 20 images that were selected to include a variety of attributes and content,
namely:
variation in scale and texture;
fine detail;
regular structure;
irregular texture;
visual clutter;
vivid, muted and varied colours;
low and mixed contrast;
complex and indistinct edges;
thin features;
long gradients;
high and low key,
human faces.
Images were selected manually (i.e. subjectively),
 although some low level image measures
(colourfulness, complexity, contrast, sharpness, lineness, noise and the mean and standard deviation of intensity)
provided guidance.
The authors applied eight NPR methods to the benchmark, and identified that some specific images were generally challenging for all the algorithms,
suggesting a suitable direction for future research.
Other groups of images were found to be very difficult for certain categories of algorithm, but not others,
indicating how the existing state-of-the-art algorithms can be best deployed according to the expected nature of the test data.

Kumar \emph{et al.}~\cite{kumar2019comprehensive} recently produced a NPR benchmark that closely follows the principles of \emph{NPRgeneral}.
It consists of 32 images, and its goal was to augment the \emph{NPRgeneral} with more varied and more complex type images.

Another, more specialised, benchmark dataset named \emph{NPRportrait0.1} has been released by Rosin \emph{et al.}~\cite{rosin2017benchmarking}.
It contains portrait images, split into two levels of difficulty, each consisting of 20 images.
The first level consists of highly constrained portrait images,
i.e. close cropping of the faces, frontal views, and simple uncluttered backgrounds.
Six NPR algorithms (both portrait-specific and general) were applied to the benchmark dataset;
all the methods worked reasonably well,
demonstrating that level one of the benchmark is tractable,
but it was evident that the domain knowledge contained by the face-specific methods enabled them to improve the quality and robustness of their stylisations,
e.g. by preserving important elements such as eyes.
The second level slightly relaxes the constraints on pose, lighting, and background, while allowing facial hair and more varied expressions.
Interestingly, stylisation results at level~2 differed from those at level~1:
the performance of the portrait-specific algorithms declined for some images
with more complex contents.
However, the general purpose algorithms were equally effective across both levels.
Compared to \emph{NPRgeneral}, \emph{NPRportrait0.1} took a more systematic approach to selecting images, using a design matrix,
and the new dataset \emph{NPRportrait1.0}
will follow that aspect of their methodology, which will be further described in section~\ref{methodology}.
Following a design matrix ensures that a balanced dataset is created.
The issue of data bias has become a hot topic in recent years, particularly for race and gender~\cite{buolamwini2018gender}.
Although the focus is normally on training data, so as to avoid biased models, in our case we are more interested in test data,
so that any biases in NPR methods (whether using machine learning or not) can be detected.

To date, these benchmark datasets have been used in a variety of ways:
to include some stylisation results from examples taken from the benchmark~\cite{azami2017detail,jing2019neural,rosin2017watercolour,wu2018field};
to provide appropriate test data as part of the optimisation of preset parameters for post-processing filters in BeCasso,
an interactive mobile iOS app for image stylisation~\cite{klingbeil2017challenges};
and to provide a competitive and common set of test images for a research course on image processing for mobile applications~\cite{trapp2018teaching}.

\subsection{Image Quality Assessment}
\label{IQA}

Previously we noted that for many computer vision tasks the computation of an error measure such as classification accuracy is straightforward.
Nevertheless, some computer vision tasks are more problematic; for example, to evaluate saliency models
many different evaluation metrics with different properties have been proposed~\cite{bylinskii2018different},
and so many researchers include several in their evaluations (e.g.,
the MIT/Tuebingen Saliency Benchmark shows seven metrics).

Evaluating NPR outputs is even more challenging, as it involves the aesthetic qualities of pictures, which is subjective, and hard to quantify.
If the evaluation task is to compare an NPR result with a ground truth result then some method for performing image comparison is required.
This is a well known computer vision task, and the literature contains a range of possible methods.
However, standard image comparison measures such as MSE, PSNR or SSIM~\cite{ssim} are too low-level,
and fail to capture important perceptual and aesthetic aspects of a stylised image.
Recent deep learning approaches have attempted to capture these perceptual characteristics (e.g. LPIPS~\cite{zhang2018perceptual}),
but while they tend to perform better than traditional measures, they still do not always follow human judgements~\cite{zamir2019vision}.
Moreover, such deep learning methods are prone to overfitting, can display a lack of robustness~\cite{kettunen2019lpips},
and have not been trained on stylised images.

The above approaches have assumed access to ground truth images, which are likely to be unavailable.
In this case, the alternative is to use a no-reference or blind image quality assessment (IQA),
of which many have been developed within the image processing community.
Early approaches to blind IQA were too restrictive since they assumed that image quality was affected by specific known types of distortions,
such as blockiness, blur, or compression artifacts.
Subsequently, regression models became more popular;
these were trained on distorted and undistorted images along with human opinion scores
and learnt to predict IQA scores from image features (e.g. DIIVINE~\cite{moorthy2011blind}, BRISQUE~\cite{mittal2012no}).
More recently, ``opinion-unaware'' methods that avoid the need for human subjective scores have been developed, which
is attractive given the difficulty in collecting enough training samples that capture the many different possible image distortion types
as well as the combinations of different distortion.
One such example is IL-NIQE~\cite{ILNIQE}, which builds a multivariate Gaussian model using natural scene statistics features
to represent clean high-quality natural image patches.
Test images are then assessed by comparing image patches against the model using a Bhattacharyya-like distance, followed by averaging the patch scores.
However, such approaches are not suitable for evaluating stylisations.

One solution that has been taken up by neural style transfer researchers
to cope with the absence of ground truth data consisting of paired before and after stylisation images,
is to use the Fr\'echet Inception Distance (FID)~\cite{heusel2017gans}.
Rather than comparing two images, two unpaired sets of images (e.g. stylised and unstylised) are compared instead.
This is done by modelling with a multivariate Gaussian for each set, the images' intermediate layer features produced by the Inception network.
The Fr\'echet (also known as Wasserstein-2) distance between the two distributions
is calculated, and involves just the means and co-variances of the distributions.
Limitations of FID are that it assumes that features have Gaussian distributions,
and the estimator of FID has a strong bias even for up to 10,000 samples~\cite{Demystifying},
and is also not trained on stylised images.
Moreover, it requires a set of ideal images in the target style, which may not be available.

\subsection{Alternative Approaches to NPR Evaluation}

The difficulties of performance evaluation in NPR have been identified and discussed thoroughly in the NPR community~\cite{Isenberg13,Hall}.
In an attempt to overcome the above difficulties, a common practice in NPR
was to employ proxy measures~\cite{Hertzmann-position} in place of directly evaluating the aesthetics of the stylised image.
Thus more easily quantifiable measures, such as performance on a memory task, a grouping task, or artist classification~\cite{sanakoyeu2018style}
could be collected.
The drawback is that the proxy measure may not directly correlate with the quality of the image stylisation.

Mould~\cite{Mould:ASE} proposed that in some situations, such as for an undirected NPR task
in which there is no clear problem statement and no available ground truth,
the researcher could carry out an authorial subjective evaluation.
This means that the author would identify important characteristics of interest,
and use these to make a (potentially more) transparent and structured visual analysis of the results.
Since authorial subjective evaluation still lacks objectivity, and does not scale up well,
it can be considered as a fallback position.

User studies are a popular alternative means of evaluation, and have the strong advantage that they
have the potential to capture all aspects of human perception including semantics, aesthetics, or art history.
They are a popular tool in the neural style transfer community;
however, the traditional NPR community has reservations on their effectiveness~\cite{Hertzmann-position,Isenberg13,Hall,Mould:ASE}.
Issues include: use of participants who are aware of the hypothesis, and provide biased responses;
study participants may be careless or insufficiently understand the task;
it can be difficult to formulate the questions or tasks in a user study;
in general it is not possible to independently verify a user study's results except by re-running the study;
and finally, it is difficult to compare results from separate user studies.
To give an example, it is difficult to ensure that participants in a user study are
assessing renderings based on aesthetics and style elements without being influenced by the source image content,
or by their preferences for certain styles
(e.g. their preferences for colour images versus black and white, or detailed versus highly abstracted).

\subsection{Portraiture in NPR}

Since the early days of NPR there has been particular interest in
generating portraits, from simple line
drawings~\cite{li1995extraction} a quarter of a century ago, to modern
state of the art methods that combine deep learning with a dataset of
artists' portraits to enable stylisation of both geometry and
texture~\cite{yaniv2019face}.  We refer the reader to Zhao and Zhu's
work~\cite{zhao2013artistic} for an
overview of portrait-specific NPR methods prior to deep
learning, and to Yaniv \emph{et al.}'s
paper~\cite{yaniv2019face} for references to more recent methods.  In
this section we briefly outline the 11 NPR algorithms (both
portrait-specific and general purpose) which will be evaluated in
section~\ref{experimentSection}.

Li and Wand's method~\cite{li2016combining} treats styles as textures,
and forces the synthesised image and the reference style image to
have the same Markovian texture statistics.
Non-parametric sampling is first used to capture patches from the
style image; patch matching and blending are then used to transfer the
style to the synthesised image. For portrait stylisation, they include
an additional content constraint that minimises the $L_2$ distance
between the CNN encoding of the portrait photo and the synthesised
image.

Berger \emph{et al.}~\cite{berger-shamir} mimic the style of
specific artists' line-drawings in a data-driven manner.  Sample
drawings of artists are collected and their statistics
are analysed.  Then, given a new portrait photograph and an artist
style, the algorithm first creates a contour image by using a variant
of the XDoG method~\cite{xdog}. Using the detected facial features, the face
geometry is modified to follow the specific artist's geometric
style. Lastly, the face contours are drawn using strokes from the
artist's stroke database following the artist's drawing statistics.

Yi \emph{et al.}~\cite{YiLLR19} proposed APDrawingGAN,
a hierarchical system of generative adversarial networks (GANs) that transforms face photographs into high-quality artistic portrait drawings.
Since artists usually use different drawing styles for different facial regions,
this hierarchical GAN model combines a global network (for fusing local parts) and six local networks (for individual facial regions).
Finally, to train this model, a novel line-promoting distance transform loss was proposed to capture the fact that an artist's drawing is usually not perfectly aligned with image features.

Rosin and Lai's algorithm~\cite{rosin-portrait} first stylises the
image with abstracted regions of flat colours plus black and white
lines~\cite{LaiRosin}, then fits a partial face model to the input
image and attempts to detect the skin region.  Shading and line
rendering is stylised in the skin region, and in addition, the face
model helps inform portrait-specific enhancements: reducing line
clutter; improving eye detail; colouring the lips and teeth; and
inserting synthesised highlights.
It is straightforward to modify this pipeline to render, in place of this ``puppet'' style,
a more abstracted version, inspired by the artist Julian Opie.

Winnem{\"oller} \emph{et al.}'s XDoG filter~\cite{xdog} can be conceptualised as the weighted sum
of a blurred source image and a scaled difference-of-Gaussians (DoG)
response of the same image, effectively applying unsharp masking to the DoG response. Combined with subsequent soft thresholding, this
computationally simple filter allows a wide range of stylistic and artistic effects, including cartoon shading, black-and-white
thresholding, and charcoal shading.
If required, local modification of filter parameters, according to facial features, would be trivial to implement.

Rosin and Lai~\cite{engraving} create an engraving style rendering of an image using a dither matrix,
which is a spatially-varying threshold.
The dither matrix has been designed so that it generates a pattern of black and white lines forming cross hatching.
The method is enhanced by using a simple cylindrical model of the face
to warp the dither matrix so that the lines curve around the face, providing a pseudo-3D effect.

Son \emph{et al.}~\cite{Son-hedcut}
proposed a novel method for hedcut, where dots and hatching lines with varying sizes are regularly spaced along local feature orientations. A smooth grid curved along the feature vector field, named a structure grid, is synthesized to store tangential and normal distances to the nearest grid intersection at each pixel. Given a structure grid, appropriate positions and attributes of primitives are determined via rapid pixel-based primitive rendering. The method works well for human faces even though it is not specially designed for portraits.


Semmo \emph{et al.}'s~\cite{semmo2016image} oil paint filter
is based on non-linear image smoothing to obtain painterly looks with a soft color blending. The method uses Gaussian-based filter kernels that are aligned to the main feature contours of an image for structure-adaptive filtering. By using the construct of the smoothed structure tensor and principles of line integral convolution to synthesize paint textures in real-time, the filter responses are locally controllable. In particular, the level of abstraction can be easily adjusted by interactive painting or could be based on facial feature masks.

Doyle \emph{et al.}'s~\cite{doyle2019automated}
pebble mosaic stylisation process begins with a superpixel
segmentation of the image, guided by an orientation field derived from the structure tensor.
Each superpixel is converted into a pebble by
first smoothing the exterior boundary and then computing a height field for the tile interior,
determined by harmonic interpolation between the
tile boundary and an interior contour placed at a set height.
The resulting 3D geometry can be conventionally rendered and textured,
using a tile color that is the average color of the pixels within the image segment.

Rosin and Lai~\cite{rosin2017watercolour} use a filter based approach to generate a watercolour stylisation.
In order to achieve the multiple characteristics of watercolour
-- namely brightening, abstraction, edge darkening, wobbling, granulation, glazing, pigment and paper variations --
they employ various steps such as smoothing, morphological opening and closing,
contrast-limited local histogram equalisation, edge detection, overlay blend, local geometric distortion,
superpixel segmentation, and level of detail masks controlled by face detection and saliency masks.

\section{Methodology}
\label{methodology}

Our initial guidelines for creating \emph{NPRportrait1.0} follow those articulated and developed for \emph{NPRgeneral}
and more particularly \emph{NPRportrait0.1}. The main principles are briefly recapped below:

\textbf{Challenging images}: The benchmark needs to include challenging images that are likely to be challenging to some extent for NPR algorithms.
Revealing weaknesses in the state-of-the-art helps drive research progress.

\textbf{Range of difficulty}: The benchmark should include images covering a range of levels of difficulty,
so as to better assess the level of performance of tested NPR algorithm, i.e. by indicating under what conditions they work, and when they fail.
Also, if all of the benchmark is too difficult then it will discourage users, and limit its take-up from the community.
To encourage widespread use, the first level should be attainable by the majority of existing methods.

\textbf{Small number of images}:
In comparison to the large benchmark datasets used in computer vision, the subjective nature of NPR evaluation
means that there will often need to be humans in the loop.
To facilitate this, the dataset should be as small as possible.
In addition, a danger is that if the dataset is too large to be manageable, then users will only use small selections,
and since different users would make different selections, the results across different papers would not be comparable,
defeating the original purpose of using a common benchmark.
Not only that, but it becomes possible for researchers to ``cherry pick'' results, which can more effectively be avoided
by creating a dataset sufficiently small that it can be treated in its entirety.
However, there is also the competing requirement that the benchmark should cover the target domain
(i.e. images that might be stylised) as thoroughly as possible.
We found that 20 images per level provided a good balance for both \emph{NPRgeneral} and \emph{NPRportrait0.1}.

\textbf{Facial characteristics}:
A number of characteristics to describe faces will be selected
to direct the construction of the design matrix in section~\ref{DesignMatrix}.
This facilitates ensuring both diversity and balance for these characteristics.
An additional benefit is that it provides a means to limit the challenge of earlier levels of the benchmark.
e.g. only allowing neutral expressions at level~1.

Some of the characteristics that we will use
have the drawback that the categories may not have precise boundaries,
and moreover that the participants in the user studies will be influenced by their cultural backgrounds,
as well as other biases.
Nevertheless, the benefits of such high-level sociological characteristics over alternative low-level features (e.g. smoothness, angularity)
is that humans have specialised mechanisms for the visual processing of faces,
and moreover develop from infancy mechanisms for making judgements about gender, ethnicity, attractiveness, etc.

\textbf{The gap between levels}:
The difficulty gap between level $n$ and level $n+1$ should not be too great
since we desire fine granularity of what conditions cause algorithms to fail.
However, again there is a trade-off, as a large number of levels would cause the benchmark to become too large and unwieldy.
\emph{NPRportrait0.1} provided 2 levels, and the authors proposed that there could be several more in the future.
In this paper we provide 3 levels for \emph{NPRportrait1.0}, which should be sufficiently demanding for the current state-of-the-art algorithms.
However, there remains scope for further levels which cover both more complicated scenes
(e.g. multiple people, full bodies, substantial occlusion, heavily cluttered background, extreme poses and expressions,
extreme perspective and other photographic distortions)
and broader coverage of portrait subjects (e.g., children, the elderly, more ethnicities).

\textbf{Variety of image sources}:
In order to provide a greater challenge to the NPR algorithms, the images should come from a wide variety of sources so as
to ensure that a variety of cameras, lighting conditions, backgrounds, poses, and varied levels of professionalism of the photographers and the subjects
are included.

\textbf{Image resolution}: Most NPR algorithms are suitable for medium resolution images,
and so all images will have a fixed height of 1024.
This also simplifies running some NPR algorithms as they may have scale parameters that can therefore be held constant across the dataset.\footnote{
However, future NPR benchmarks should expand on the issue of image resolution.
Many commercial stylisation apps need to operate on images of arbitrary sizes.
Moreover, they typically provide a lower resolution preview (e.g. when changing interactive settings).
Thus a good stylisation algorithm would ideally be resolution-independent.
}

\textbf{Copyright clearance}: Since (manual) visual evaluation of results remains an important part of NPR,
the benchmark images should have copyright clearance so that they can be published along with the derived results.

\subsection{Design Matrix}
\label{DesignMatrix}

For each of the benchmark levels, a set of desired characteristics will be defined
that all the images should satisfy (e.g., frontal view).
There is also another set of desired characteristics which should vary
(e.g., subjects' gender, ethnicity, expression), and these will be constrained to a set of categories
(e.g., \{young adult, middle-aged adult\}).
With 20 images in a benchmark level, it is not possible to cover
all combinations of these characteristics.
Instead, treating the characteristics as independent,
we will use the methodology of generating a ``nearly orthogonal design matrix''
to capture a good representative set of images, rather than rely on a full factorial design.
We use the optFederov function from the R package AlgDesign~\cite{wheeler2014algdesign},
which allows a number of runs (in our case, images) to be specified,
as well as allowing for different numbers of values for each of the input variables.

\subsection{Level 1}

\begin{table}[t]
\centering
\caption{Design matrix for level 1.}
\label{matrix1}
\small
\begin{tabular}{|c|c|c|c|}
\hline
\bf gender & \bf age & \bf attractiveness & \bf ethnicity \\ \hline \hline
female & middle & average & black \\ \hline
female & young & average & black \\ \hline
male & middle & below & black \\ \hline
female & young & below & black \\ \hline
male & middle & above & black \\ \hline
male & young & above & black \\ \hline
male & middle & average & South Asian \\ \hline
male & young & average & South Asian \\ \hline
female & young & below & South Asian \\ \hline
female & middle & above & South Asian \\ \hline
female & middle & average & East Asian \\ \hline
male & middle & average & East Asian \\ \hline
female & middle & below & East Asian \\ \hline
male & young & below & East Asian \\ \hline
female & young & above & East Asian \\ \hline
male & young & above & East Asian \\ \hline
female & young & average & white \\ \hline
male & young & average & white \\ \hline
male & middle & below & white \\ \hline
female & middle & above & white \\ \hline
\end{tabular}
\end{table}
Level~1 is intended to be straightforward to stylise, and thus  many restrictions are imposed.
Each image should contain only a frontal, approximately upright, and unoccluded view of a single face which has a forwards gaze direction.
The images must contain essentially no background objects or clutter,
effectively providing a clear separation of the face from the background.
The backgrounds are homogeneous, but natural -- they were not manually masked out.
The images should be dominated by the face, which should fill most of the image and be cropped approximately at the neck
so as to include only minimal clothing; other body parts such as the hands are excluded.
To further simplify the task of stylisation,
the subject in the portrait should not have facial hair or long hair that partly covers the face,
should not wear jewellery or other accessories such as a pipe,
glasses, or hat.
Harsh or complex lighting is avoided, and only soft lighting used.
Finally, all the subjects should have approximately neutral expressions.

\emph{NPRportrait0.1} included \emph{face shape} as
a variable characteristic,
identified using the following set of descriptors: \{round, square, oval, heart, long\}.
At the time it was noted that these were not strictly defined, and that due to the differences between some shapes being subtle, it meant that
the attribution of face shape to images was only approximate.
One of the differences in construction between \emph{NPRportrait1.0} and \emph{NPRportrait0.1} is that
the characteristics of images are now more rigorously checked by running user studies for validation.
We found in preliminary tests that face shape could not be reliably determined,
and so this characteristic has been excluded from the current benchmark.

Another change for the new benchmark is that ethnicity has been expanded from three to four categories,
with Asian being split into East Asian (e.g. Chinese) and South Asian (e.g. Indian).

The remaining characteristics that appear in the design matrix are the
same as before: gender, age, and attractiveness.  There are two
categories for gender, \{male and female\},\footnote{
Gender was assessed by the authors and the participants of the user studies as a binary label,
based on visual characteristics, and is not necessarily aligned with the subject's personal gender identification.
}
and for age, \{young adult, middle-aged adult\}.
Finally, we have specified three levels
of attractiveness: \{below average, average, above average\}.  It is
important to control attractiveness since there is a tendency in the
NPR literature to use aesthetically pleasing images with attractive
and/or interesting faces.  However, stylisation should also be
effective for unattractive or ordinary faces.

\subsection{Level 2}

\begin{table}[t]
\centering
\caption{Design matrix for level 2.}
\label{matrix2}
\small
\begin{tabular}{|c|c|c|}
\hline
\bf gender & \bf expression & \bf facial hair \\ \hline
\hline
male & negative & none \\ \hline
male & neutral & none \\ \hline
female & neutral & --- \\ \hline
female & positive & --- \\ \hline
male & negative & moustache \\ \hline
female & neutral & --- \\ \hline
male & positive & moustache \\ \hline
female & positive & --- \\ \hline
male & negative & beard \\ \hline
female & negative & --- \\ \hline
male & neutral & beard \\ \hline
female & positive & --- \\ \hline
female & negative & --- \\ \hline
male & neutral & goatee \\ \hline
female & neutral & --- \\ \hline
male & positive & goatee \\ \hline
female & negative & --- \\ \hline
male & neutral & stubble \\ \hline
female & neutral & --- \\ \hline
male & positive & stubble \\ \hline
\end{tabular}
\end{table}

The criteria and design matrix for level~2 are unchanged from that in \emph{NPRportrait0.1}.
Level 2 retains many of the restrictions enforced in level~1: each
image contains a frontal, approximately upright, unoccluded view of a
single face that fills most of the image, is cropped to include
minimal clothing, and does not include hands or other body parts. The
background should be relatively plain, but since this requirement is
not as strict as for level~1, some mostly unobtrusive background
content is present. The requirement for unadorned faces is also
relaxed, and so some jewellery is allowed. Likewise,
level~1's requirement for moderate lighting is maintained, but relaxed
a little.  Gaze direction is mostly forwards, but not exclusively.
Ages are again restricted to adult, but are not considered as a
control variable for this level.

Regarding desirable variations, like level~1 an equal distribution of gender is maintained.
Facial expressions have been broadened from neutral in level~1 to three categories: \{negative, neutral, positive\},
but extreme versions of these facial expressions should be avoided.
The latter restriction is imposed as otherwise the fitting of face models (used by the face-specific NPR algorithms) becomes unreliable,
and also it avoids the stylisation task becoming too challenging
(i.e. the gap between levels~1 and~2 should not be large).
The final factor to control at level~2 is to include varieties of facial hair; we used the following categories: \{none, moustache, beard, goatee,
stubble\}, and assumed that females had no facial hair.

Unlike level~1, for practical reasons the design matrix does not include controls for age, attractiveness or ethnicity.
As more control factors are applied, then it becomes progressively more difficult to source images that satisfy all these constraints.
However, where images are available, we try to maintain a reasonable spread of these characteristics.

\subsection{Level 3}

\begin{table}[t]
\centering
\caption{Design matrix for level 3.}
\label{matrix3}
\small
\begin{tabular}{|c|c|c|c|}
\hline
\bf gender & \bf lighting & \bf expression / & \bf skin / \\
& & \bf eyes & \bf occlusion \\ \hline \hline
male & complex & extreme & skin marking \\ \hline
female & complex & extreme & skin marking \\ \hline
female & complex & regular & skin marking \\ \hline
male & simple & regular & skin marking \\ \hline
male & complex & odd & skin marking \\ \hline
male & simple & eyes & skin marking \\ \hline
female & simple & eyes & skin marking \\ \hline
female & simple & extreme & occlusion \\ \hline
male & complex & extreme & occlusion \\ \hline
female & complex & regular & occlusion \\ \hline
male & simple & odd & occlusion \\ \hline
female & simple & odd & occlusion \\ \hline
male & complex & eyes & occlusion \\ \hline
female & complex & eyes & occlusion \\ \hline
male & simple & extreme & regular \\ \hline
female & simple & regular & regular \\ \hline
male & complex & regular & regular \\ \hline
male & complex & odd & regular \\ \hline
female & complex & odd & regular \\ \hline
female & complex & eyes & regular \\ \hline
\end{tabular}
\end{table}

Level~3 roughly maintains the previous criteria, but is not as strict.
The cropping can be less tight, the pose can be
less frontal, and there can be background clutter.  Several other
factors are relaxed in a systematic manner via the design matrix.  A
variety of lighting effects are allowed, and are categorised in the
design matrix as \{simple, complex\}, where ``simple'' indicates the
soft frontal lighting that has been used in the previous two levels,
and ``complex'' encompasses anything else such as side lighting, back
lighting, strong lighting, strong shadows, or unusual lighting effects.
There are now four categories of expression: \{regular, extreme, odd,
eyes\}, where ``eyes'' indicates that eyes are not open and
forward facing as before.  The final variations concern either
additions to or occlusions of the face;
additions typically mean  skin markings such scars, tattoos,
freckles, strong makeup, strong specularities, while occlusions
are caused by objects such as jewellery, hats, glasses, or hands.  This level is
less strict on viewpoint, but initial attempts to systematically
sample different viewpoints in the design matrix were abandoned due to
difficulties in sourcing sufficient images that also satisfied the
other conditions.

\subsection{Image Selection}

Following the criteria for the three levels of the benchmark,
the resulting nearly orthogonal design matrices for levels~1, 2, and~3 are
shown in Tables~\ref{matrix1},~\ref{matrix2} and~\ref{matrix3}.
The next step is to acquire images that satisfy these design matrices,
and are also consistent with our goals of using a variety of image sources
and of course have copyright clearance and sufficient image resolution.
This was found to be challenging, and even after collecting hundreds of images that were potentially suitable,
it was difficult to satisfy the design matrices.
As noted in~\cite{rosin2017benchmarking}, when constructing
\emph{NPRportrait0.1} it was found that the majority of photographs available online
were taken under uncontrolled conditions, and hence have complicated
backgrounds, harsh lighting, non-frontal view, occlusion or other
factors that often made then unsuitable; moreover, many do not provide
sufficient or explicit copyright clearance.

\subsubsection{Level 1}

Since level~1 contains the most tightly controlled images, this required the most amount of work to ensure that suitable images were selected.
First a collection of 540 photos was acquired from sources such as Wikimedia Commons, Flickr, and unsplash, as well as photographs from the authors' own collections.
A user study was carried out to collect the main characteristics of the faces that appear in the design matrix:
age, attractiveness, and ethnicity.
Note that here and in later user studies, users were given a choice of four categories for the question about age,
even though we only aim to capture portraits for two age groups.
These groups were bracketed above and below by the categories \emph{child} and \emph{old}
so that we could reject unsuitable images.
Even when characteristics are well defined, as age is, we do not have access to any ground truth, and so all the characteristics
are determined from the appearances in the images.
Due to the large number of images, each participant only saw a small proportion of the images,
namely 49, so that users could complete the study within an acceptable time period.

The most uncertain (or contentious) characteristic is attractiveness, since the perception of attractiveness is very subjective,
and varies widely across participants,
depending on many factors such as age and gender~\cite{mclellan1993effects}, ethnicity, cultural background,
rural versus urban living~\cite{batres2017familiarity},
and even just recent experiences~\cite{cooper2008influence}.
Moreover, if we consider that the level of attractiveness is a normally distributed random variable,
then it follows that the majority of the population will lie close to the average,
and so our collection of $N=540$ images will contain relatively few
faces that are significantly above or below average attractiveness.

We took the approach of assigning an attractiveness score to each face, calculated as the mean user judgement,
where the user judgements are scored as \{-1, 0, +1,\}  for \{below average, average, above average\}.
The images are then ranked according to the users' mean judgement.
Two thresholds were set on the ranks, $T_1=85$ and $T_2=166$ such that images ranked below $T_1$ or above $N-T_1$
were considered to be significantly below or above average attractiveness respectively,
and images ranked in the range $T_2$ to $N-T_2$ were considered to be of average attractiveness.
Treating the distribution of attractiveness scores as normal with zero mean,
this is equivalent to setting the thresholds
such that images that appear in the distribution in the ranges $ [ -\infty,\sigma ] $ and $ [\sigma, \infty] $
are selected as having below and above average attractiveness respectively,
while images in the range $[ \frac{N}{2}-\frac{\sigma}{2}, \frac{N}{2}+\frac{\sigma}{2} ]$
are treated as having average attractiveness.
Note that the three ranges were kept disjoint so that the three categories should appear distinct.
Ideally we would have preferred to make the threshold for $T_1$ based on a value larger than $\sigma$ (e.g. $2\sigma$ or $3\sigma$),
but this was not possible as we were then unable to fill all the rows of the design matrix with candidate images.

A further consideration at this stage was that images were retained for consideration in the design matrix
only if the majority response from the user study was consistent.

We did not include gender in this study as it is a less subjective
quantity, and omitting it reduced demands on the users.  At this stage
the assessment of gender was done by the authors; however, a later
user study will provide further validation all four characteristics,
including gender.

\subsubsection{Level 2}

Since the design matrix for level~2 did not change, the images
previously used in \emph{NPRportrait0.1} could be potentially retained.
However, the characteristics of expressions are subtle,
and so a second
user study was carried out to determine if the perceived facial
expressions were correct.  Initial tests showed problems with some
images, and so the full user study eventually included the 20 images
from \emph{NPRportrait0.1} plus another 13 images.  All 22 participants
saw all the 33 images.  The result was that four of the original
images have now been replaced with new images that the user study
confirmed display the appropriate expression (i.e. negative, neutral,
or positive) more consistently.  In addition, one image was moved (from
row 13 to row 15) since it was considered to have a neutral rather
than negative expression.

\subsubsection{Level 3}

Since the characteristics of this level are straightforward,
and also since the level is less tightly controlled,
we did not consider it necessary to run a user study for the characteristics specific to this level.

\subsubsection{The full three-level benchmark}

\begin{figure*}[!t]
\renewcommand*\thesubfigure{\arabic{subfigure}}

\centering
\subfloat[]{\includegraphics[height=2.2cm]{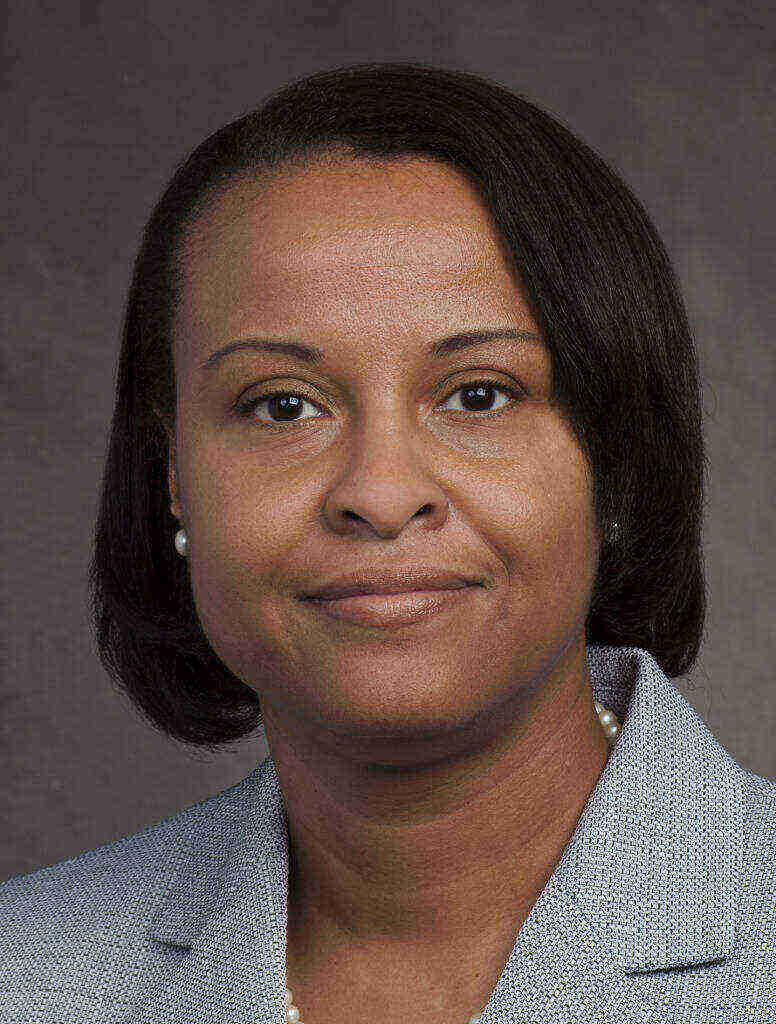}}
\subfloat[]{\includegraphics[height=2.2cm]{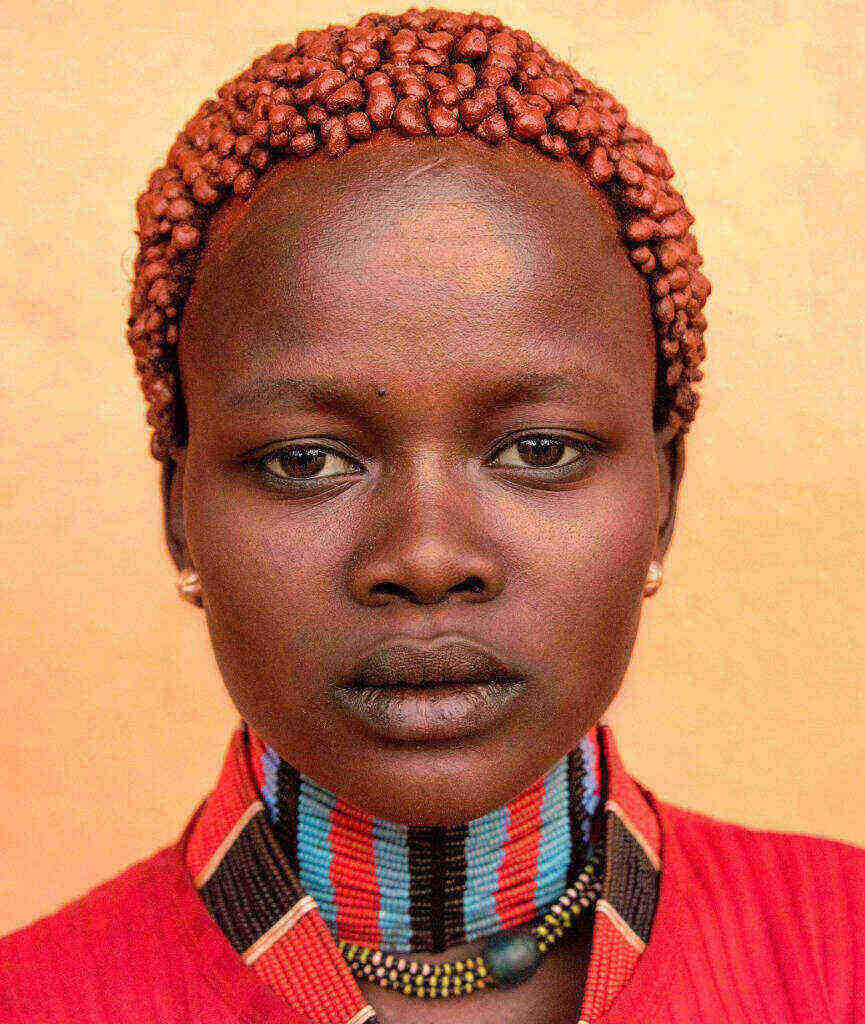}}
\subfloat[]{\includegraphics[height=2.2cm]{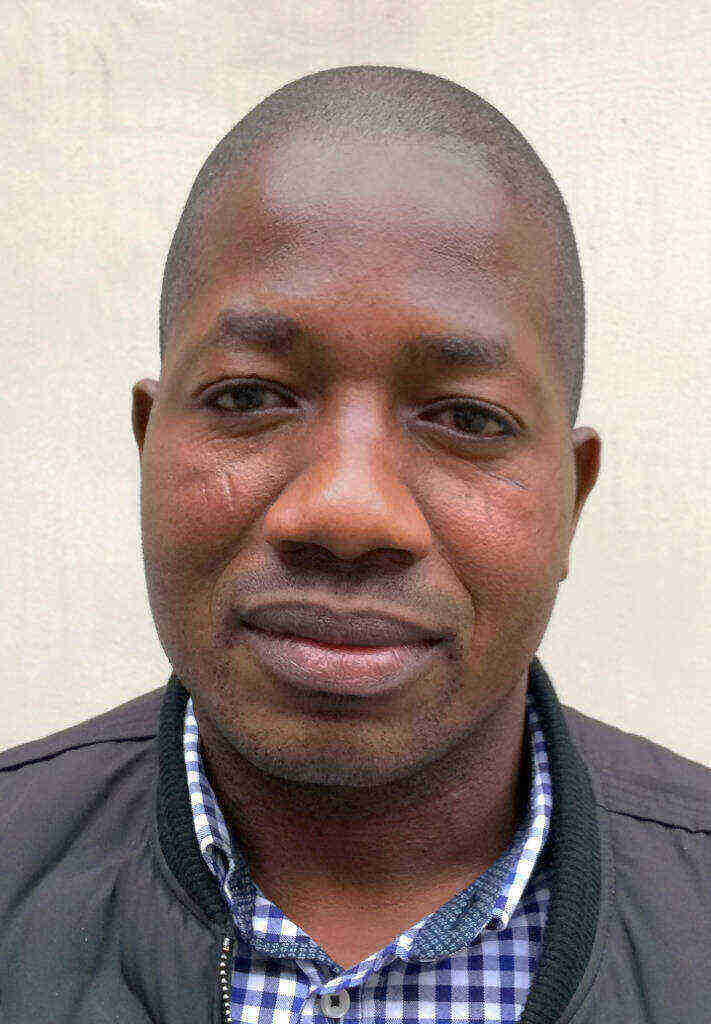}}
\subfloat[]{\includegraphics[height=2.2cm]{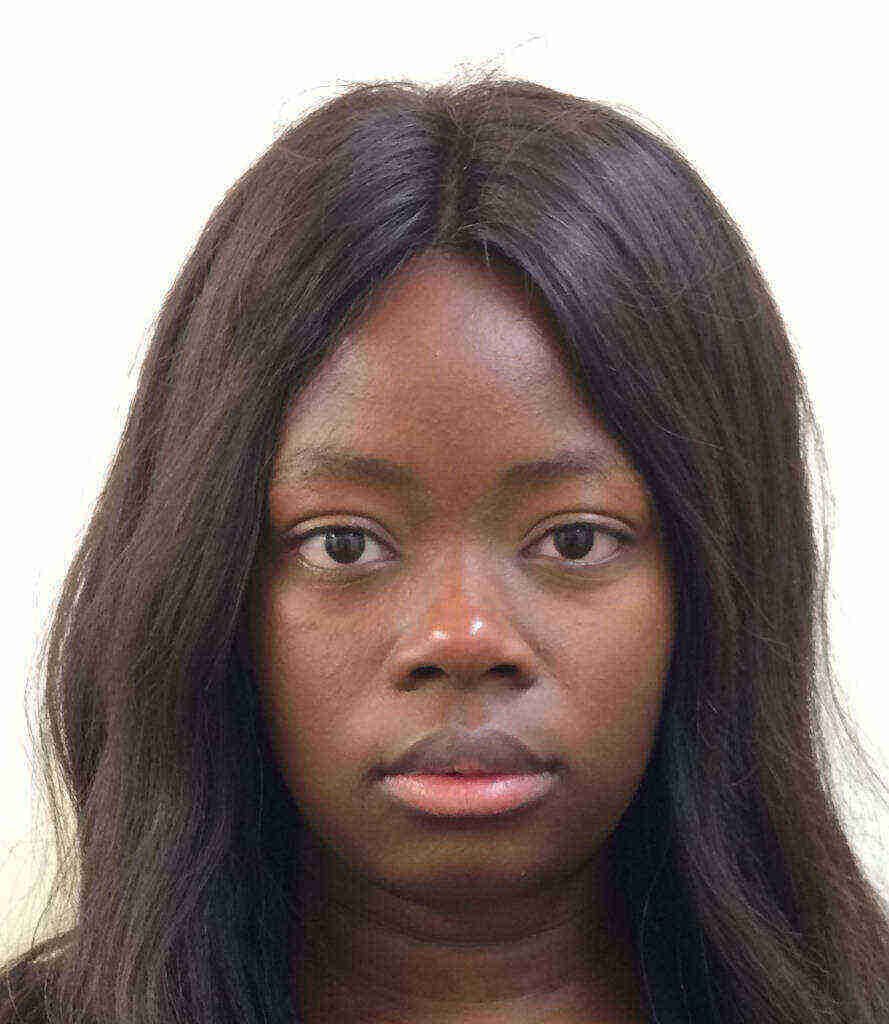}}
\subfloat[]{\includegraphics[height=2.2cm]{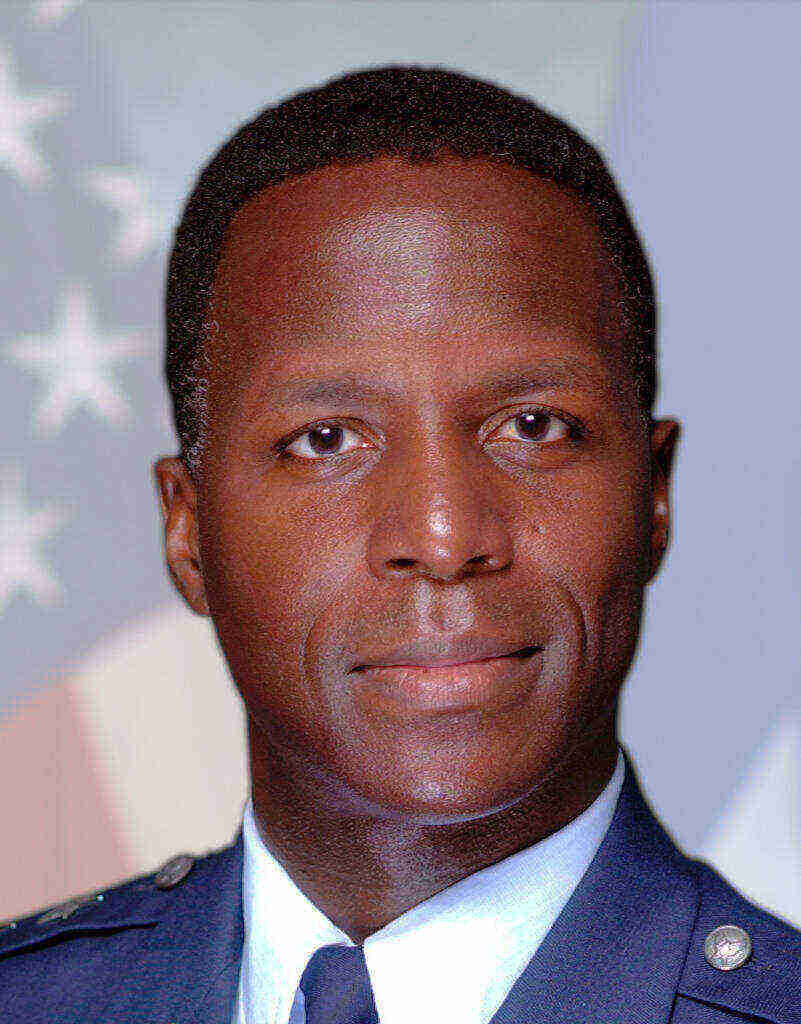}}
\subfloat[]{\includegraphics[height=2.2cm]{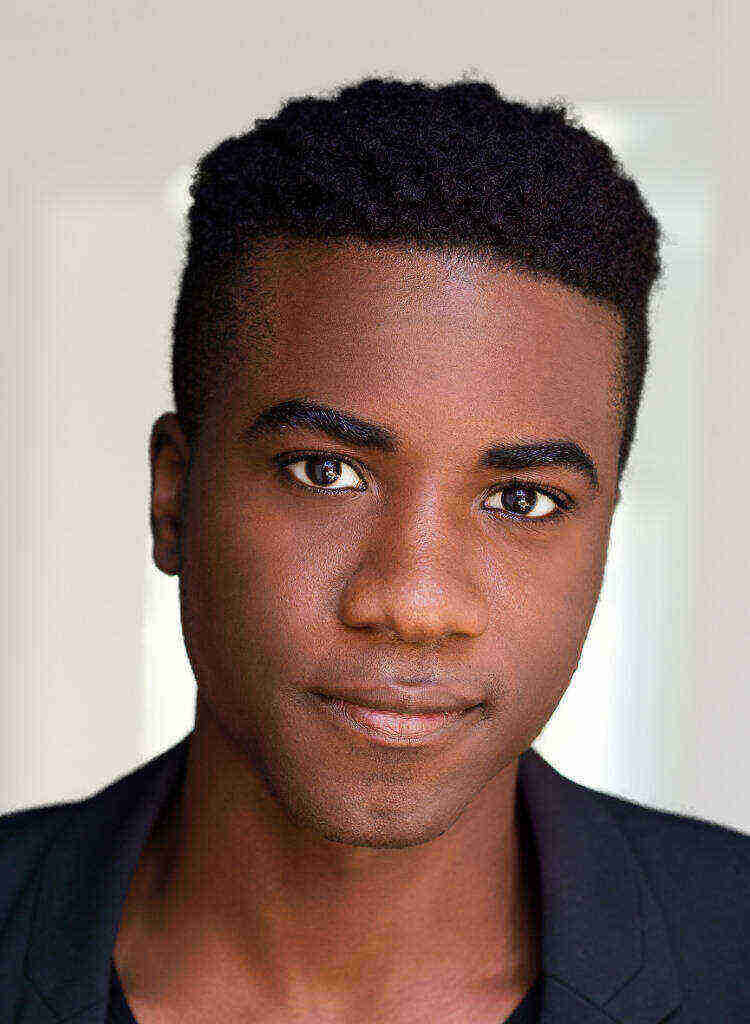}}
\subfloat[]{\includegraphics[height=2.2cm]{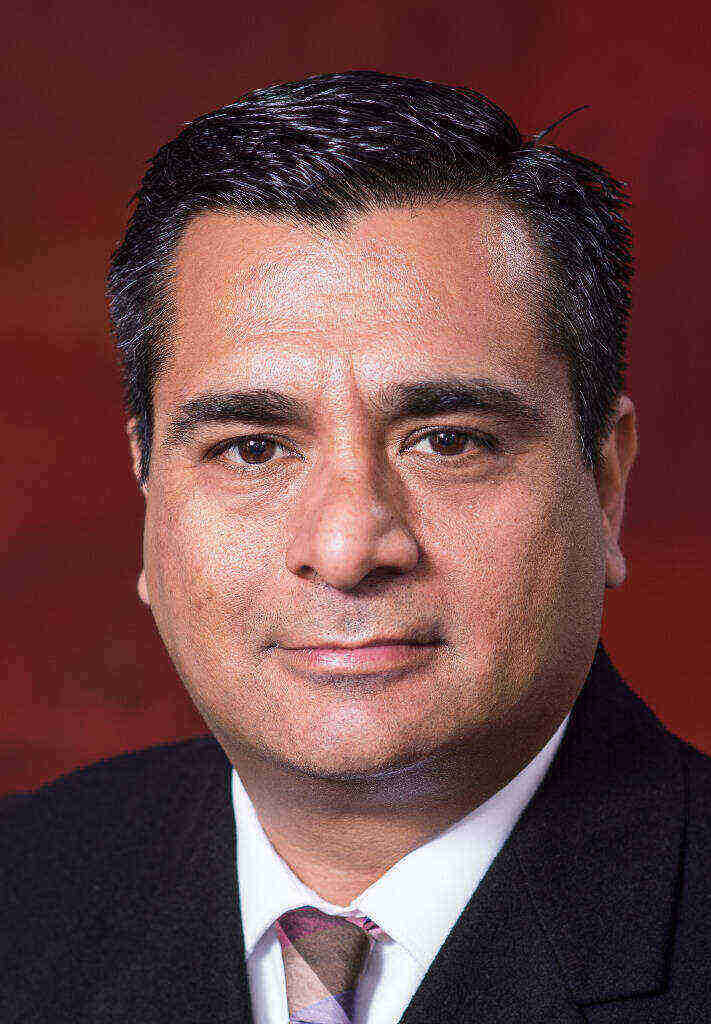}}
\subfloat[]{\includegraphics[height=2.2cm]{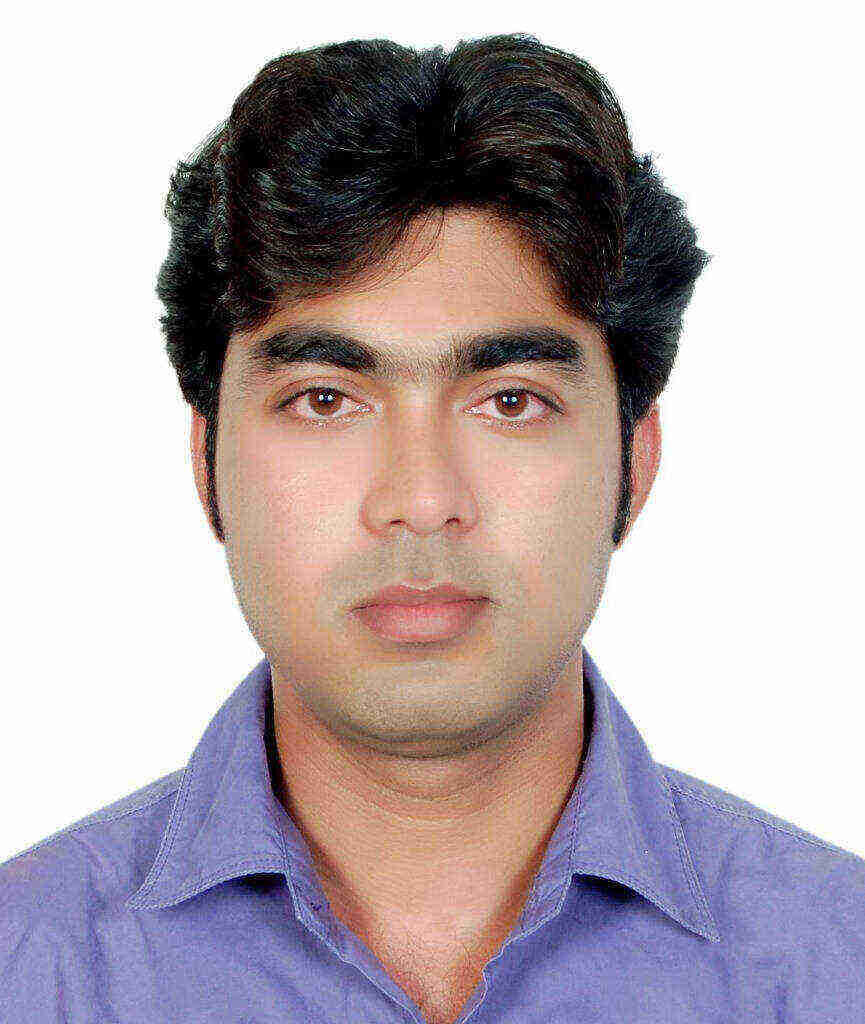}}
\subfloat[]{\includegraphics[height=2.2cm]{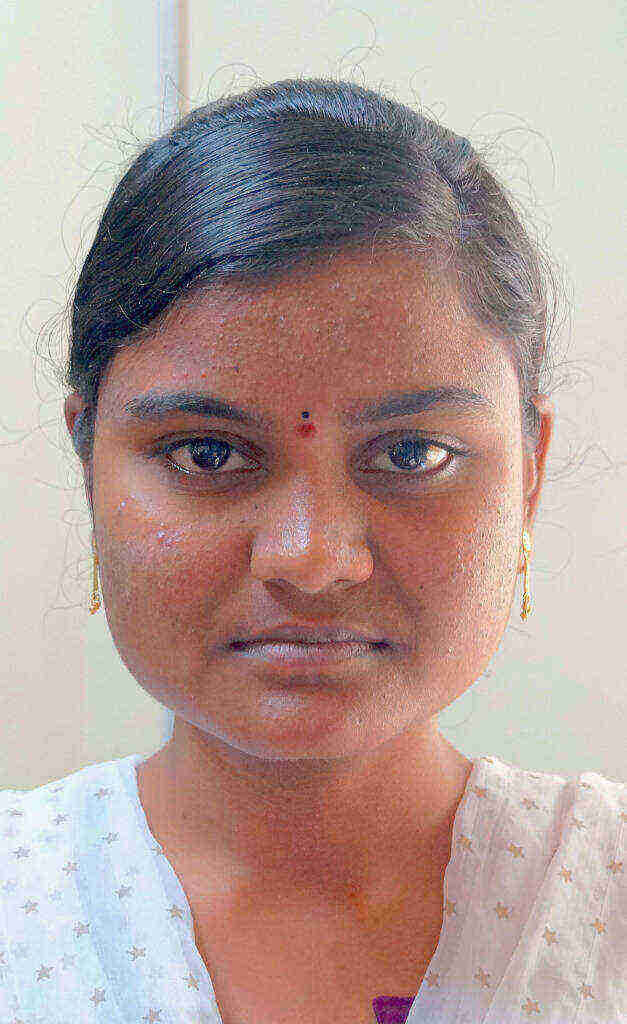}}
\subfloat[]{\includegraphics[height=2.2cm]{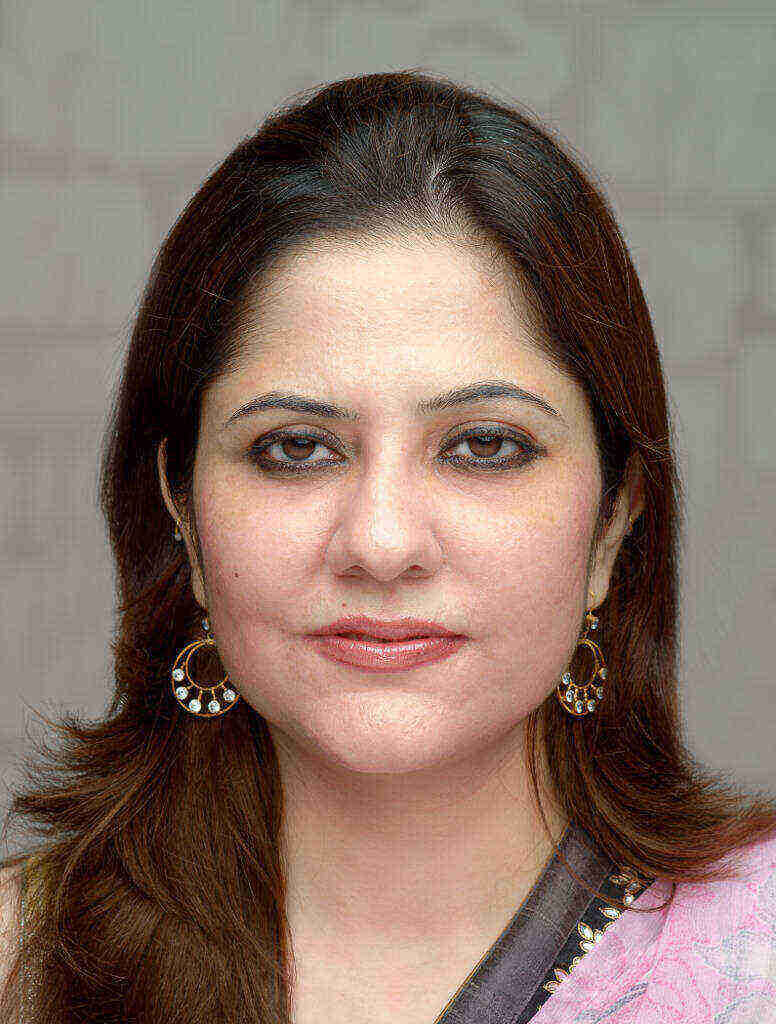}}\\[-1ex]

\subfloat[]{\includegraphics[height=2.2cm]{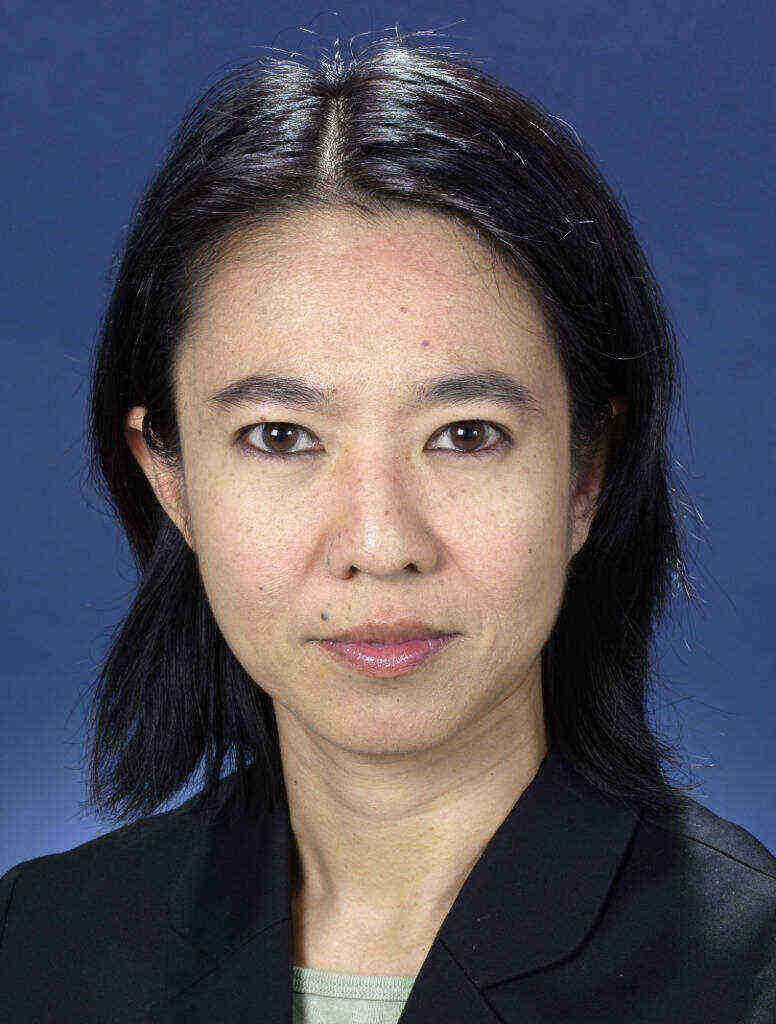}}
\subfloat[]{\includegraphics[height=2.2cm]{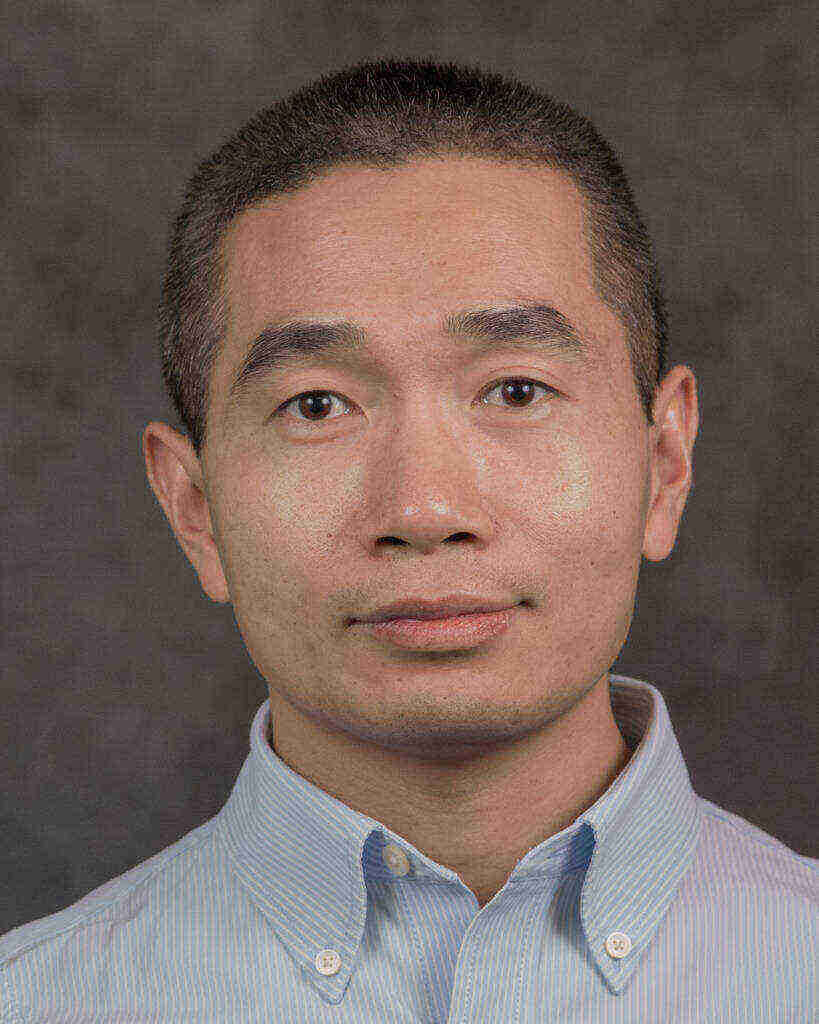}}
\subfloat[]{\includegraphics[height=2.2cm]{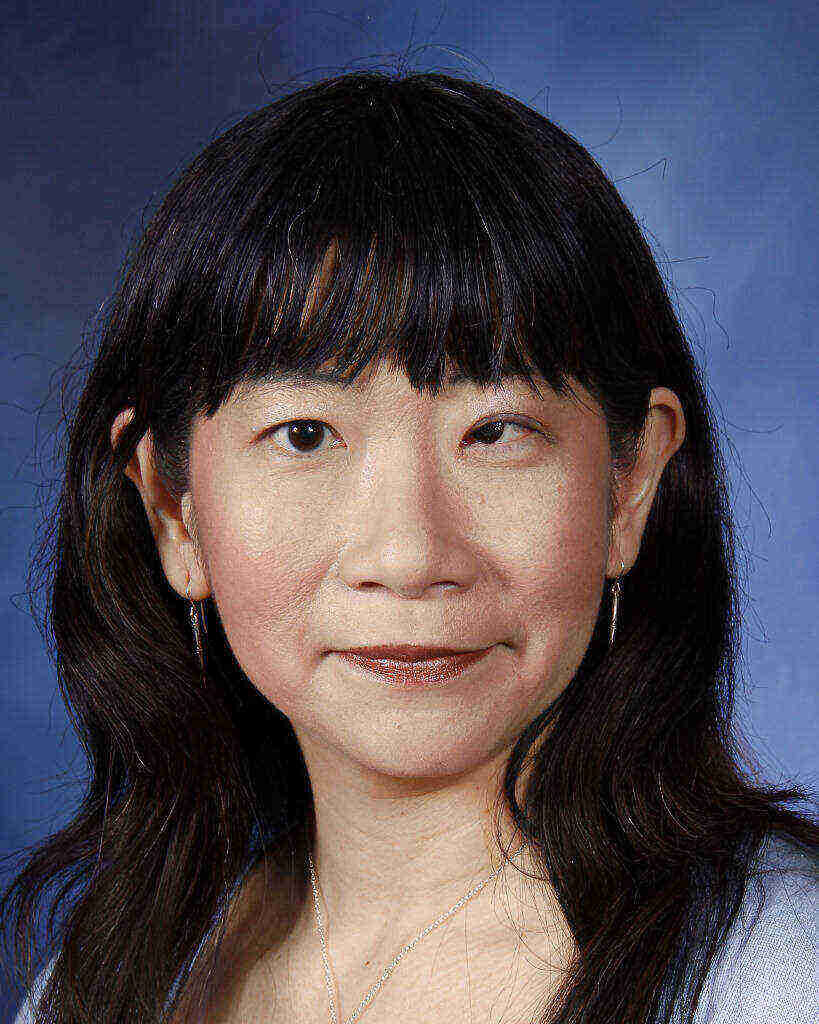}}
\subfloat[]{\includegraphics[height=2.2cm]{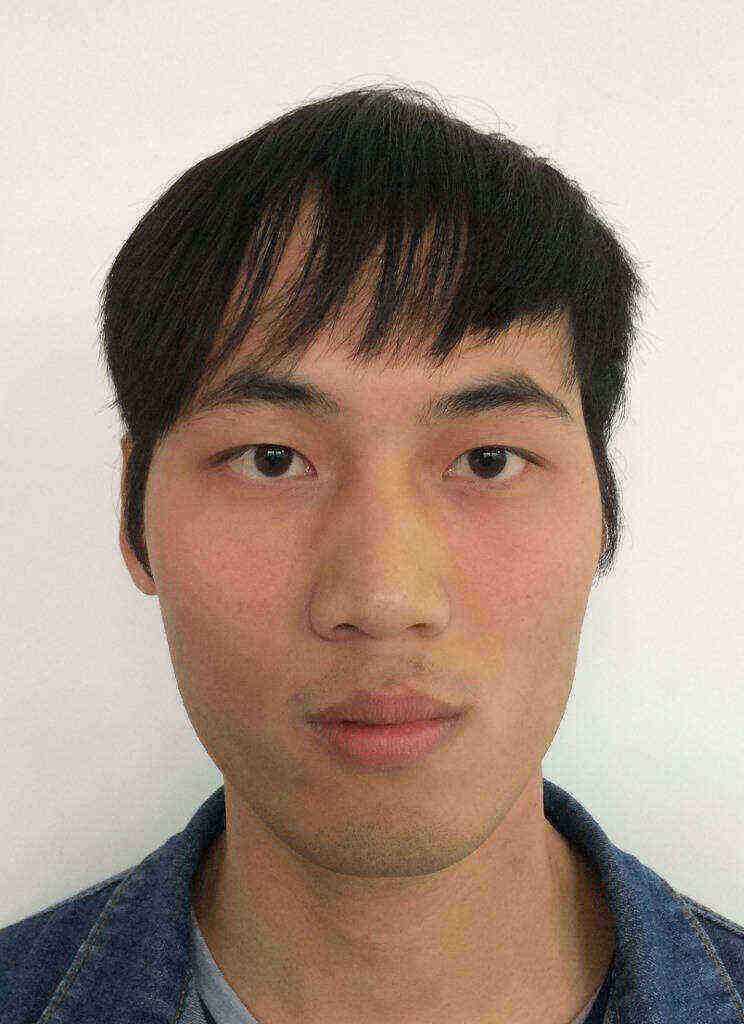}}
\subfloat[]{\includegraphics[height=2.2cm]{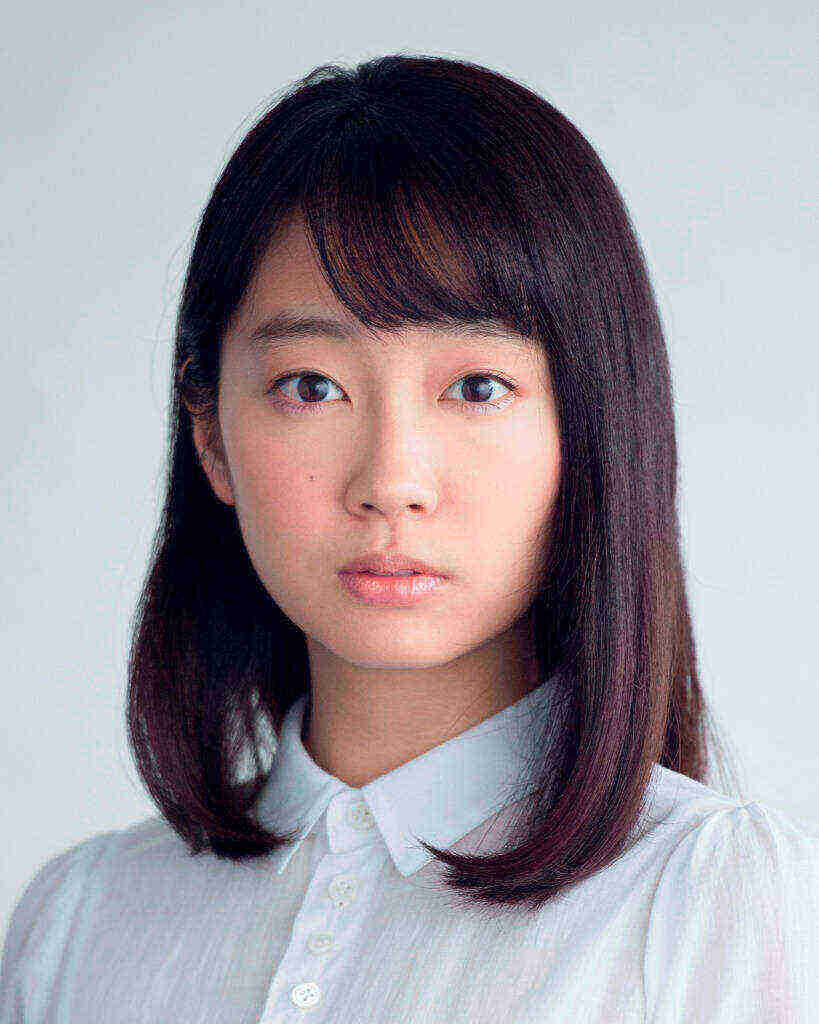}}
\subfloat[]{\includegraphics[height=2.2cm]{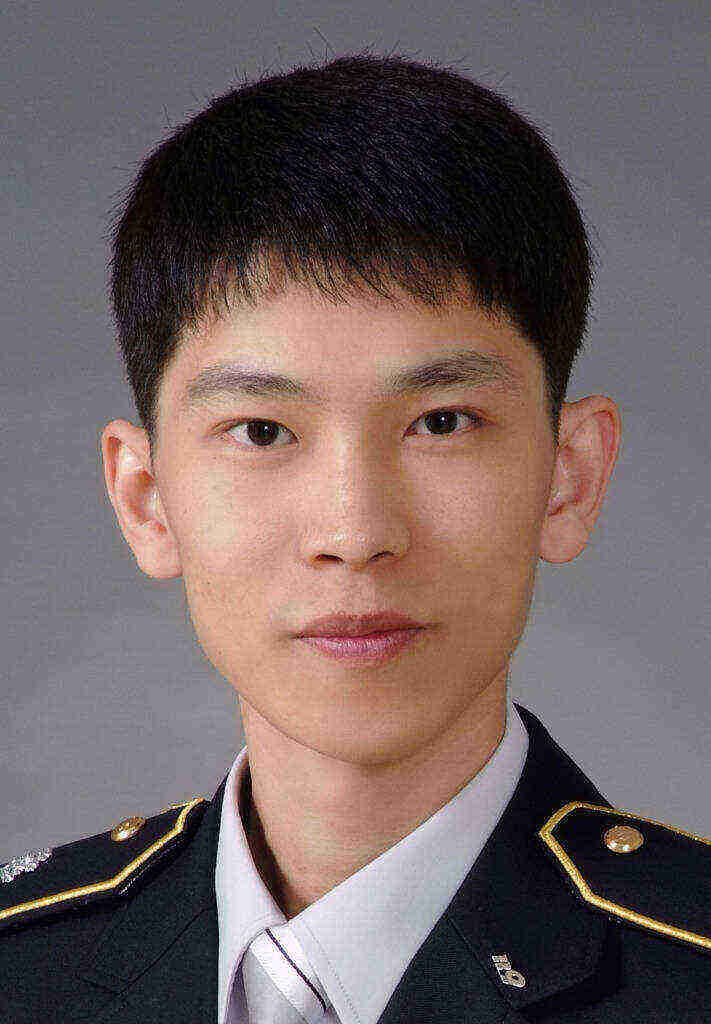}}
\subfloat[]{\includegraphics[height=2.2cm]{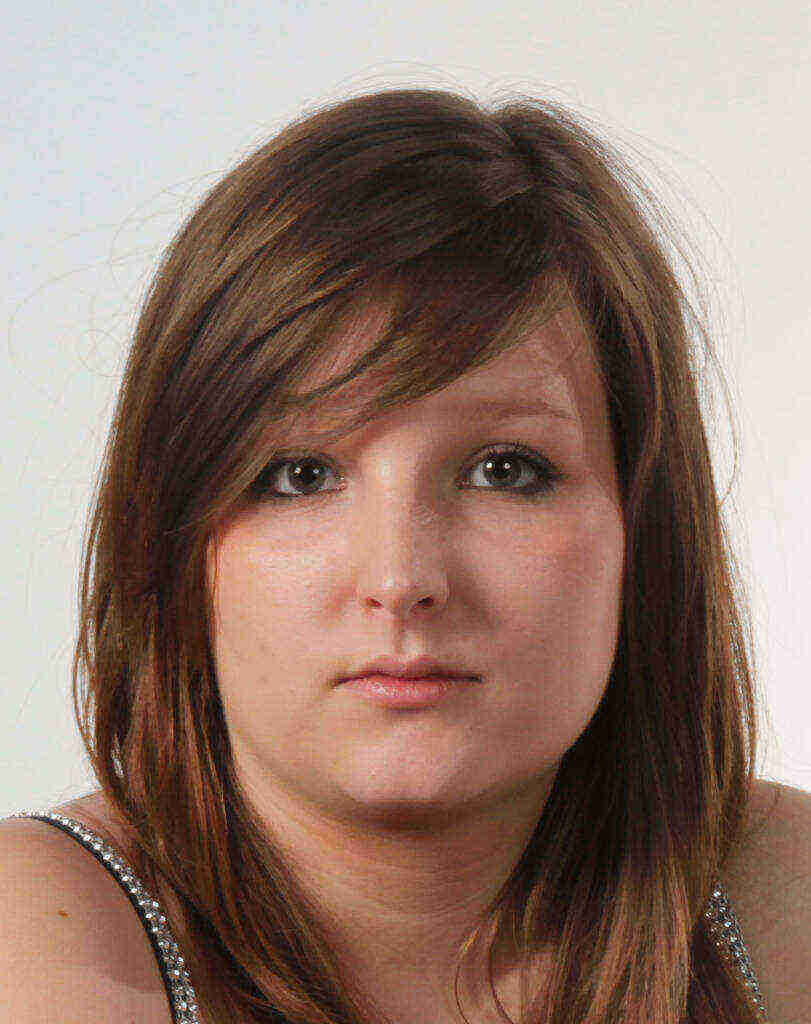}}
\subfloat[]{\includegraphics[height=2.2cm]{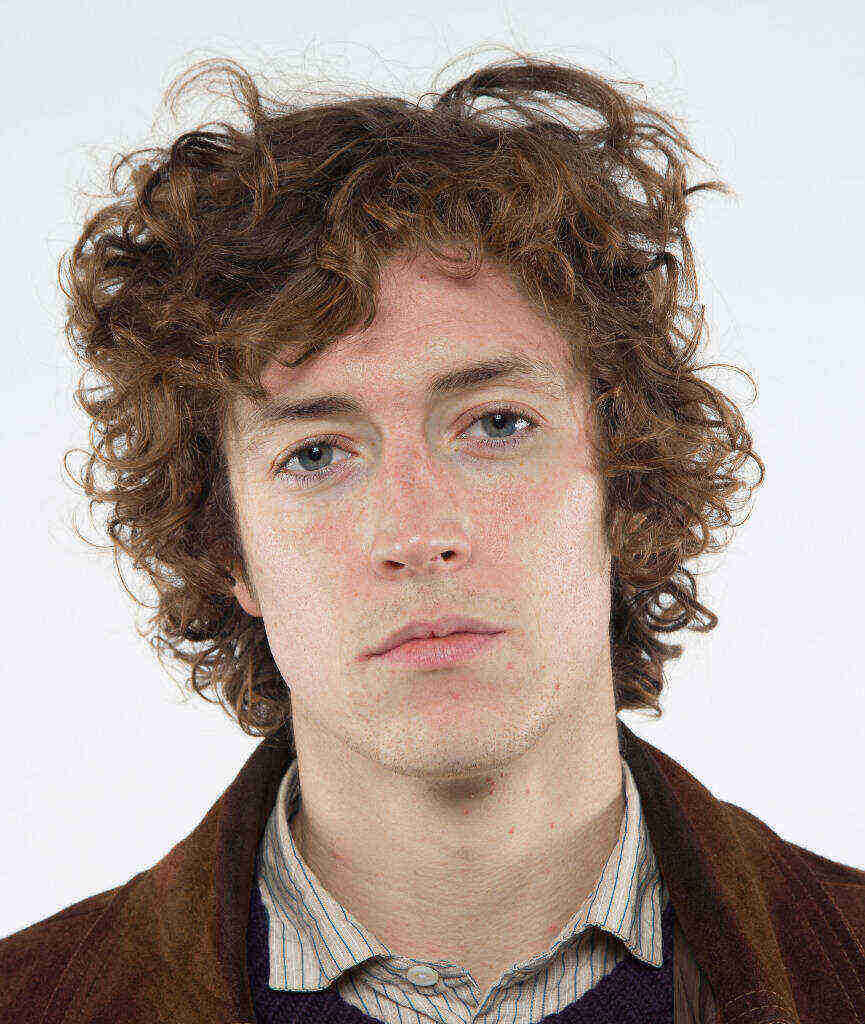}}
\subfloat[]{\includegraphics[height=2.2cm]{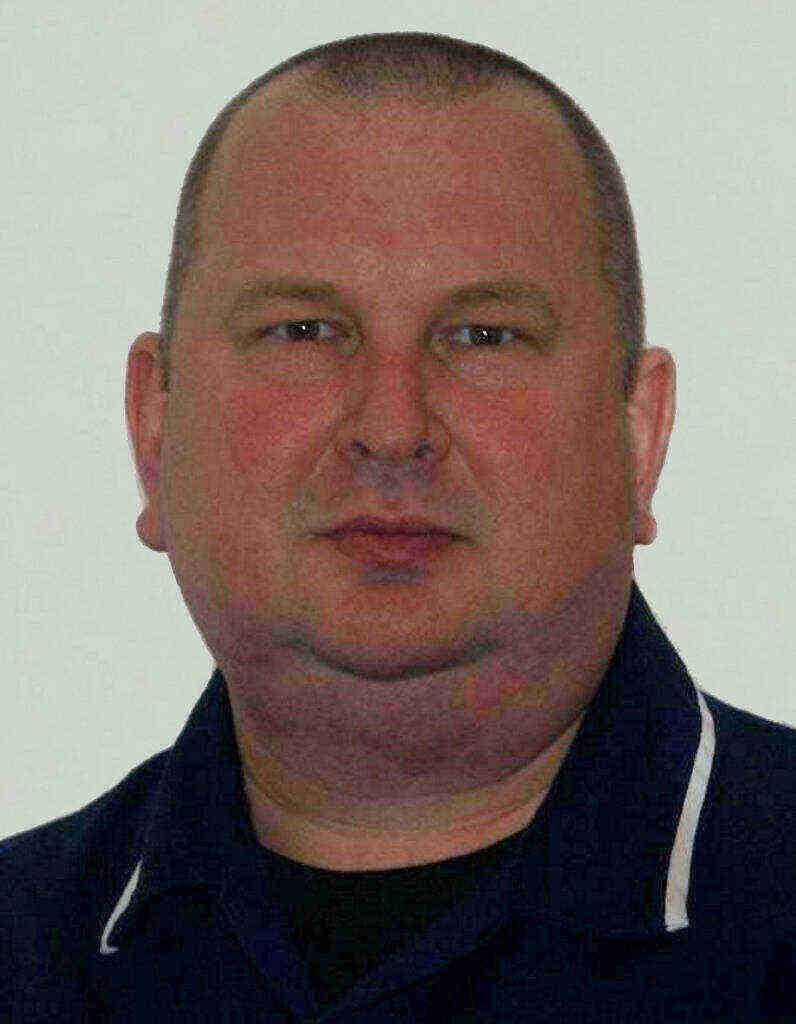}}
\subfloat[]{\includegraphics[height=2.2cm]{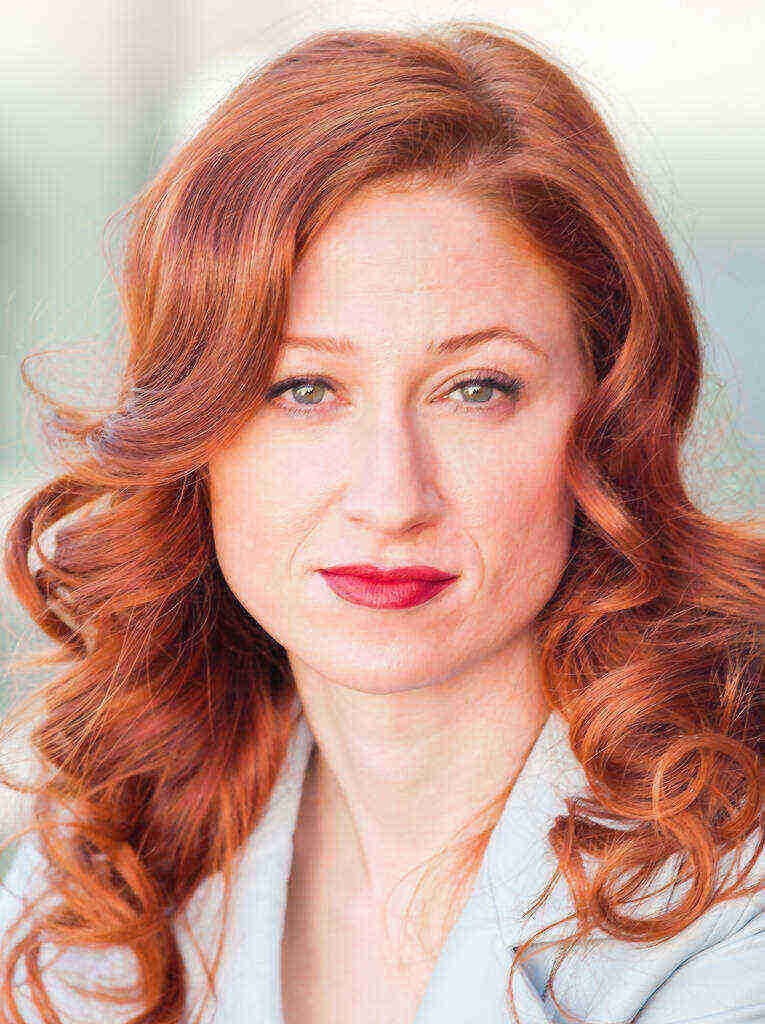}}

\centerline{Level 1}
\smallskip
\setcounter{subfigure}{0}

\subfloat[]{\includegraphics[height=2.1cm]{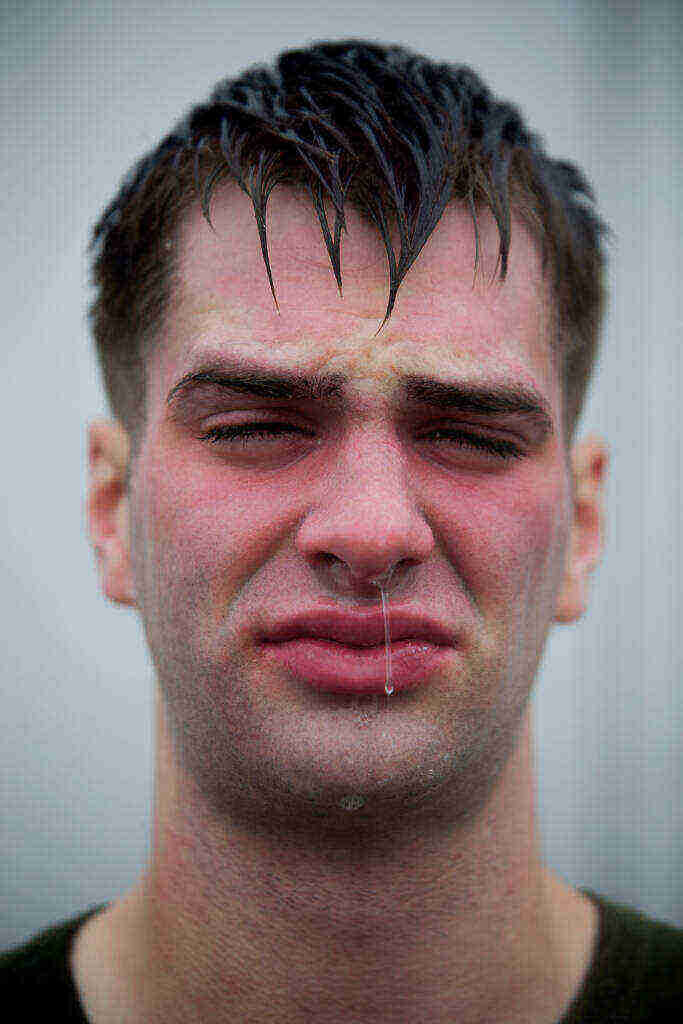}}
\subfloat[]{\includegraphics[height=2.1cm]{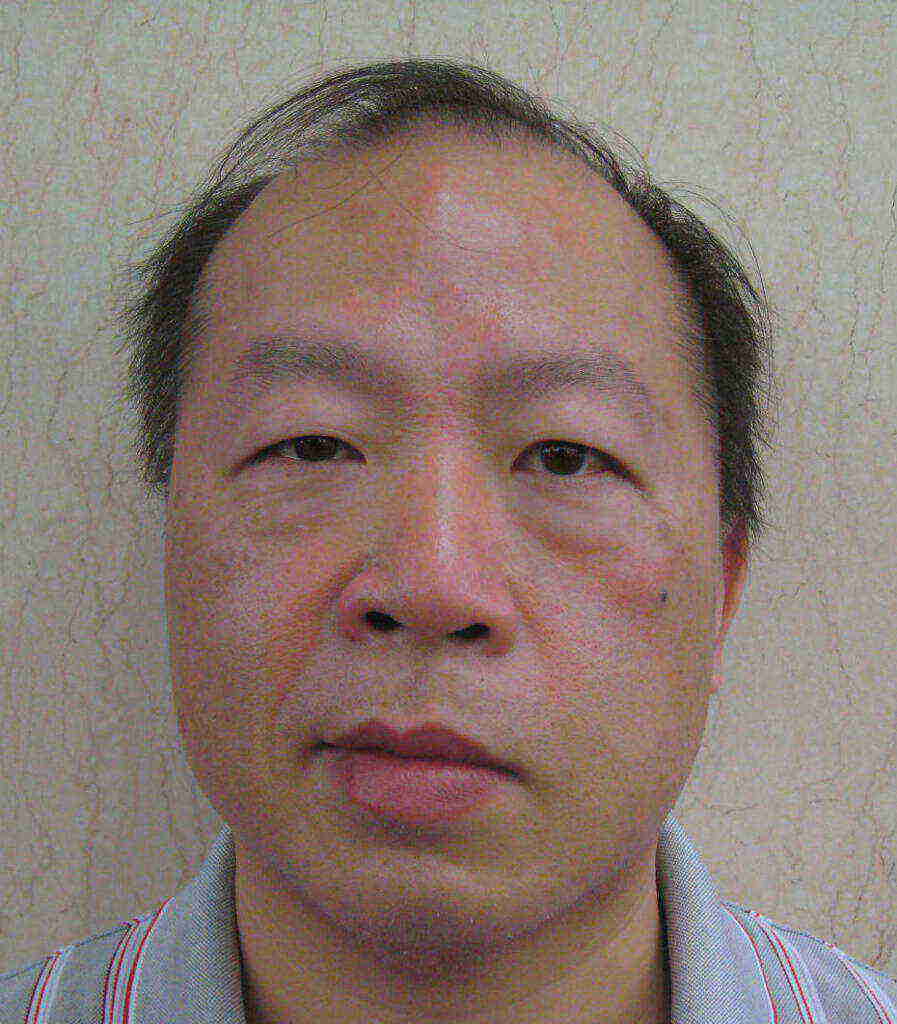}}
\subfloat[]{\includegraphics[height=2.1cm]{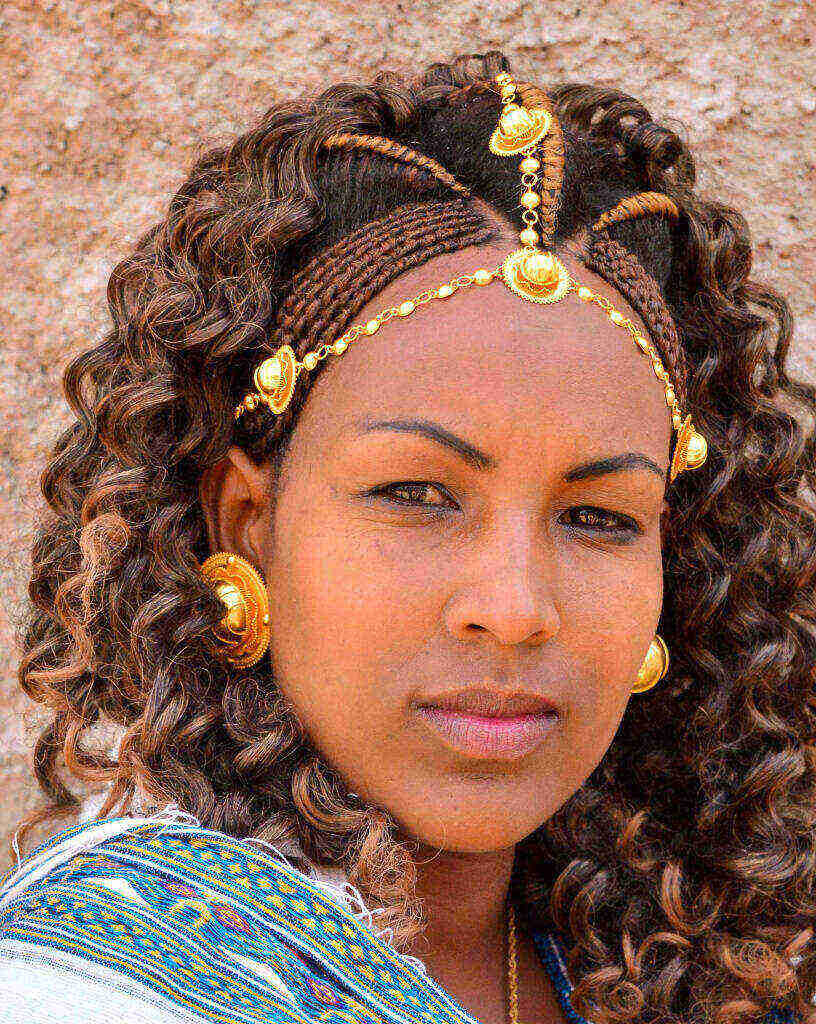}}
\subfloat[]{\includegraphics[height=2.1cm]{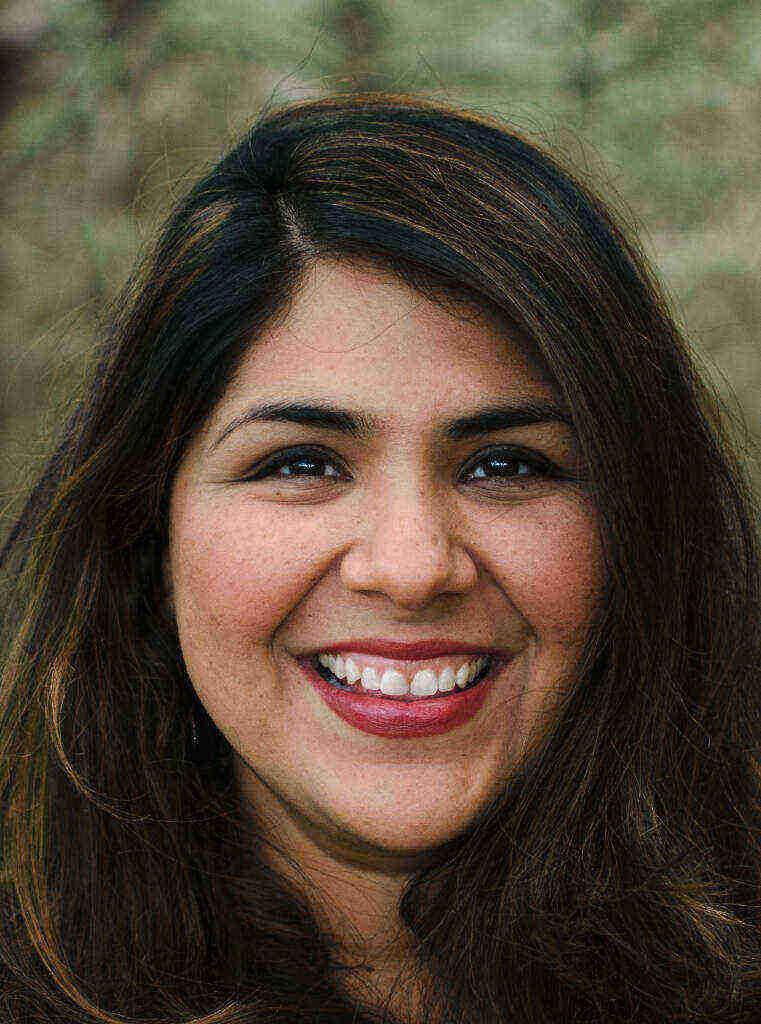}}
\subfloat[]{\includegraphics[height=2.1cm]{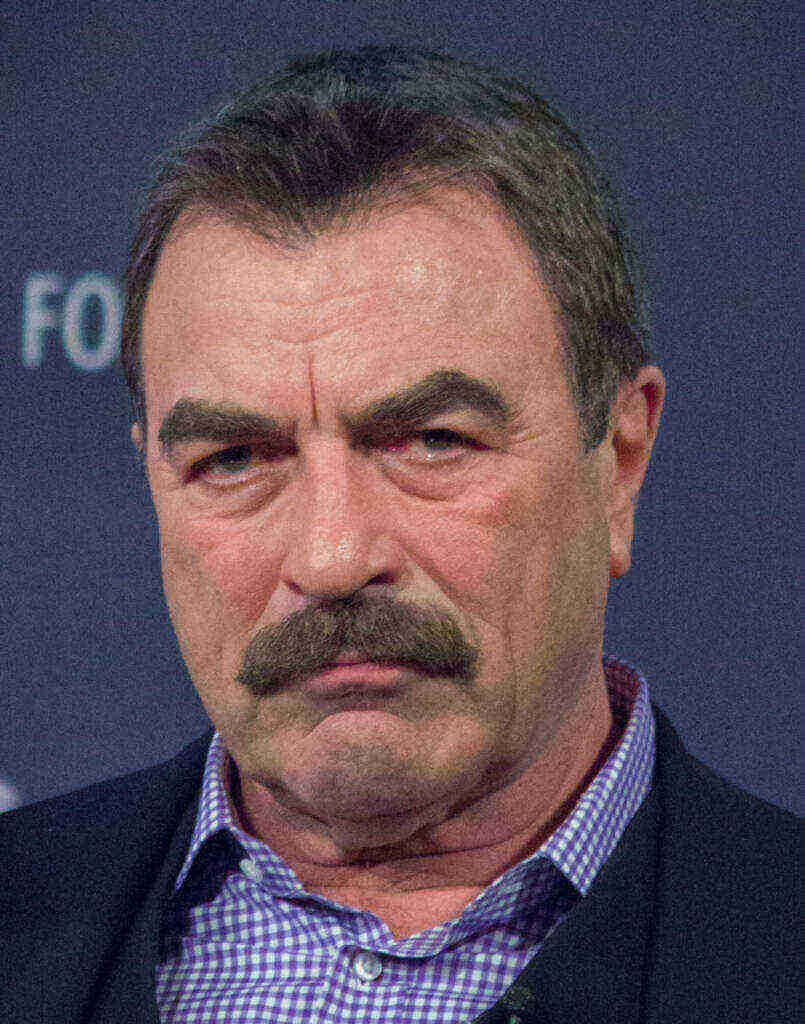}}
\subfloat[]{\includegraphics[height=2.1cm]{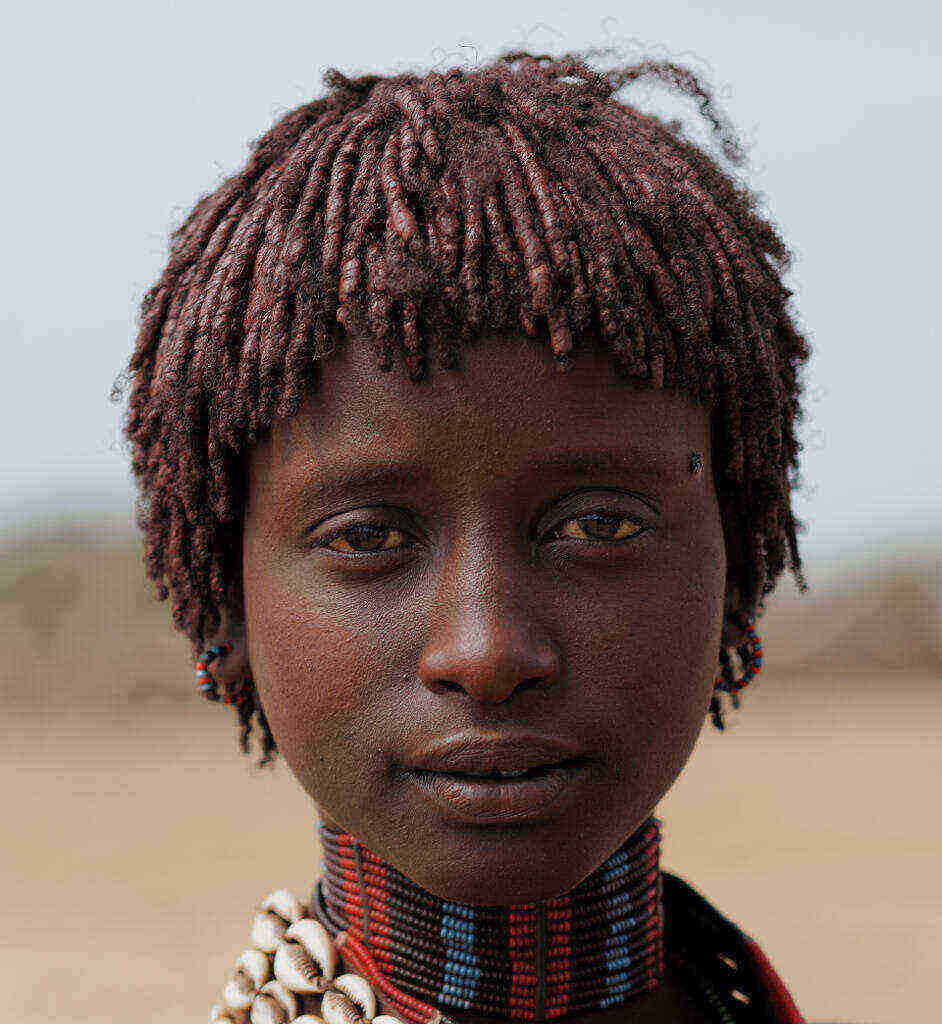}}
\subfloat[]{\includegraphics[height=2.1cm]{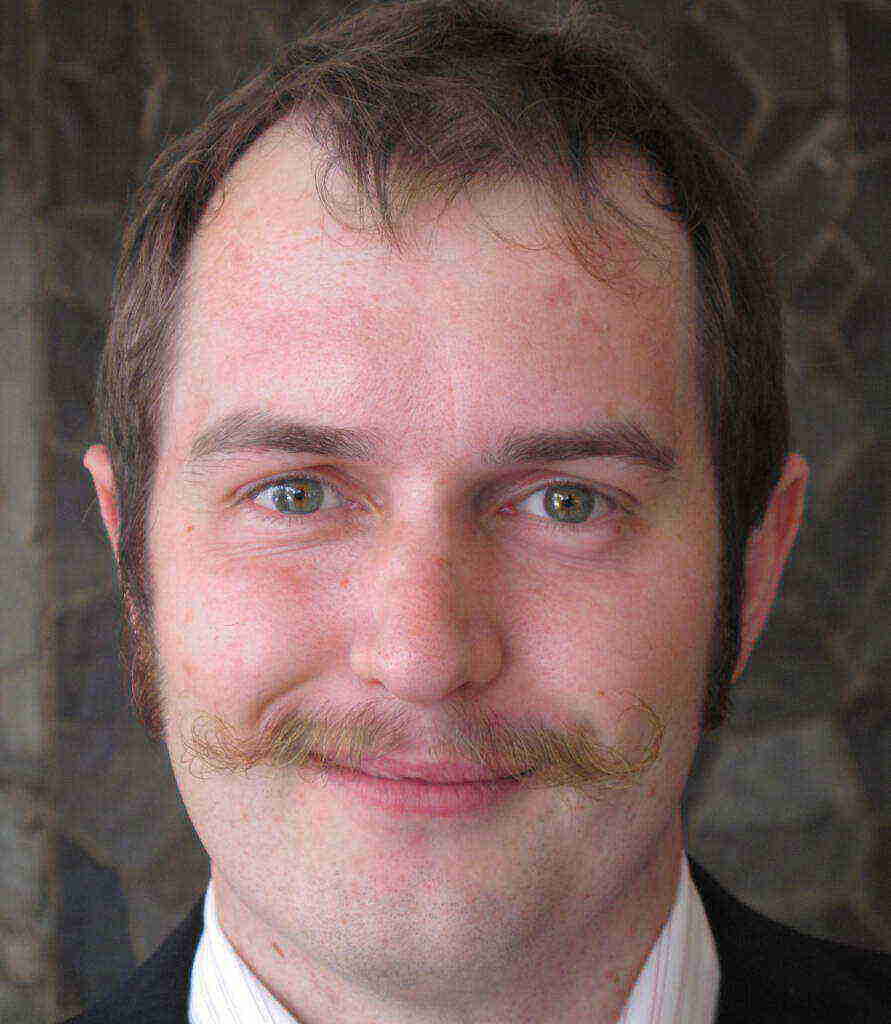}}
\subfloat[]{\includegraphics[height=2.1cm]{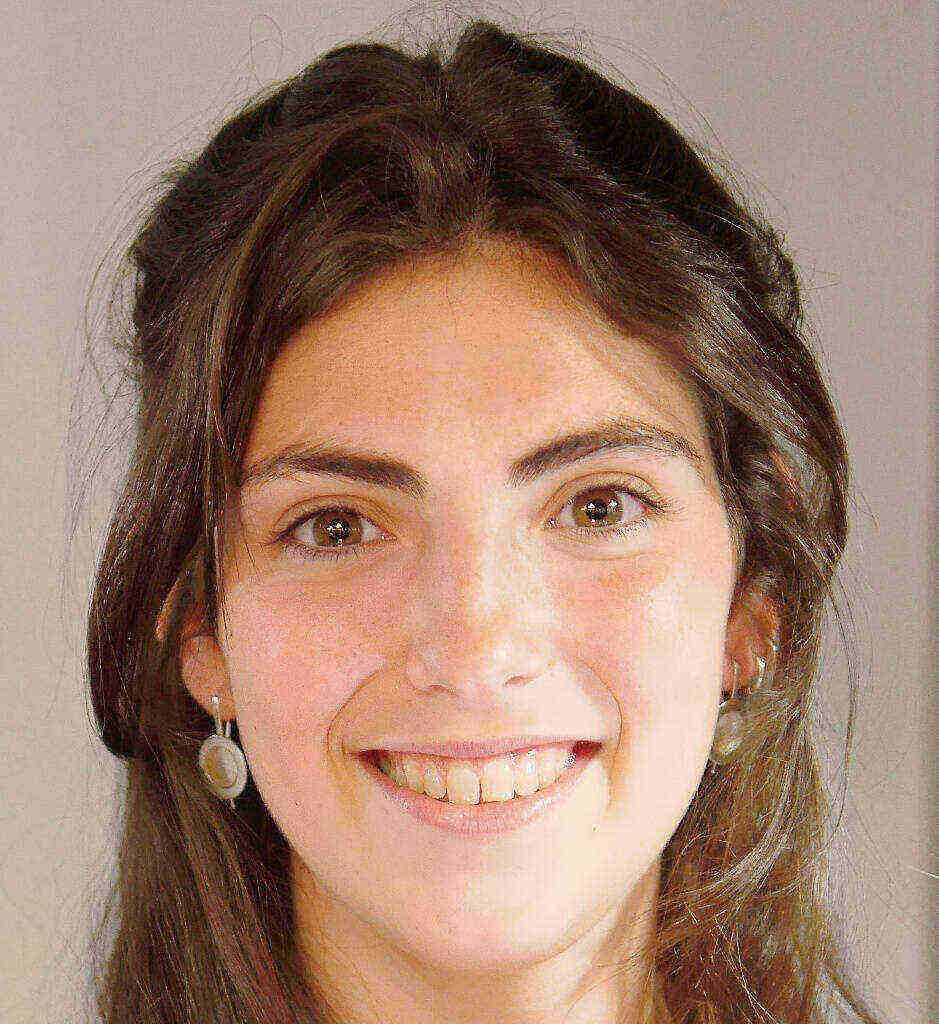}}
\subfloat[]{\includegraphics[height=2.1cm]{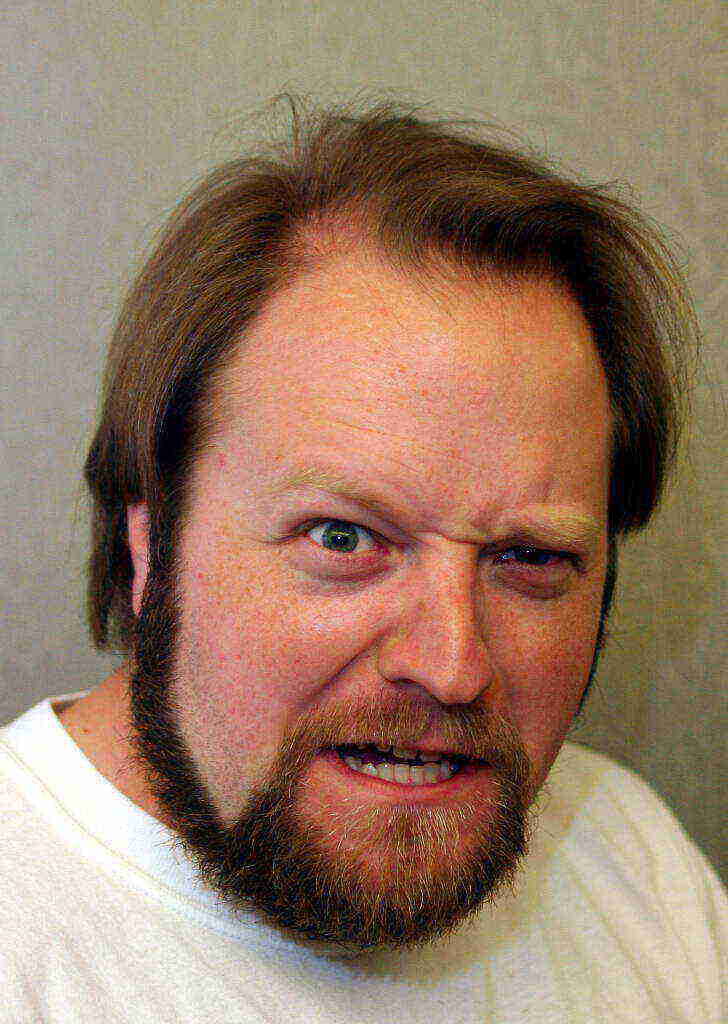}}
\subfloat[]{\includegraphics[height=2.1cm]{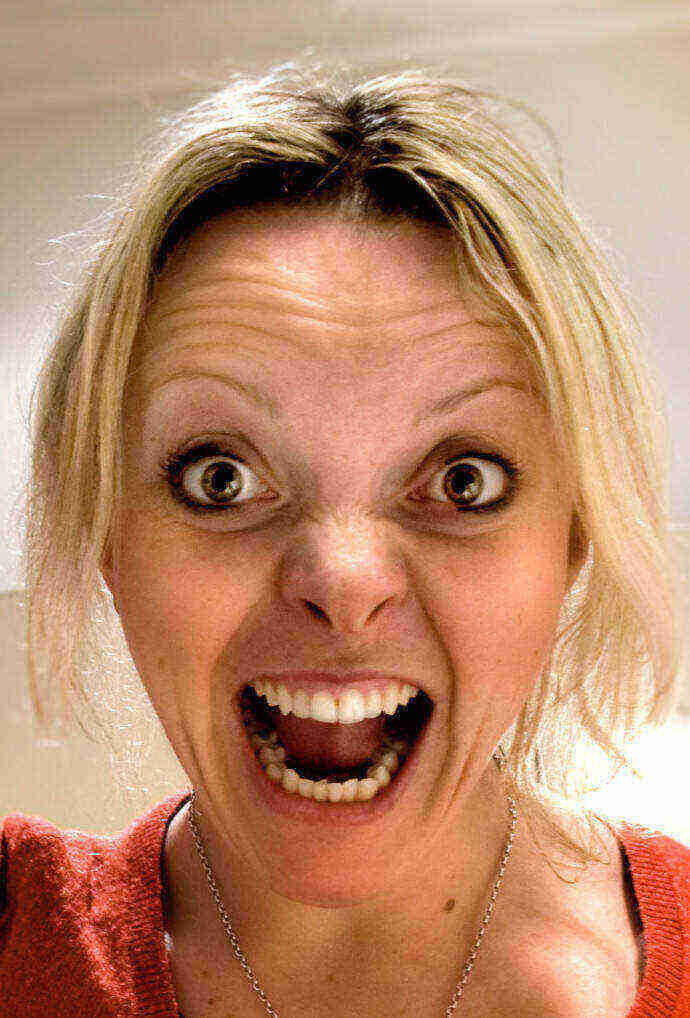}}\\[-1ex]

\subfloat[]{\includegraphics[height=2.1cm]{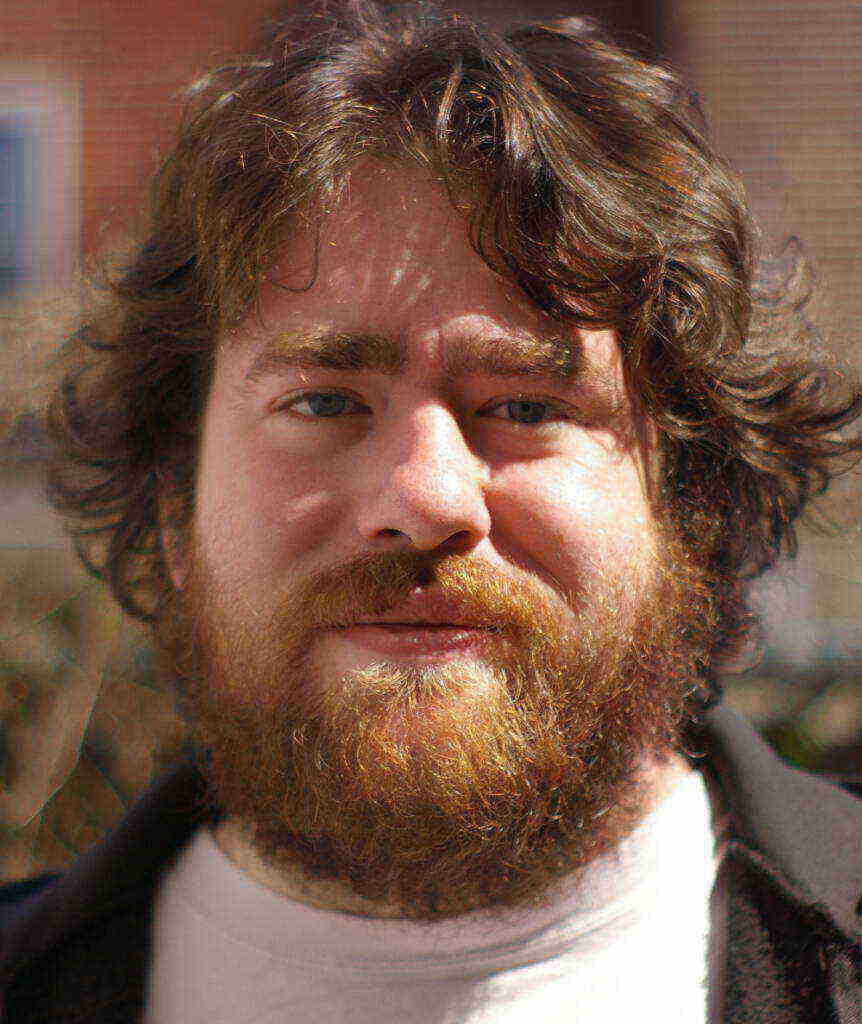}}
\subfloat[]{\includegraphics[height=2.1cm]{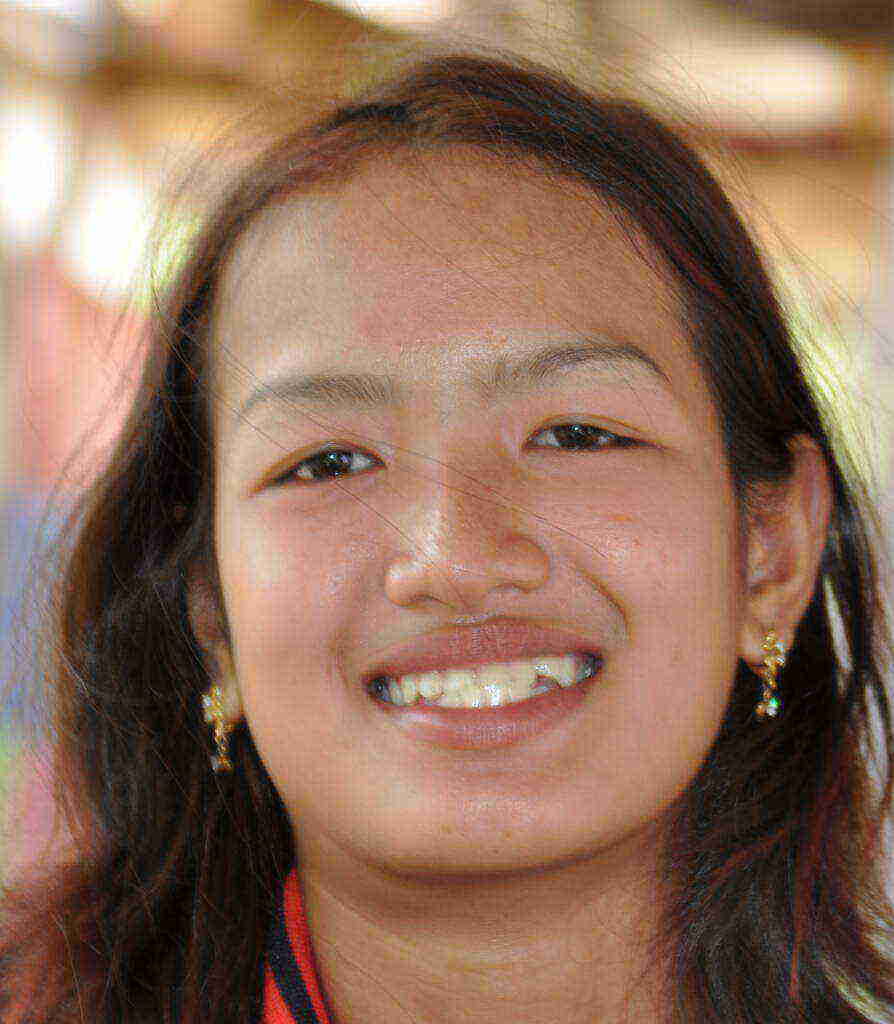}}
\subfloat[]{\includegraphics[height=2.1cm]{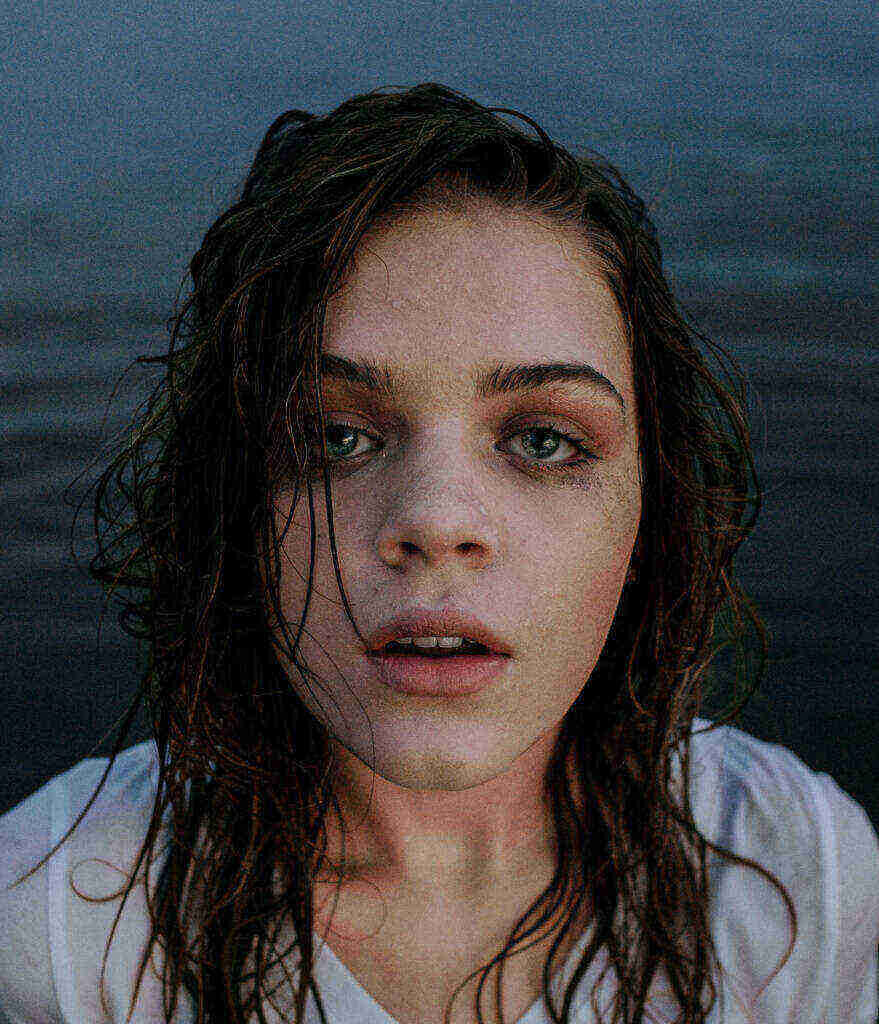}}
\subfloat[]{\includegraphics[height=2.1cm]{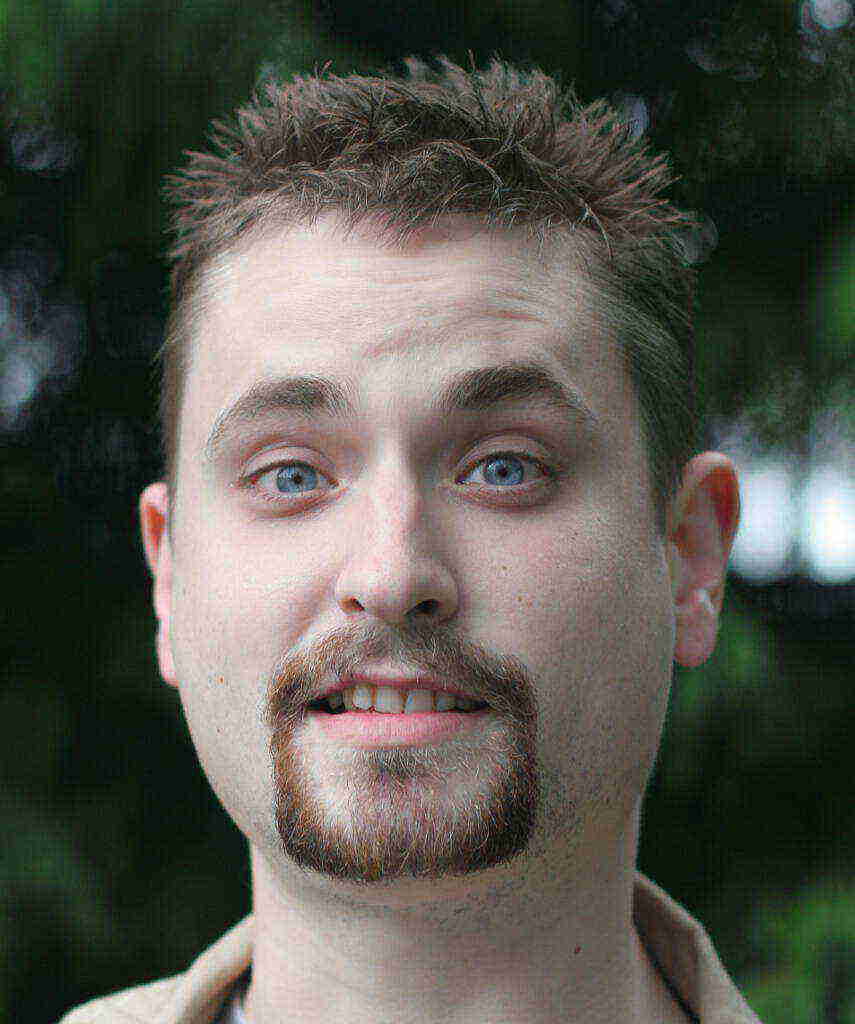}}
\subfloat[]{\includegraphics[height=2.1cm]{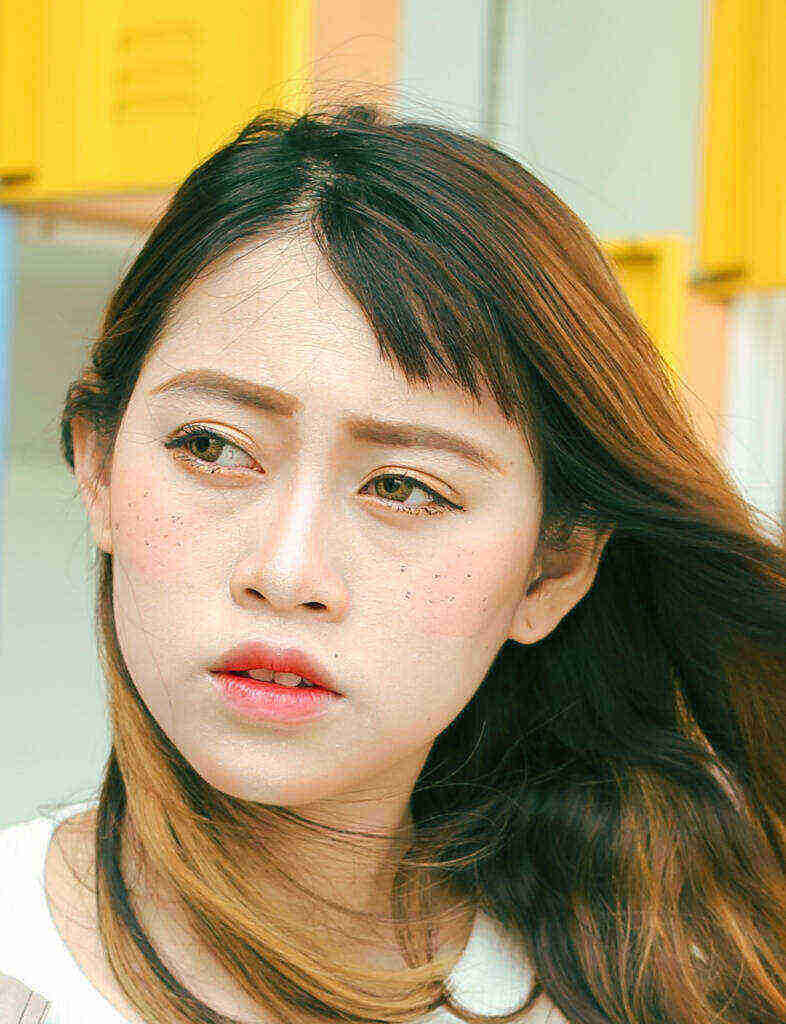}}
\subfloat[]{\includegraphics[height=2.1cm]{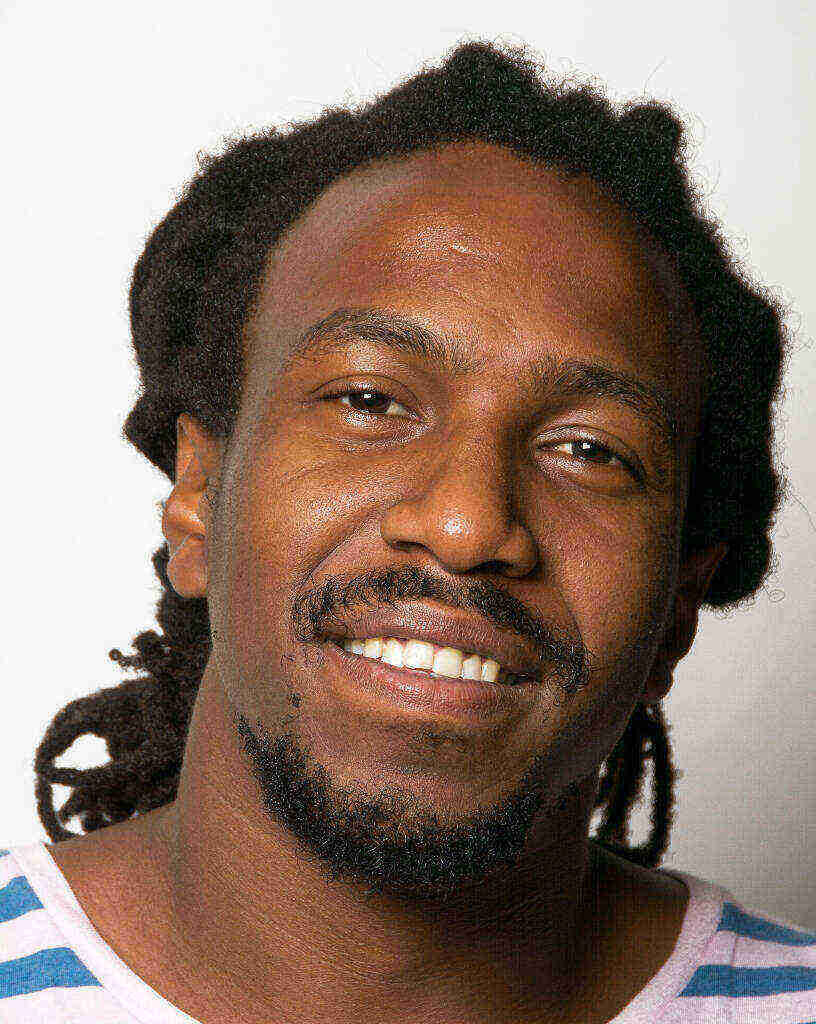}}
\subfloat[]{\includegraphics[height=2.1cm]{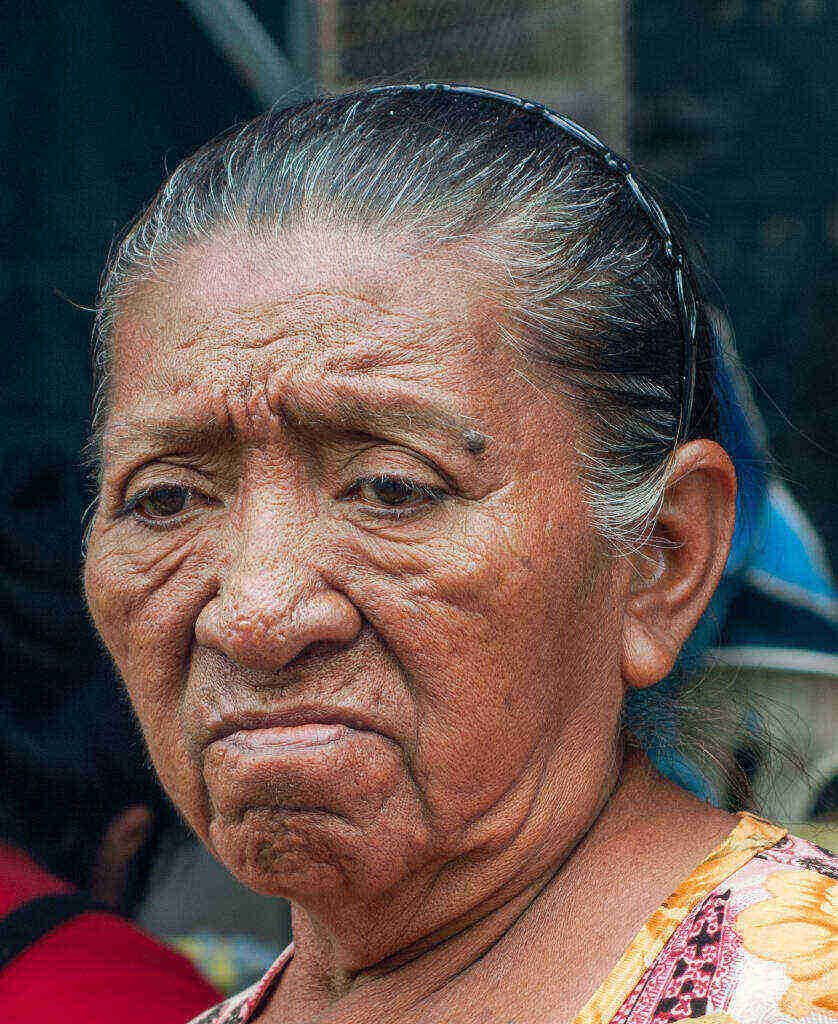}}
\subfloat[]{\includegraphics[height=2.1cm]{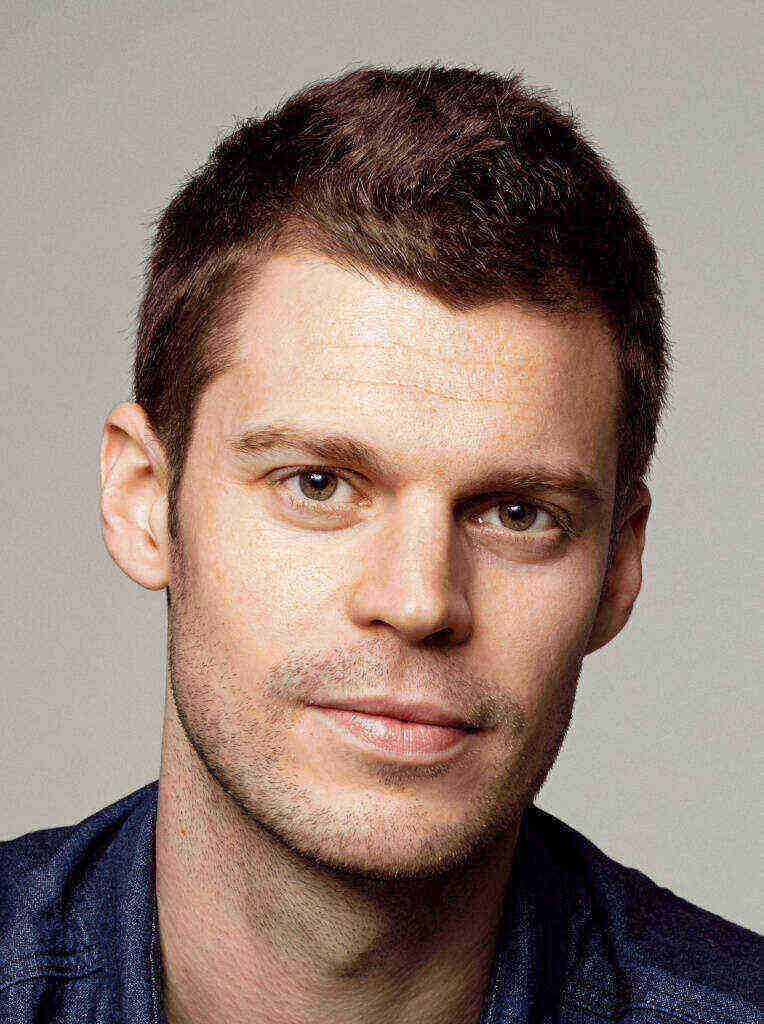}}
\subfloat[]{\includegraphics[height=2.1cm]{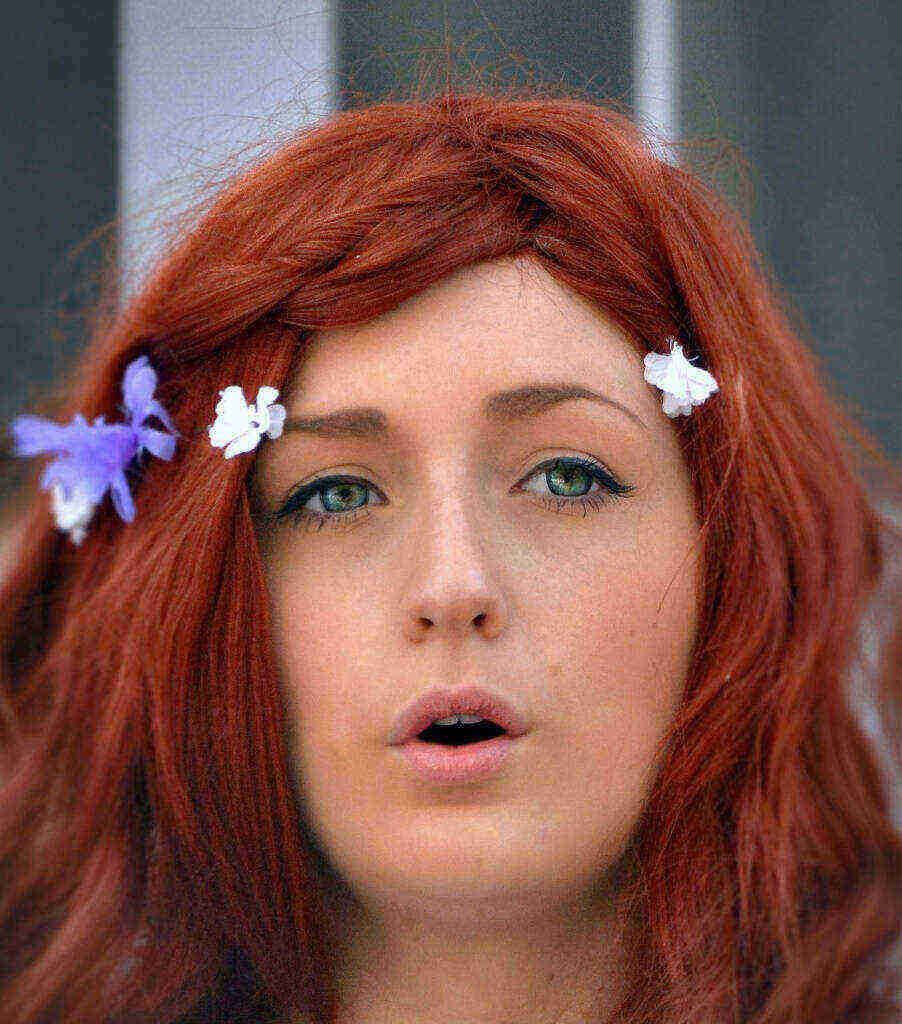}}
\subfloat[]{\includegraphics[height=2.1cm]{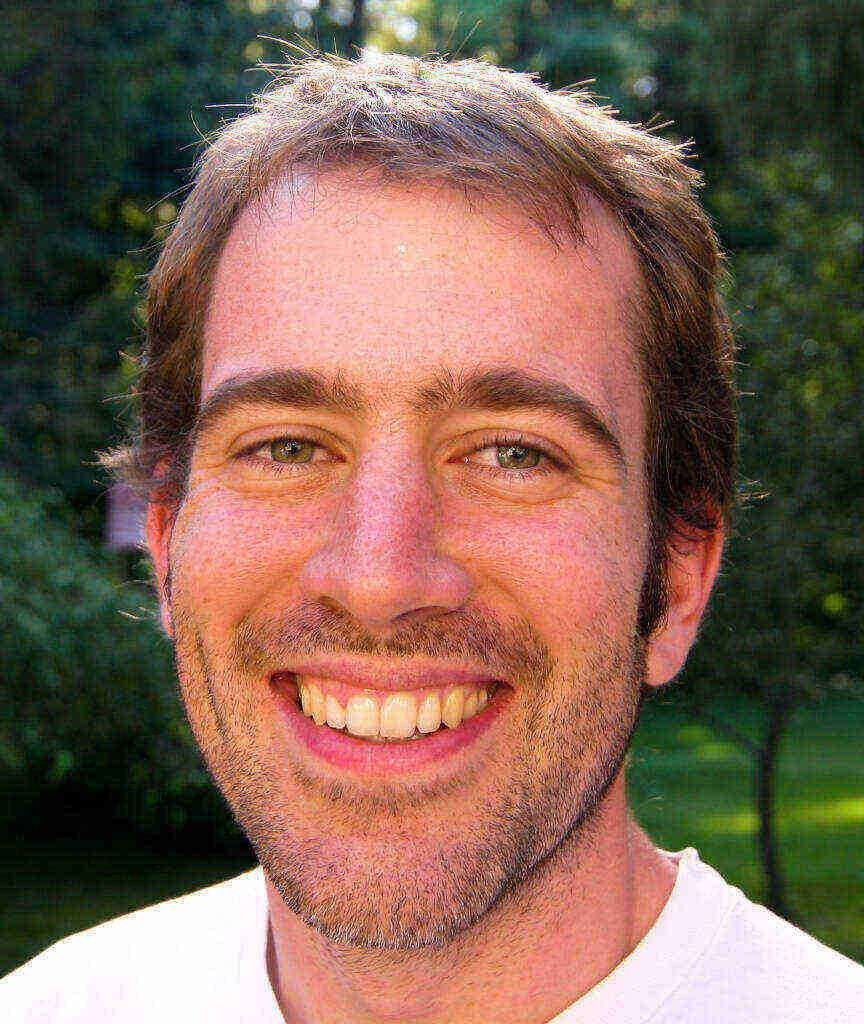}}

\centerline{Level 2}
\medskip
\setcounter{subfigure}{0}

\subfloat[]{\includegraphics[height=1.95cm]{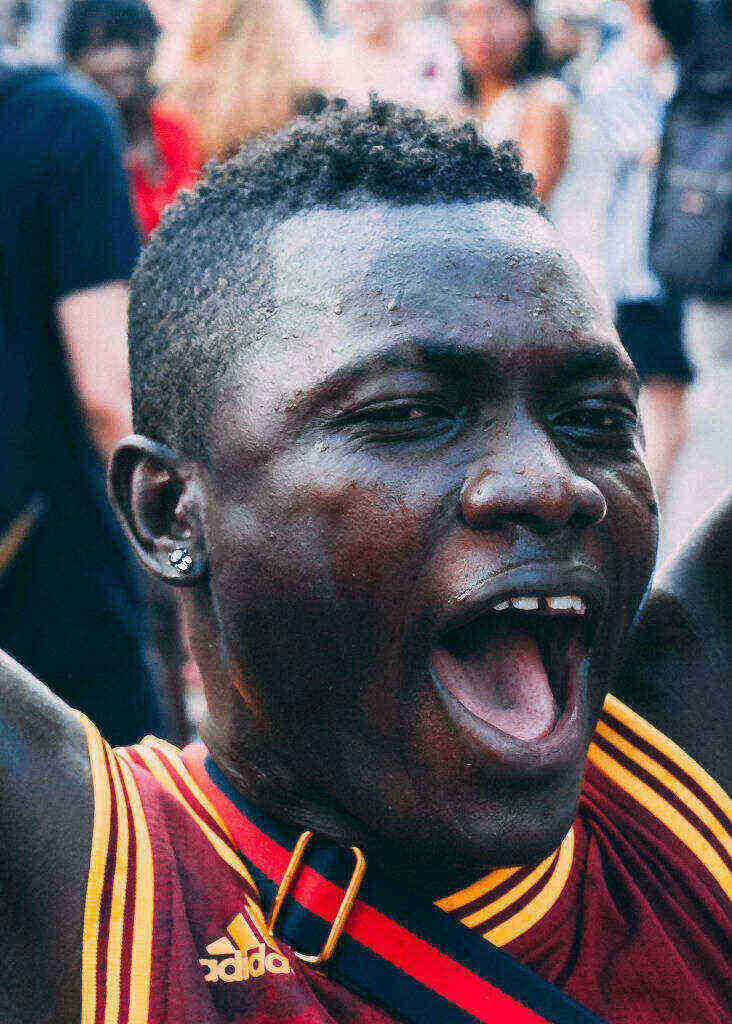}}
\subfloat[]{\includegraphics[height=1.95cm]{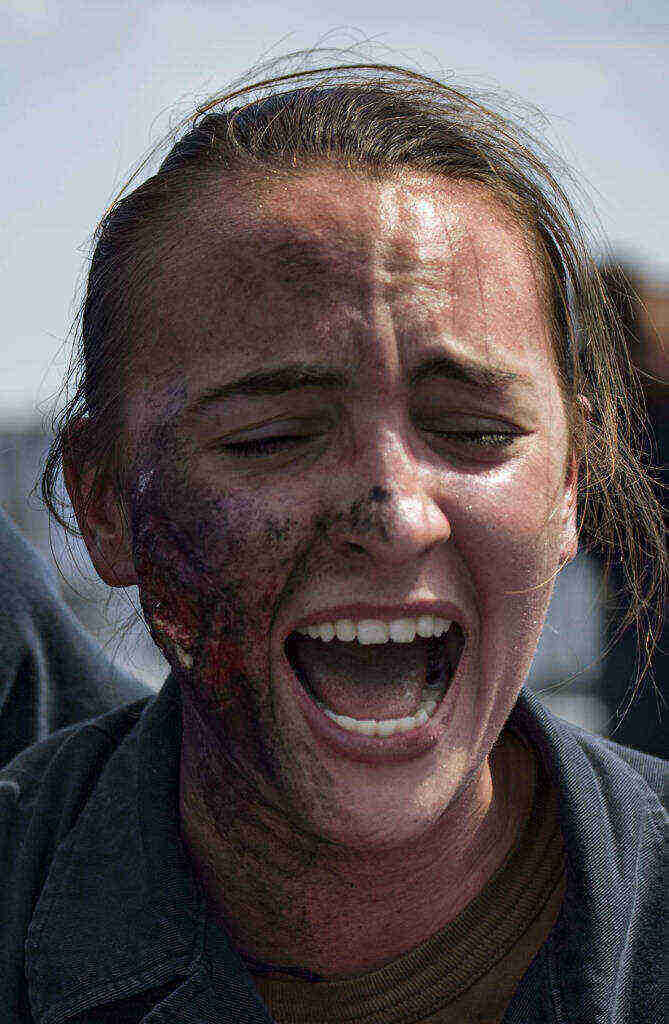}}
\subfloat[]{\includegraphics[height=1.95cm]{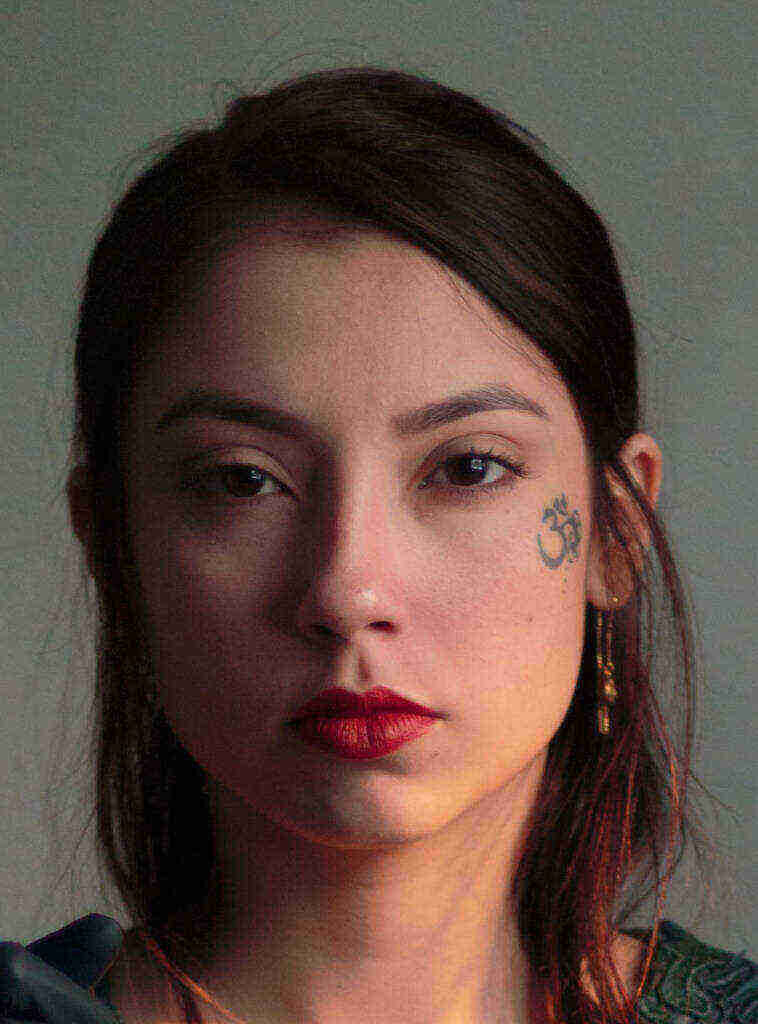}}
\subfloat[]{\includegraphics[height=1.95cm]{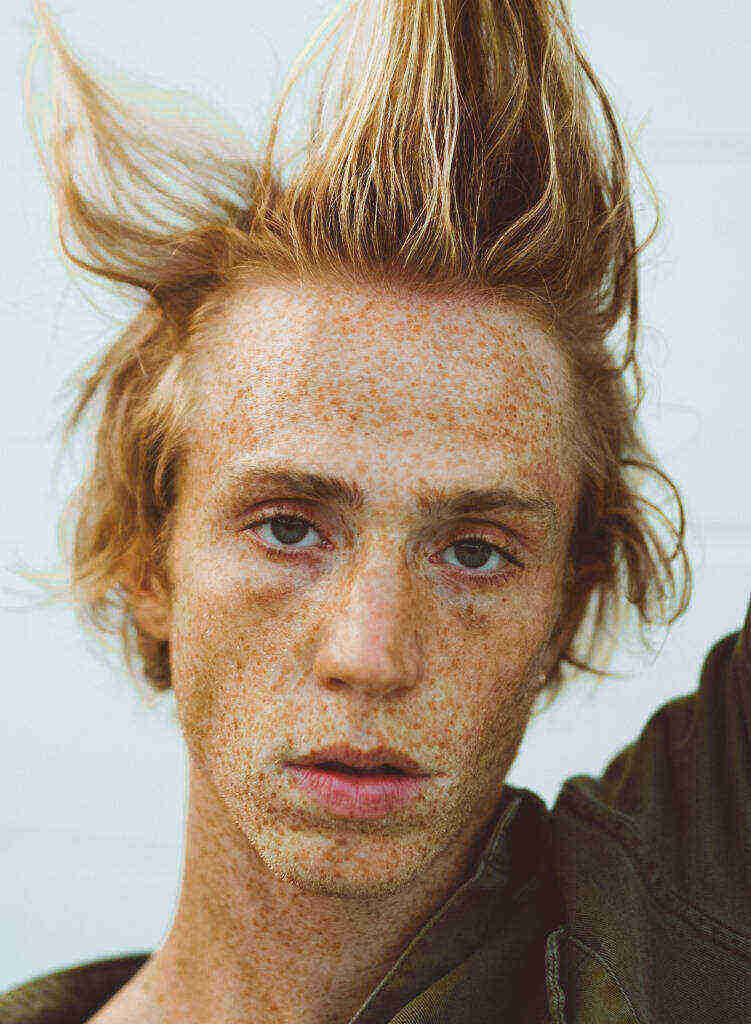}}
\subfloat[]{\includegraphics[height=1.95cm]{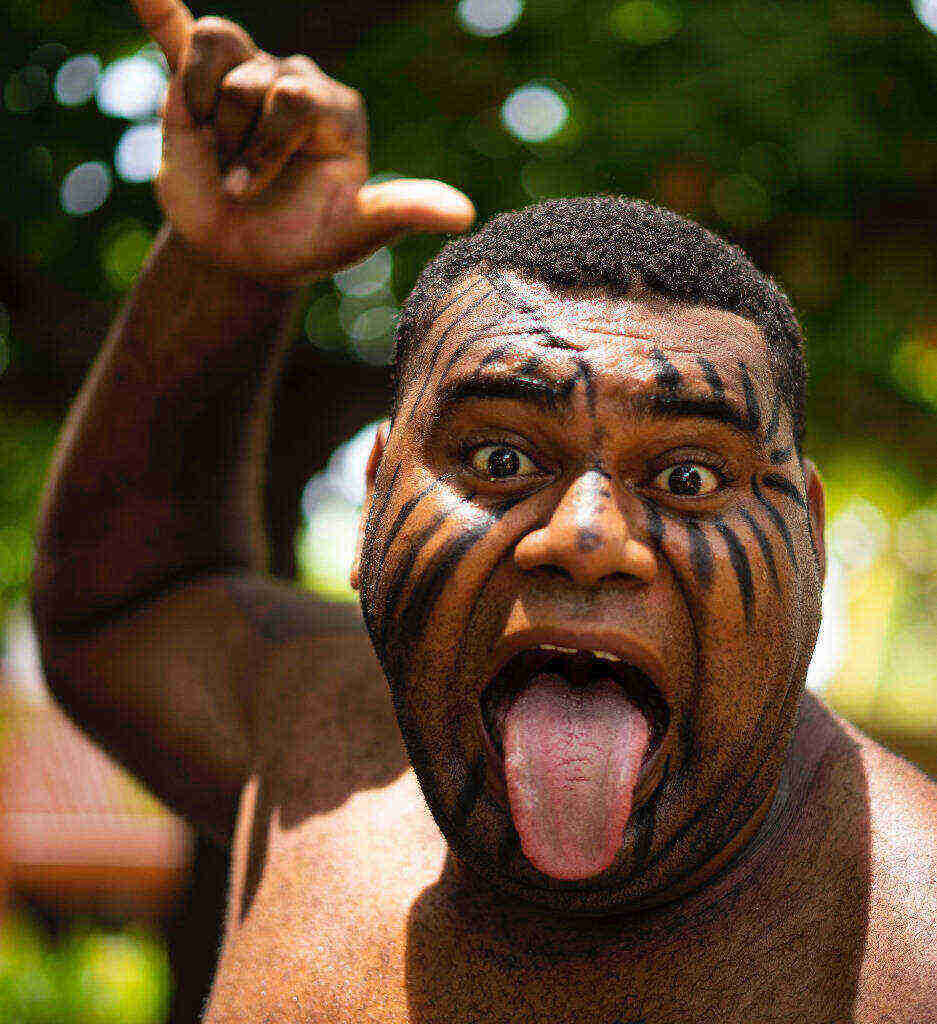}}
\subfloat[]{\includegraphics[height=1.95cm]{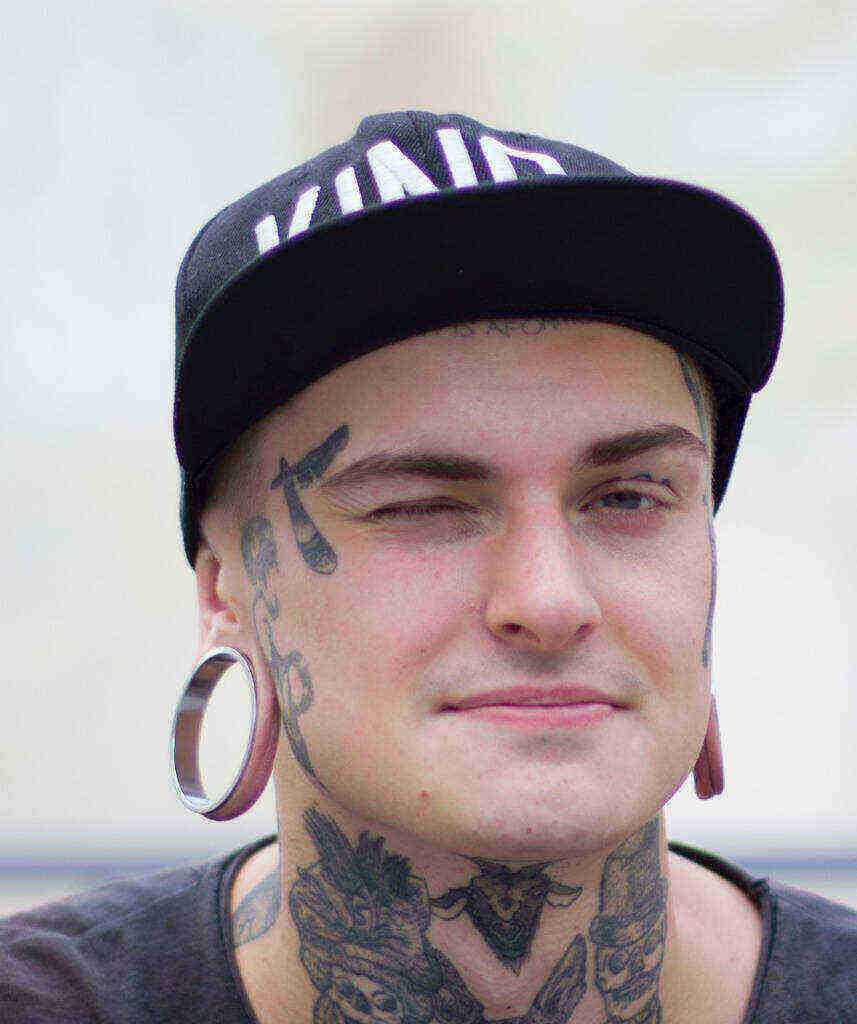}}
\subfloat[]{\includegraphics[height=1.95cm]{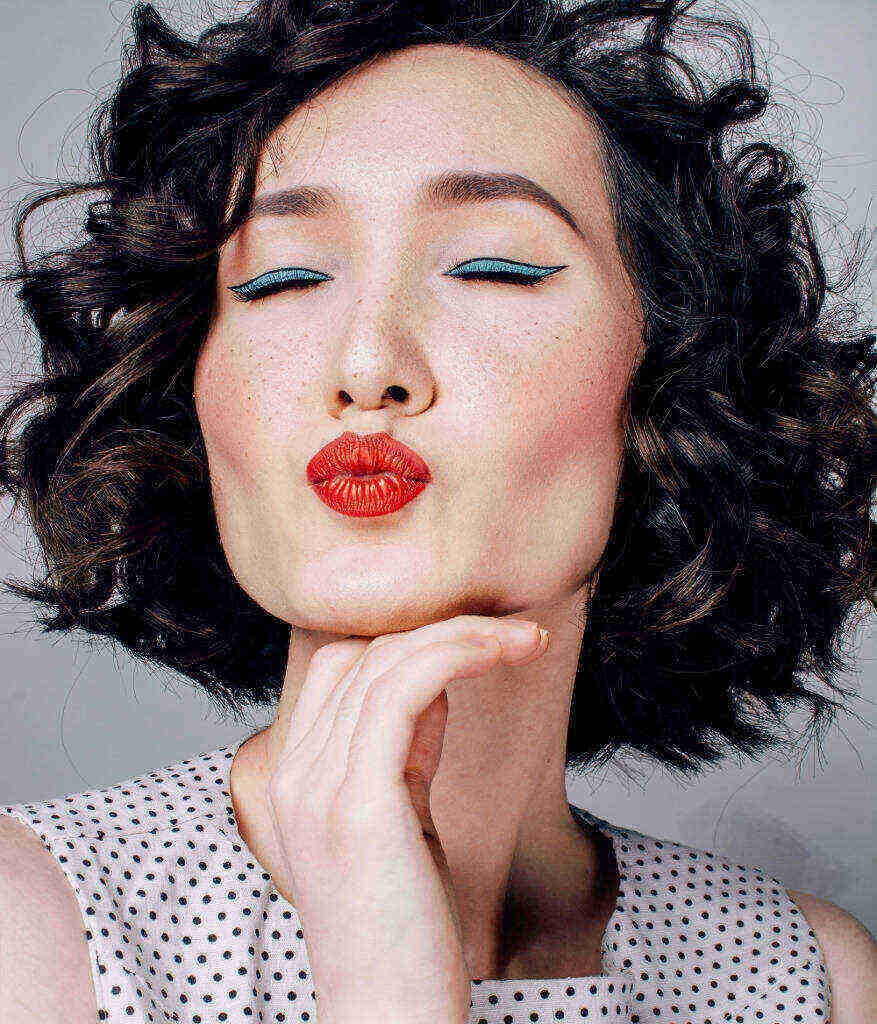}}
\subfloat[]{\includegraphics[height=1.95cm]{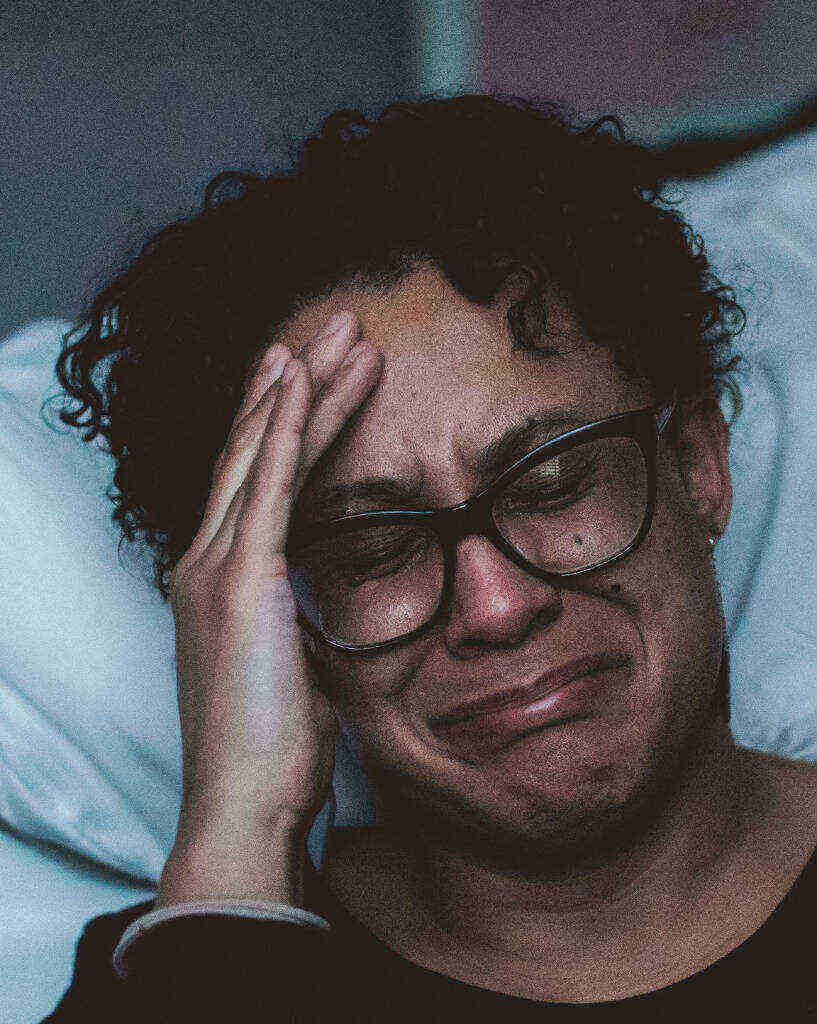}}
\subfloat[]{\includegraphics[height=1.95cm]{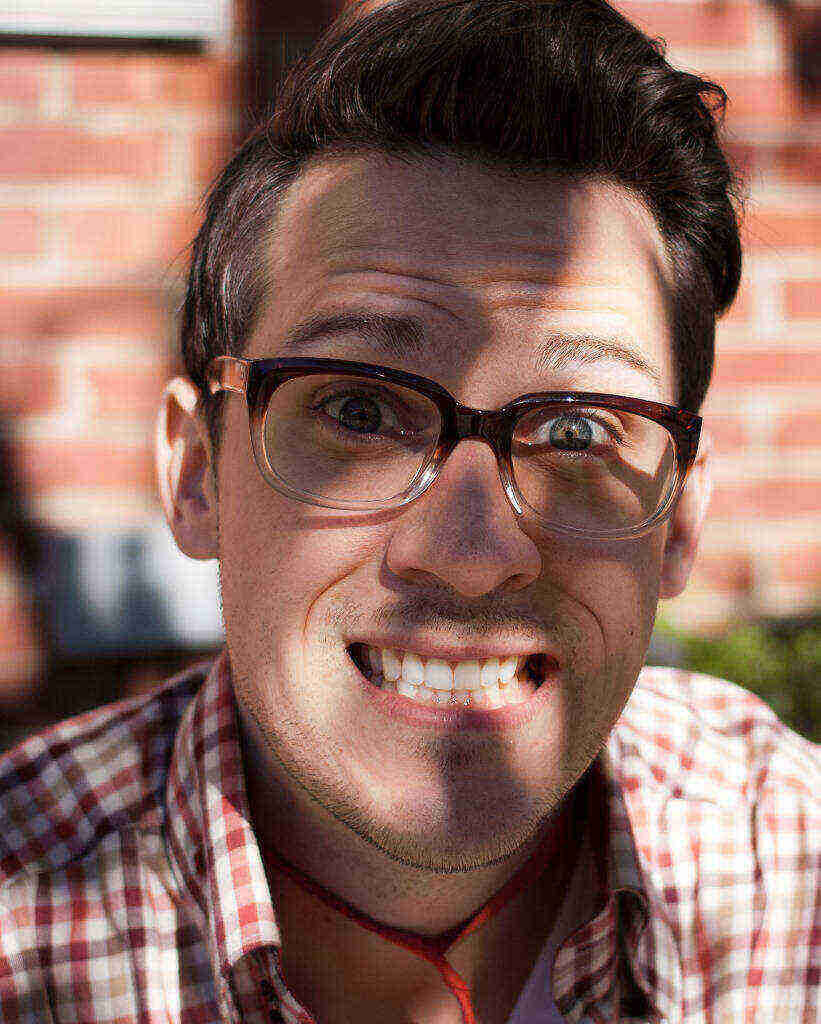}}
\subfloat[]{\includegraphics[height=1.95cm]{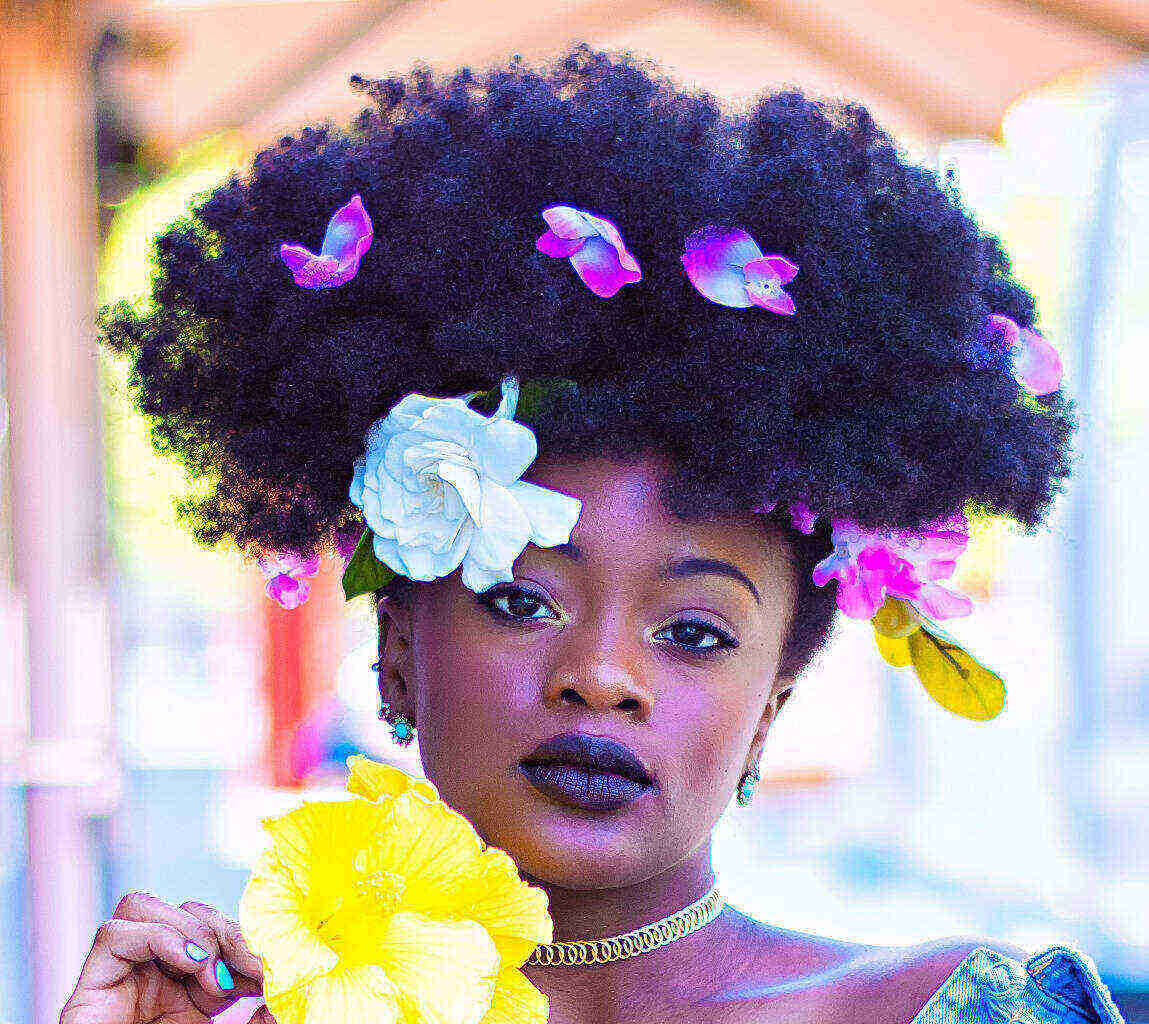}}
\subfloat[]{\includegraphics[height=1.95cm]{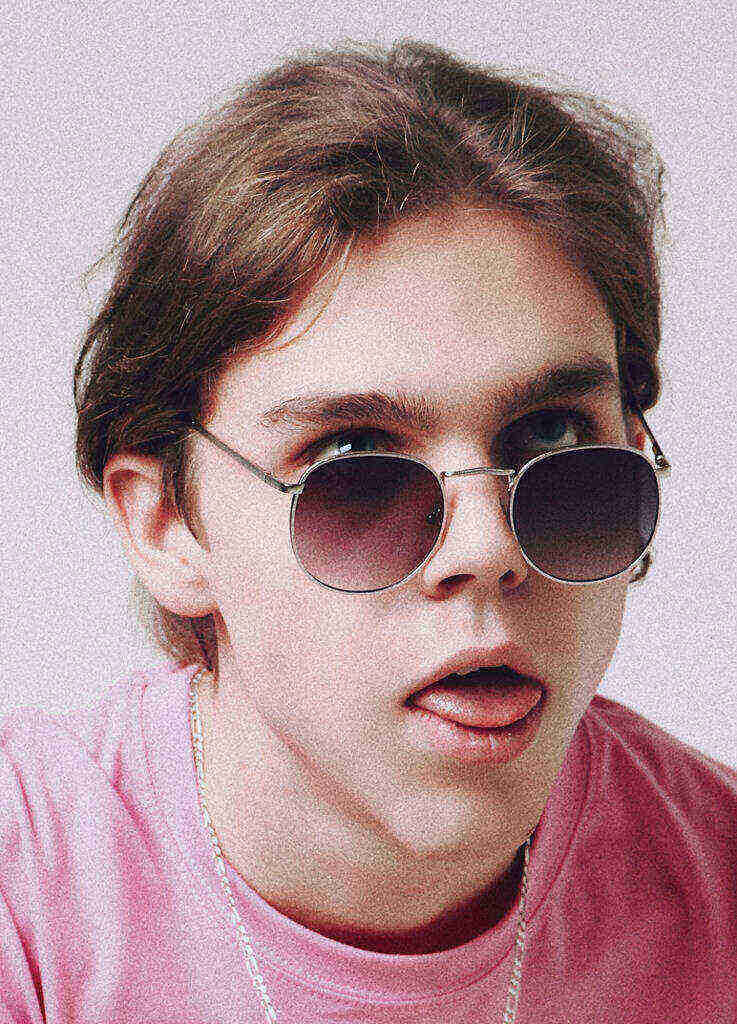}}\\[-1ex]

\subfloat[]{\includegraphics[height=1.95cm]{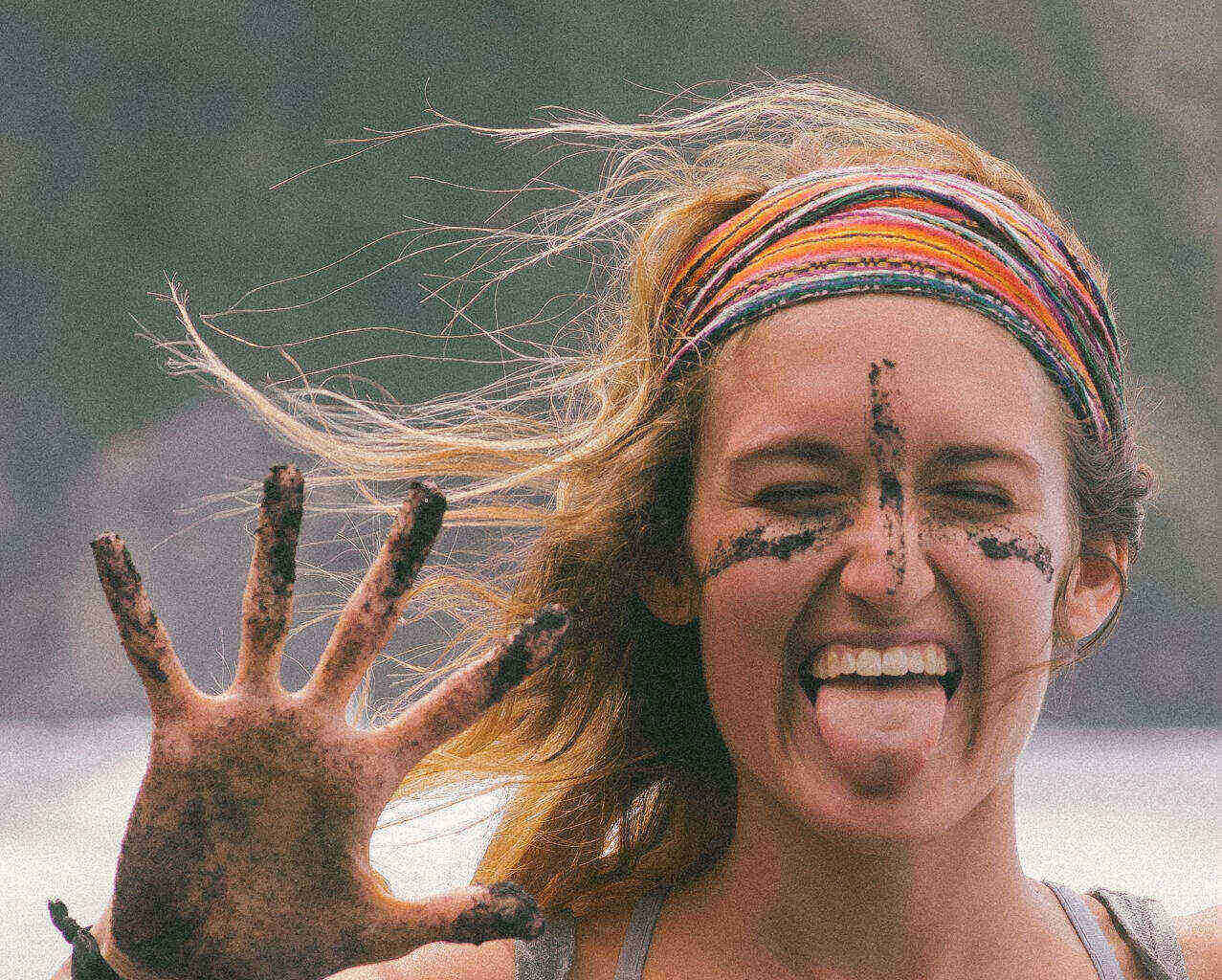}}
\subfloat[]{\includegraphics[height=1.95cm]{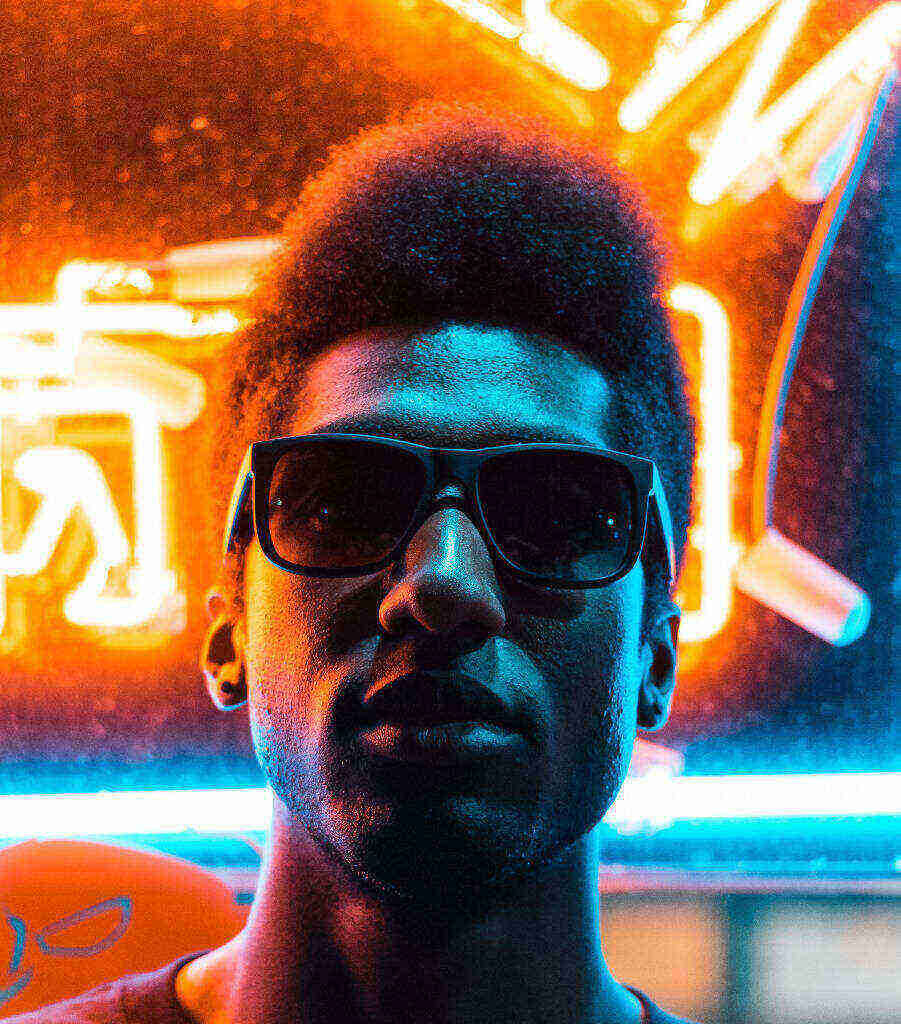}}
\subfloat[]{\includegraphics[height=1.95cm]{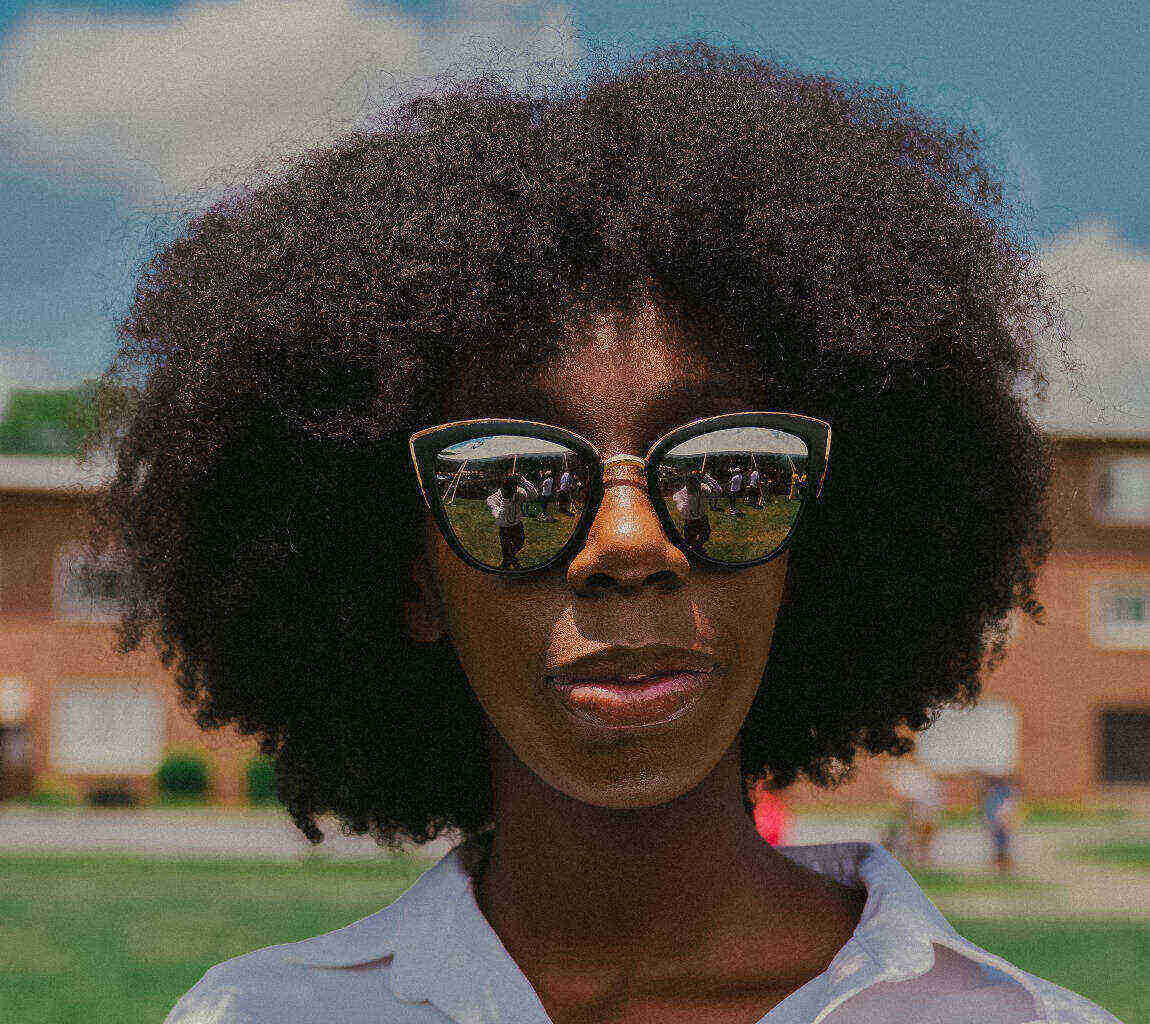}}
\subfloat[]{\includegraphics[height=1.95cm]{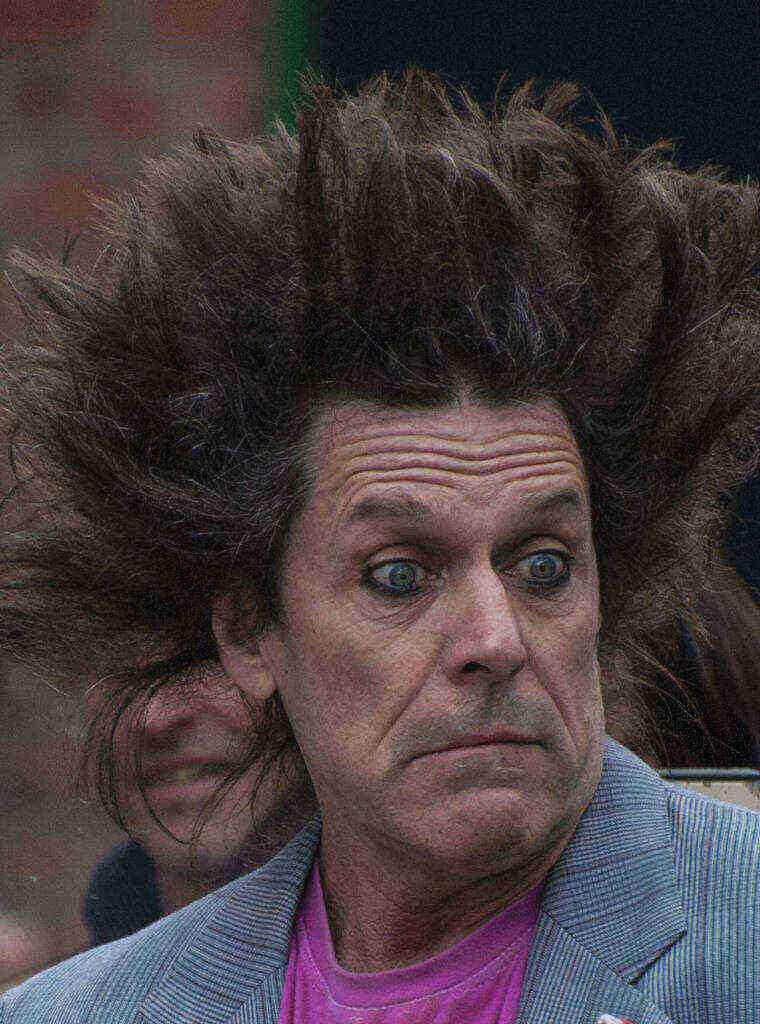}}
\subfloat[]{\includegraphics[height=1.95cm]{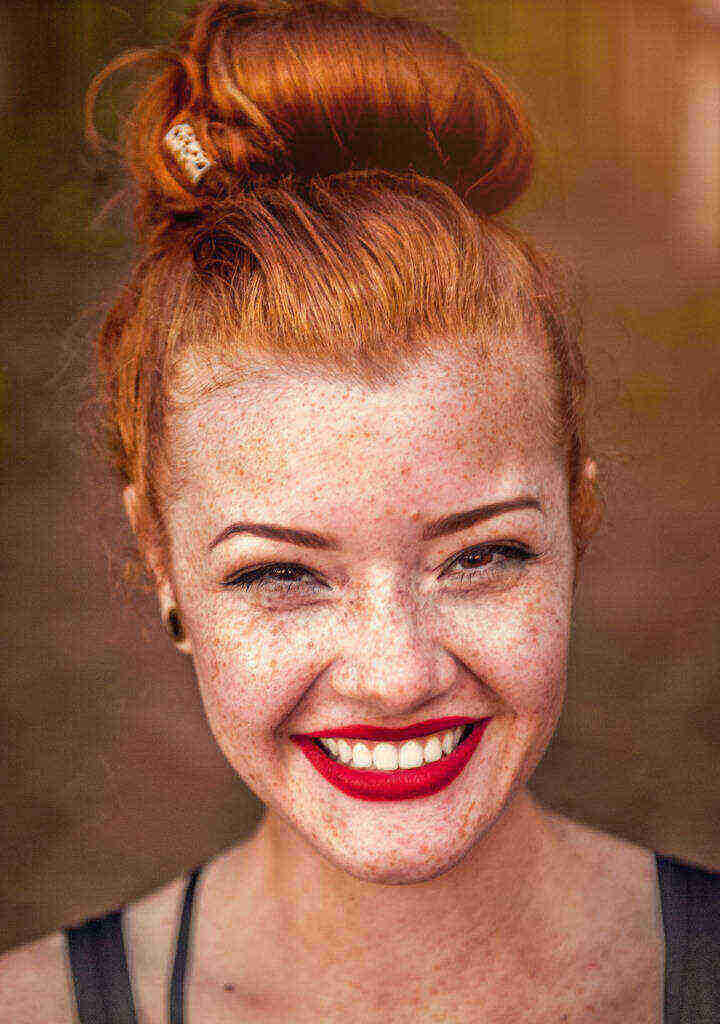}}
\subfloat[]{\includegraphics[height=1.95cm]{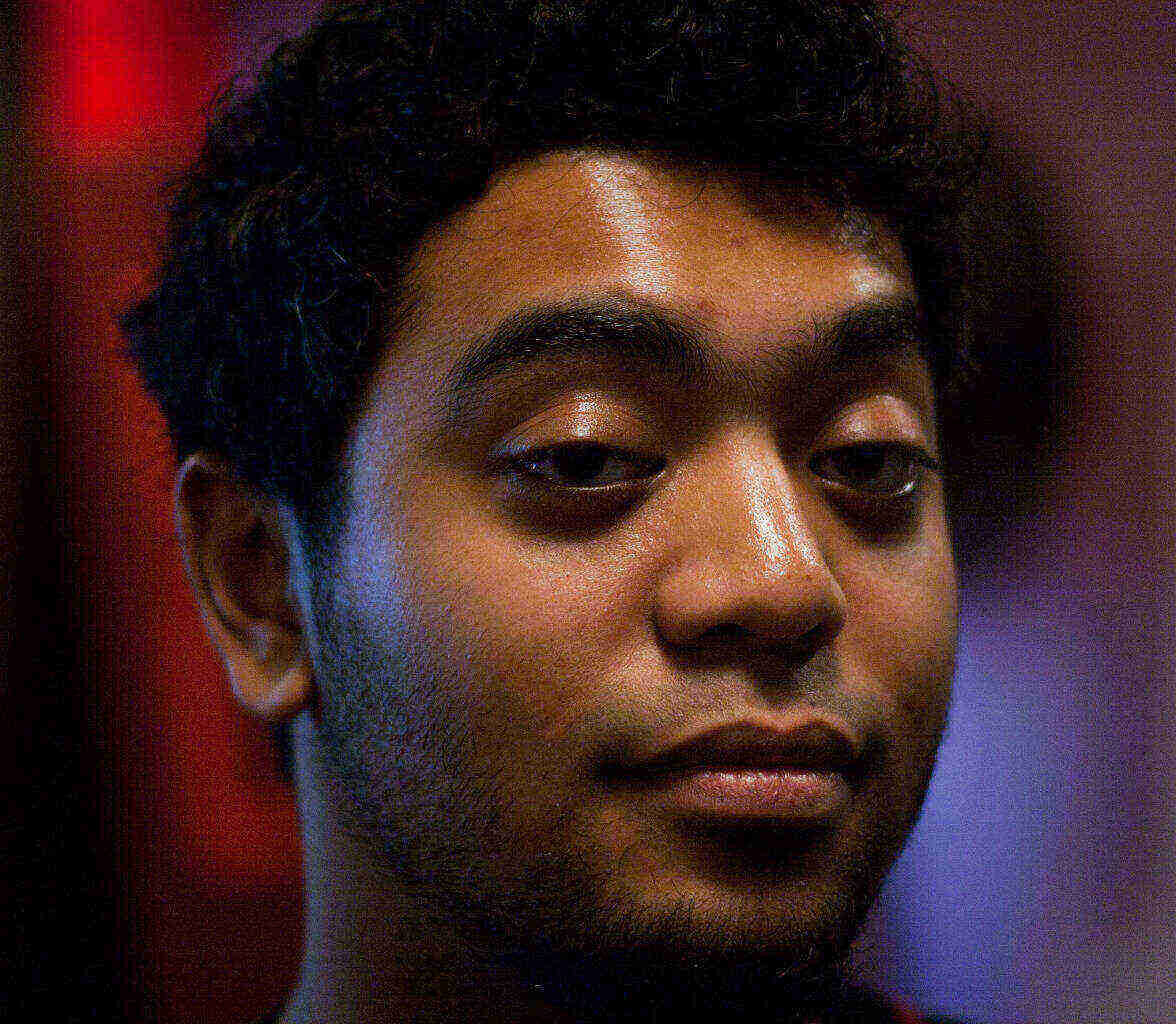}}
\subfloat[]{\includegraphics[height=1.95cm]{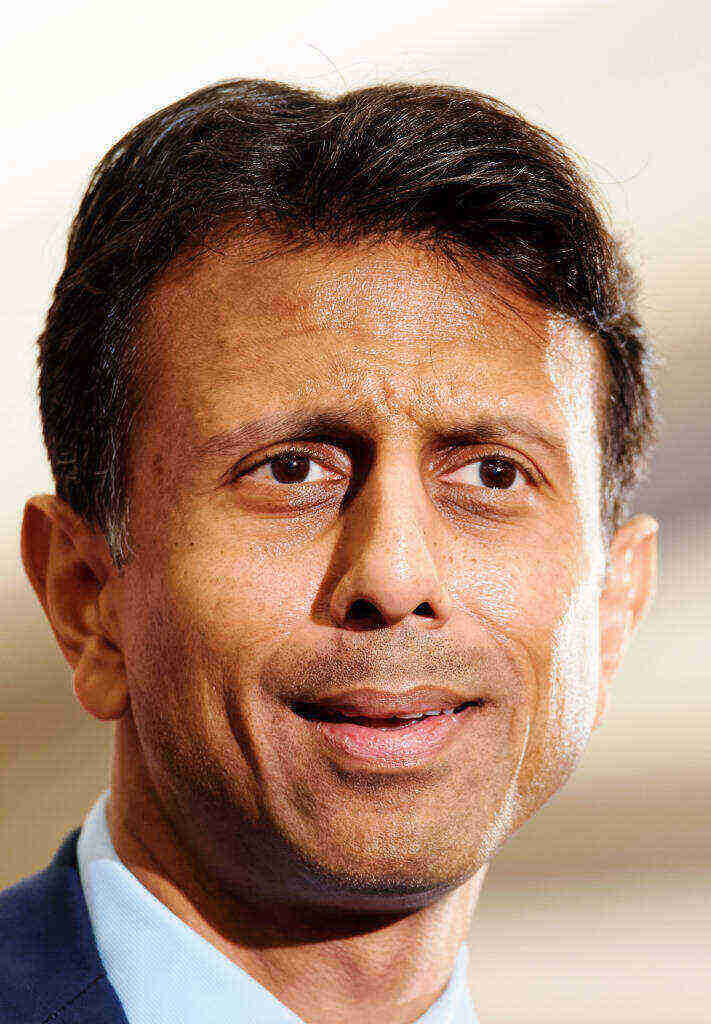}}
\subfloat[]{\includegraphics[height=1.95cm]{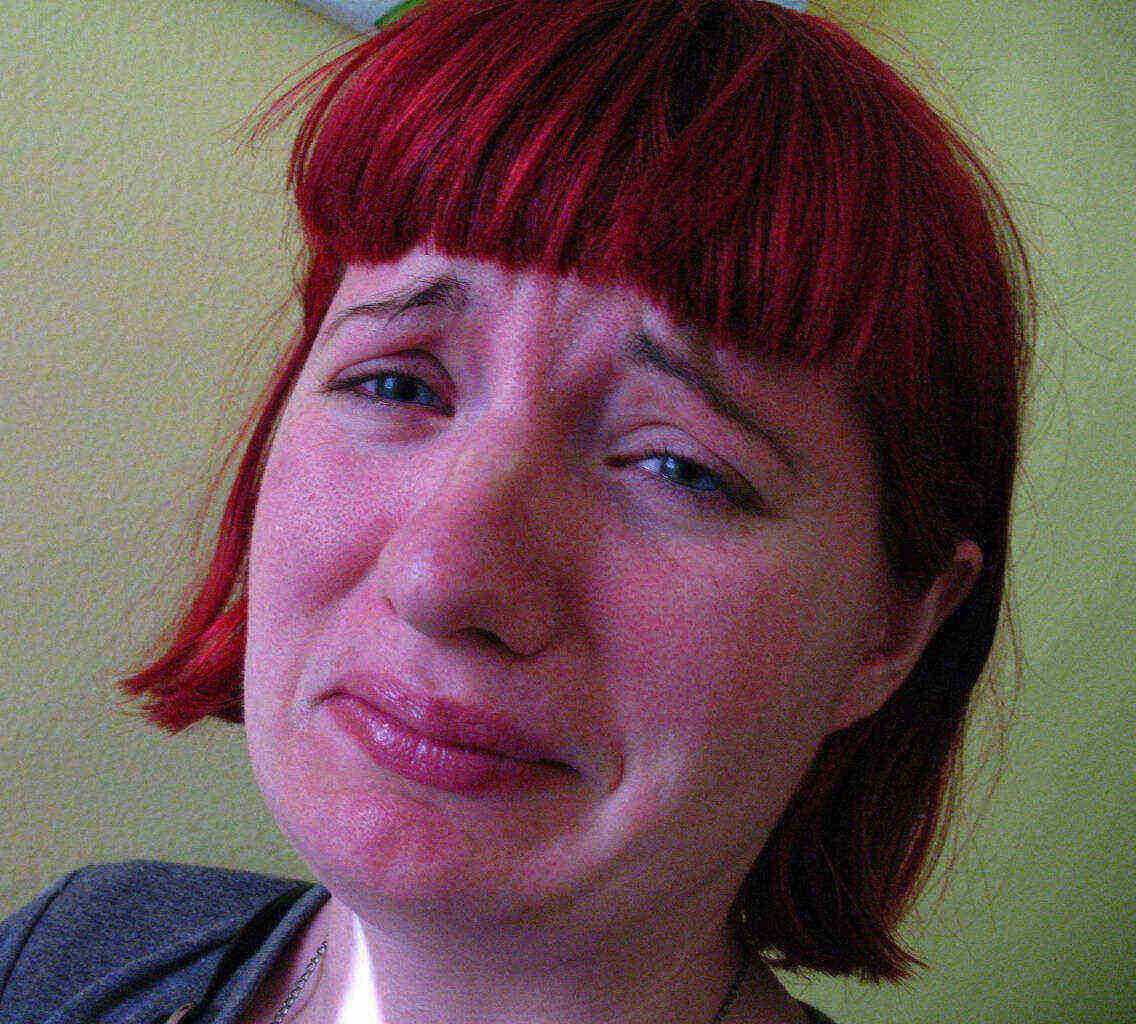}}
\subfloat[]{\includegraphics[height=1.95cm]{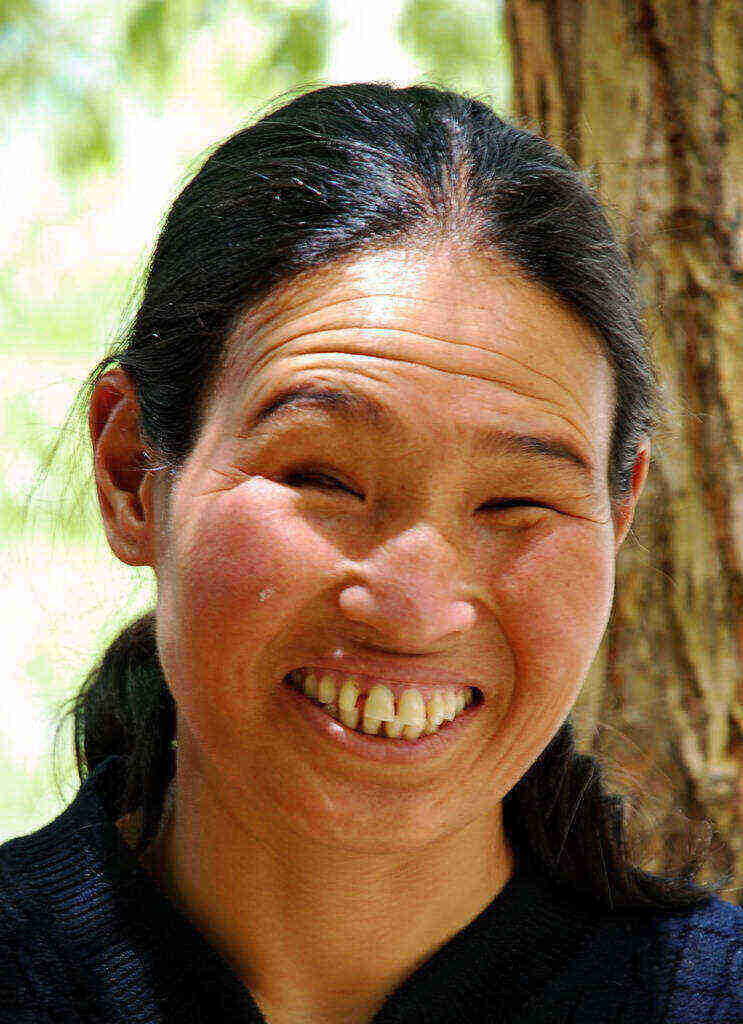}}

\centerline{Level 3}
\medskip
\caption{Images comprising levels 1, 2 and 3 of the \emph{NPRportrait1.0} benchmark.}
\label{benchmark}
\end{figure*}

The full set of 60 images selected for the three levels of the \emph{NPRportrait1.0} benchmark are shown in figure~\ref{benchmark}.
A further user study in which 56 participants were shown all 60 images was carried out to check the four characteristics of
gender, age, attractiveness, and ethnicity.
Not only did this confirm that the image labels were assigned correctly,
but it gave us user responses be used later in experimental evaluation of NPR stylisations.

\subsection{Evaluation of Stylisations}
\label{experiments}

Our benchmark allows researchers to use carefully chosen images to
test out their NPR algorithms, but as discussed in section~\ref{IQA},
carrying out the next step of evaluation is not straightforward, especially if it is to be quantitative.
In the context of an application, a stylisation may have some precise goal
(e.g. mimicking an existing artist, or enabling the viewer to identify the rendered object quickly),
which allows for a task-performance metric.
However, in this paper we do not assume that such a goal is known (or even exists).
To avoid the difficulty of
directly comparing outputs of one algorithm against another algorithm,
we formulate several experiments which are either based on the
aesthetics from single stylisation algorithms, or else operate
indirectly on the aesthetic aspects, using the four facial
characteristics with which the benchmark dataset is annotated: gender,
age, attractiveness, and ethnicity.  The rationale for the latter
approach is that it is better to ask users to make decisions about
such characteristics rather than asking them to score the quality of a
stylisation.  Asking about stylisation quality involves making
aesthetic judgements; not only is this difficult for users and
subjective, but the task is often ill-defined
given the multiple and interacting factors of
content, style, and level of abstraction.
In contrast, the four facial characteristics we use are extremely familiar to all study participants.

\textbf{Experiment 1: Correctness of facial characteristics.}
NPR algorithms are evaluated by measuring the differences between
estimates of four facial characteristics (gender, age, attractiveness,
and ethnicity) which were captured from the user studies.  The
estimates from the source images are taken as a good approximation of
ground truth, and it is expected that good stylisations can preserve
these characteristics, although this may not hold for highly
abstracted styles.  Since the responses in the user studies are not
totally consistent, a distribution is captured for each question.
Therefore, when comparing image characteristics, the Earth Mover's
Distance (EMD) is applied to the ordinal scales (gender, age,
attractiveness) and $L_1$ distance is applied to ethnicity.  In
addition to the traditional unsigned EMD, it is interesting to consider
a signed version, which can be simply done by modifying Cha and
Srihari's~\cite{EMD} Algorithm~1 to accumulate the signed prefix sum
rather than the absolute prefix sum.

\textbf{Experiment 2: Quality of stylisation across levels.}
This experiment checks the robustness of an NPR algorithm by directly
looking at its stylisations across the three benchmark levels.  One
possibility would be to perform a user study involving a grouping task
on the stylised photographs, but our user studies were carried out
remotely, and 60 images is too many to view simultaneously on a
screen.  As an alternative, we ask users to view a triple of stylised
images (all from the same NPR algorithm) and rank them according to
the quality of the stylisations.  The triples are generated randomly,
and contains one image from level~1, another from level~2, and the
last one from level~3 (although the users are not aware of the three
benchmark levels).  The correlation between the set of user rankings
and the benchmark levels is then computed.  Restricting the elements
of the triples such that they are drawn from different levels implies
that their stylisations should be more distinct, and this has a double
benefit.  First, by avoiding a fine-grained task it makes the user's
task in the study easier, as trying to choose between similar quality
stylisations is difficult and frustrating.  Second, it makes the user
study more efficient as the user can answer the questions more quickly
and more reliably.  User responses to similar quality stylisations are
likely to be random, and so such triplets provide little useful
information.

\section{Experiments}
\label{experimentSection}

\begin{table*}[htbp]
\centering
\caption{Variability of user judgements of face characteristics from source images in the \emph{NPRportrait1.0} benchmark;
standard deviations for gender, age and attractiveness, and the index of dispersion for ethnicity.}
\label{SDdispersion}
\begin{tabular}{|c||c|c|c||c|c|c||c|c|c||c|c|c|} \hline
characteristic & \multicolumn{3}{c|}{gender} & \multicolumn{3}{c||}{age} & \multicolumn{3}{c||}{attractiveness} & \multicolumn{3}{c|}{ethnicity} \\ \hline
level & 1 & 2 & 3 & 1 & 2 & 3 & 1 & 2 & 3 & 1 & 2 & 3 \\ \hline \hline
& 0.070 & 0.069 & 0.087 & 0.459 & 0.464 & 0.486 & 0.563 & 0.580 & 0.603 & 0.188 & 0.220 & 0.301 \\ \hline
\end{tabular}
\end{table*}

Since not all of the four facial characteristics were carefully
controlled at all three levels in the benchmark, we first check on the
consistency of the participants. For each image the standard
deviation of the user responses was calculated, and averaged over
the 20 images in each level.
This was done for gender, age and attractiveness, which can be treated
as numerical values, with each possible value in the user study mapped
to $\mathbb{N}$.  For example, attractiveness values \{below average,
average, above average\} are mapped to \{1, 2, 3\}.  For ethnicity,
which is a nominal value, the index of dispersion was used
instead.\footnote{ A version of the index of dispersion can be applied
to nominal values, and is computed as $ D = \frac{k(N^2-\sum_c
f_c^2)}{N^2(k-1)} $ where $k$ = number of categories, $N$ = number of
samples, and $f_c$ = frequency of $c$'th category.  }
Table~\ref{SDdispersion} shows that gender and age have standard
deviations below 0.5; that is, a clear majority of responses fall into
the same category.  The standard deviations
for attractiveness are a little higher, which is to be expected since
this is a more subjective characteristic.  The index of dispersion
values range from zero (all ratings fall into the same category) to
one (all ratings are equally divided between all the categories).
Since the dispersion values are less intuitive to understand than
standard deviation, we look at two examples.  The image with highest
ethnicity dispersion is the eighth image in level~3, Since there was
no control over ethnicity at level~3 such ambiguities are expected.
For this image the user responses for ethnicity were as follows: South Asian: 13,
East Asian: 4, White: 3, Black: 22, other: 14.  The resulting dispersion score
is 0.9, which reflects that the mode response (39\%) was below an
absolute majority.  It can be seen that the image is challenging (as
befits level~3): the figure
in the portrait has closed eyes, exhibits a strong expression, and the
lighting level is low.

  Since level~1 is controlled for ethnicity,
images with significant ambiguity of this characteristic should have
been avoided.  This is confirmed by noting that the image with largest
dispersion score in level 1 is the first image with a score of 0.6.
The user responses were: South Asian: 15, East Asian: 2, White: 0,
Black: 37, other: 2. Thus, a majority of users agreed.
Overall, in Table~\ref{SDdispersion} we see that for all four face
characteristics, in all but one case the variations increase slightly
as the levels increase, which is in line with the greater
variability in the images.

\begin{table*}[tbp]
\footnotesize
\centering
\caption{Evaluation of facial characteristics of 11 NPR algorithms. Errors for gender, age, attractiveness are \emph{signed} EMD distances;
for age and attractiveness positive values indicates an increase in judged value after stylisation, while for gender it indicates increased likelihood of assignment as female rather than male.
Larger absolute errors are marked in red: gender$\ge1$, age$\ge7$, attractiveness$\ge6$.
Yellow highlights indicate significant differences between levels for an NPR method (ANOVA at 0.05 level).
}
\label{signedEMD}
\begin{tabular}{|c||c|c|c||c|c|c||c|c|c|} \hline
characteristic & \multicolumn{3}{c||}{gender} & \multicolumn{3}{c||}{age} & \multicolumn{3}{c||}{attractiveness} \\ \hline
level & 1 & 2 & 3 & 1 & 2 & 3 & 1 & 2 & 3 \\ \hline \hline
neural style transfer~\cite{li2016combining} & 0.55 & \color{red}1.15 & \color{red}2.36 & \color{red}7.01 & \color{red}9.19 & \color{red}10.10 & \color{red}-6.32 & \color{red}-8.71 & \color{red}-8.11 \\ \hline
artistic sketch method~\cite{berger-shamir} & 0.17 & -0.33 & -1.32 & \cellcolor{yellow} 0.04 & 5.05 & \cellcolor{yellow} \color{red}7.96 & -2.42 & -3.62 & -2.65 \\ \hline
APDrawingGAN~\cite{YiLLR19} & -0.45 & 0.16 & 0.02 & \cellcolor{yellow} 0.79 & 3.85 & \cellcolor{yellow} \color{red}7.12 & -0.19 & -1.49 & -2.64 \\ \hline
puppet style~\cite{rosin-portrait} & 0.19 & -0.24 & 0.55 & -0.61 & 3.45 & 2.06 & 0.32 & -1.29 & -0.08 \\ \hline
XDoG~\cite{xdog} & -0.29 & -0.40 & -0.51 & 2.44 & 2.42 & -0.03 & 2.09 & \cellcolor{yellow} 0.21 & \cellcolor{yellow} 3.85 \\ \hline
engraving~\cite{engraving} & -0.25 & -0.05 & 0.34 & -2.20 & 0.02 & -0.88 & 1.37 & -0.36 & 3.76 \\ \hline
hedcut~\cite{Son-hedcut} & 0.45 & \cellcolor{yellow} -0.41 & \cellcolor{yellow} \color{red}1.27 & 0.24 & 1.59 & 2.50 & -0.80 & -1.58 & 0.88 \\ \hline
oil painting~\cite{semmo2016image} & -0.38 & -0.34 & 0.52 & -1.42 & 0.55 & -0.79 & 4.25 & 2.06 & 2.86 \\ \hline
Julian Opie style~\cite{rosin-portrait} & \color{red}-1.68 & -0.94 & \color{red}-2.76 & -3.53 & -2.79 & -3.74 & -2.90 & -3.06 & -0.44 \\ \hline
pebble mosaic~\cite{doyle2019automated} & 0.03 & -0.77 & 0.73 & 0.26 & 2.45 & -0.69 & 2.42 & 1.44 & 1.06 \\ \hline
watercolour~\cite{rosin2017watercolour} & 0.03 & -0.24 & 0.31 & -3.16 & -2.91 & -0.61 & 2.72 & 0.51 & 3.90 \\ \hline
\end{tabular} \end{table*}

\begin{table*}[tbp]
\footnotesize
\centering
\caption{Evaluation of facial characteristics of 11 NPR algorithms. Errors for gender, age, attractiveness are \emph{unsigned} EMD distances.
Larger errors are marked in red: gender$\ge2$, age$\ge7$, attractiveness$\ge7$, ethnicity$\ge15$.
Yellow highlights indicate significant differences between levels for an NPR method (ANOVA at 0.05 level).
}
\label{unsignedEMD}
\begin{tabular}{|c||c|c|c||c|c|c||c|c|c|} \hline
characteristic & \multicolumn{3}{c||}{gender} & \multicolumn{3}{c||}{age} & \multicolumn{3}{c||}{attractiveness} \\ \hline
level & 1 & 2 & 3 & 1 & 2 & 3 & 1 & 2 & 3 \\ \hline \hline
neural style transfer~\cite{li2016combining} & 1.49 & \color{red}2.02 & \color{red}3.53  & \color{red}8.90 & \color{red}11.03 & \color{red}11.23 & \color{red}8.49 & \color{red}10.18 & \color{red}8.58 \\ \hline
artistic sketch method~\cite{berger-shamir} & \cellcolor{yellow} \color{red}2.21 & \cellcolor{yellow} \color{red}2.00 & \cellcolor{yellow} \color{red}4.82 & \color{red}7.57 & \color{red}8.70 & \color{red}11.72 & 6.94 & 6.01 & 6.29 \\ \hline
APDrawingGAN~\cite{YiLLR19} & 0.97 & \cellcolor{yellow} 0.55 & \cellcolor{yellow} \color{red}2.06 & 5.50 & 6.13 & \color{red}9.07 & \cellcolor{yellow} 3.81 & 4.90 & \cellcolor{yellow} \color{red}7.10 \\ \hline
puppet style~\cite{rosin-portrait} & 0.59 & 0.73 & 1.13 & 6.19 & 4.74 & \color{red}7.51 & 5.33 & 5.06 & 4.69 \\ \hline
XDoG~\cite{xdog} & 0.90 & 0.76 & 0.99 & 5.11 & 5.05 & 5.02 & 5.39 & 4.45 & 6.25 \\ \hline
engraving~\cite{engraving} & 0.63 & 0.61 & 0.74 & 4.17 & 4.34 & 4.27 & 4.98 & 5.29 & 5.32 \\ \hline
hedcut~\cite{Son-hedcut} & 1.03 & 1.24 & 1.45 & 5.63 & 4.19 & 6.31 & 4.37 & 4.78 & 4.79 \\ \hline
oil painting~\cite{semmo2016image} & 0.65 & 0.52 & 0.96 & 4.23 & 3.95 & 3.05 & 5.32 & 4.48 & 4.37 \\ \hline
Julian Opie style~\cite{rosin-portrait} & 1.97 & 1.09 & \color{red}3.91 & 6.37 & 5.88 & \color{red}7.84 & 6.64 & 6.16 & \color{red}7.43 \\ \hline
pebble mosaic~\cite{doyle2019automated} & 0.51 & 1.07 & 1.81 & 5.19 & 4.75 & 5.87 & 4.42 & 5.25 & 5.98 \\ \hline
watercolour~\cite{rosin2017watercolour} & 0.49 & 0.66 & 0.86 & \cellcolor{yellow} 5.49 & \cellcolor{yellow} 3.40 & 4.74 & 4.62 & 4.75 & 5.70 \\ \hline
\end{tabular}
\end{table*}

\begin{table}[tbp]
\footnotesize
\centering
\caption{Evaluation of facial characteristic of 11 NPR algorithms: ethnicity.
Error is measured using the$L_1$ distance, and larger errors (ethnicity$\ge15$) are marked in red.
Yellow highlights indicate significant differences between levels for an NPR method (ANOVA at 0.05 level).
}
\label{ethnicity}
\begin{tabular}{|c||c|c|c|} \hline
characteristic & \multicolumn{3}{c|}{ethnicity} \\ \hline
level & 1 & 2 & 3 \\ \hline \hline
neural style transfer~\cite{li2016combining} & \color{red}20.93 & \color{red}16.10 & \color{red}17.15 \\ \hline
artistic sketch method~\cite{berger-shamir} & \color{red}18.14 & \color{red}15.51 & \color{red}18.75 \\ \hline
APDrawingGAN~\cite{YiLLR19} & \cellcolor{yellow} 11.08 & 12.64 & \cellcolor{yellow} \color{red}17.50 \\ \hline
puppet style~\cite{rosin-portrait} & 10.83 & \color{red}14.84 & \color{red}17.81 \\ \hline
XDoG~\cite{xdog} & 8.90 & 8.02 & 9.75 \\ \hline
engraving~\cite{engraving} & 6.37 & 6.54 & 7.88 \\ \hline
hedcut~\cite{Son-hedcut} & 9.20 & 8.48 & 9.58 \\ \hline
oil painting~\cite{semmo2016image} & 4.74 & 4.51 & 7.58 \\ \hline
Julian Opie style~\cite{rosin-portrait} & \color{red}15.82 & \color{red}17.03 & \color{red}16.79 \\ \hline
pebble mosaic~\cite{doyle2019automated} & \cellcolor{yellow} 6.10 & \cellcolor{yellow} 5.91 & \cellcolor{yellow} 12.66 \\ \hline
watercolour~\cite{rosin2017watercolour} & 5.77 & 6.22 & 5.66 \\ \hline
\end{tabular}
\end{table}

\subsection{Experiment 1: Correctness of facial characteristics}

We conducted Experiment~1 described in section~\ref{experiments} and
applied it to 11 NPR algorithms which cover a wide range of styles and
methods: neural style transfer~\cite{li2016combining},
XDoG~\cite{xdog}, oil painting~\cite{semmo2016image}, pebble
mosaic~\cite{doyle2019automated}, artistic sketch
method~\cite{berger-shamir}, APDrawingGAN~\cite{YiLLR19}, puppet
style~\cite{rosin-portrait} engraving~\cite{engraving},
hedcut~\cite{Son-hedcut}, Julian Opie style~\cite{rosin-portrait},
watercolour~\cite{rosin2017watercolour}.  The 11 NPR algorithms are
run on the full 60-image benchmark and so the first user study to
collect the four face characteristics contained 660 stylised photos.
There were 225 participants in the study, and they viewed randomly
generated subsets of 30 stylised images.

Tables~\ref{signedEMD}, \ref{unsignedEMD} and~\ref{ethnicity} list the
errors in the face characteristics of the stylised images compared to
the original portraits.  Note that since not all images had the
same number of user responses, the histograms are standardised to unit
area before computing distances.  The signed EMD distances are useful
in showing trends in the signs of differences.
For instance, the neural style transfer~\cite{li2016combining}
stylisation has a slight trend to make people look more
feminine\footnote{ The value of 2.36 for shift in gender at level~3 is
mostly accounted for by five of the images that had movements of
between one and three quarters of their distribution from male to
female.  Three of these images had a change in the majority gender
compared to the ground truth.  }, older, and less attractive.  On the
other hand, the Julian Opie style~\cite{rosin-portrait} tends to make
people look more masculine and a little younger.

Under
the signed EMD distance, opposite sign movements (differences)
cancel out, so it is useful to look at the unsigned EMD distances to
check the overall error.  Table~\ref{unsignedEMD} shows that both the
neural style transfer~\cite{li2016combining} and the artistic sketch
method~\cite{berger-shamir} produce renderings that differ
substantially from the ground truth on all the face characteristics.
This is due to their highly stylised output, which has elements of
strong geometric abstraction and distortion.  Of course, this
distortion is carried out deliberately to match the geometric style of
the artist, so it is natural that this will impact the perception of
facial characteristics.  APDrawingGAN~\cite{YiLLR19} is seen to be
sensitive to the complexity of the input; its errors are reasonably
low for level~1, but double at level~3 for some characteristics.
Table~\ref{ethnicity} shows that ethnicity is poorly recognised on
outputs from the puppet style~\cite{rosin-portrait}, which is due to
low lighting levels causing the shading effect to make the faces dark.
For instance, in level~3 the main error came from five such images
which were unambiguously classified as white from the source
portraits, but between 44\% and 88\% users classified them as black
from the puppet stylised versions.  Significant errors were also made
in determining ethnicity from the Julian Opie
style~\cite{rosin-portrait}.  This is unsurprising given the strong
level of abstraction.

We applied ANOVA tests to the signed and unsigned distances to check
for significant differences between levels for each characteristic and
stylisation.  This allows us to check the effects of increasing the
complexity of the source images on the NPR algorithms.  Both the
artistic sketch method~\cite{berger-shamir} and
APDrawingGAN~\cite{YiLLR19} show significant increase in the perceived
age of the portraits when the image complexity increases.  This is
probably due to the increased difficulty in generating clean
renderings, and the increased number of fragmented and
distracting lines that appear in the renderings.  Although
Table~\ref{signedEMD} shows two other instances of statistically
significant differences between levels (increased attractiveness for
XDoG~\cite{xdog} and increased femininity for
hedcut~\cite{Son-hedcut}) the trends are not consistent across all
three levels.  Table~\ref{unsignedEMD} indicates that the perceived
attractiveness of images stylised by APDrawingGAN~\cite{YiLLR19}
exhibits a consistently increasing divergence from the original photos
across levels, and that this is statistically significant.  Although
the pebble mosaic stylisation~\cite{doyle2019automated} generally
produces less discrepancies for ethnicity than most of the other
stylisations,
we see a statistically significant increase in these errors
as the image complexity increases.  This may be due to the constant
colour mosaic boundaries, which effectively dilute skin tone and
thereby potentially cause confusion under challenging lighting
conditions.

\subsection{Experiment 2: Quality of stylisation across levels}

\begin{table}[t]
\centering
\caption{Correlation coefficients between triplet rankings and benchmark levels.}
\label{correlation}
\small
\begin{tabular}{|c||c|c|}
\hline
\bf method & \bf Pearson & \bf Kendall \\ \hline \hline
neural style transfer~\cite{li2016combining} & 0.400& 0.363\\ \hline
artistic sketch method~\cite{berger-shamir} & 0.337 & 0.306\\ \hline
APDrawingGAN~\cite{YiLLR19} & 0.384 & 0.346 \\ \hline
puppet style~\cite{rosin-portrait} & 0.316 & 0.284\\ \hline
XDoG~\cite{xdog} & 0.145 & 0.130\\ \hline
engraving~\cite{engraving} & 0.170& 0.154\\ \hline
hedcut~\cite{Son-hedcut} & 0.222& 0.202 \\ \hline
oil painting~\cite{semmo2016image} & -0.019 & -0.017\\ \hline
Julian Opie style~\cite{rosin-portrait} & 0.296 & 0.266\\ \hline
pebble mosaic~\cite{doyle2019automated} & 0.232 & 0.207\\ \hline
watercolour~\cite{rosin2017watercolour} & 0.126& 0.113 \\ \hline
\end{tabular}
\end{table}

\begin{figure}[!t]
\centering
\includegraphics[height=1.6cm]{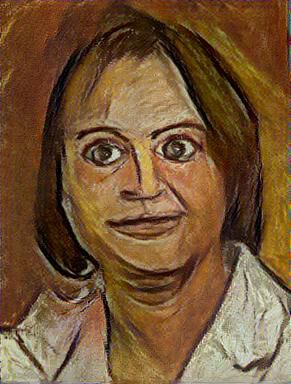}
\includegraphics[height=1.6cm]{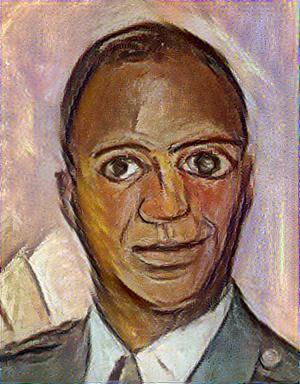}
\includegraphics[height=1.6cm]{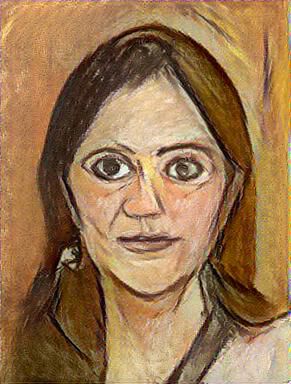}
\includegraphics[height=1.6cm]{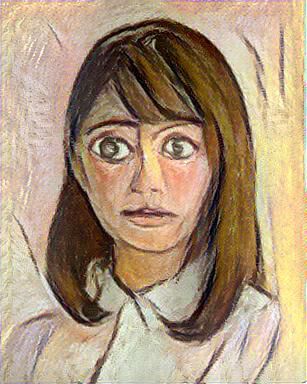}
\includegraphics[height=1.6cm]{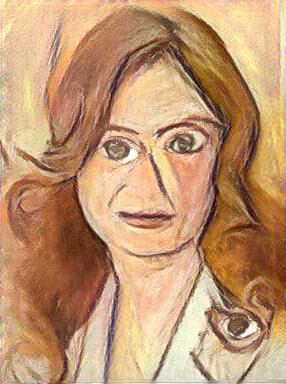}

\centerline{Level 1}
\medskip

\includegraphics[height=1.6cm]{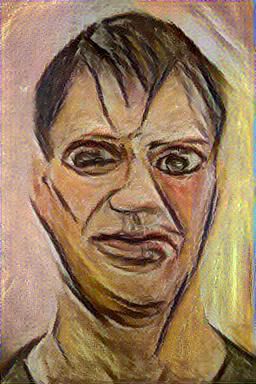}
\includegraphics[height=1.6cm]{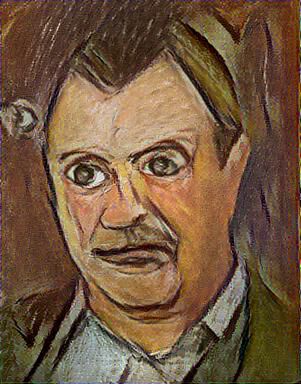}
\includegraphics[height=1.6cm]{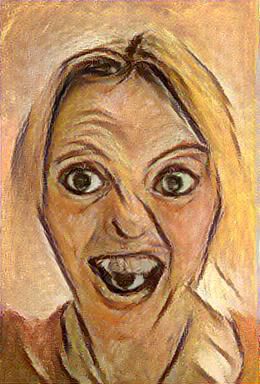}
\includegraphics[height=1.6cm]{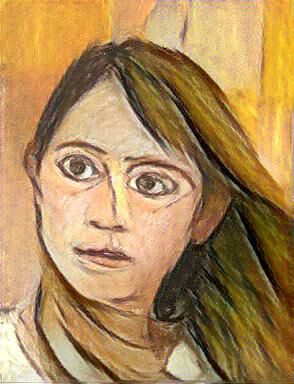}
\includegraphics[height=1.6cm]{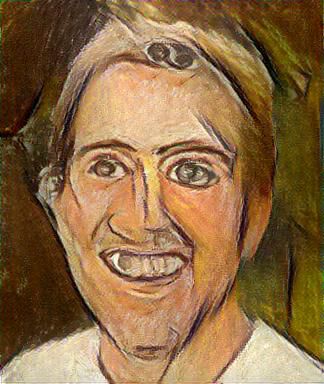}

\centerline{Level 2}
\medskip

\includegraphics[height=1.6cm]{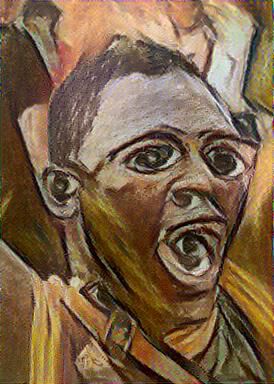}
\includegraphics[height=1.6cm]{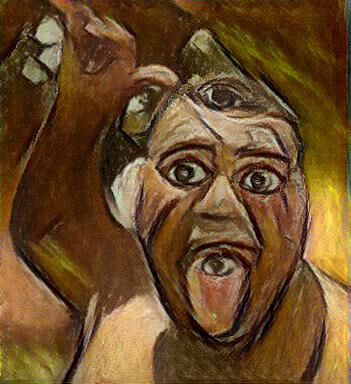}
\includegraphics[height=1.6cm]{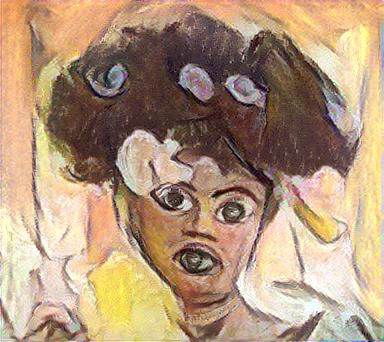}
\includegraphics[height=1.6cm]{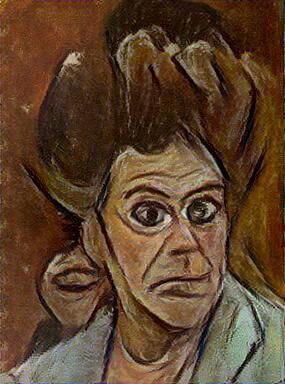}
\includegraphics[height=1.6cm]{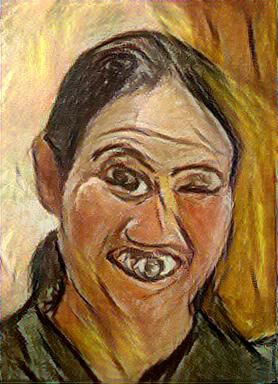}

\centerline{Level 3}
\medskip
\caption{Images from the \emph{NPRportrait1.0} benchmark stylised using neural style transfer: Li and Wand~\cite{li2016combining}}
\label{resultsChuan}
\end{figure}


\begin{figure}[!t]
\centering
\includegraphics[height=1.6cm]{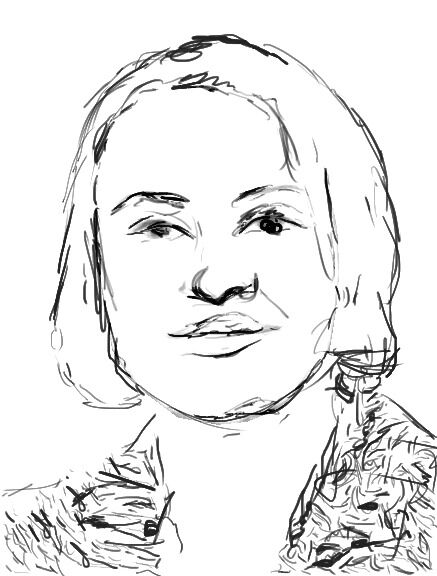}
\includegraphics[height=1.6cm]{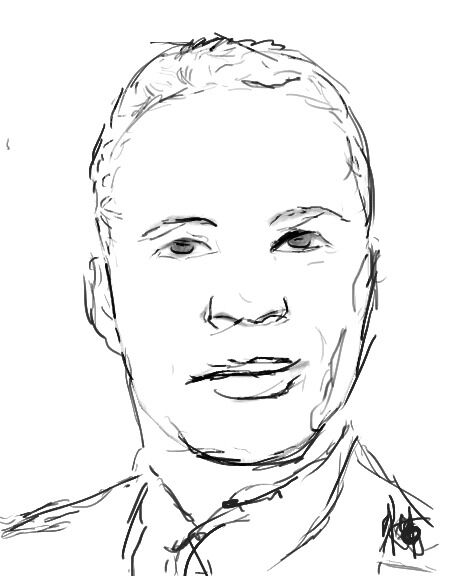}
\includegraphics[height=1.6cm]{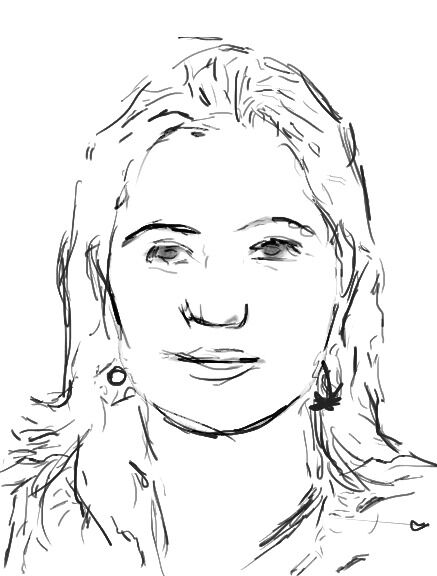}
\includegraphics[height=1.6cm]{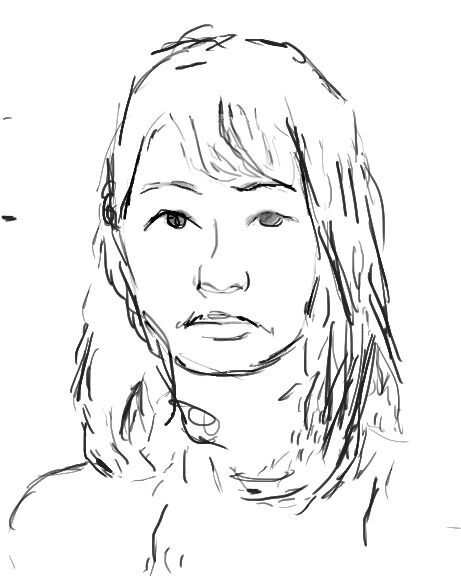}
\includegraphics[height=1.6cm]{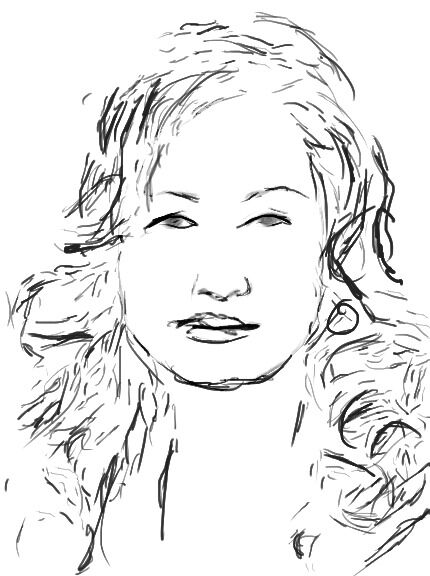}

\centerline{Level 1}
\medskip

\includegraphics[height=1.6cm]{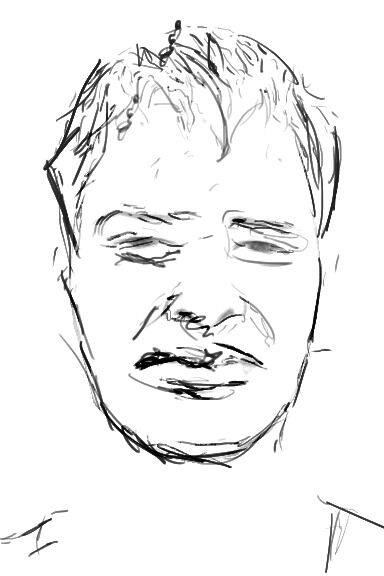}
\includegraphics[height=1.6cm]{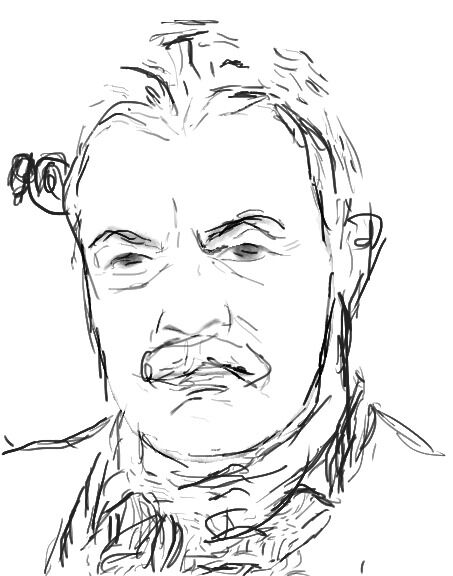}
\includegraphics[height=1.6cm]{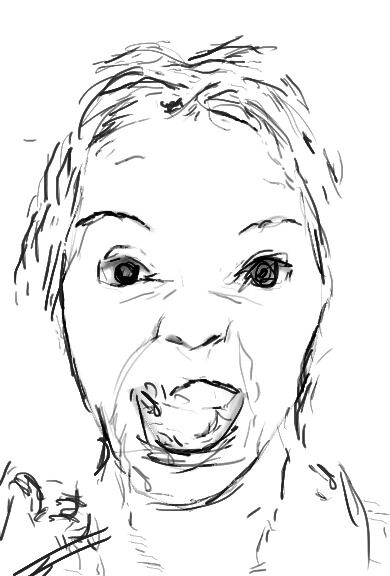}
\includegraphics[height=1.6cm]{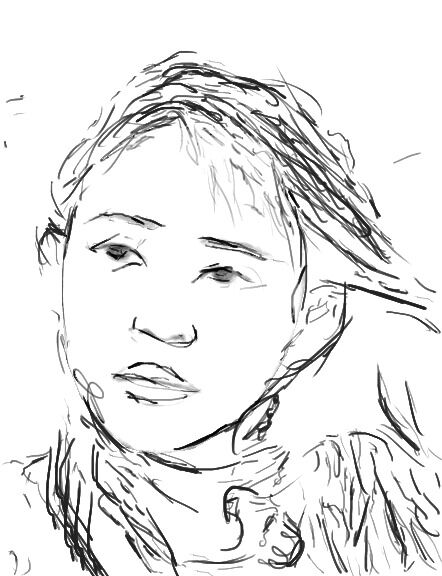}
\includegraphics[height=1.6cm]{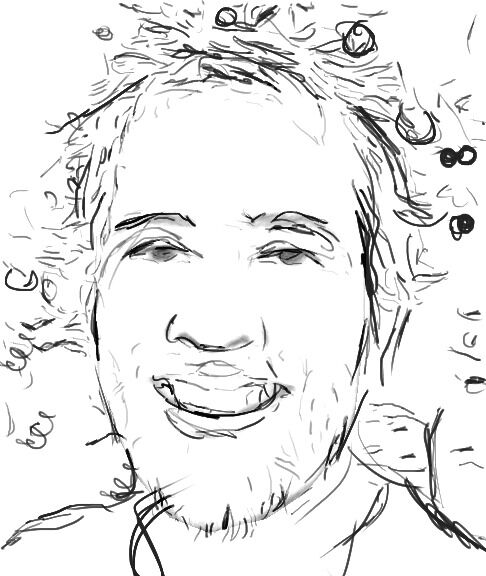}

\centerline{Level 2}
\medskip

\includegraphics[height=1.6cm]{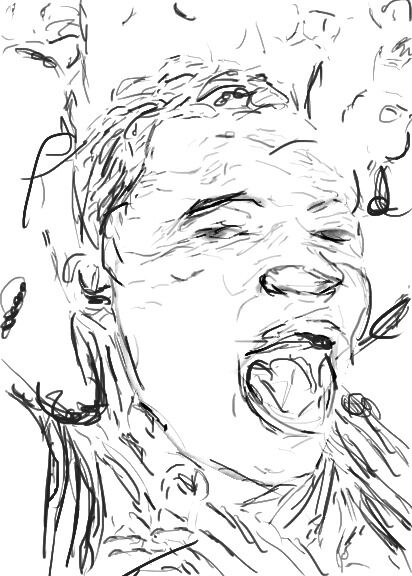}
\includegraphics[height=1.6cm]{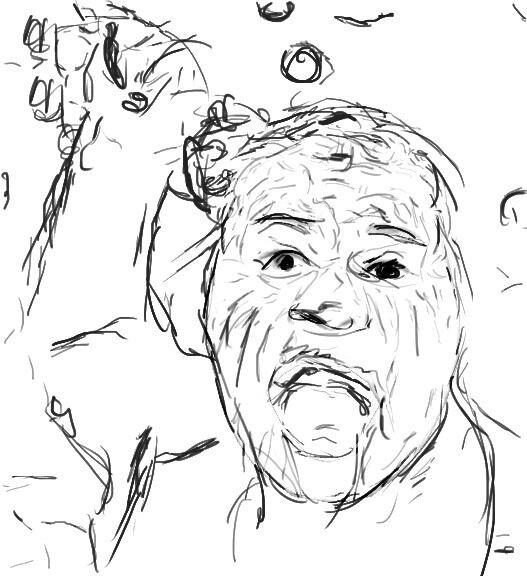}
\includegraphics[height=1.6cm]{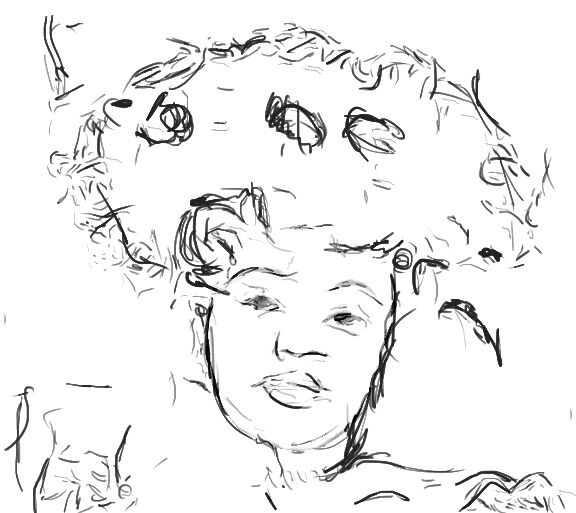}
\includegraphics[height=1.6cm]{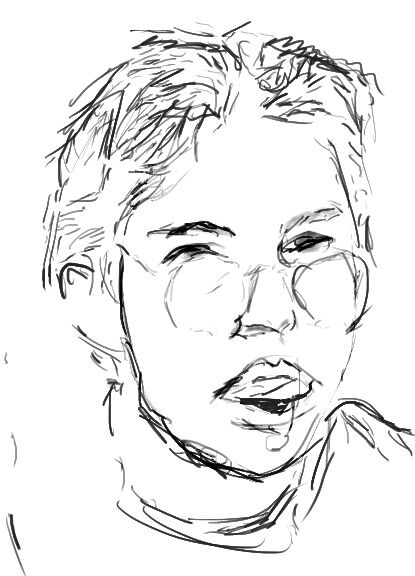}
\includegraphics[height=1.6cm]{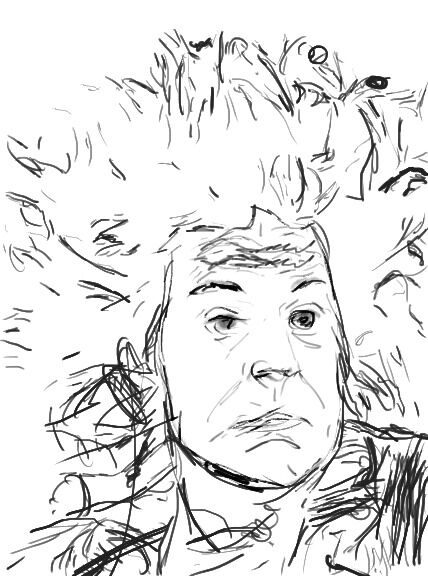}
\includegraphics[height=1.6cm]{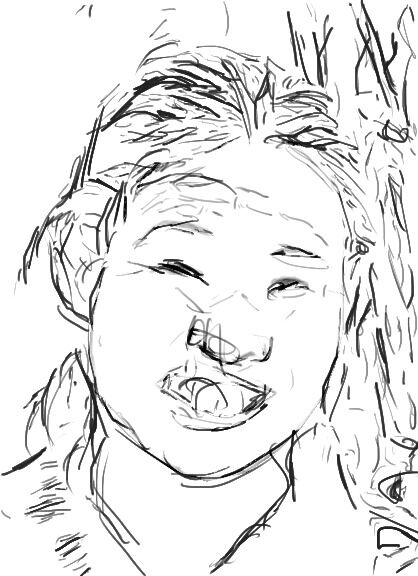}

\centerline{Level 3}
\medskip
\caption{Images from the \emph{NPRportrait1.0} benchmark stylised by the artistic sketch method:
Berger \emph{et al.}~\cite{berger-shamir}}
\label{resultsItamar}
\end{figure}


\begin{figure}[!t]
\centering
\includegraphics[height=1.6cm]{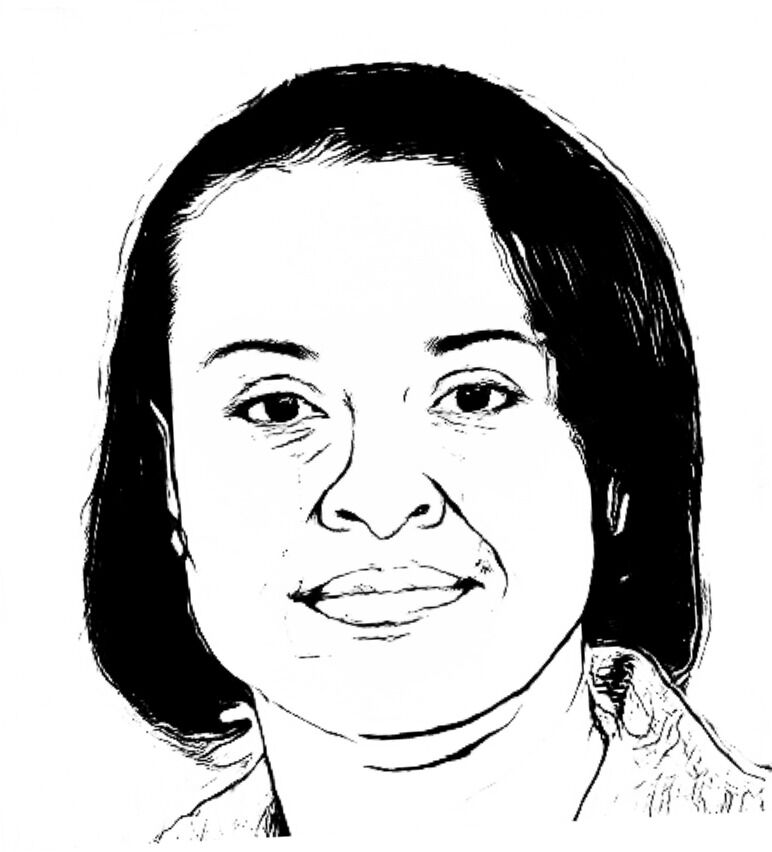}
\includegraphics[height=1.6cm]{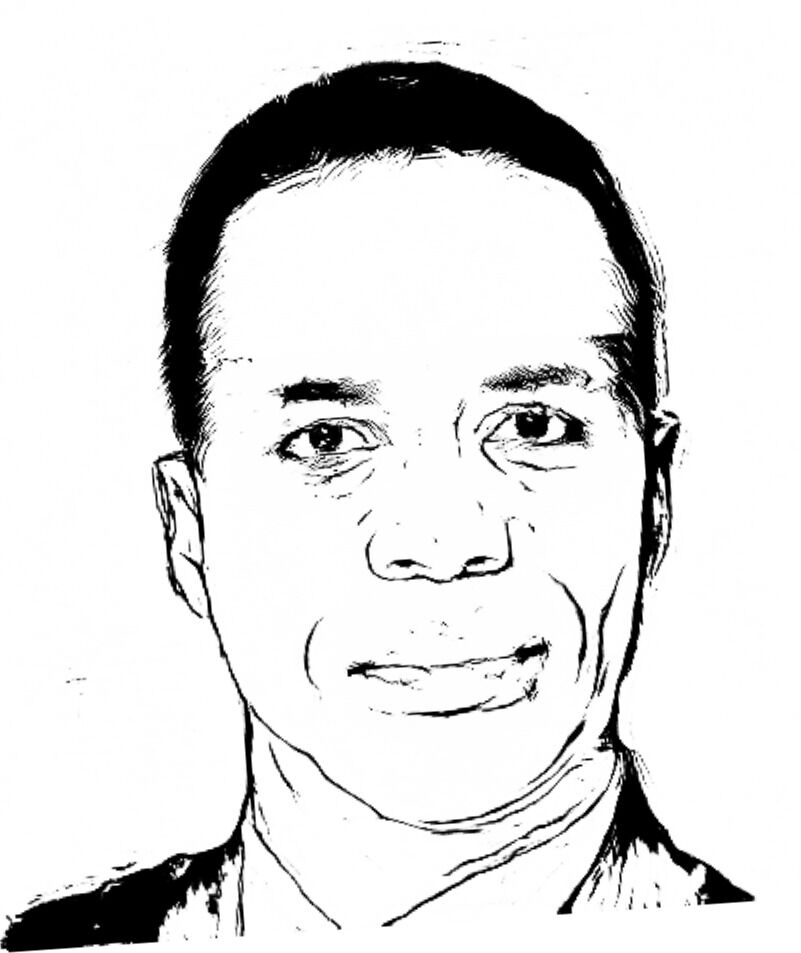}
\includegraphics[height=1.6cm]{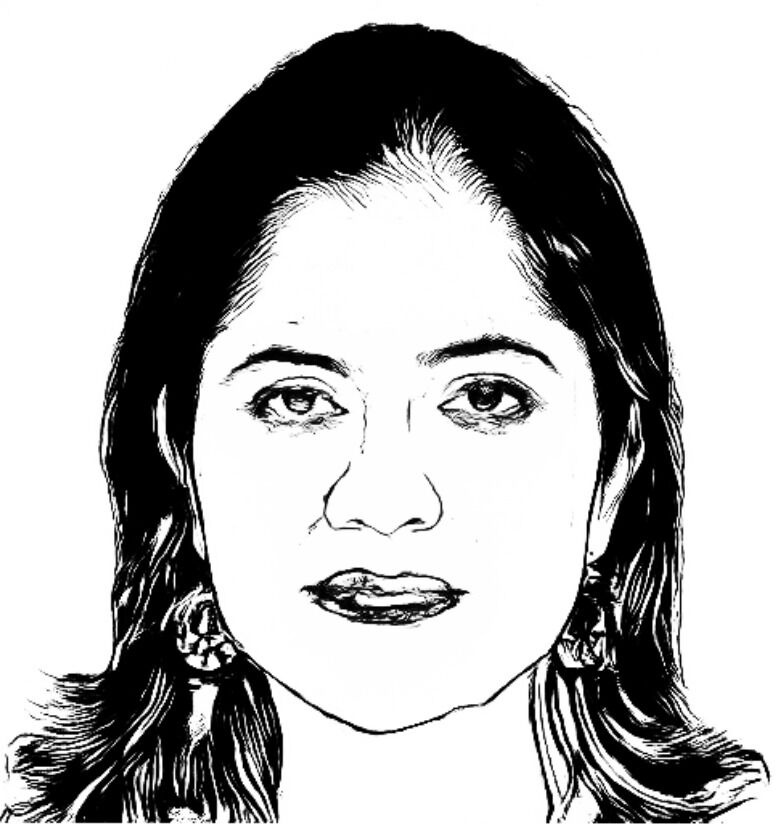}
\includegraphics[height=1.6cm]{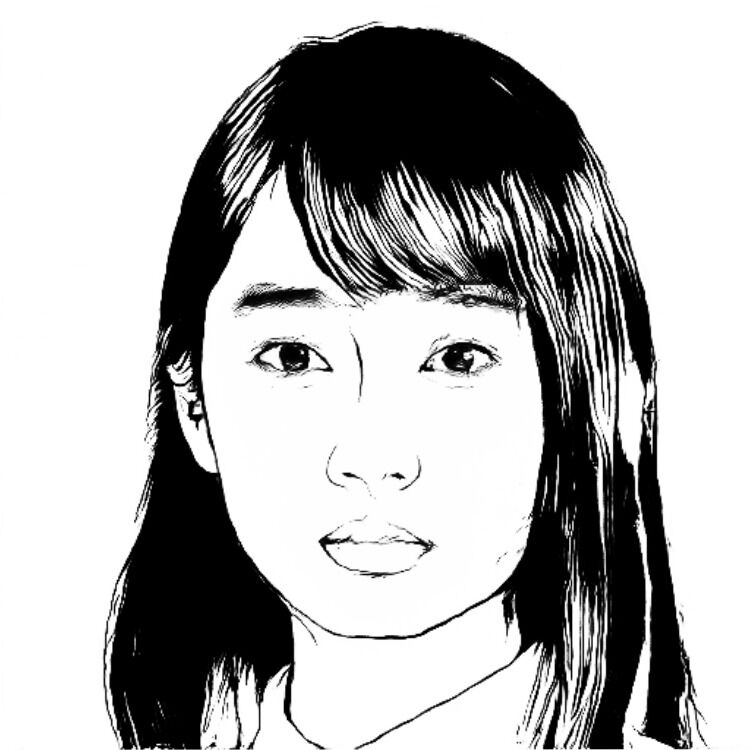}
\includegraphics[height=1.6cm]{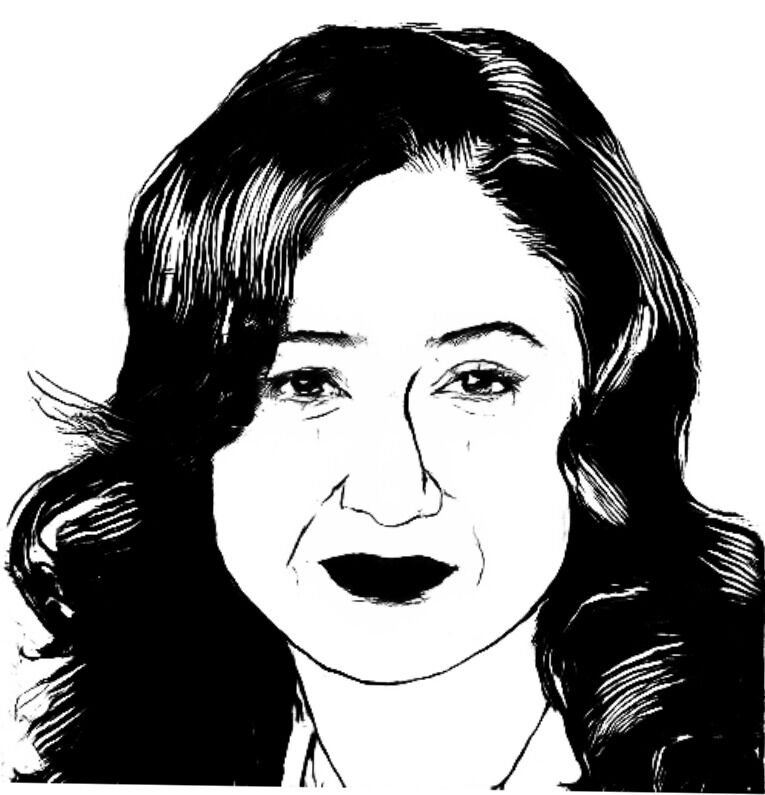}

\centerline{Level 1}
\medskip

\includegraphics[height=1.6cm]{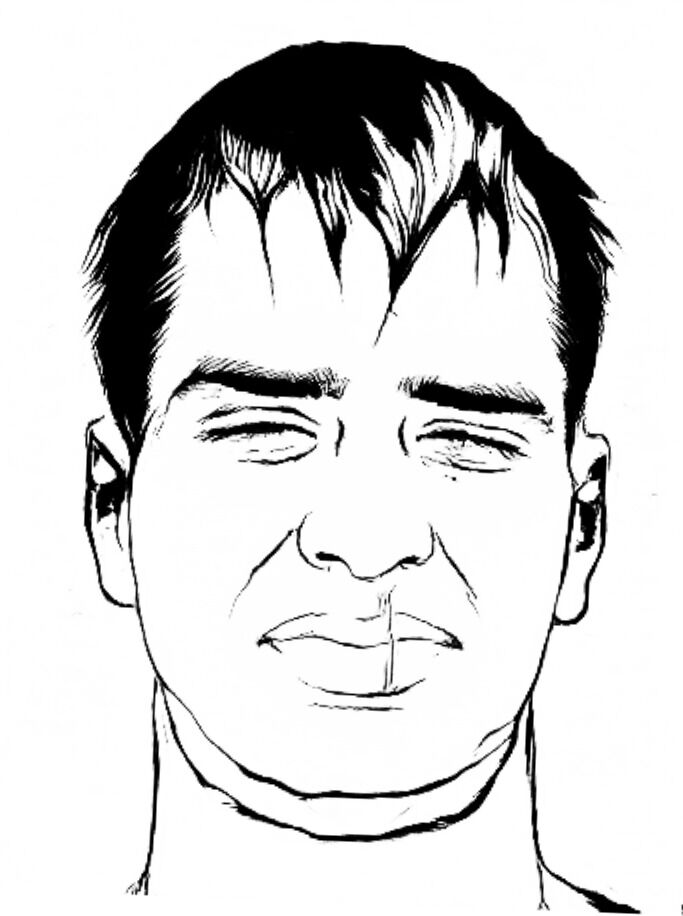}
\includegraphics[height=1.6cm]{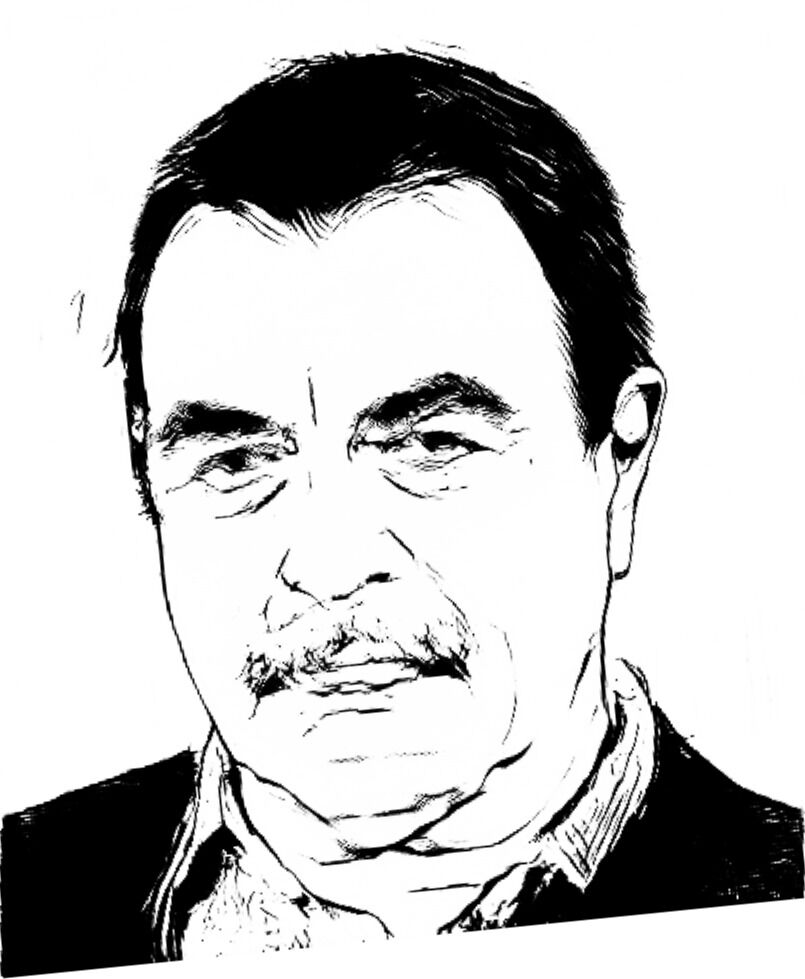}
\includegraphics[height=1.6cm]{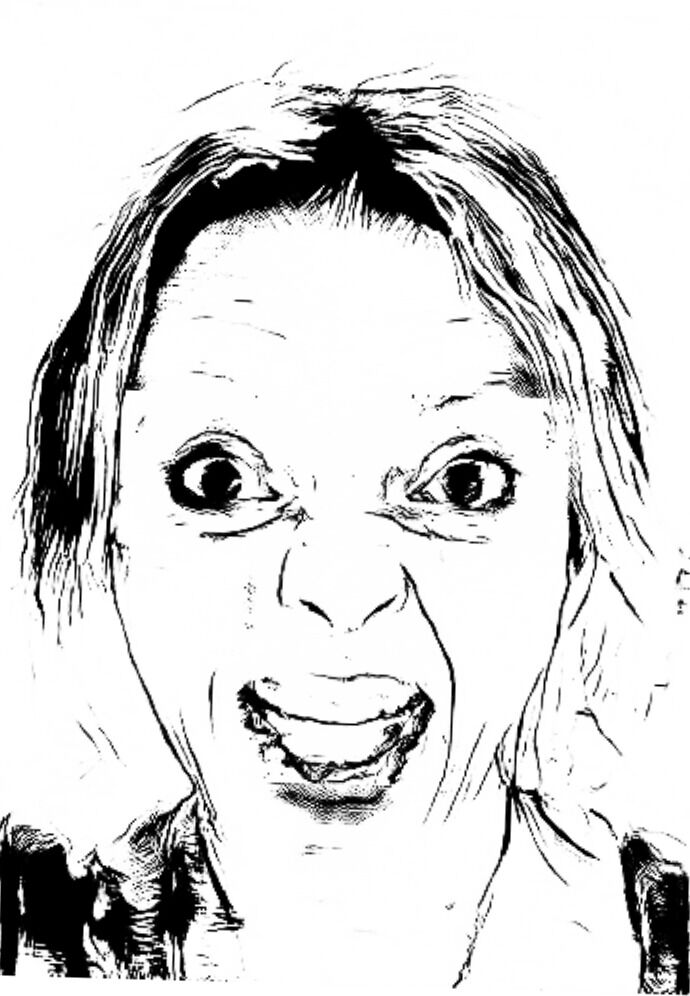}
\includegraphics[height=1.6cm]{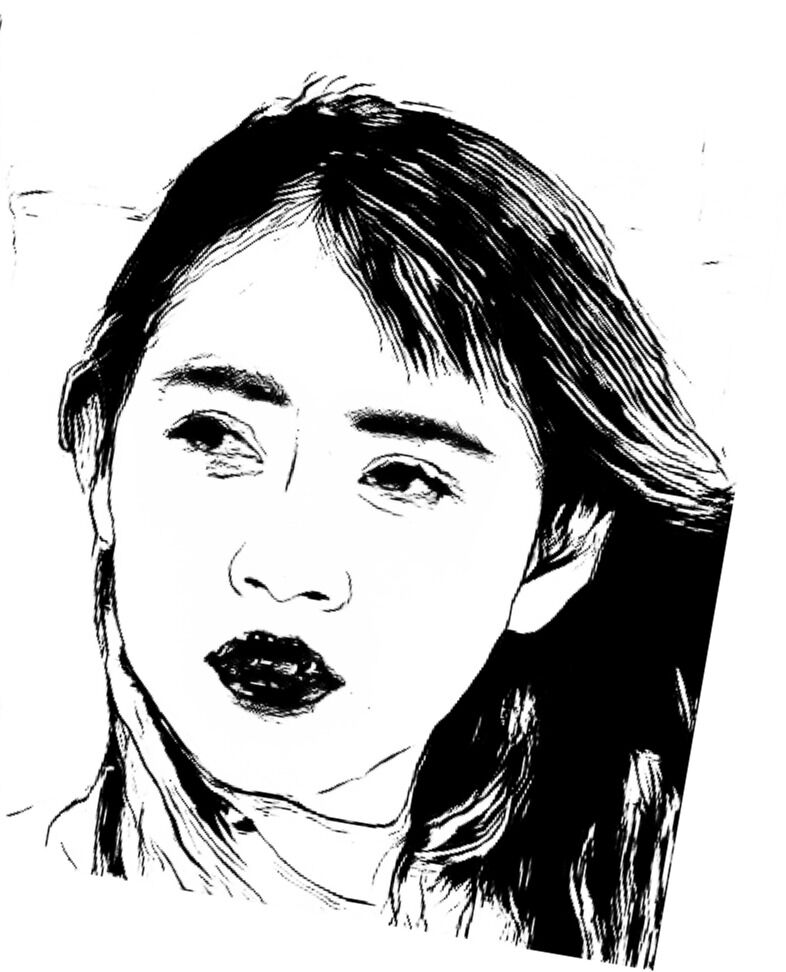}
\includegraphics[height=1.6cm]{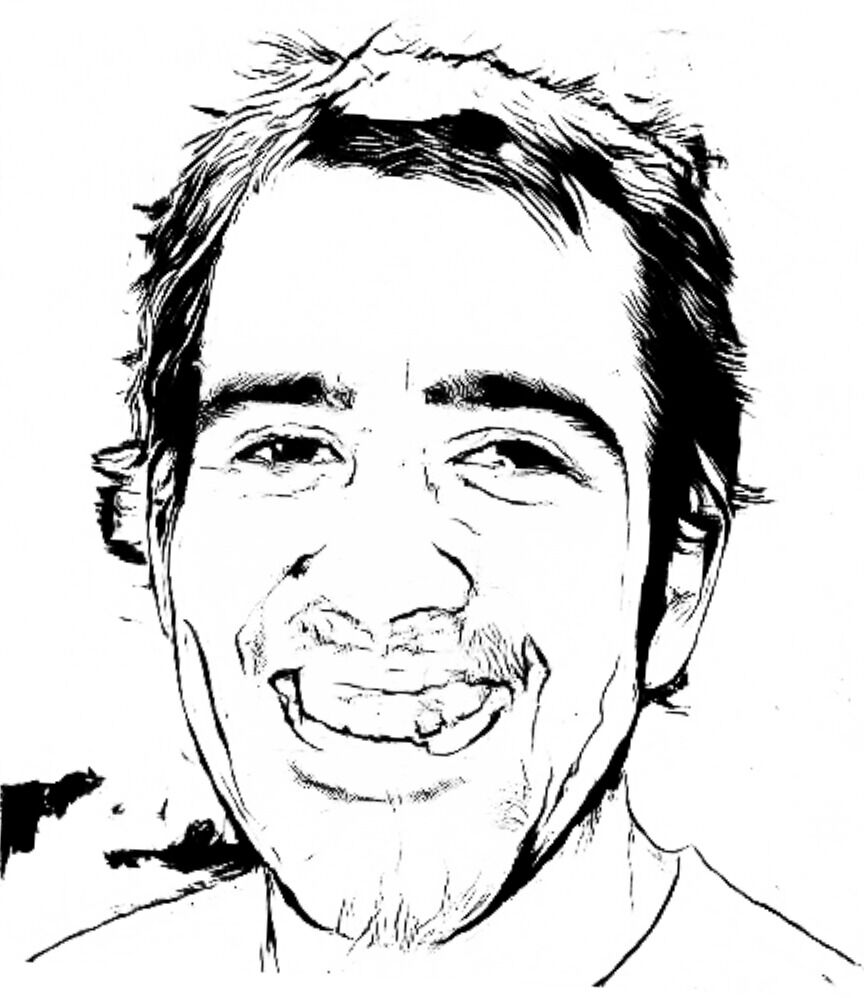}

\centerline{Level 2}
\medskip

\includegraphics[height=1.6cm]{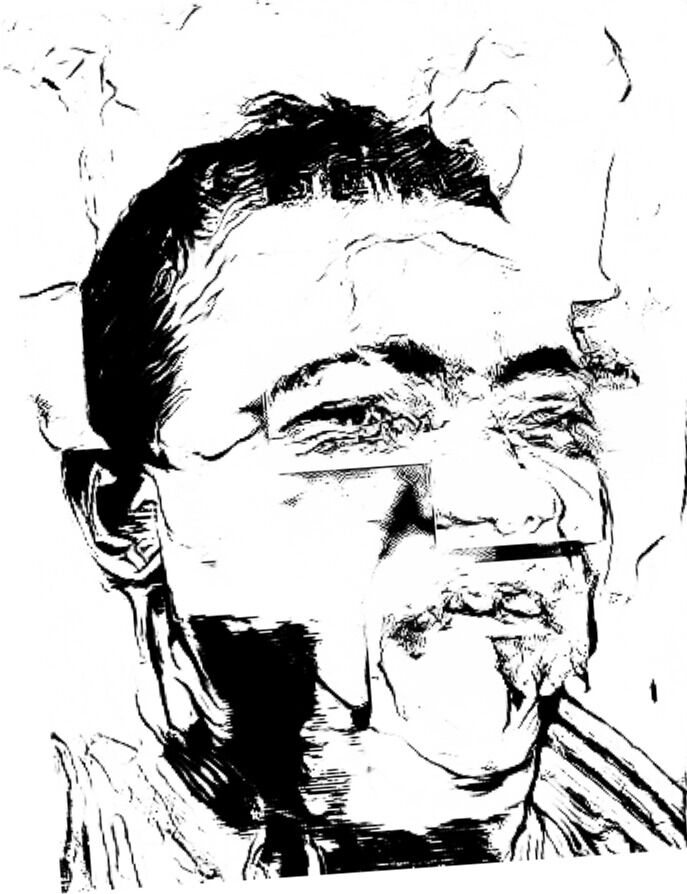}
\includegraphics[height=1.6cm]{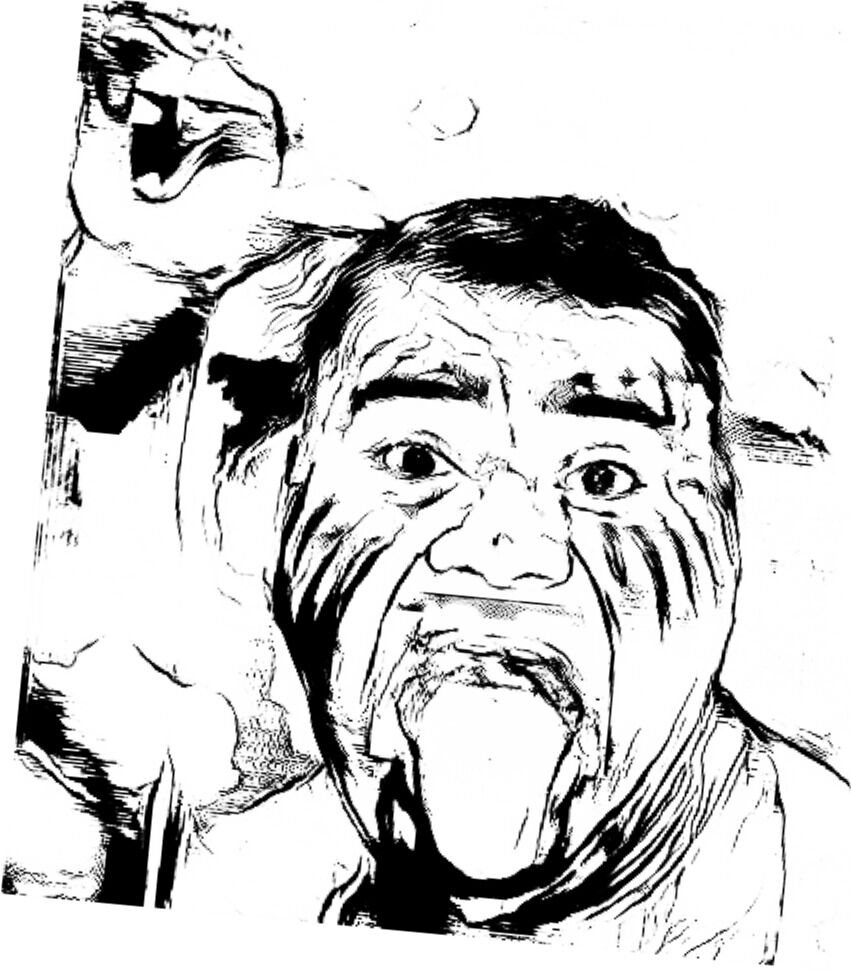}
\includegraphics[height=1.6cm]{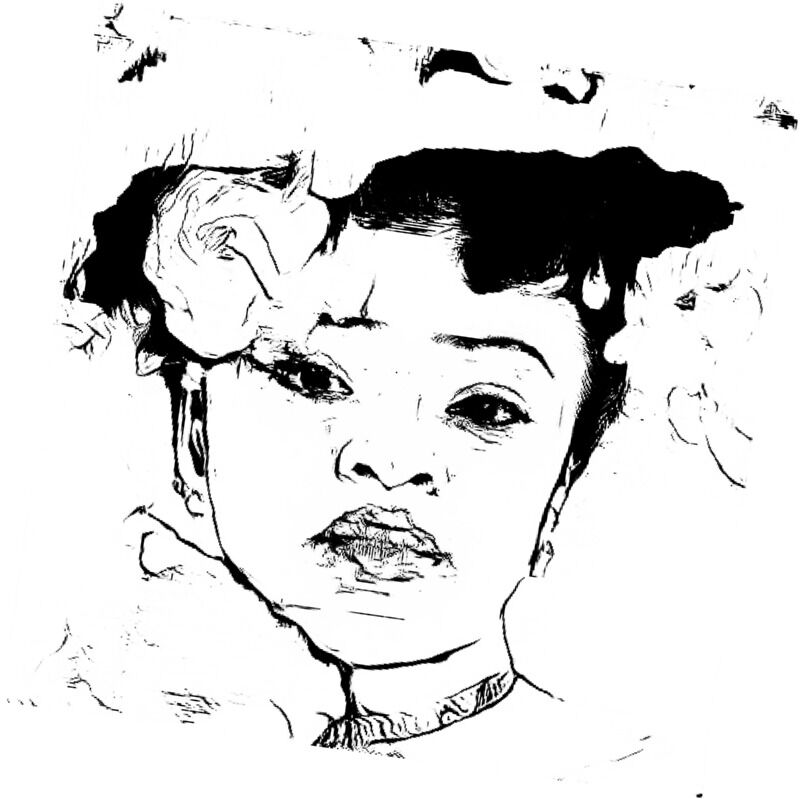}
\includegraphics[height=1.6cm]{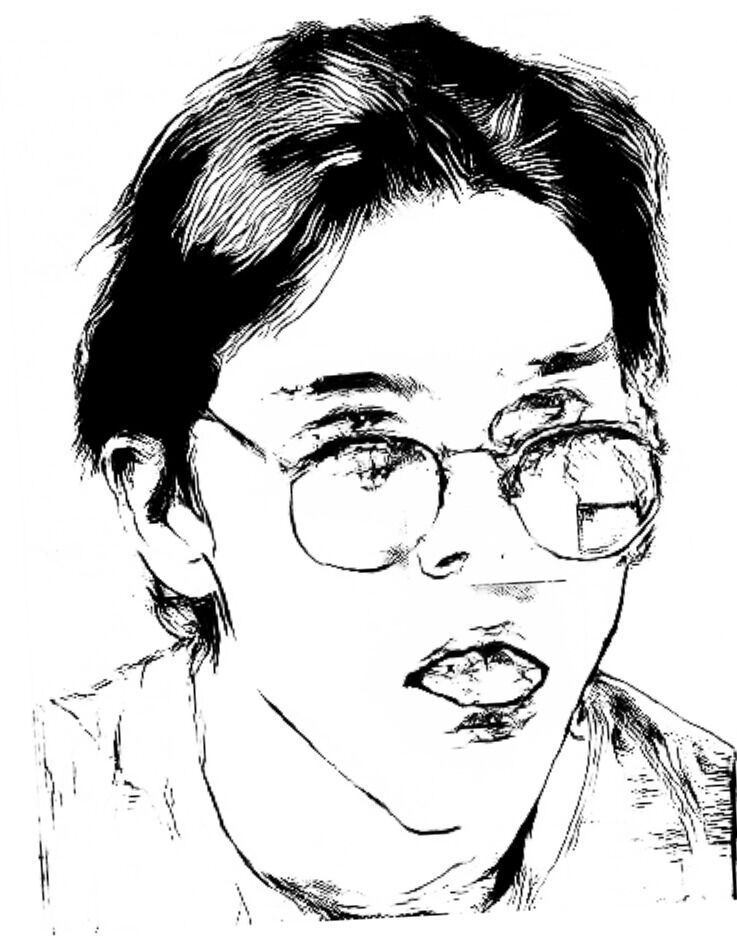}
\includegraphics[height=1.6cm]{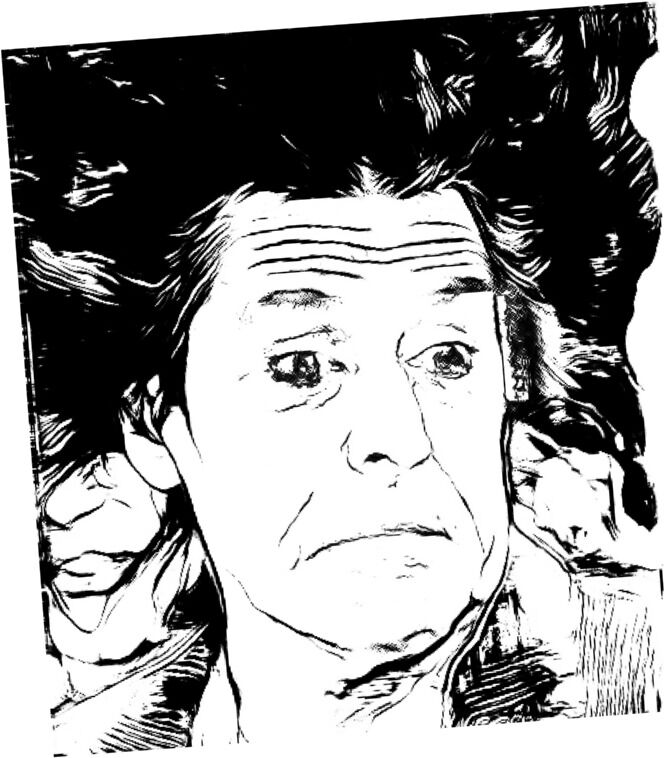}
\includegraphics[height=1.6cm]{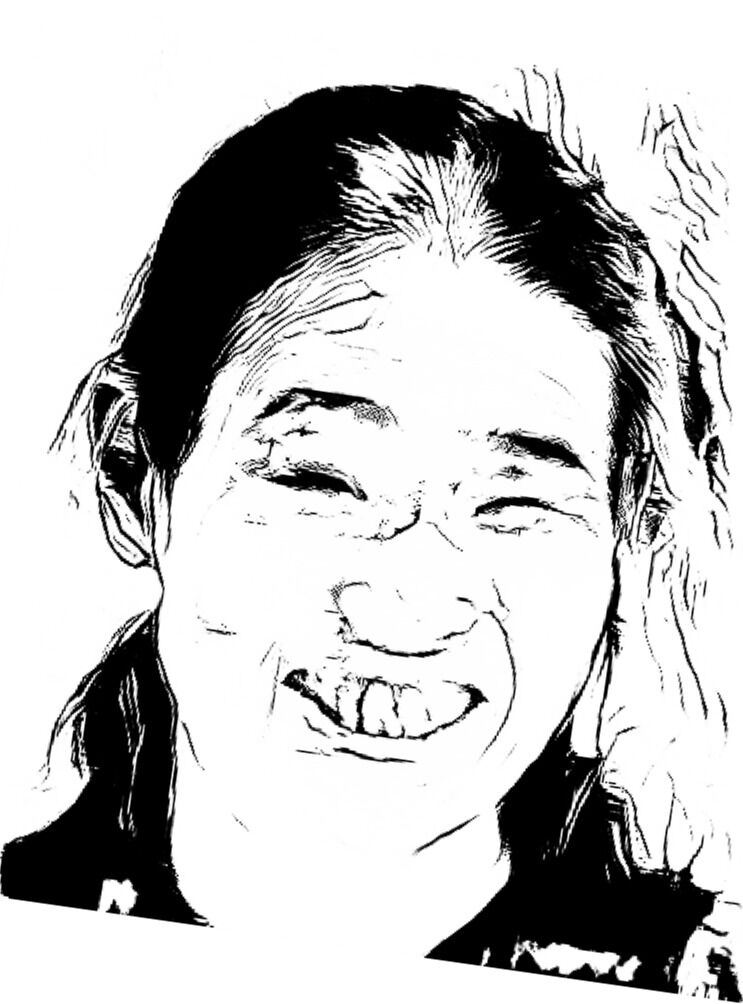}

\centerline{Level 3}
\medskip
\caption{Images from the \emph{NPRportrait1.0} benchmark stylised by APDrawingGAN: Yi \emph{et al.}~\cite{YiLLR19}}
\label{resultsRan}
\end{figure}


\begin{figure}[!t]
\centering
\includegraphics[height=1.6cm]{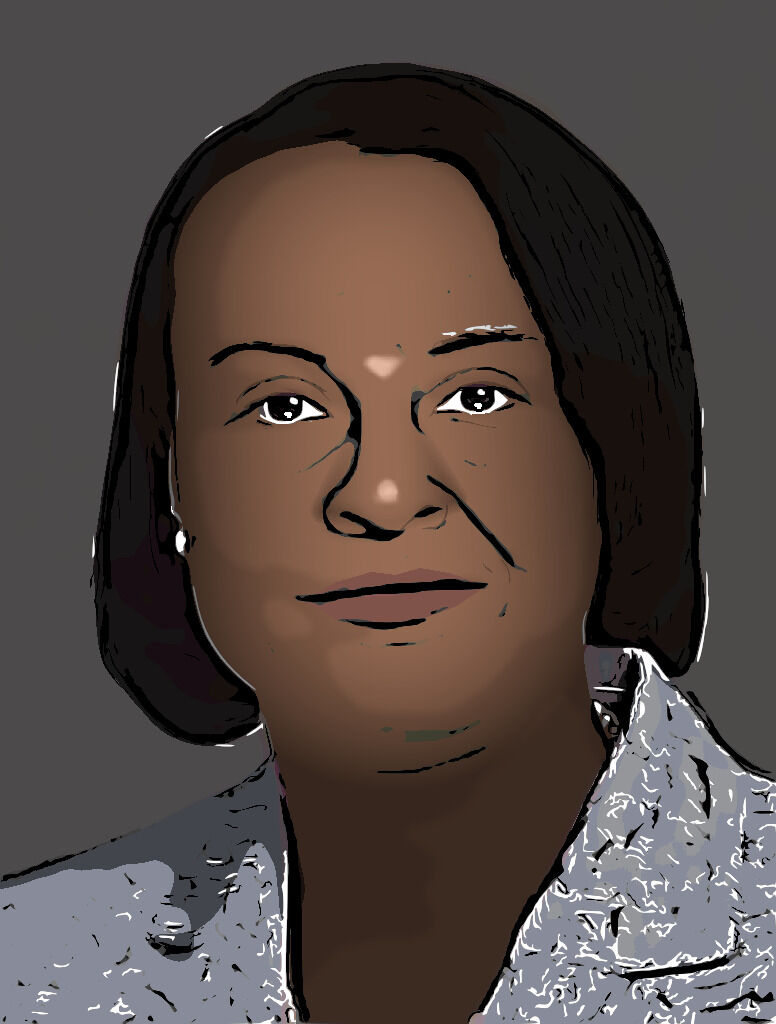}
\includegraphics[height=1.6cm]{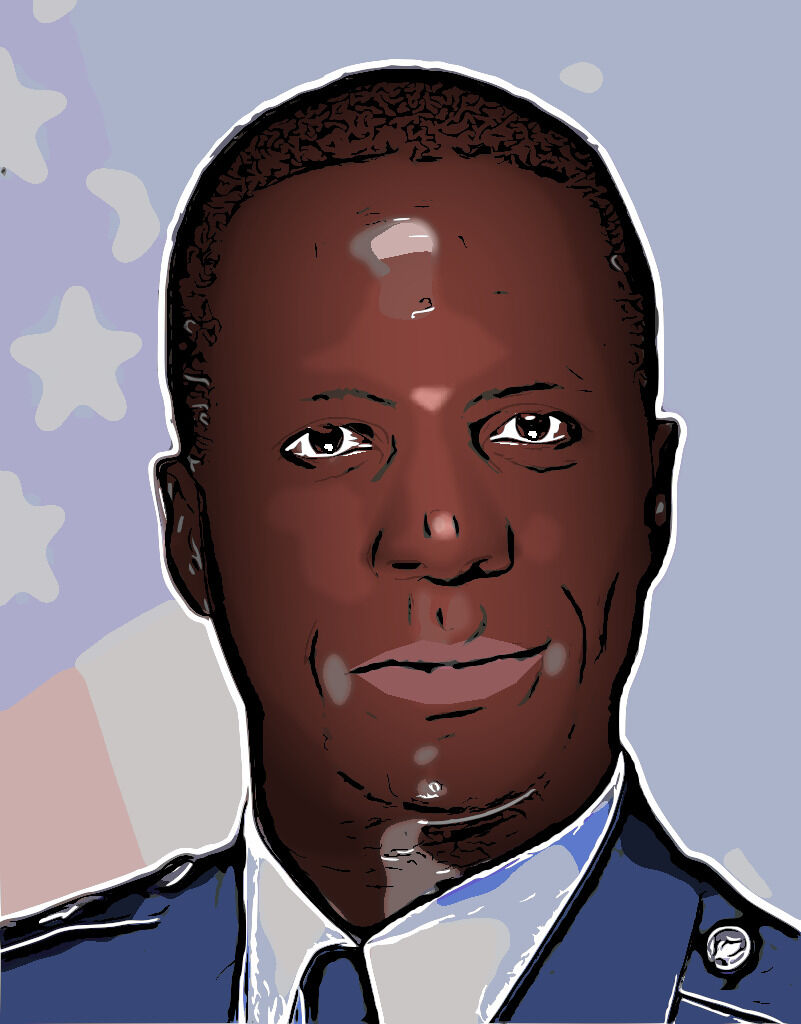}
\includegraphics[height=1.6cm]{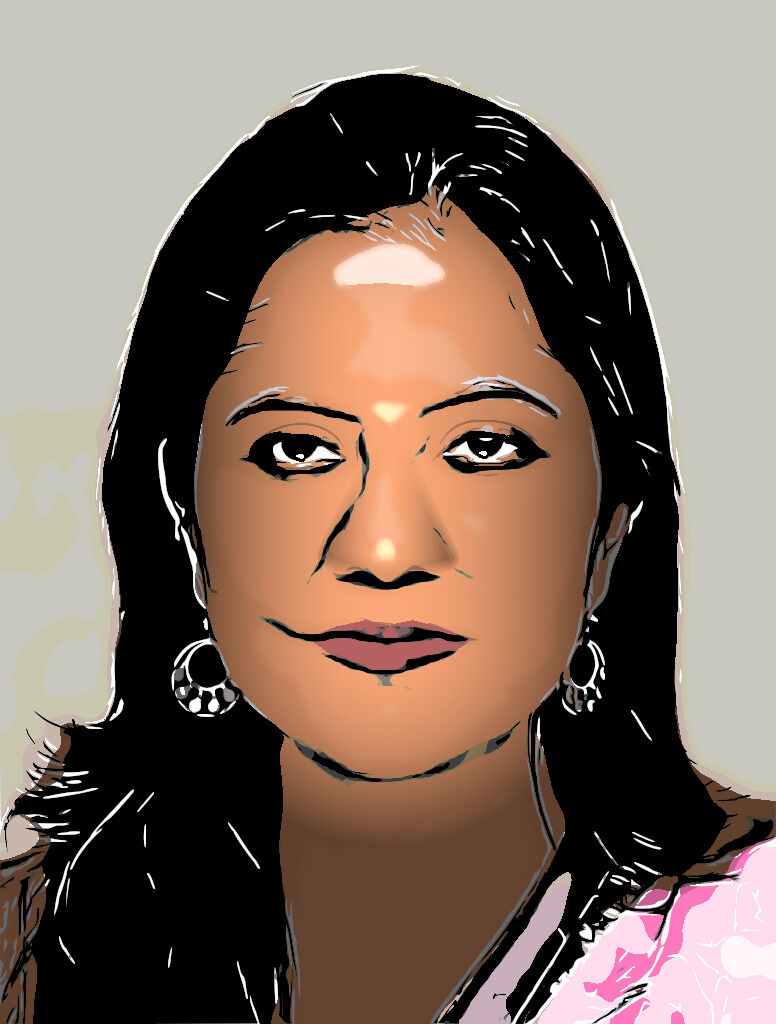}
\includegraphics[height=1.6cm]{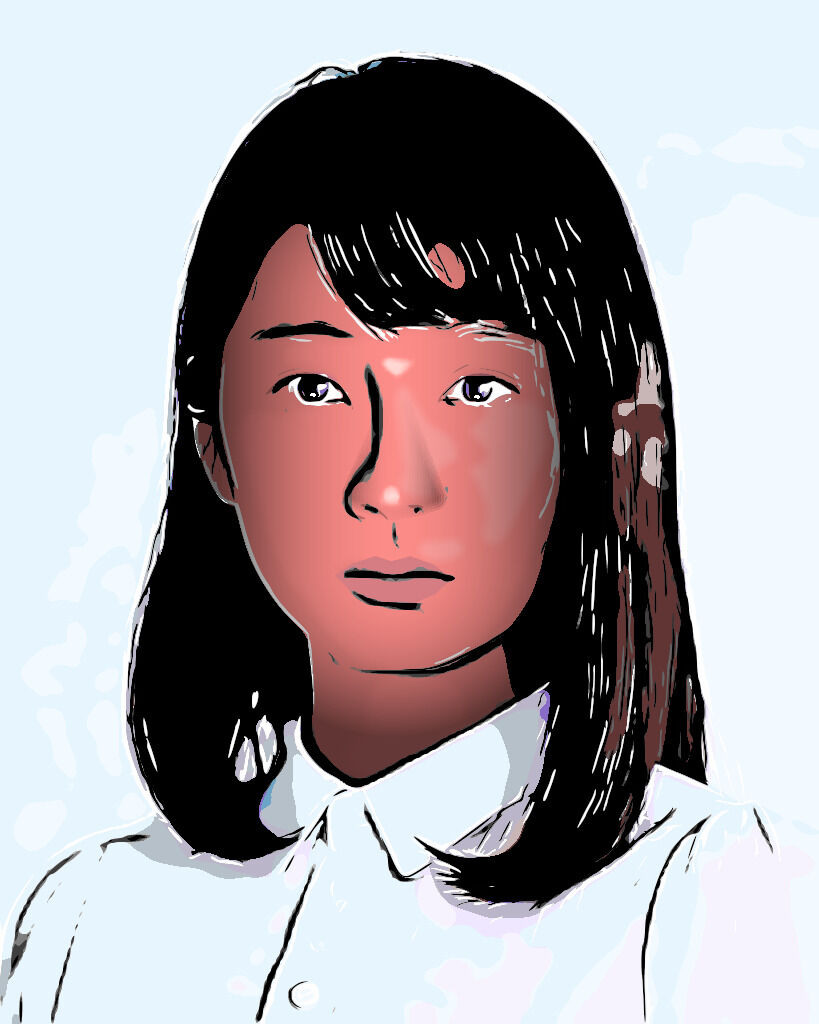}
\includegraphics[height=1.6cm]{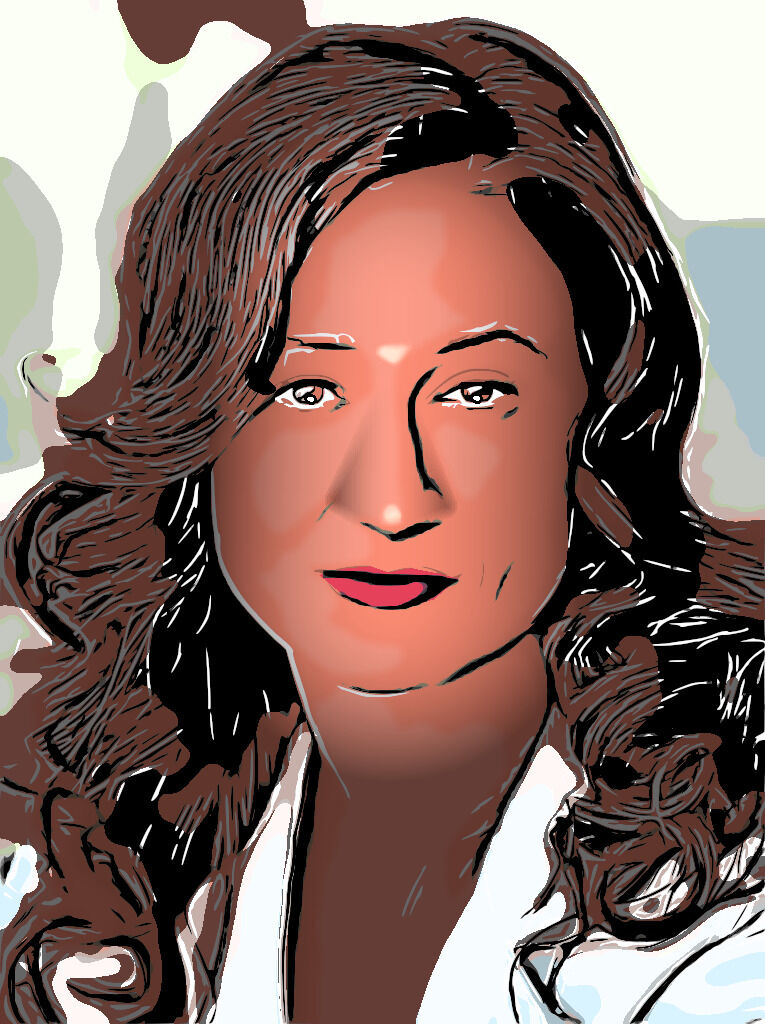}

\centerline{Level 1}
\medskip

\includegraphics[height=1.6cm]{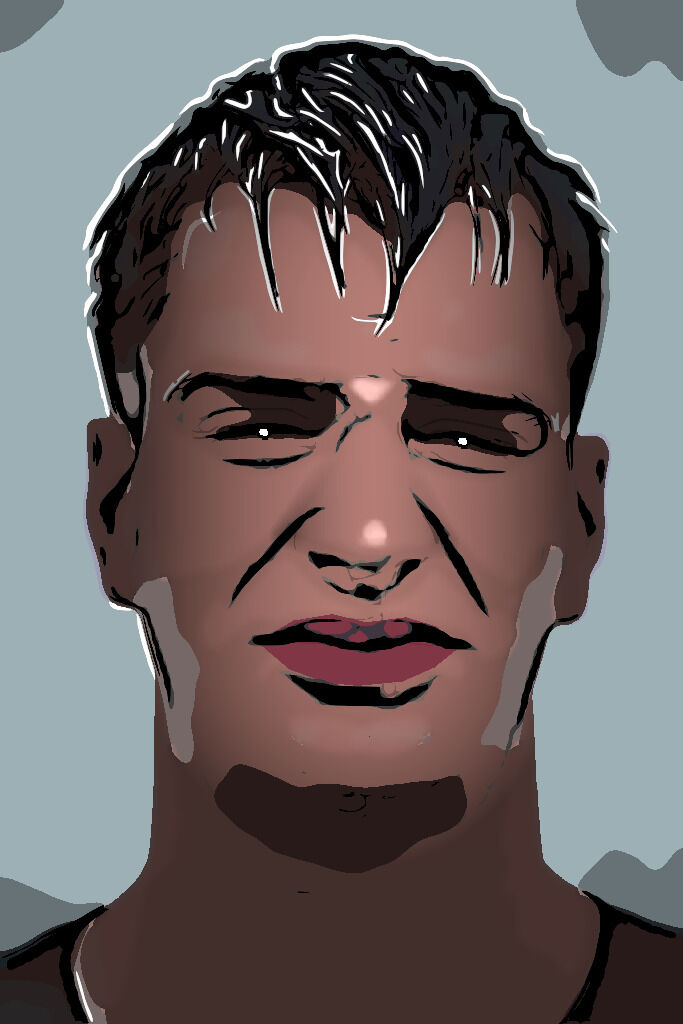}
\includegraphics[height=1.6cm]{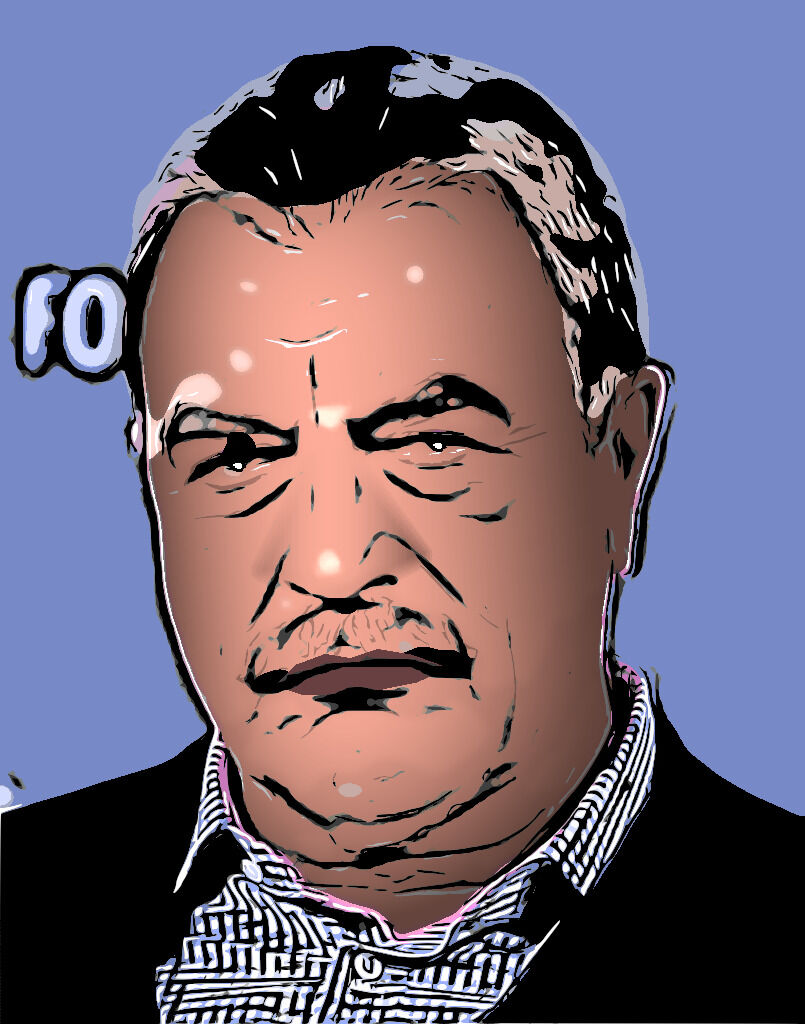}
\includegraphics[height=1.6cm]{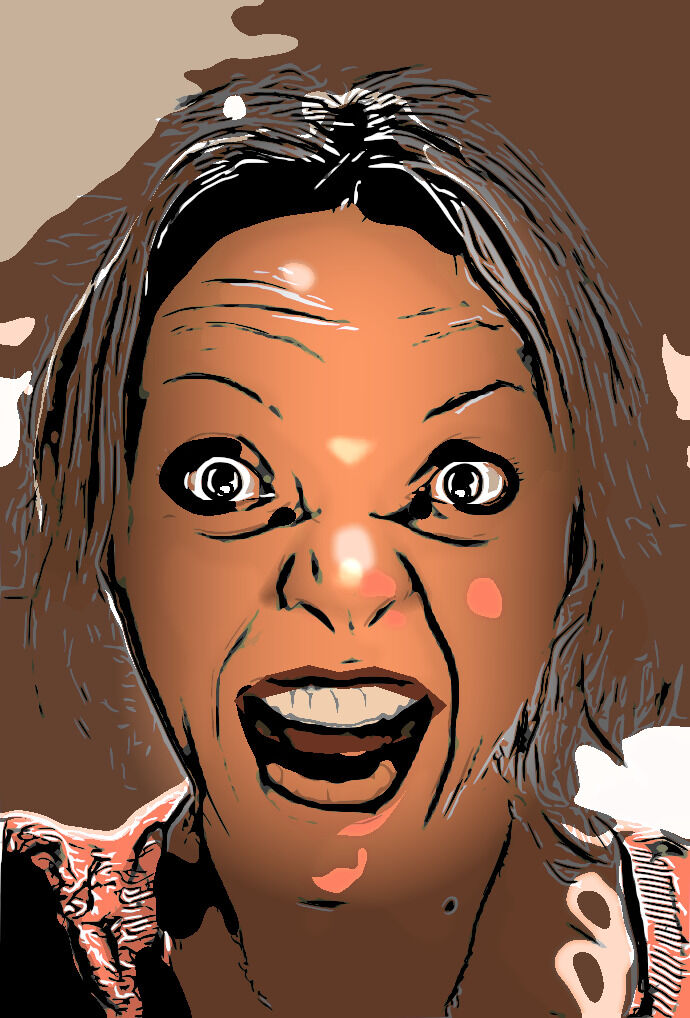}
\includegraphics[height=1.6cm]{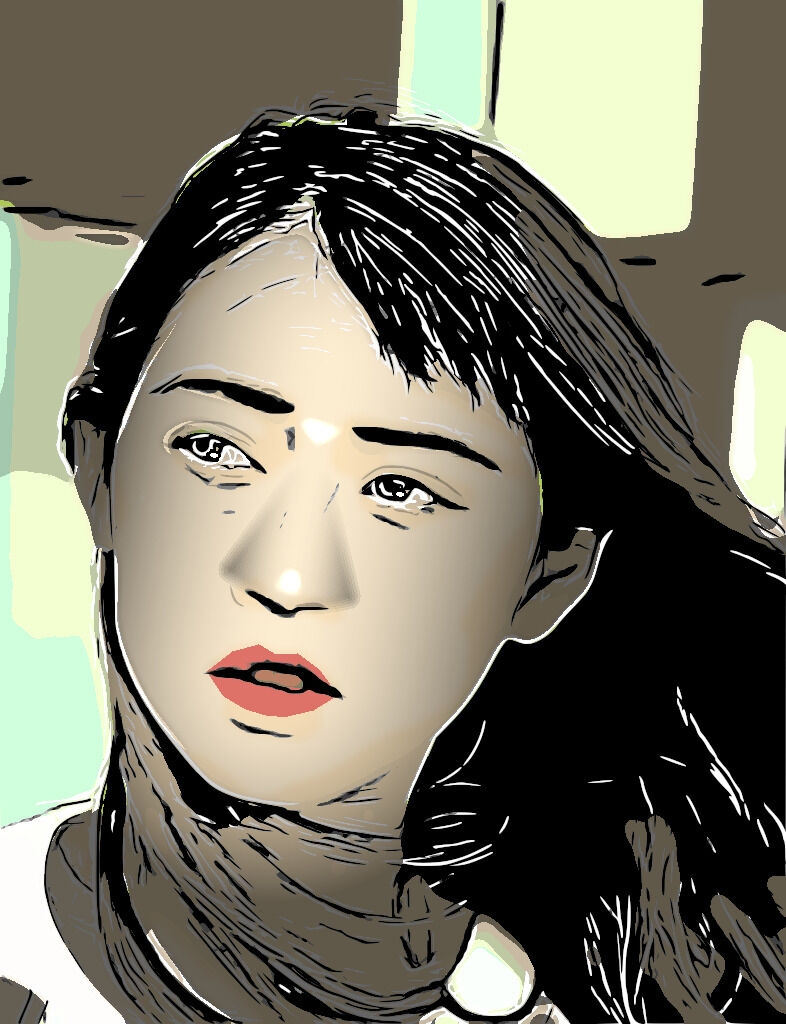}
\includegraphics[height=1.6cm]{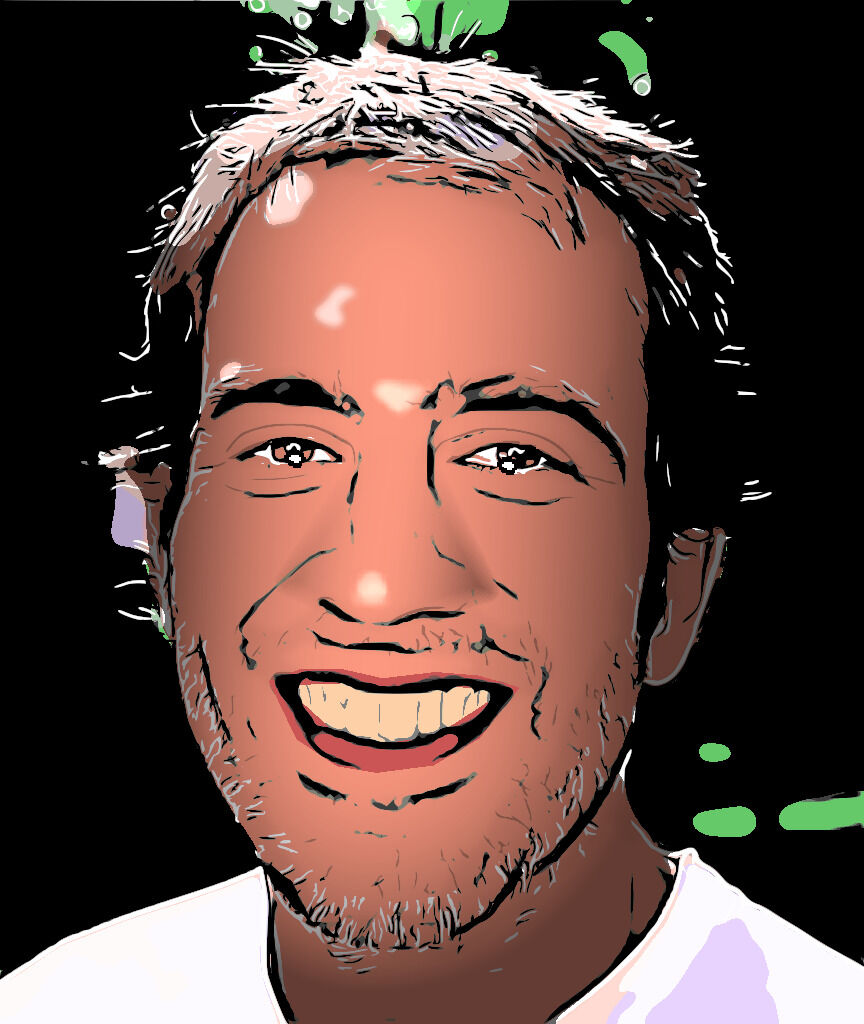}

\centerline{Level 2}
\medskip

\includegraphics[height=1.6cm]{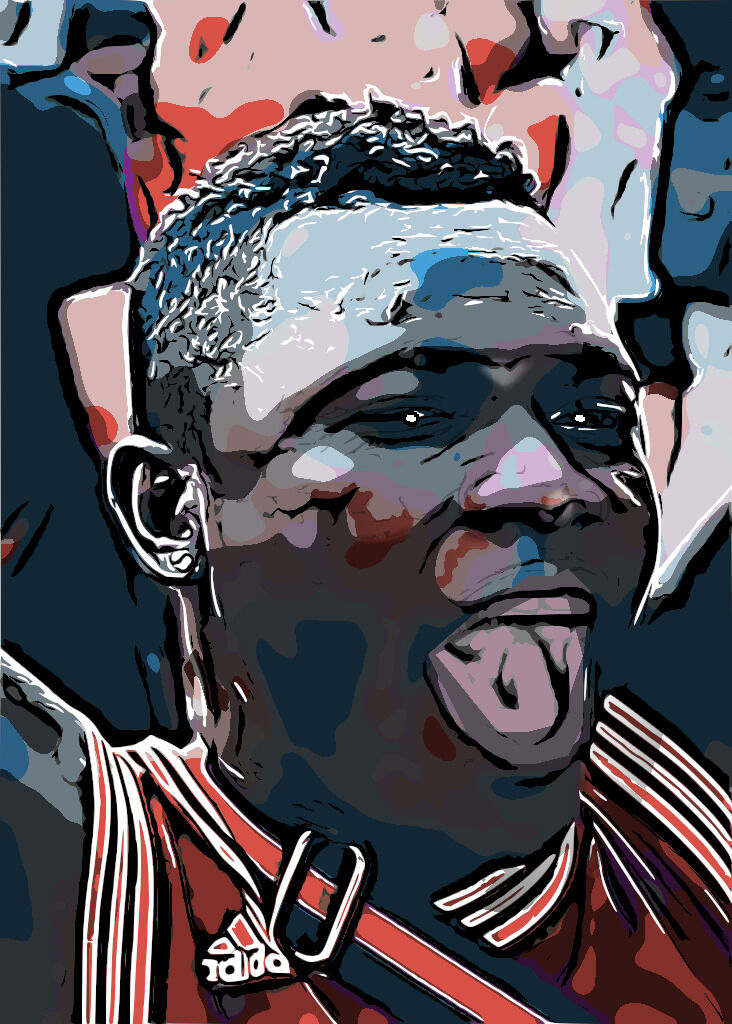}
\includegraphics[height=1.6cm]{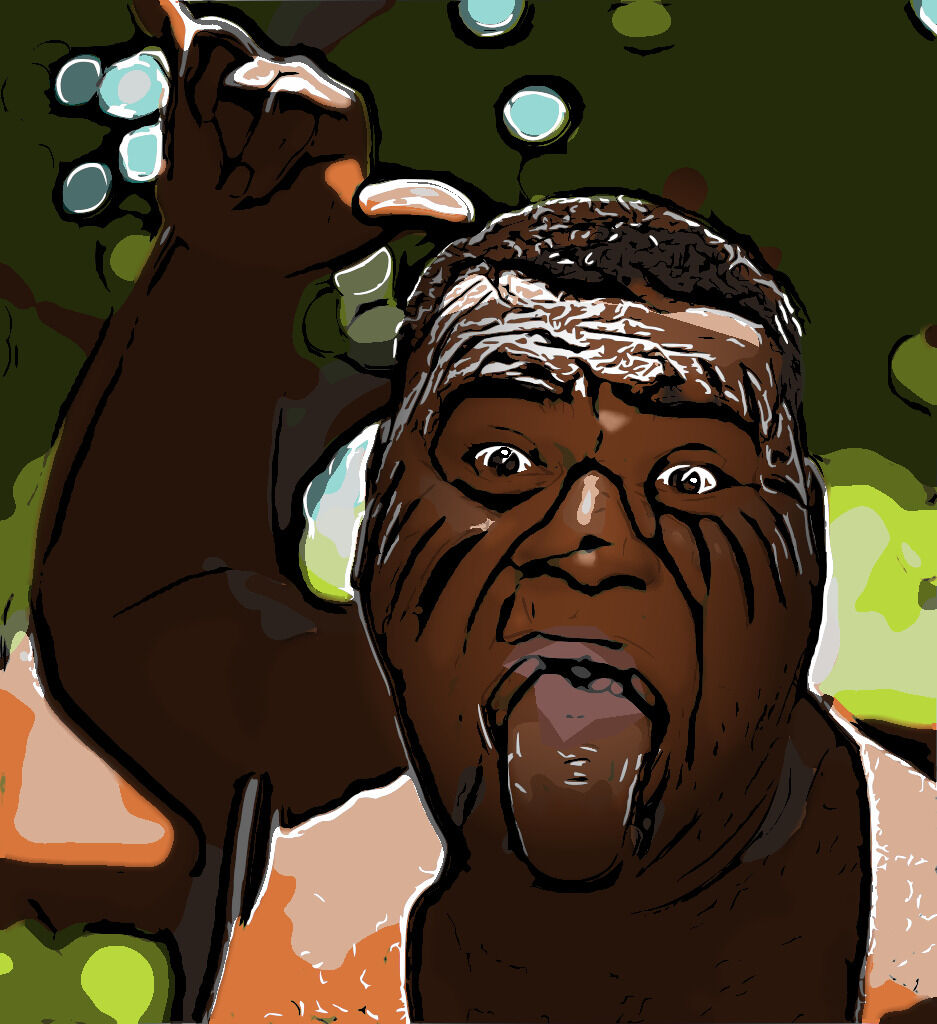}
\includegraphics[height=1.6cm]{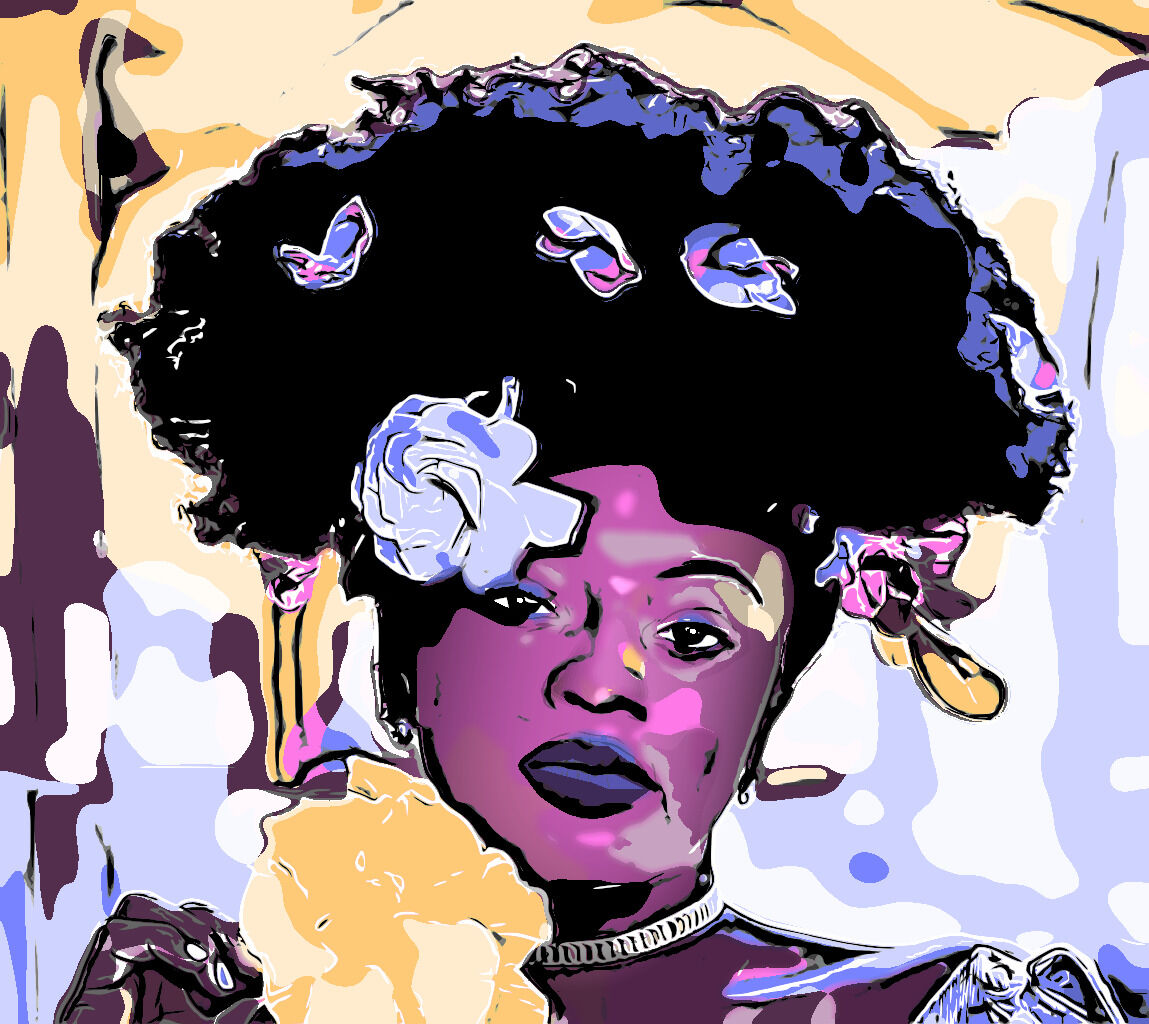}
\includegraphics[height=1.6cm]{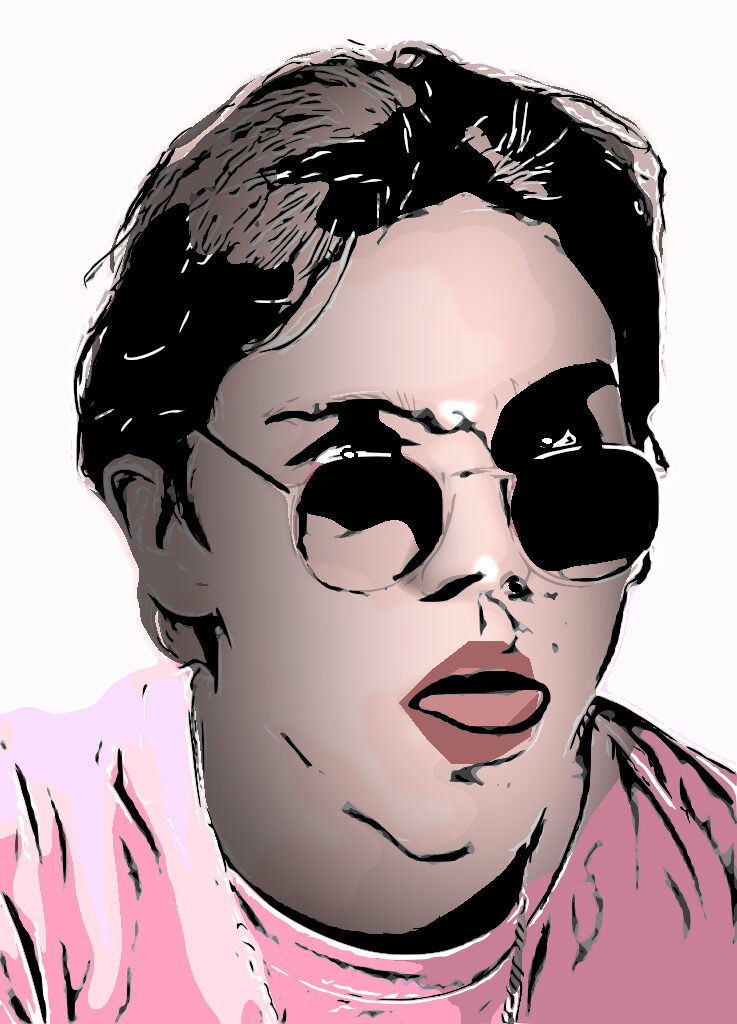}
\includegraphics[height=1.6cm]{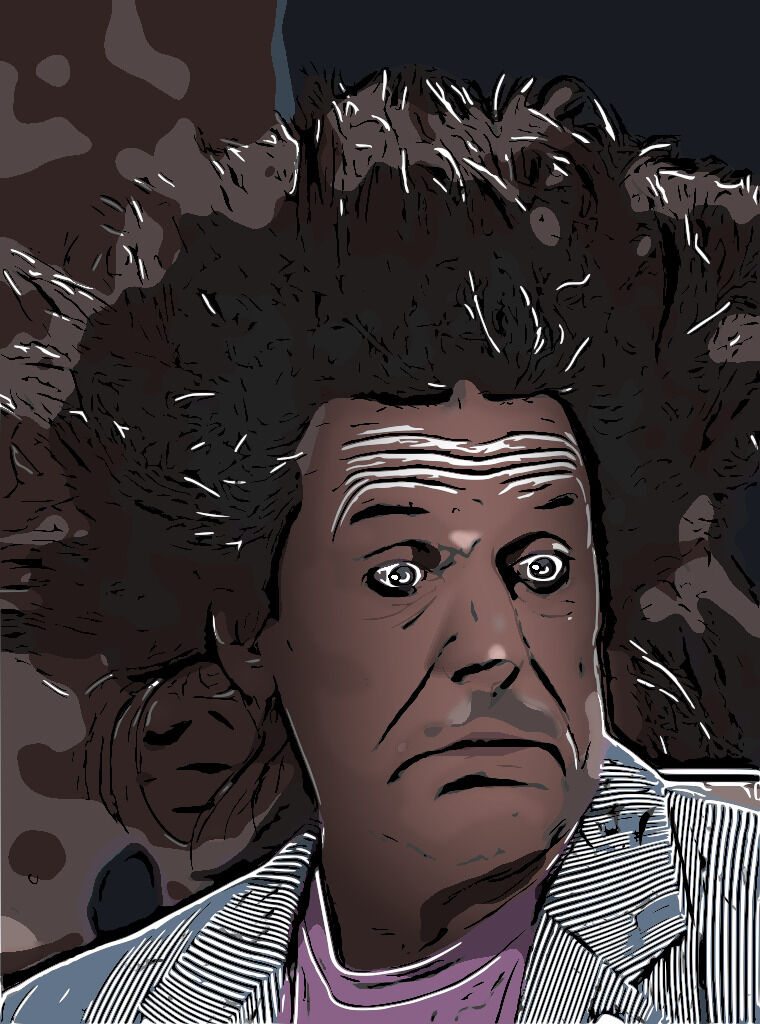}
\includegraphics[height=1.6cm]{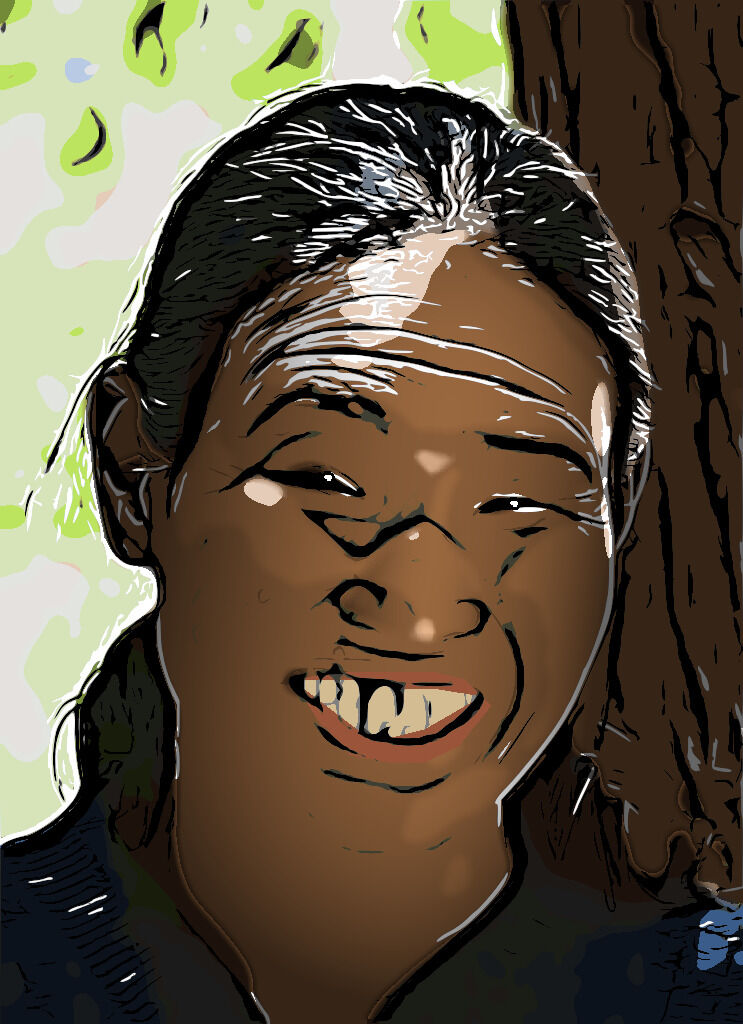}

\centerline{Level 3}
\medskip
\caption{Images from the \emph{NPRportrait1.0} benchmark stylised as puppets: Rosin and Lai~\cite{rosin-portrait}}
\label{resultsRosinLai}
\end{figure}


\begin{figure}[!t]
\centering
\includegraphics[height=1.6cm]{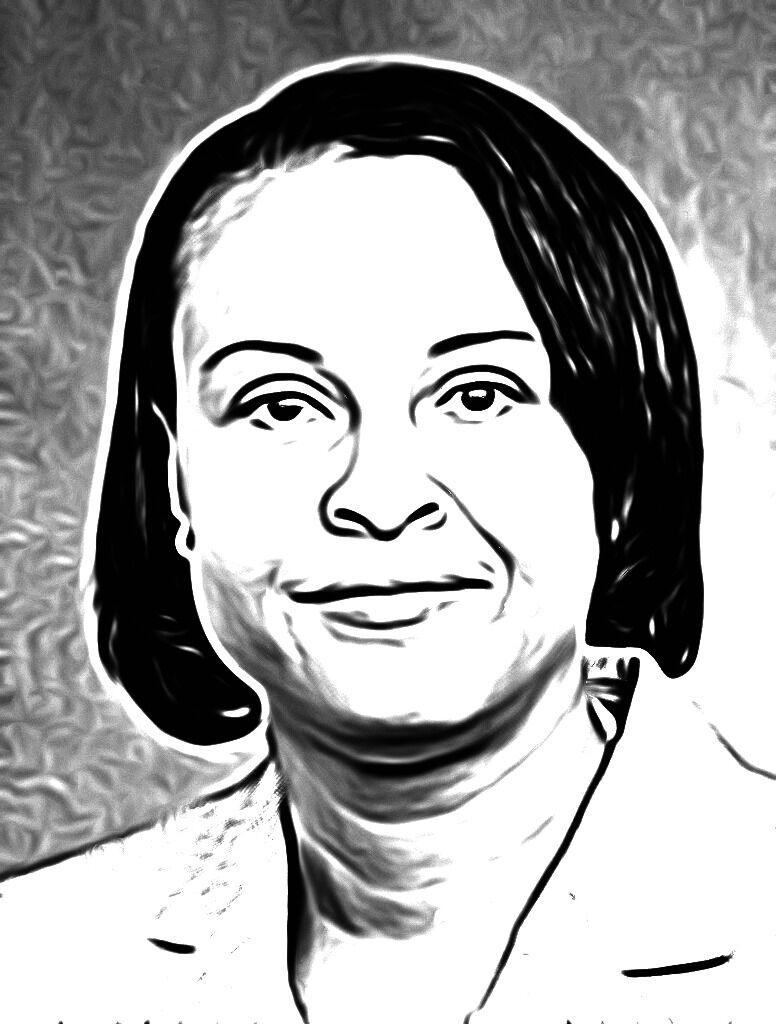}
\includegraphics[height=1.6cm]{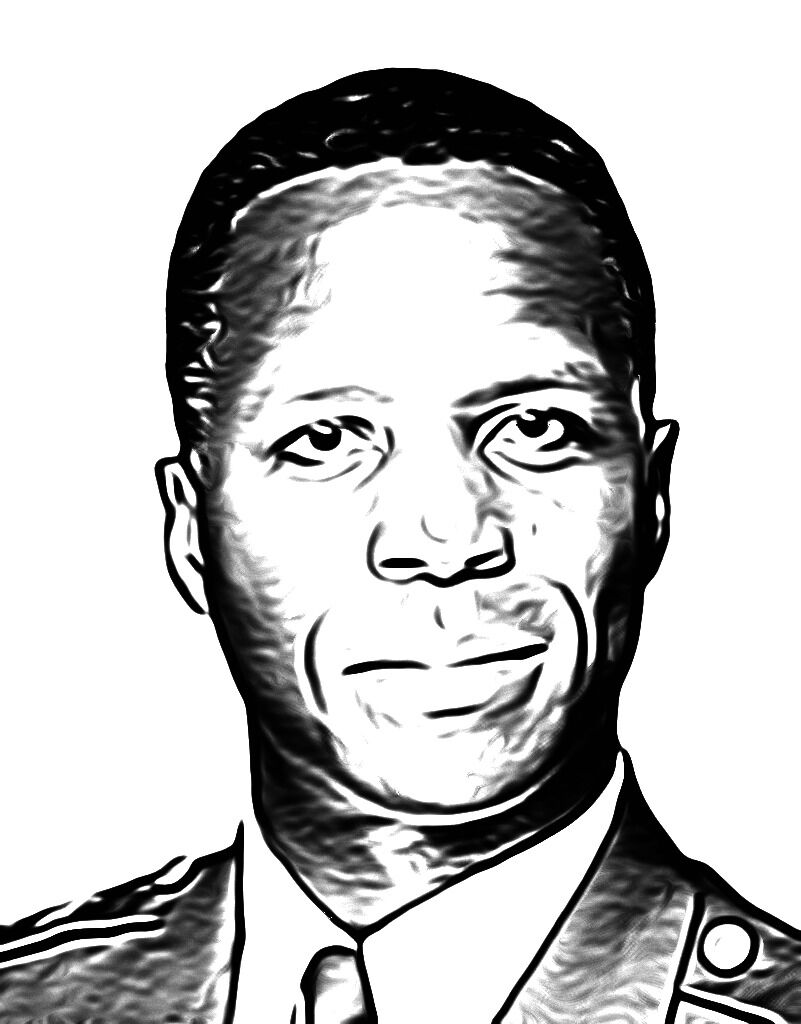}
\includegraphics[height=1.6cm]{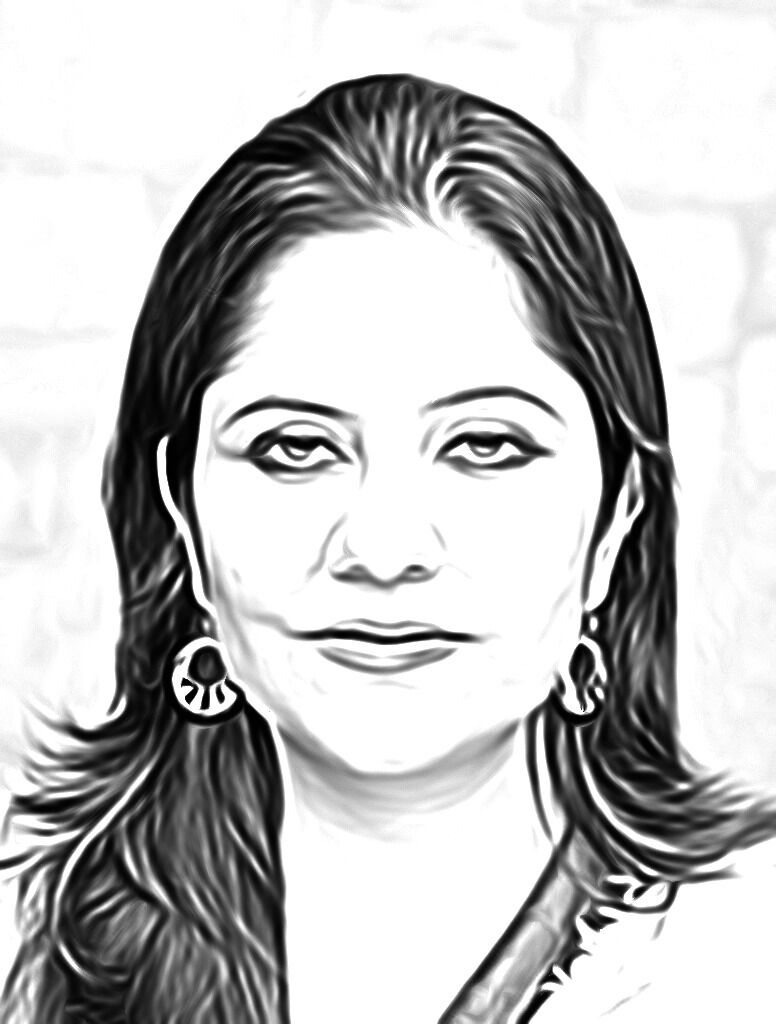}
\includegraphics[height=1.6cm]{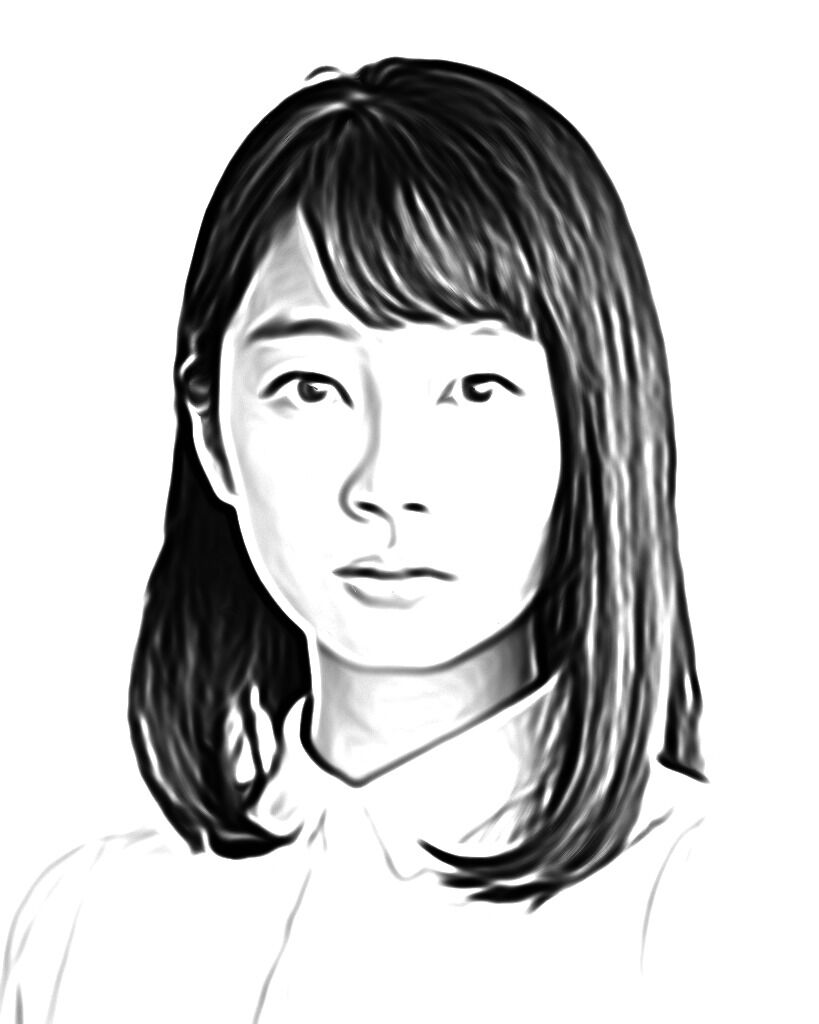}
\includegraphics[height=1.6cm]{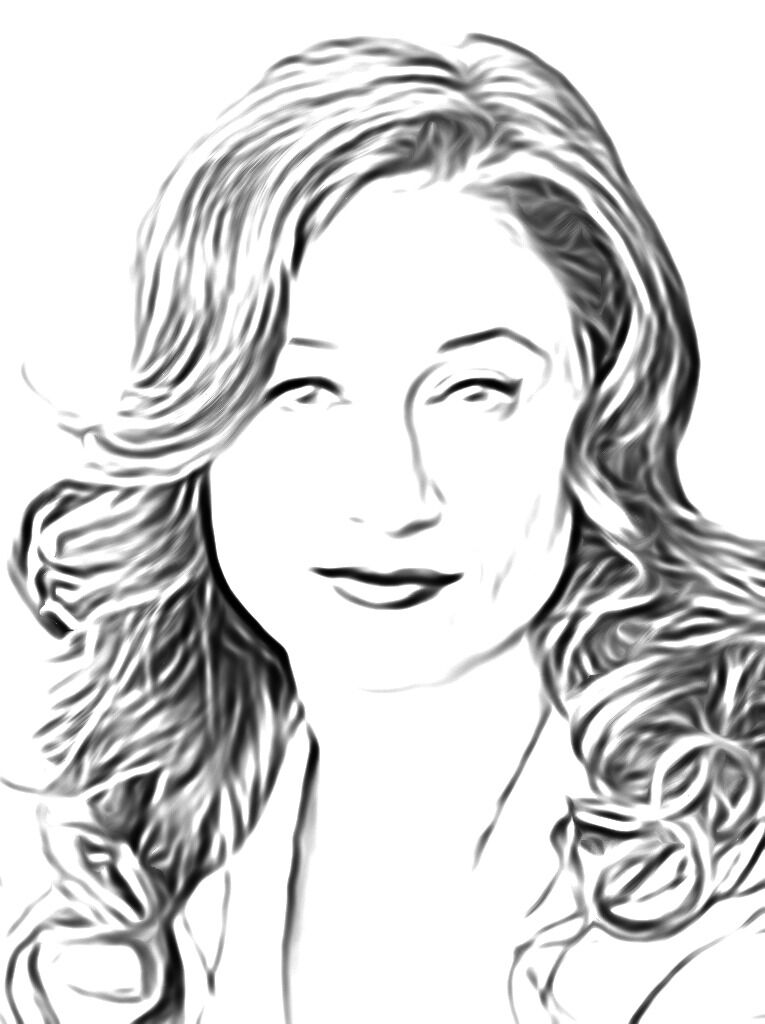}

\centerline{Level 1}
\medskip

\includegraphics[height=1.6cm]{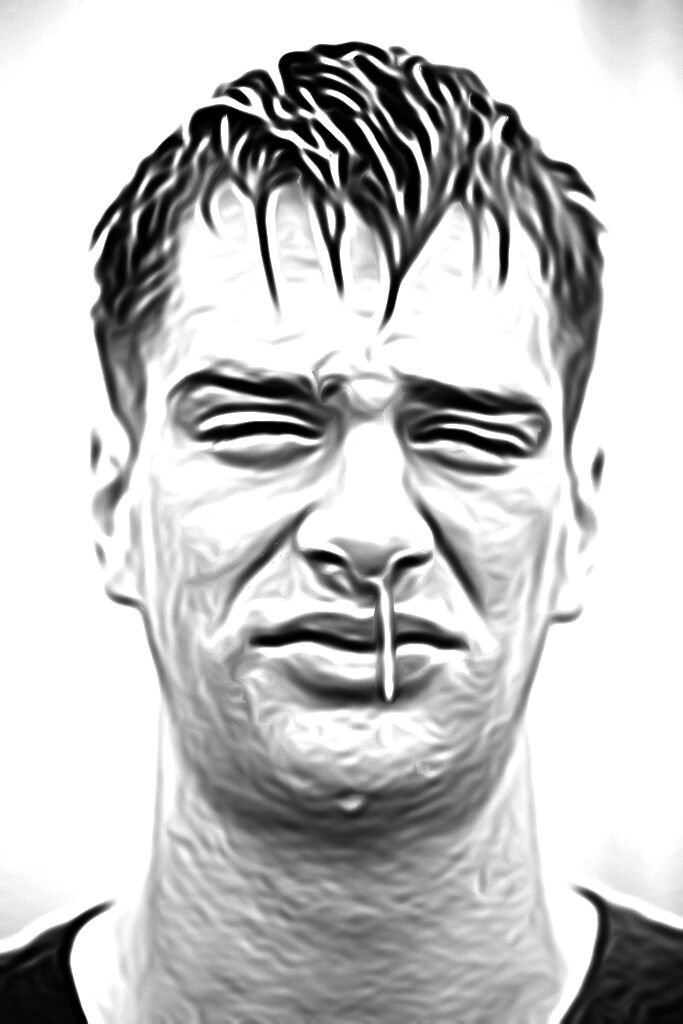}
\includegraphics[height=1.6cm]{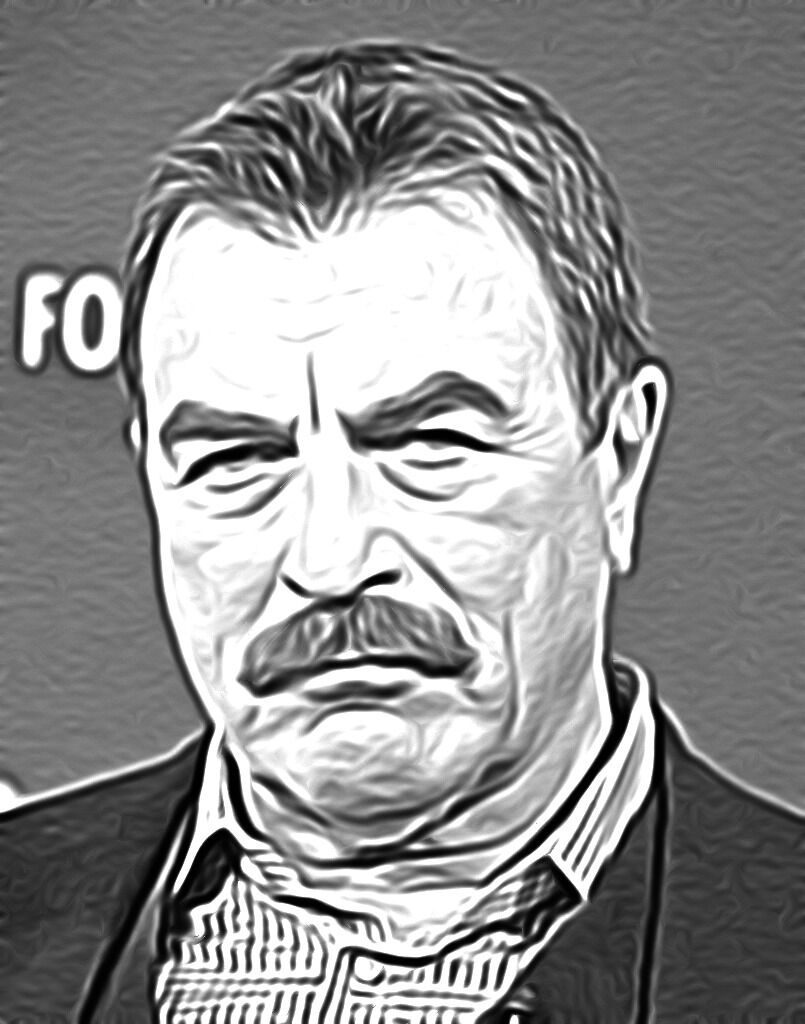}
\includegraphics[height=1.6cm]{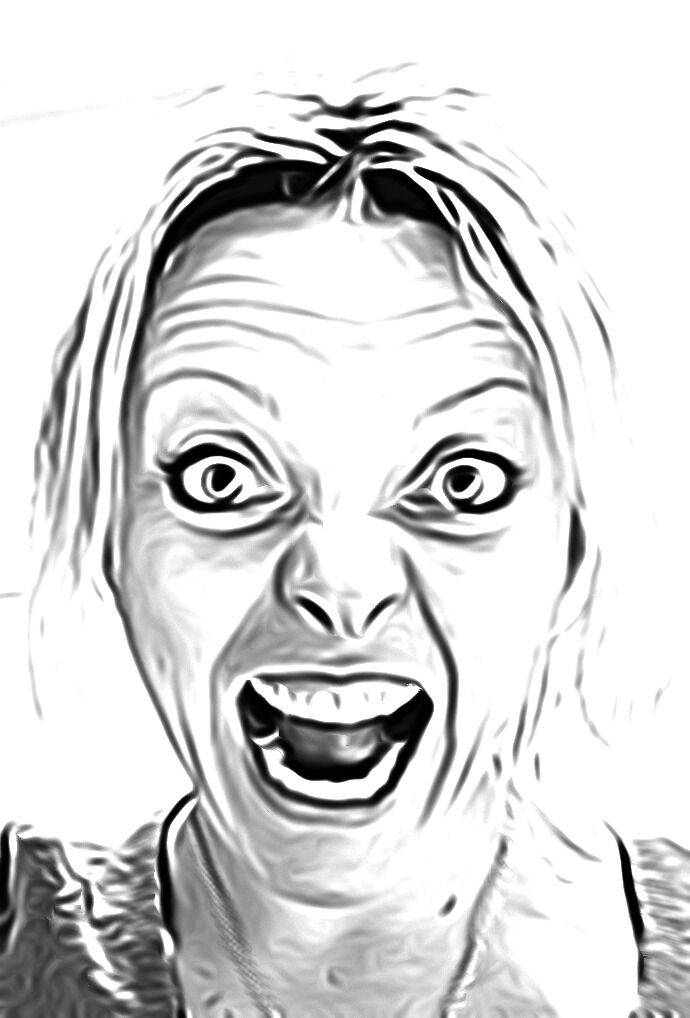}
\includegraphics[height=1.6cm]{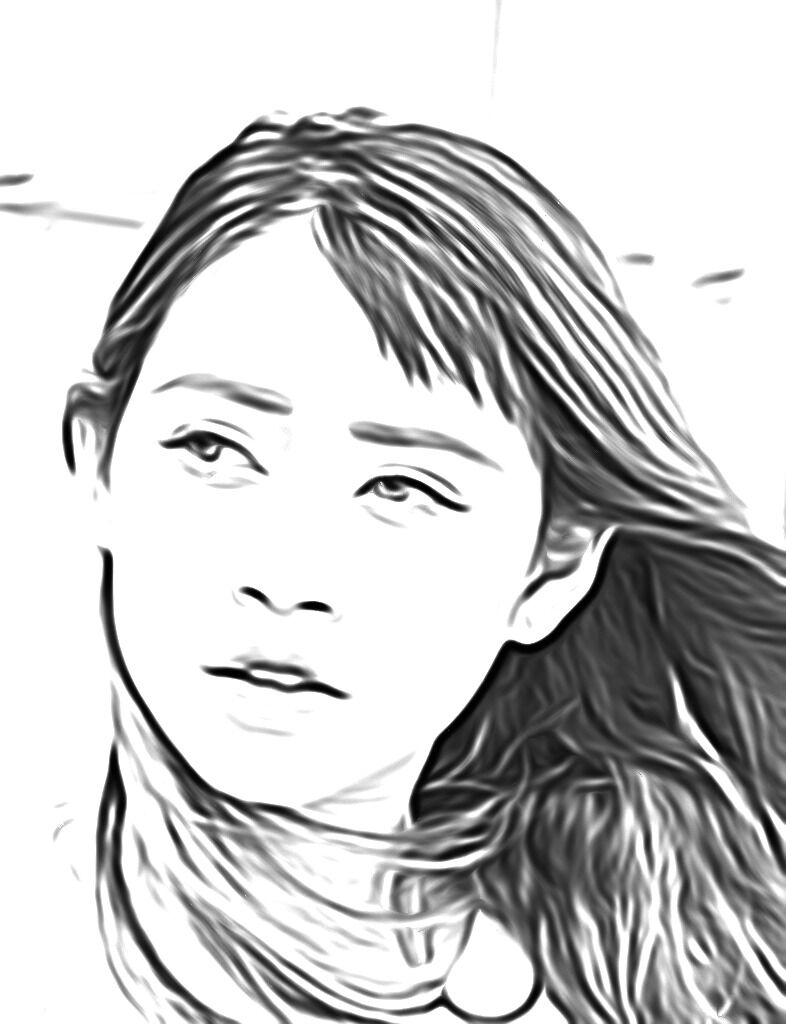}
\includegraphics[height=1.6cm]{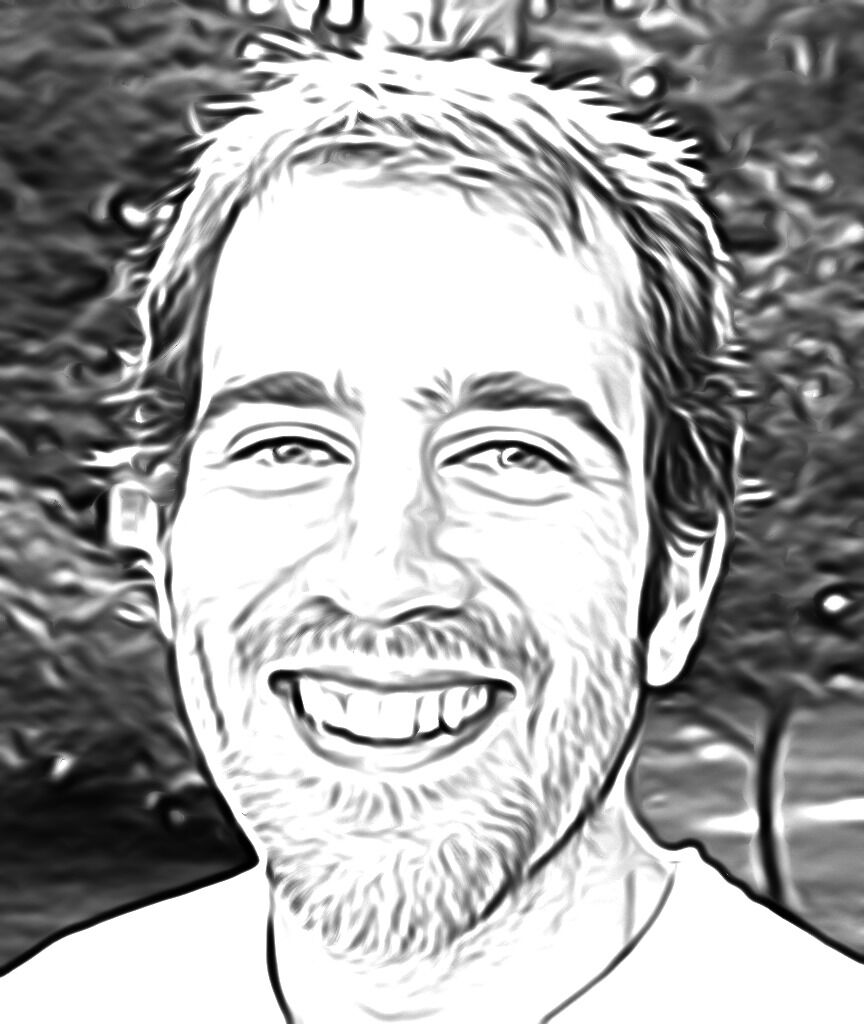}

\centerline{Level 2}
\medskip

\includegraphics[height=1.6cm]{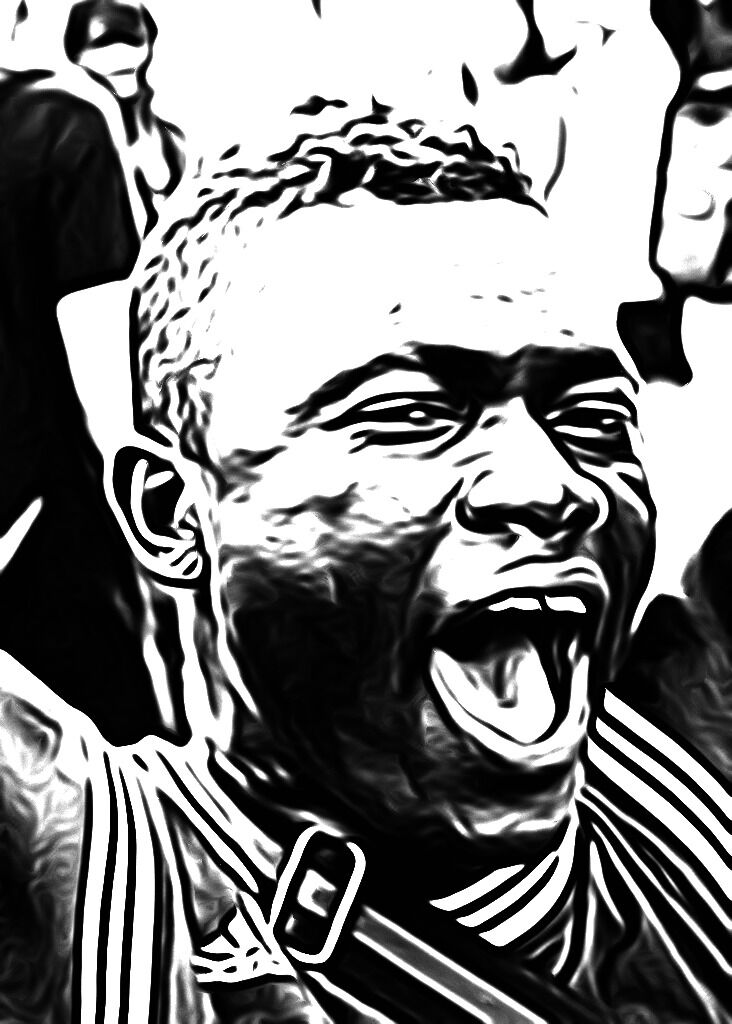}
\includegraphics[height=1.6cm]{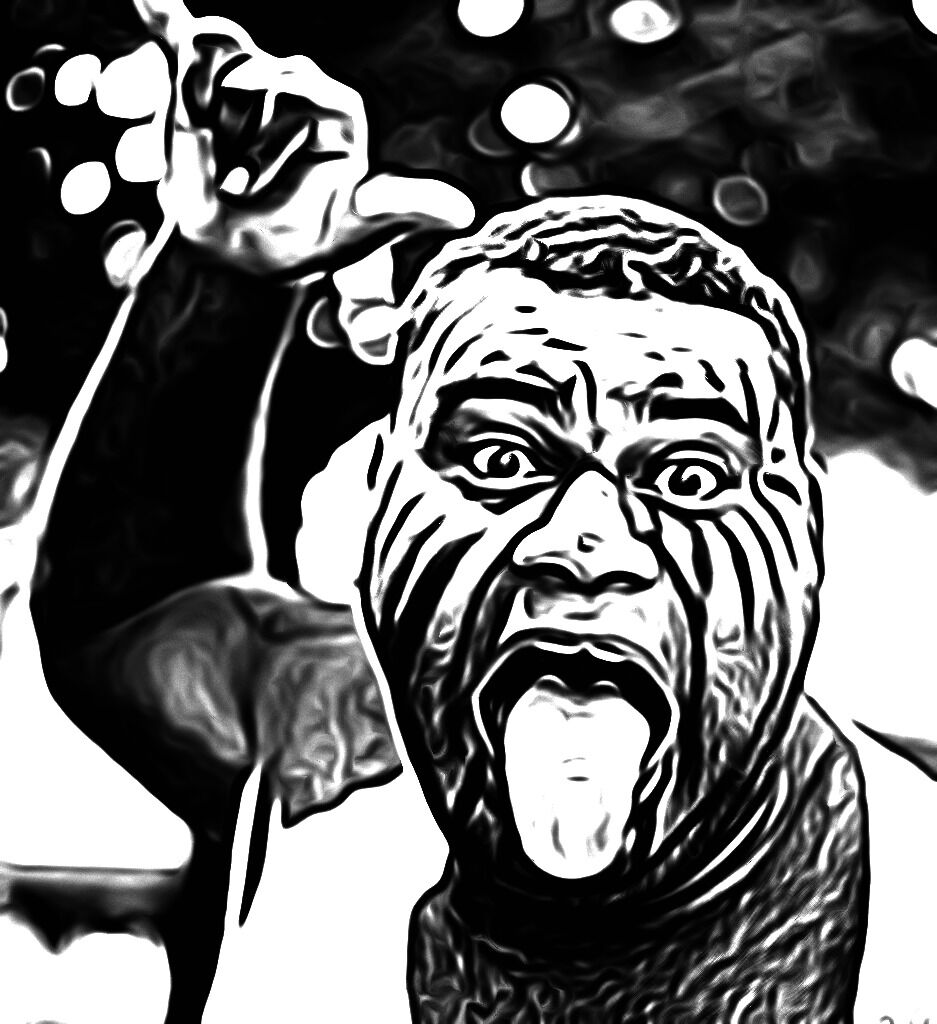}
\includegraphics[height=1.6cm]{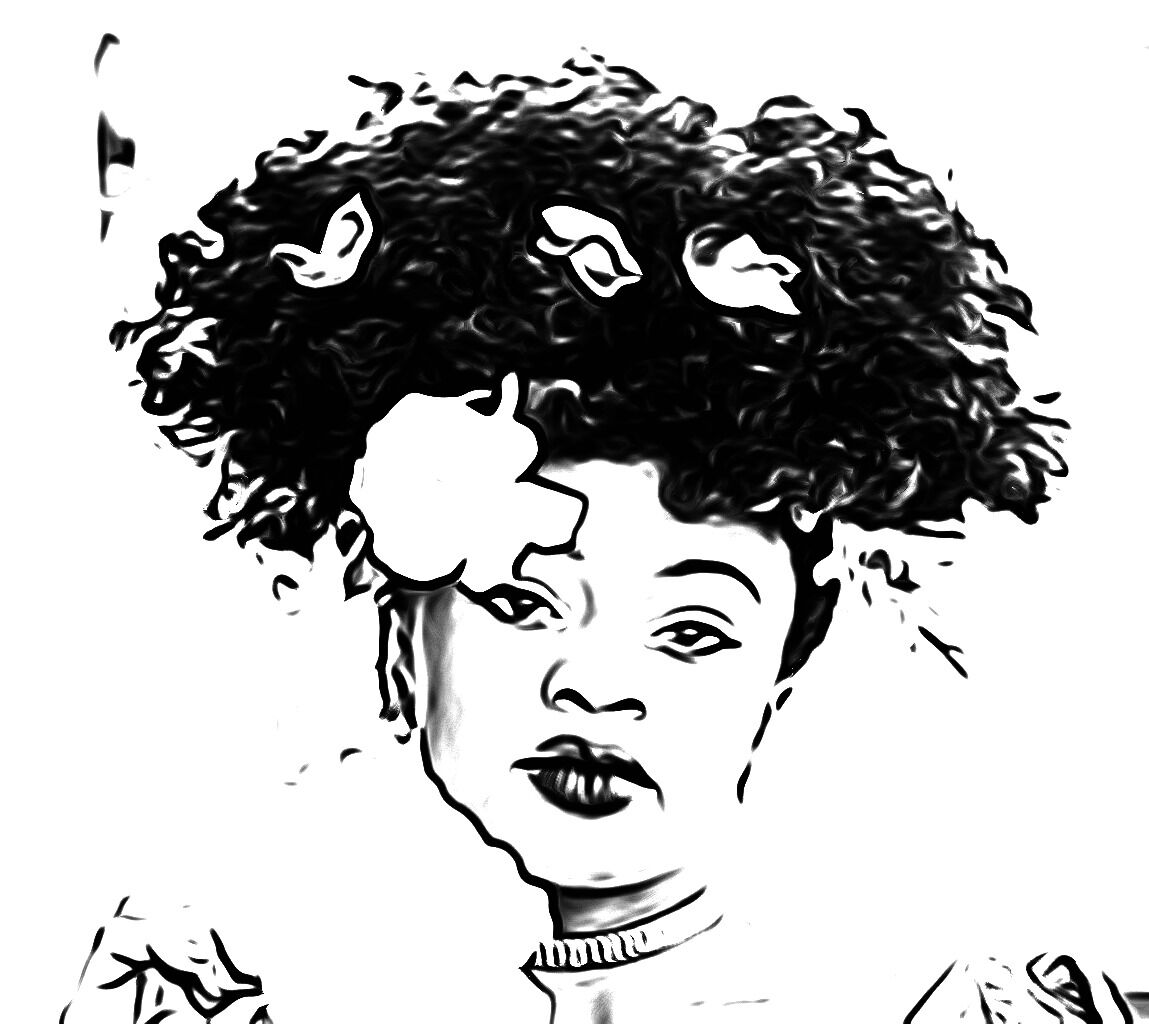}
\includegraphics[height=1.6cm]{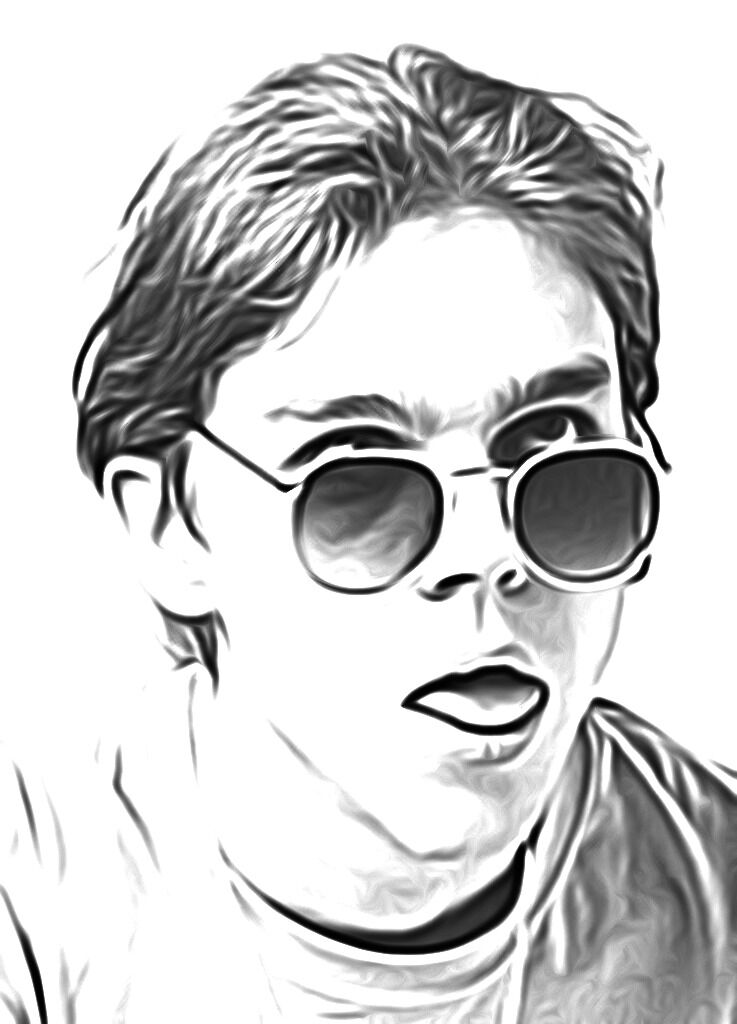}
\includegraphics[height=1.6cm]{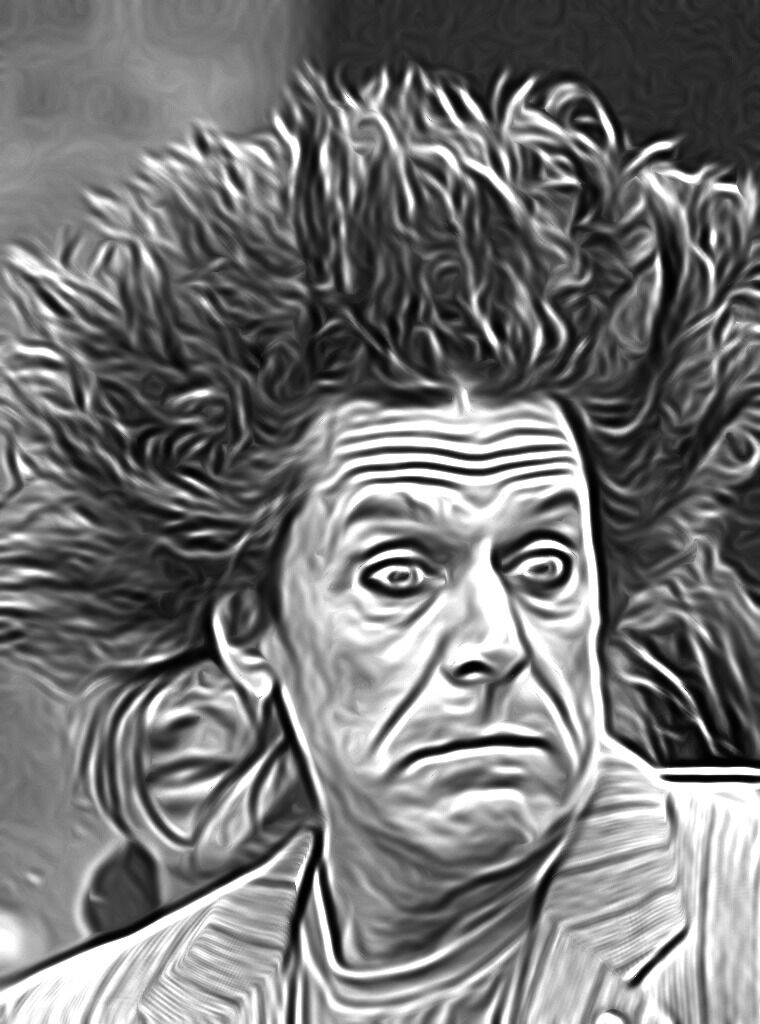}
\includegraphics[height=1.6cm]{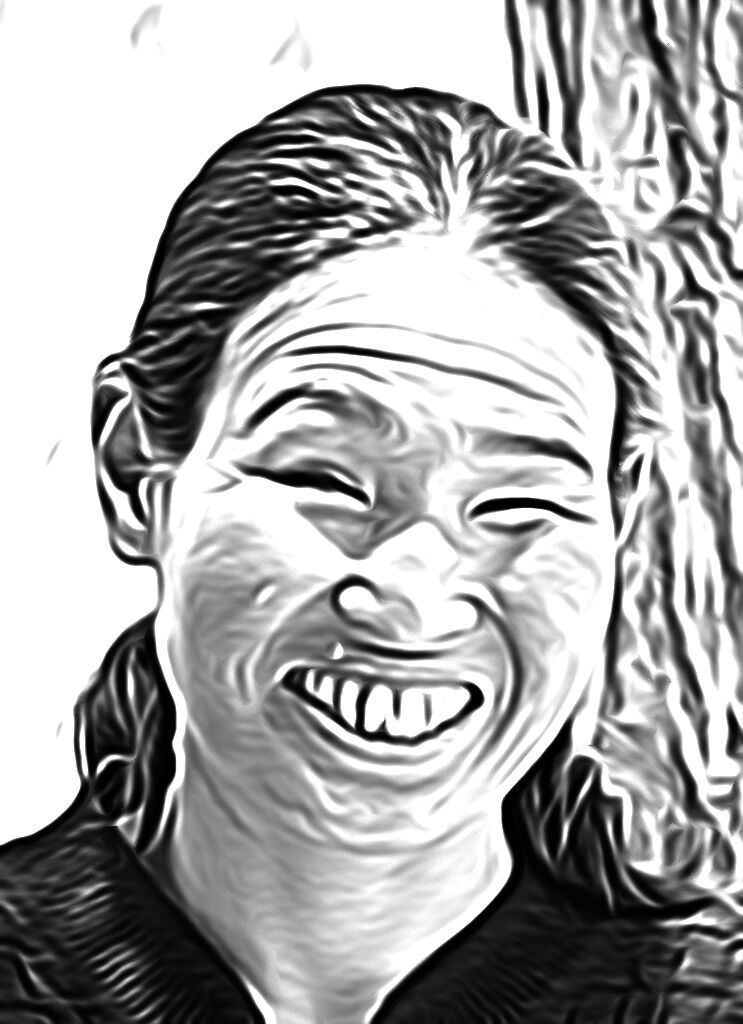}

\centerline{Level 3}
\medskip
\caption{Images from the \emph{NPRportrait1.0} benchmark stylised by XDoG: Winnem{\"oller} \emph{et al.}~\cite{xdog}}
\label{resultsXDoG}
\end{figure}


\begin{figure}[!t]
\centering
\includegraphics[height=1.6cm]{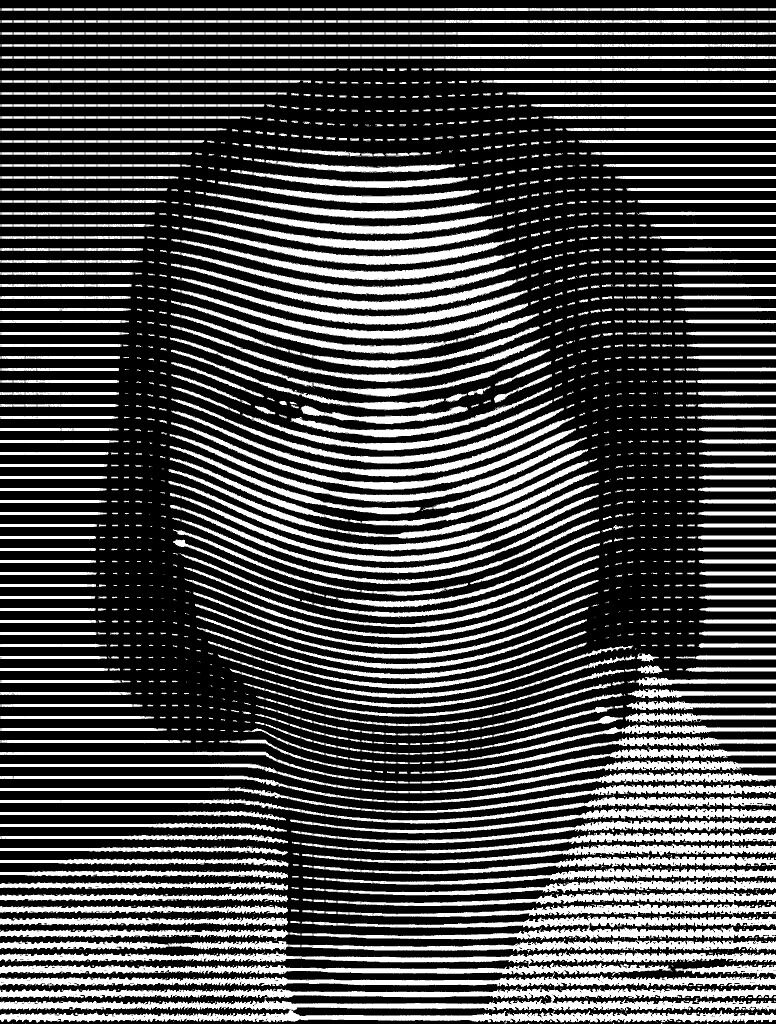}
\includegraphics[height=1.6cm]{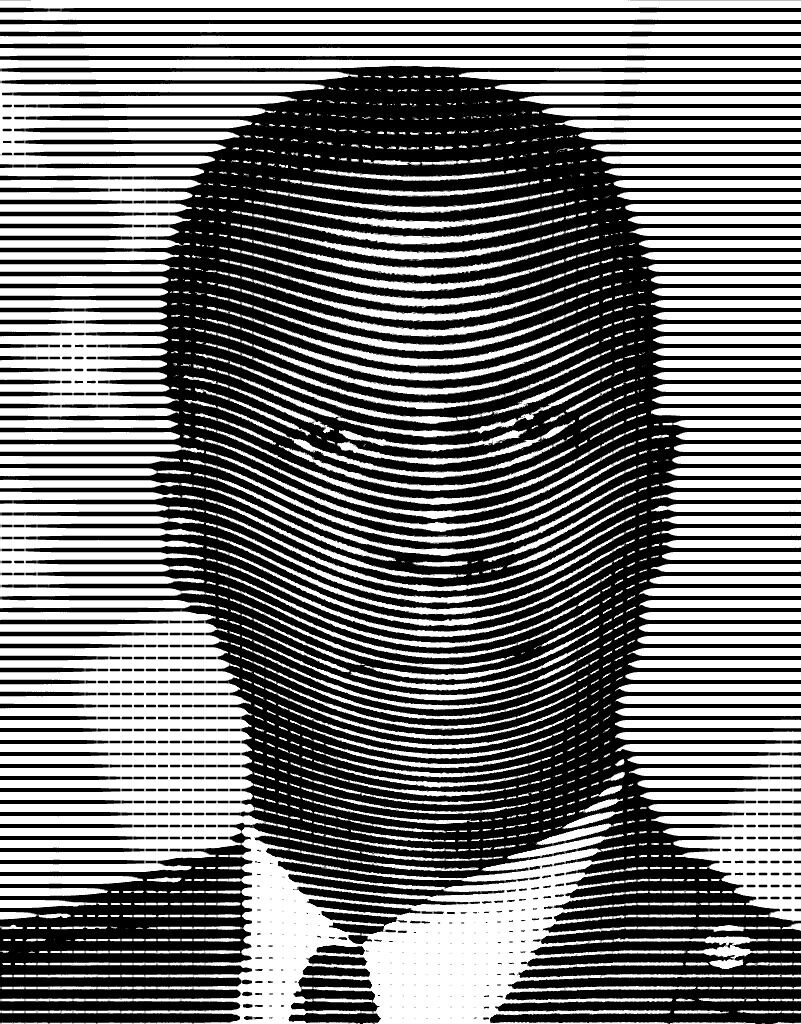}
\includegraphics[height=1.6cm]{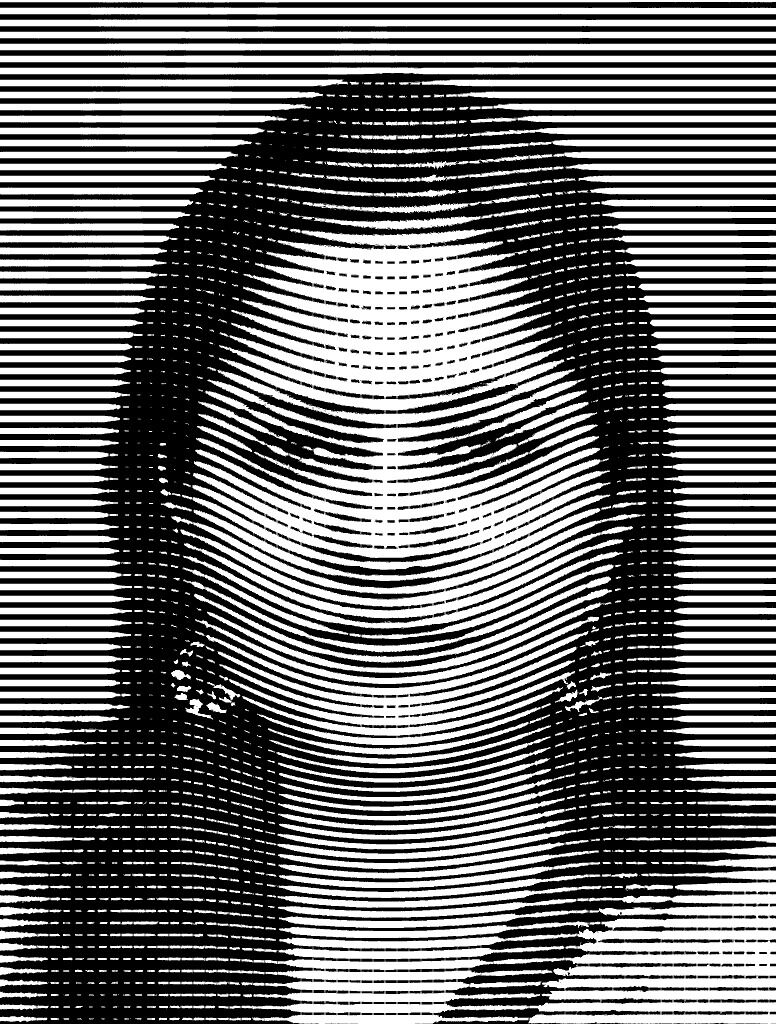}
\includegraphics[height=1.6cm]{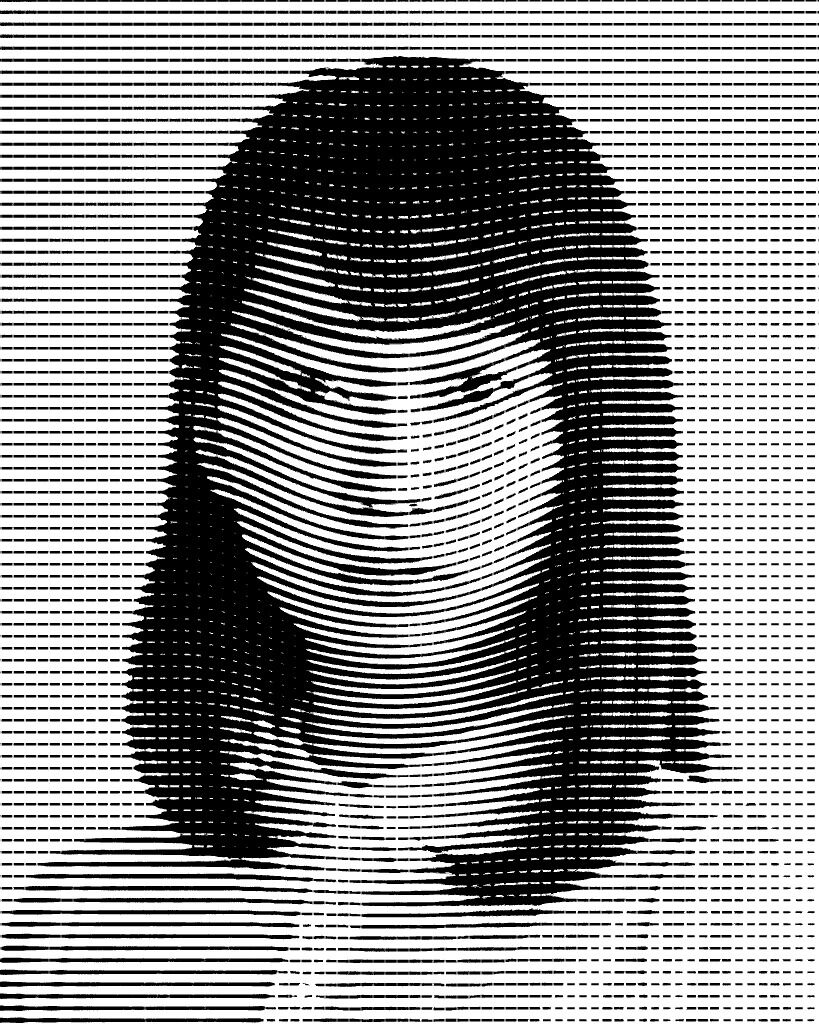}
\includegraphics[height=1.6cm]{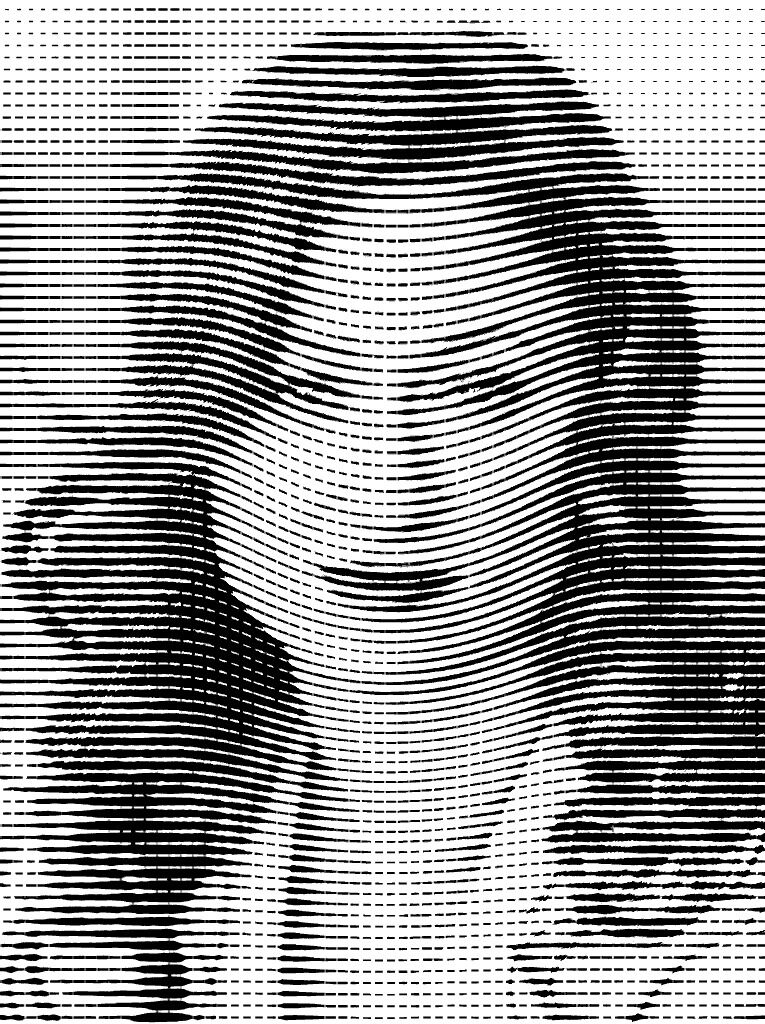}

\centerline{Level 1}
\medskip

\includegraphics[height=1.6cm]{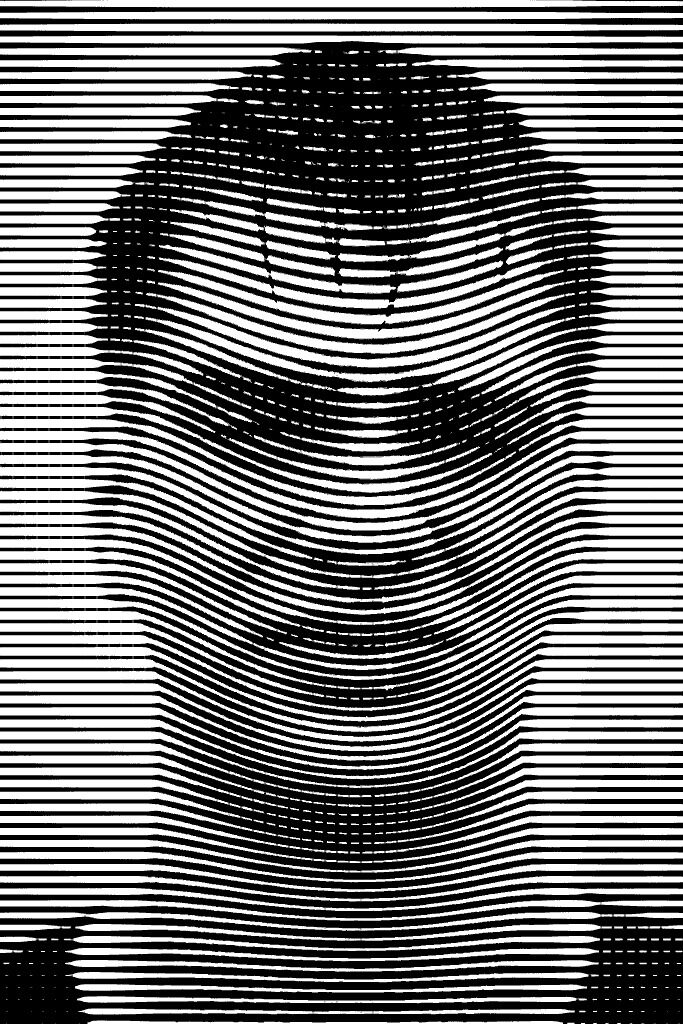}
\includegraphics[height=1.6cm]{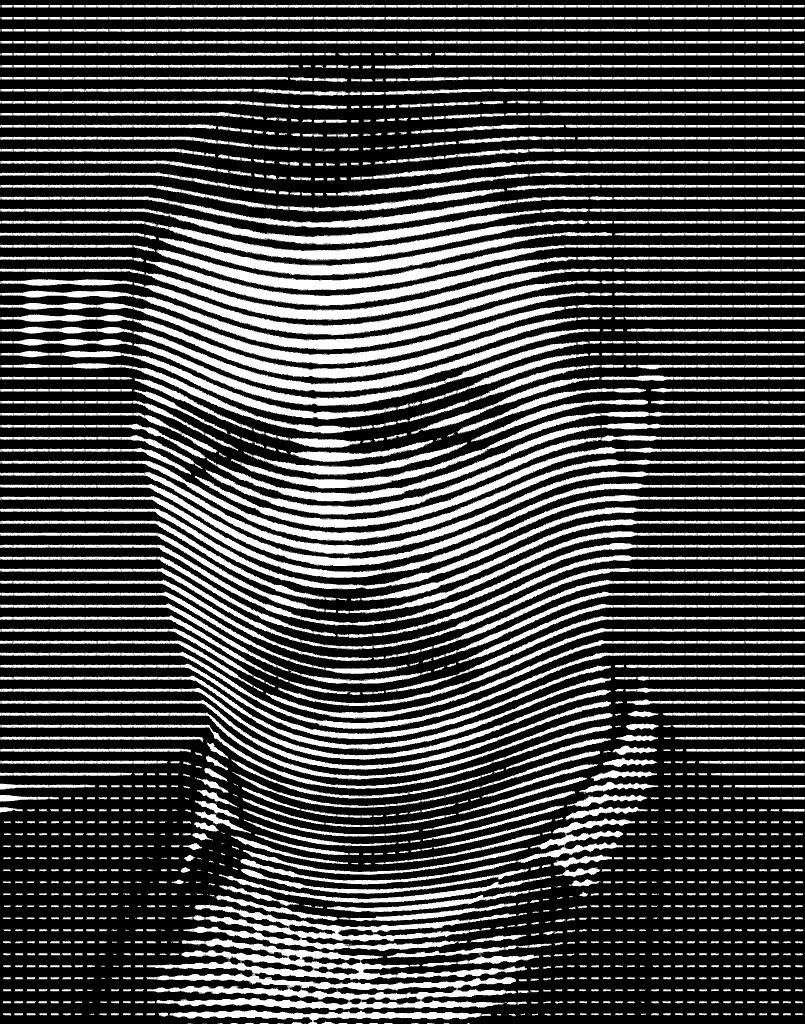}
\includegraphics[height=1.6cm]{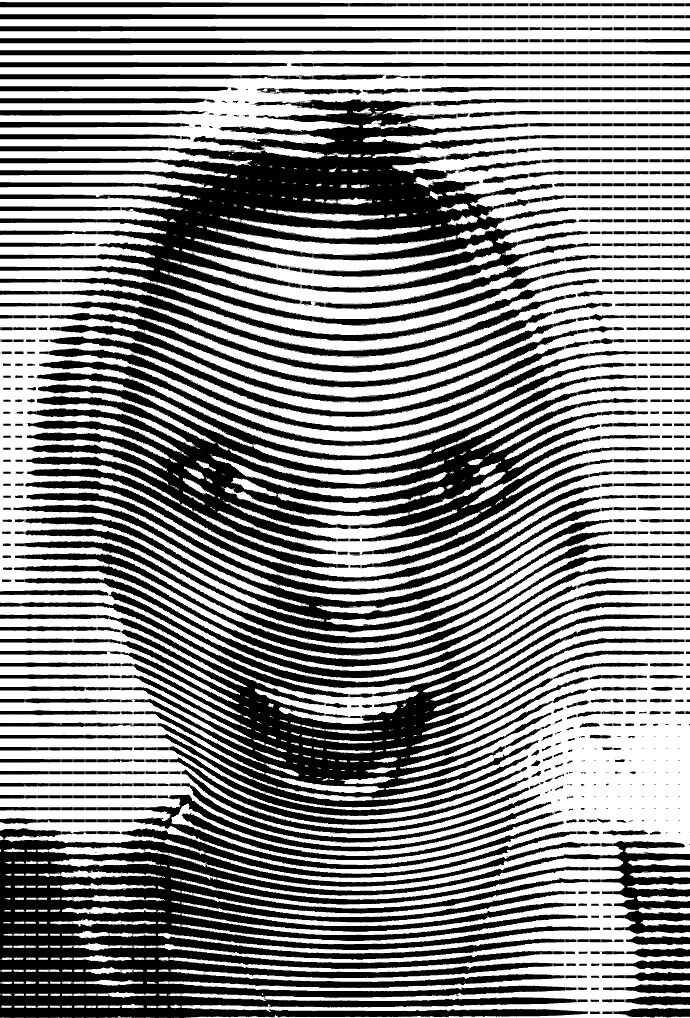}
\includegraphics[height=1.6cm]{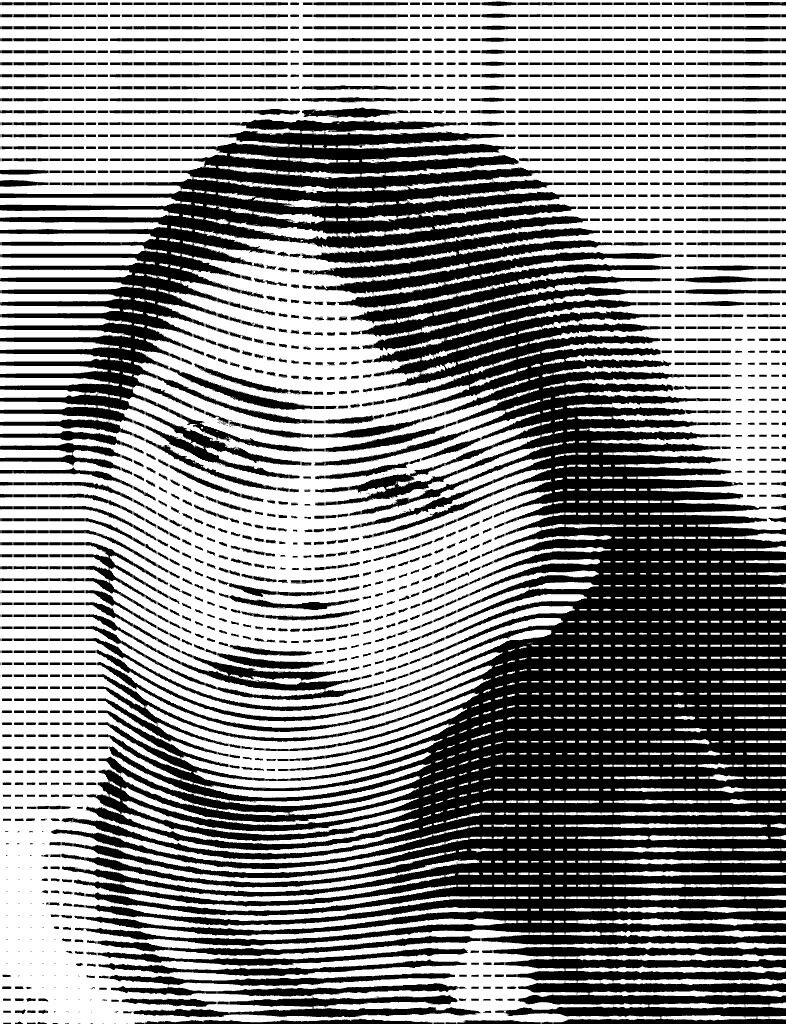}
\includegraphics[height=1.6cm]{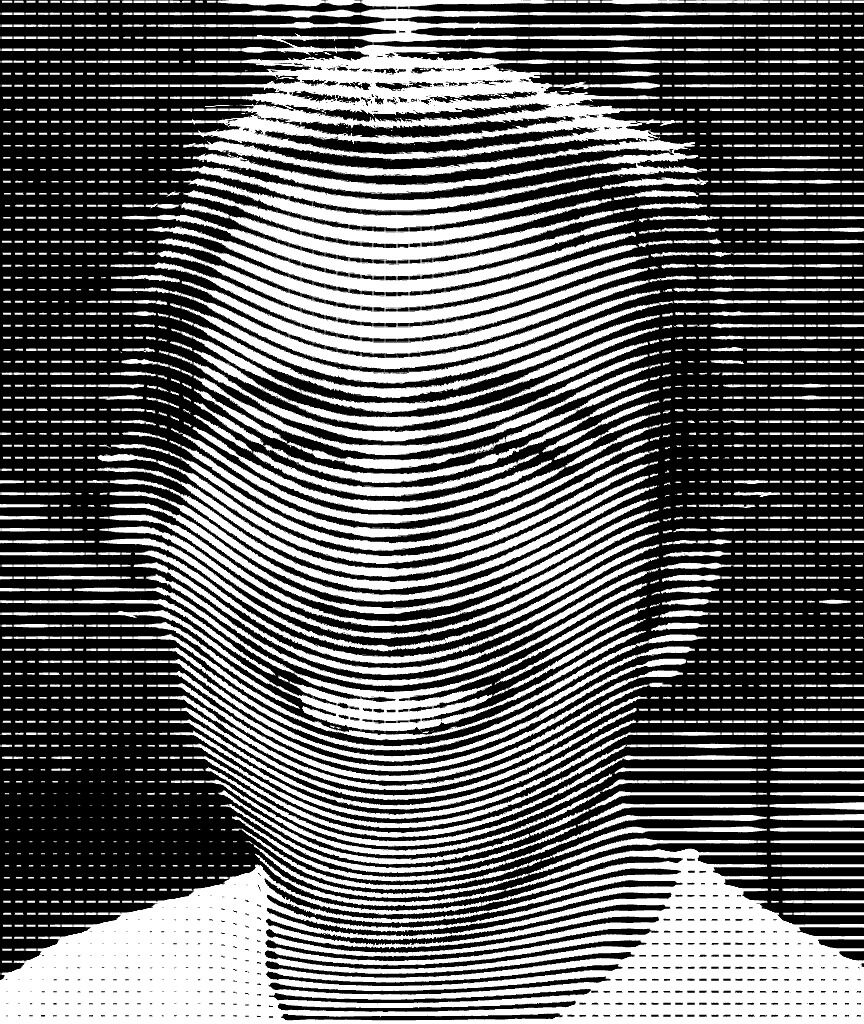}

\centerline{Level 2}
\medskip

\includegraphics[height=1.6cm]{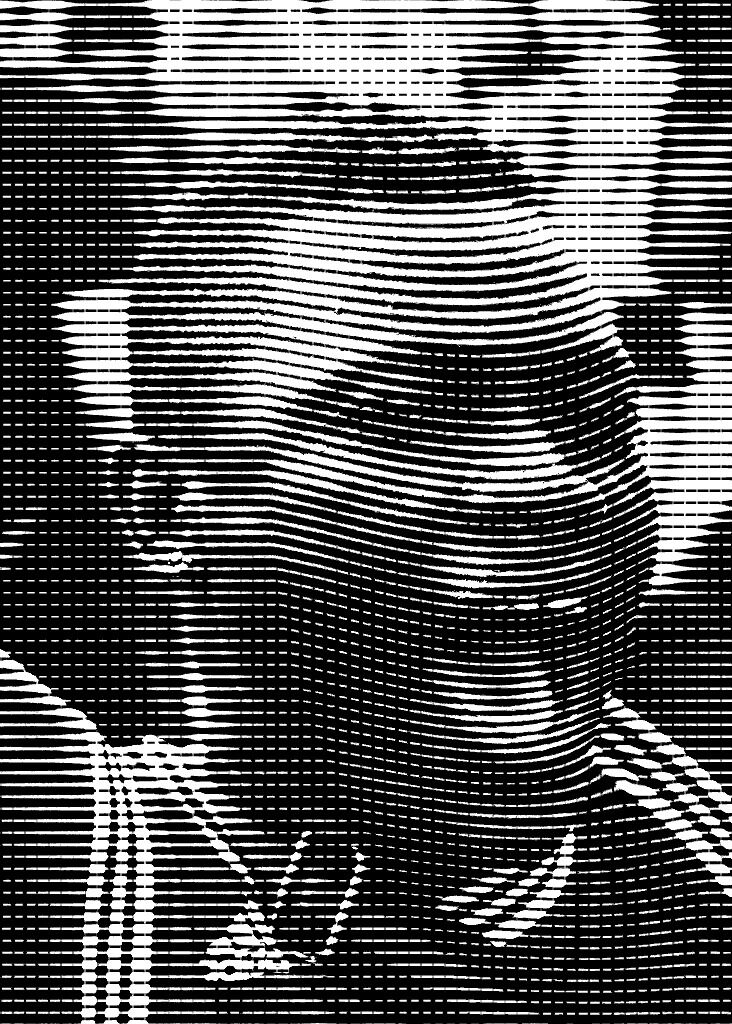}
\includegraphics[height=1.6cm]{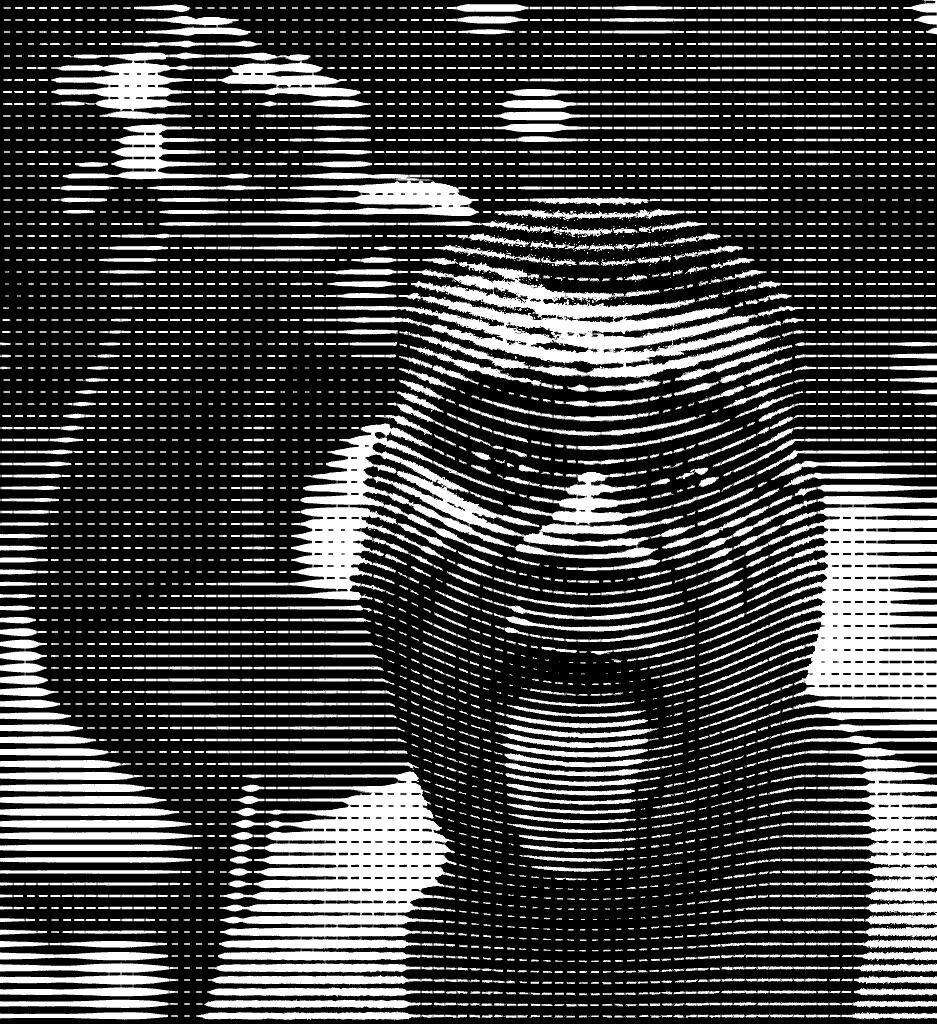}
\includegraphics[height=1.6cm]{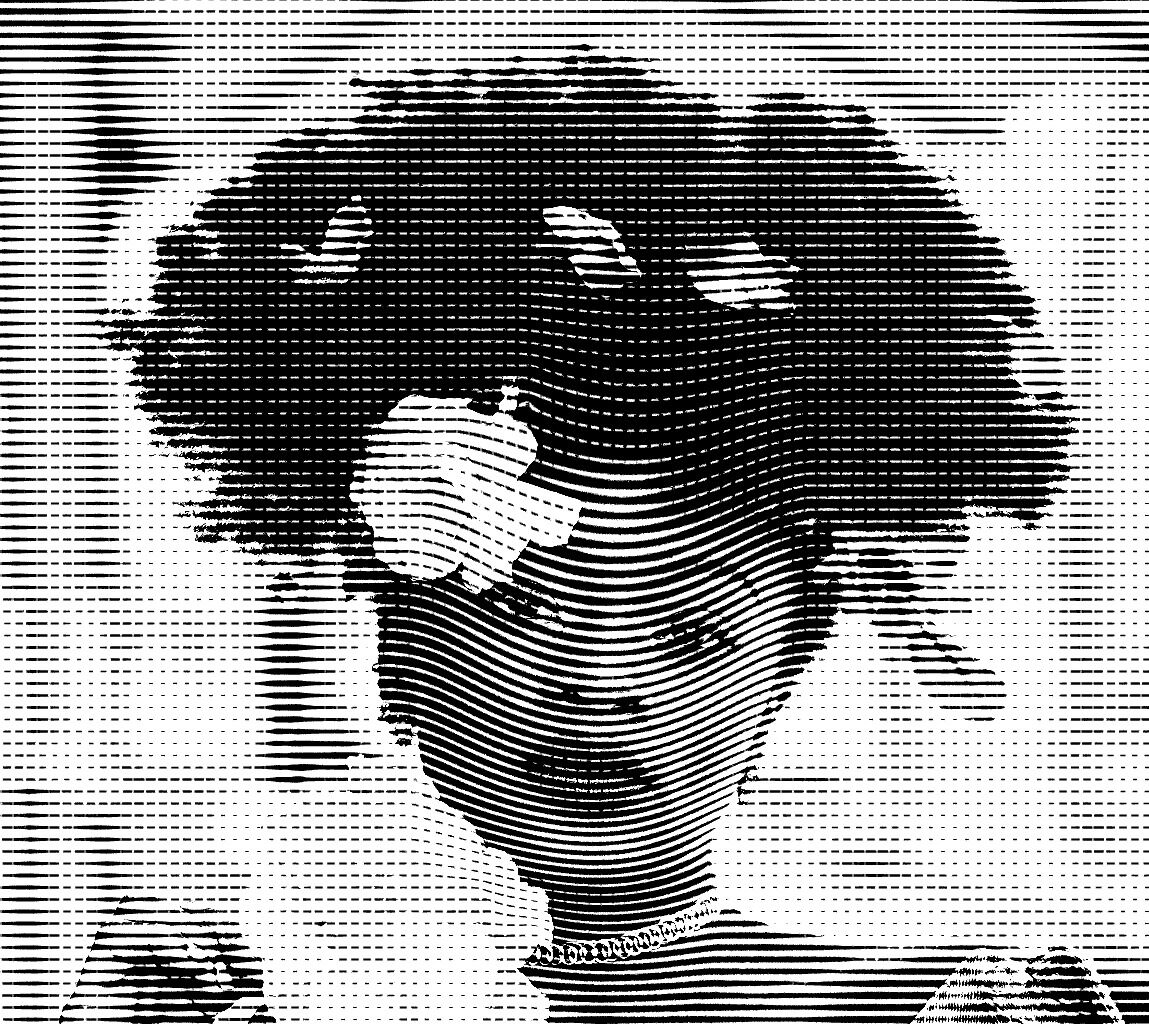}
\includegraphics[height=1.6cm]{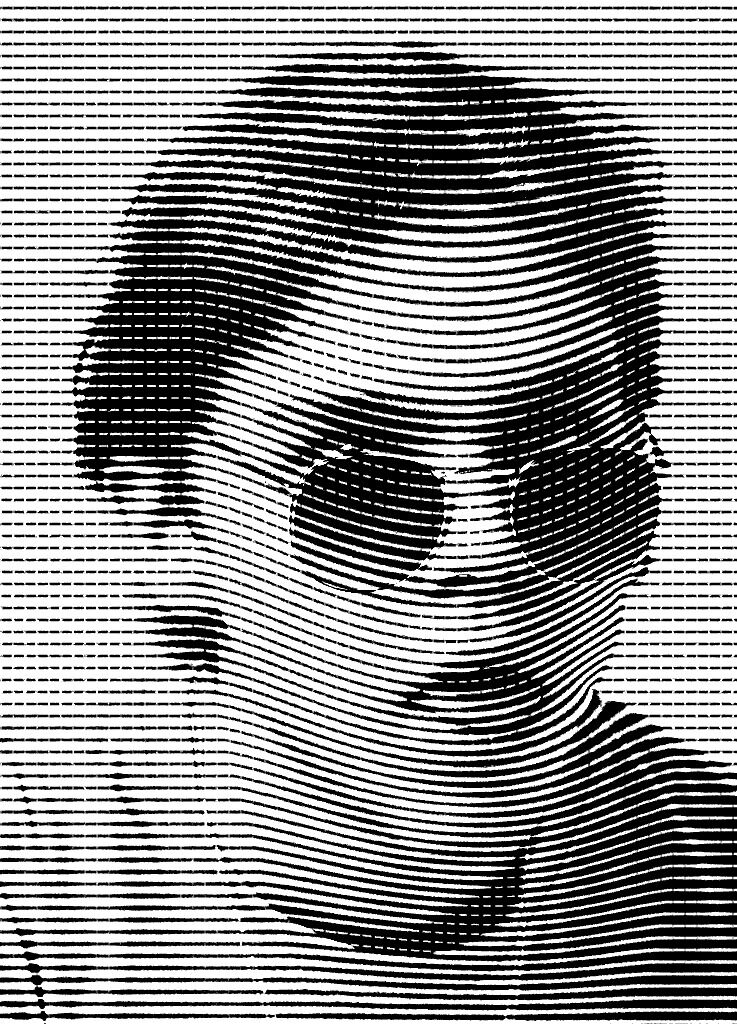}
\includegraphics[height=1.6cm]{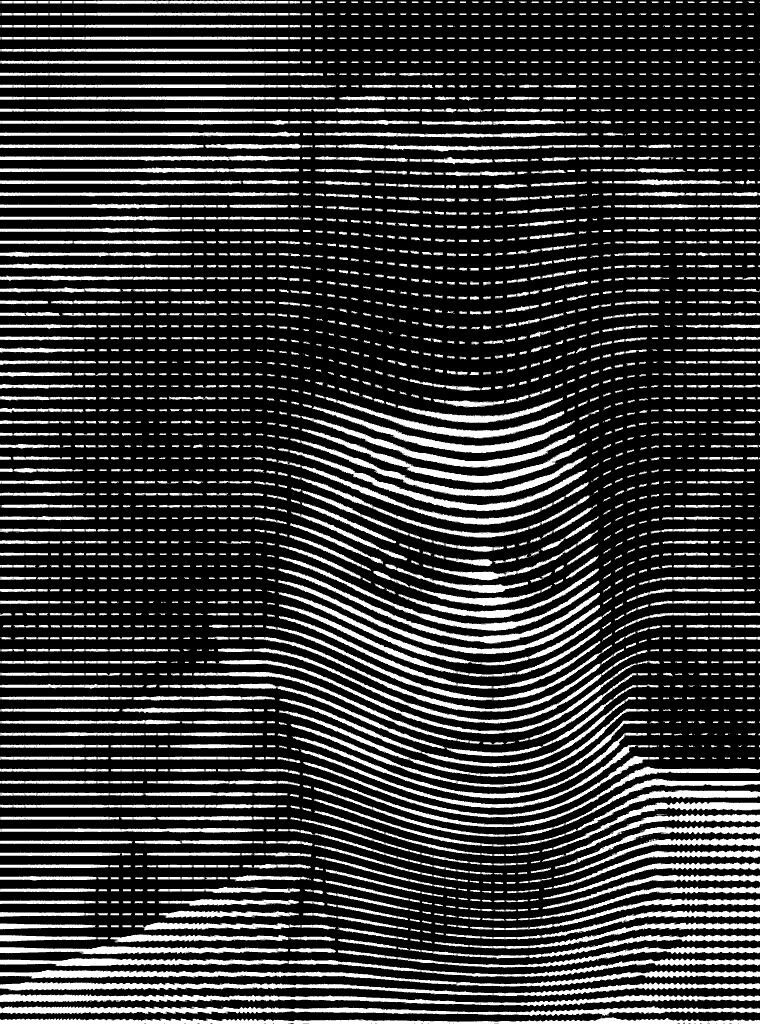}
\includegraphics[height=1.6cm]{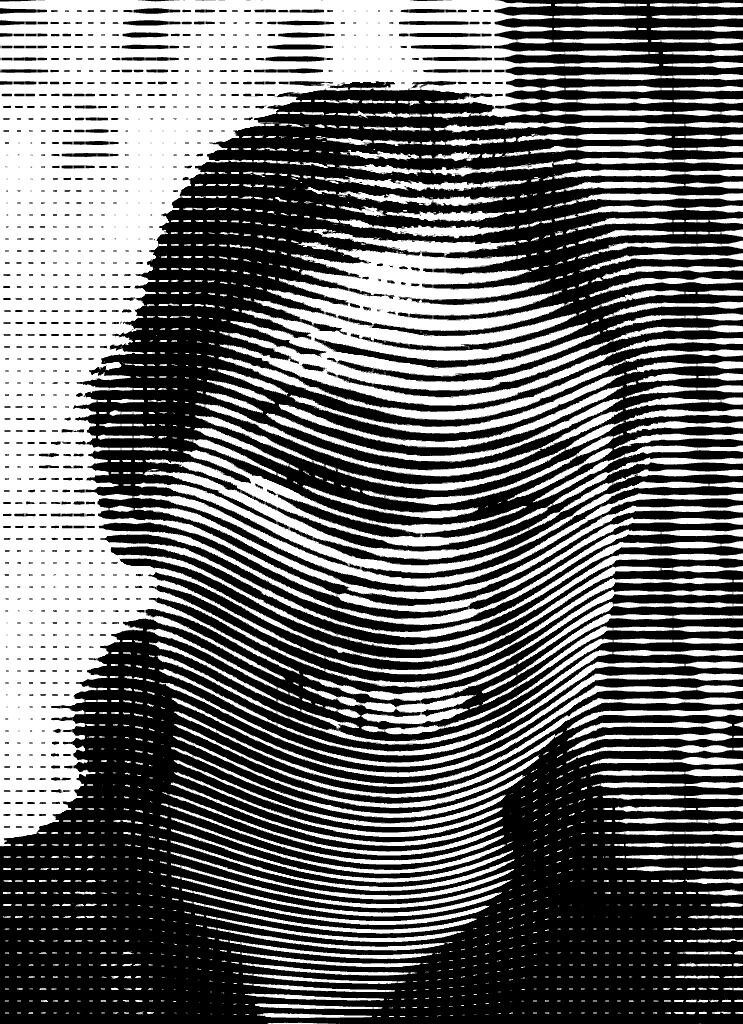}

\centerline{Level 3}
\medskip
\caption{Images from the \emph{NPRportrait1.0} benchmark stylised as engravings: Rosin and Lai~\cite{engraving}}
\label{resultsengraving}
\end{figure}


\begin{figure}[!t]
\centering
\includegraphics[height=1.6cm]{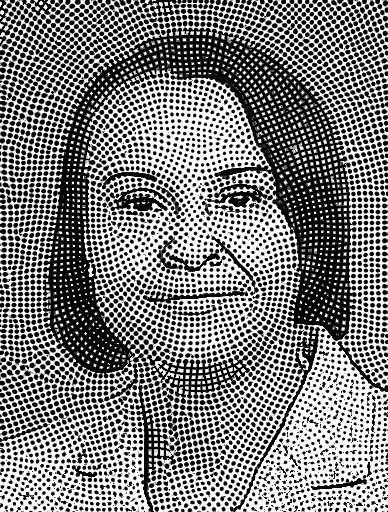}
\includegraphics[height=1.6cm]{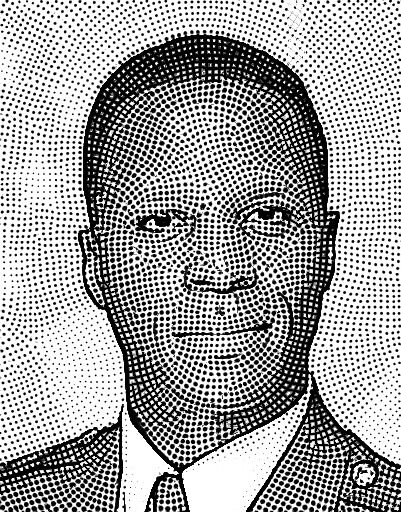}
\includegraphics[height=1.6cm]{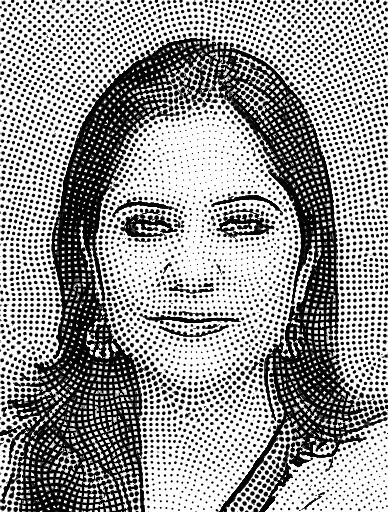}
\includegraphics[height=1.6cm]{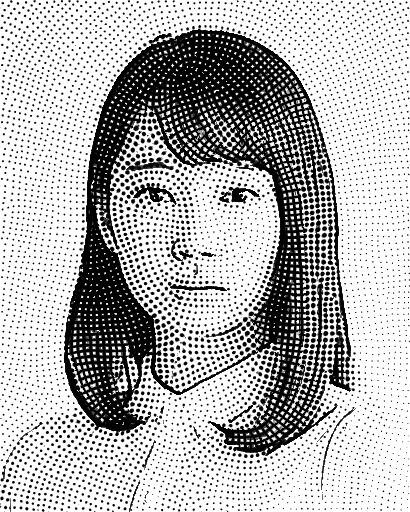}
\includegraphics[height=1.6cm]{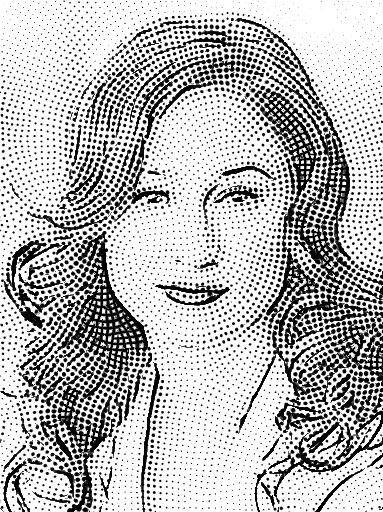}

\centerline{Level 1}
\medskip

\includegraphics[height=1.6cm]{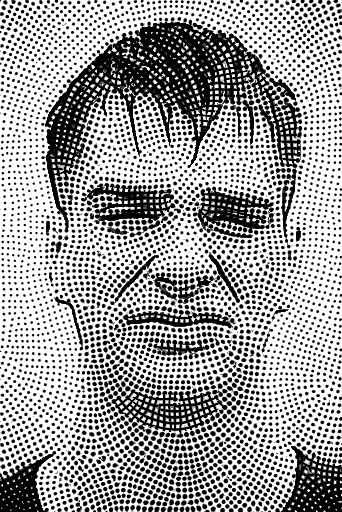}
\includegraphics[height=1.6cm]{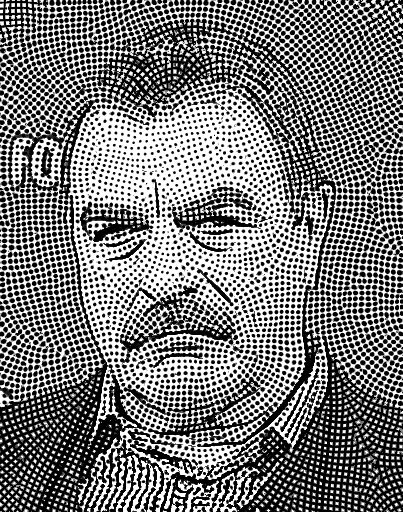}
\includegraphics[height=1.6cm]{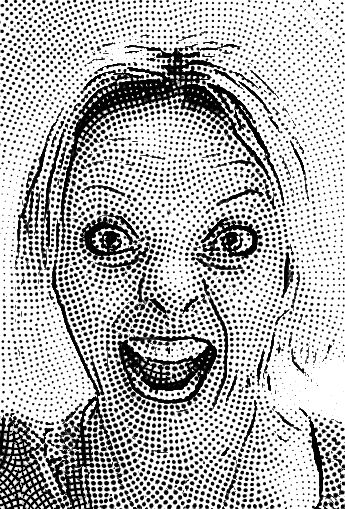}
\includegraphics[height=1.6cm]{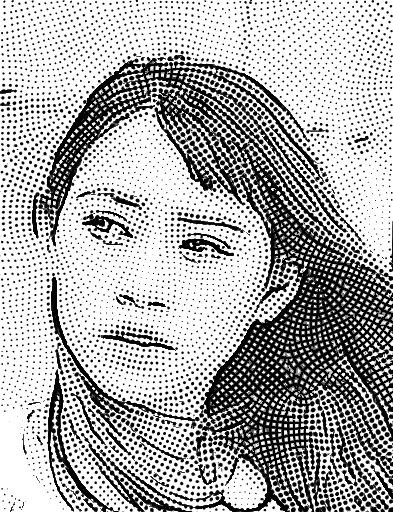}
\includegraphics[height=1.6cm]{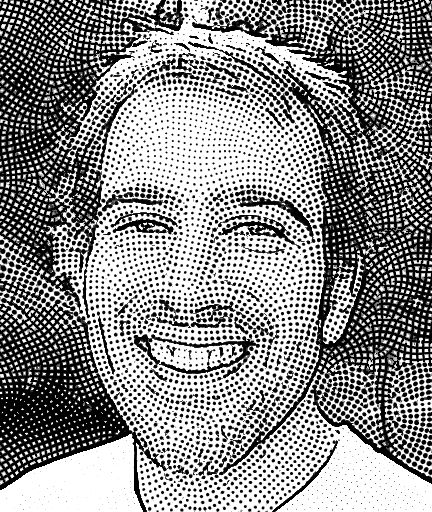}

\centerline{Level 2}
\medskip

\includegraphics[height=1.6cm]{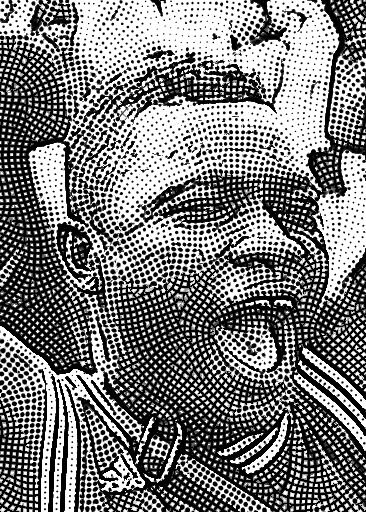}
\includegraphics[height=1.6cm]{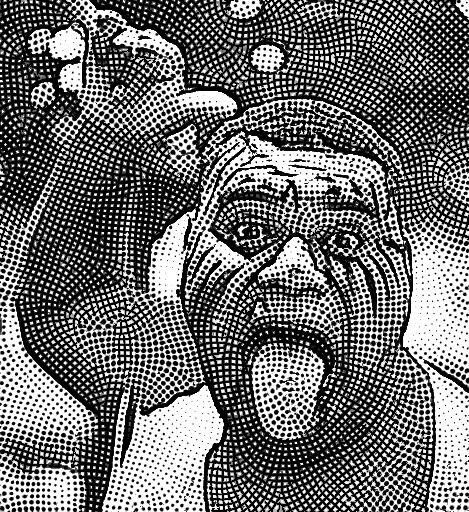}
\includegraphics[height=1.6cm]{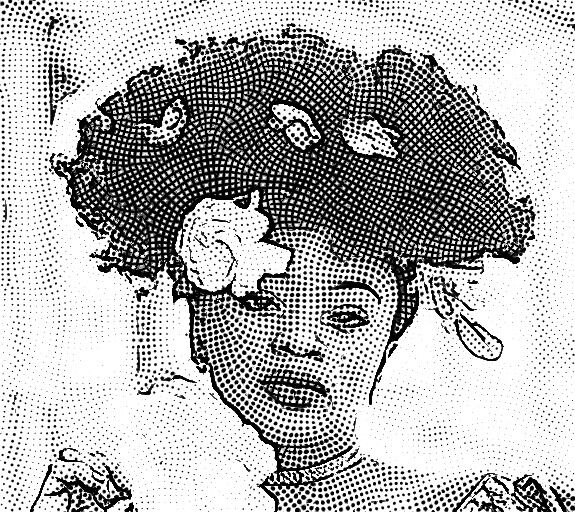}
\includegraphics[height=1.6cm]{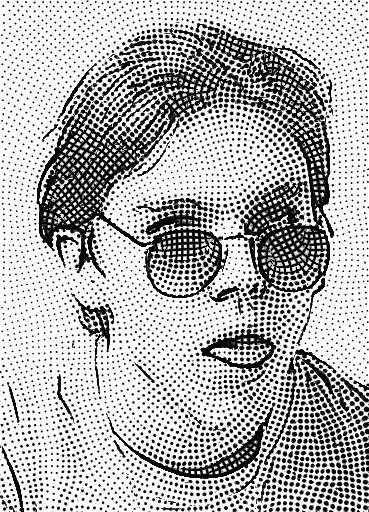}
\includegraphics[height=1.6cm]{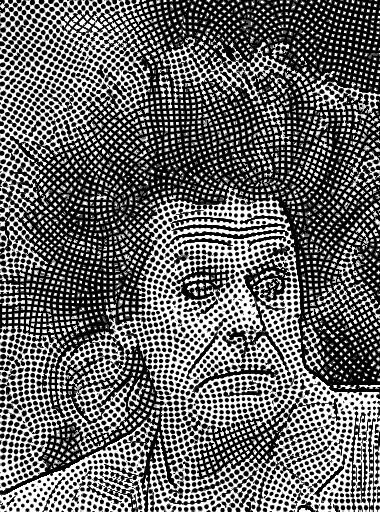}
\includegraphics[height=1.6cm]{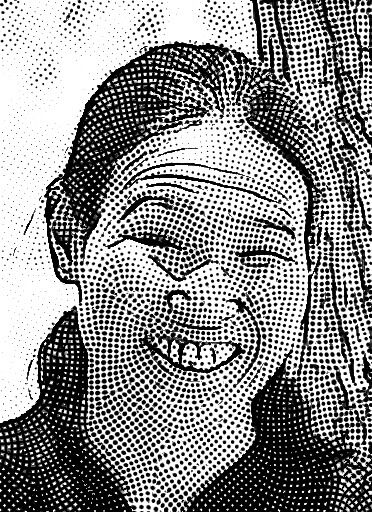}

\centerline{Level 3}
\medskip
\caption{Images from the \emph{NPRportrait1.0} benchmark stylised as hedcuts: Son \emph{et al.}~\cite{Son-hedcut}}
\label{resultshedcut}
\end{figure}


\begin{figure}[!t]
\centering
\includegraphics[height=1.6cm]{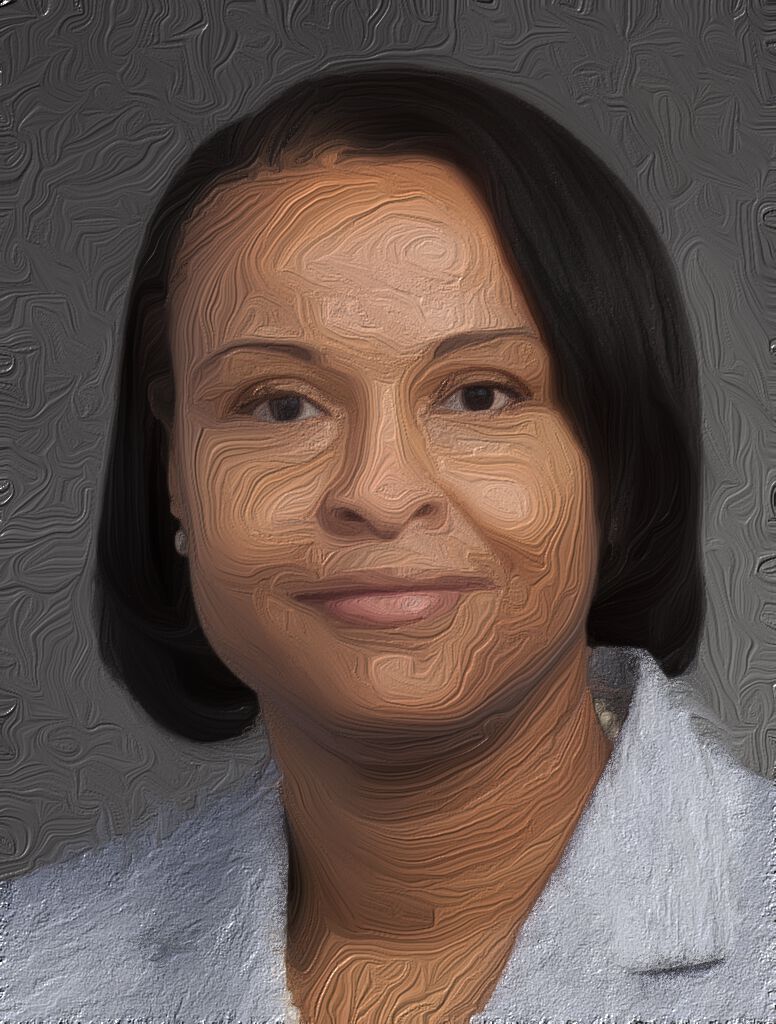}
\includegraphics[height=1.6cm]{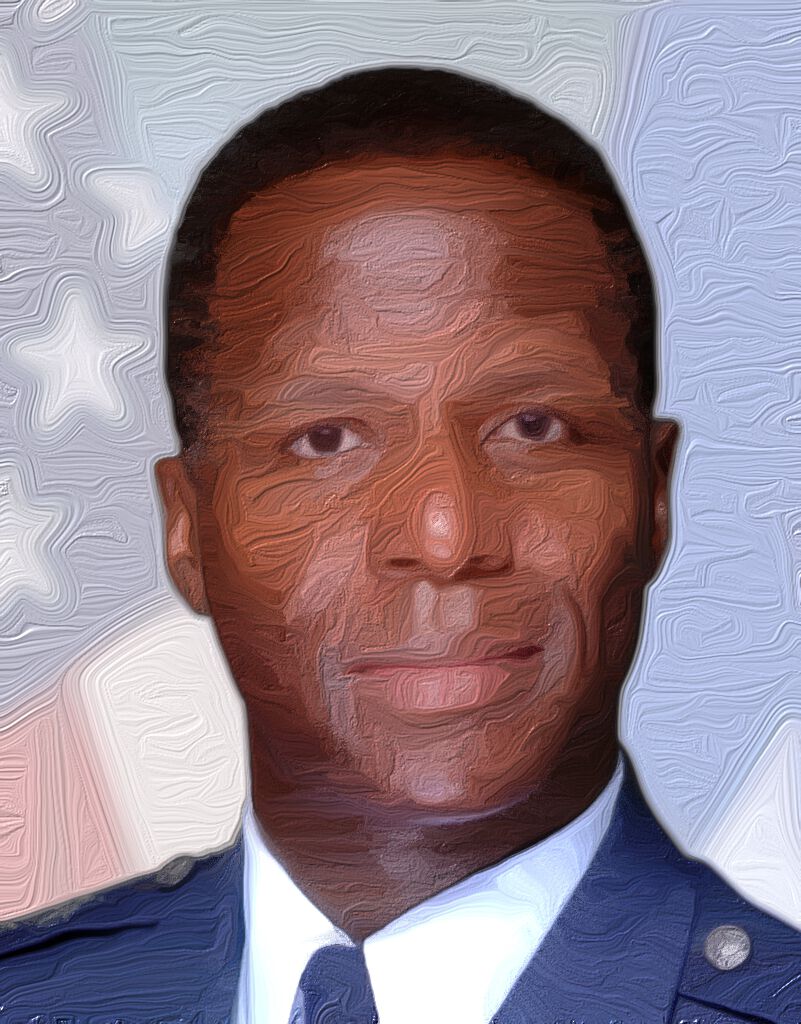}
\includegraphics[height=1.6cm]{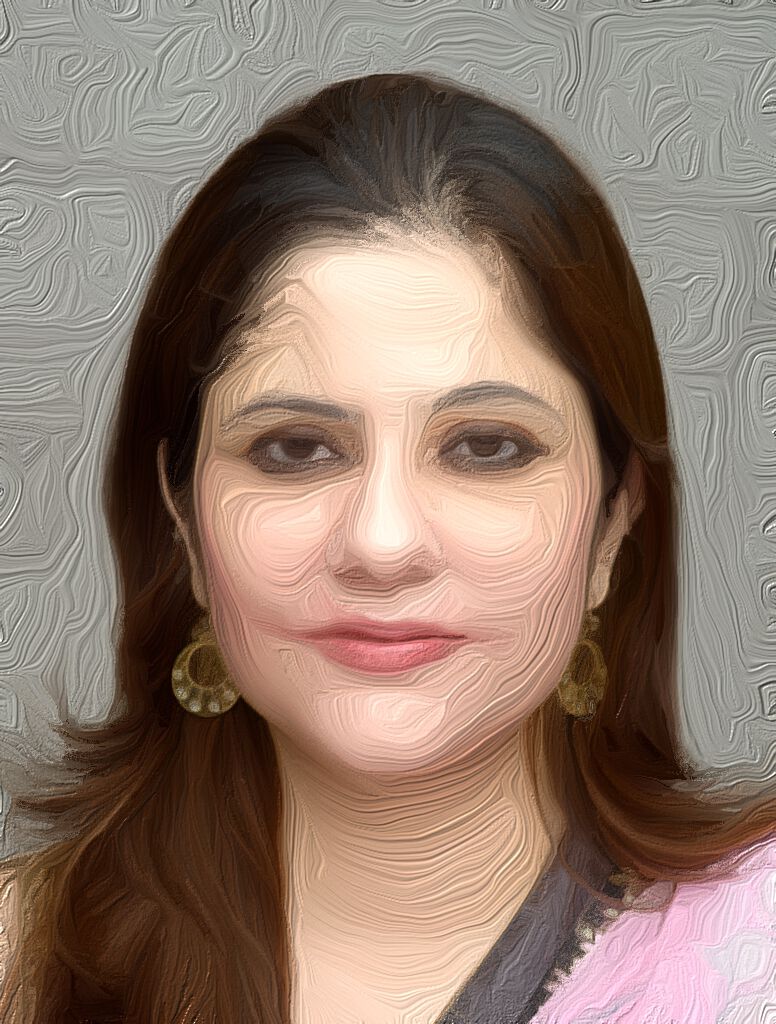}
\includegraphics[height=1.6cm]{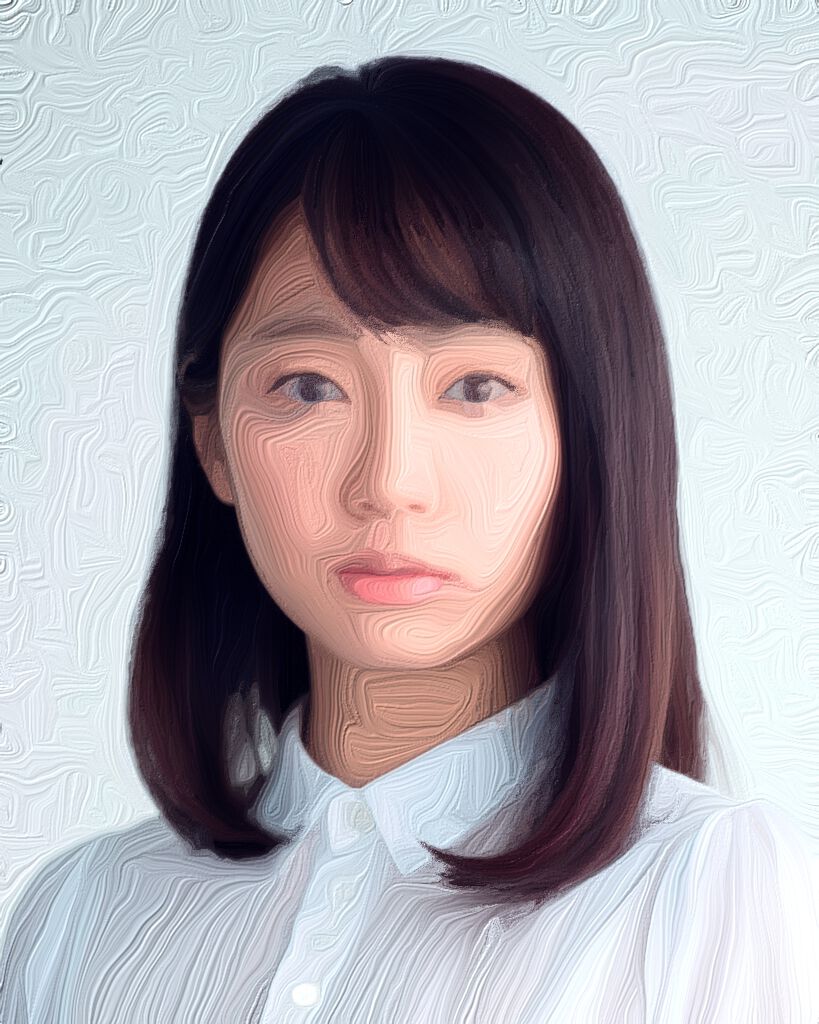}
\includegraphics[height=1.6cm]{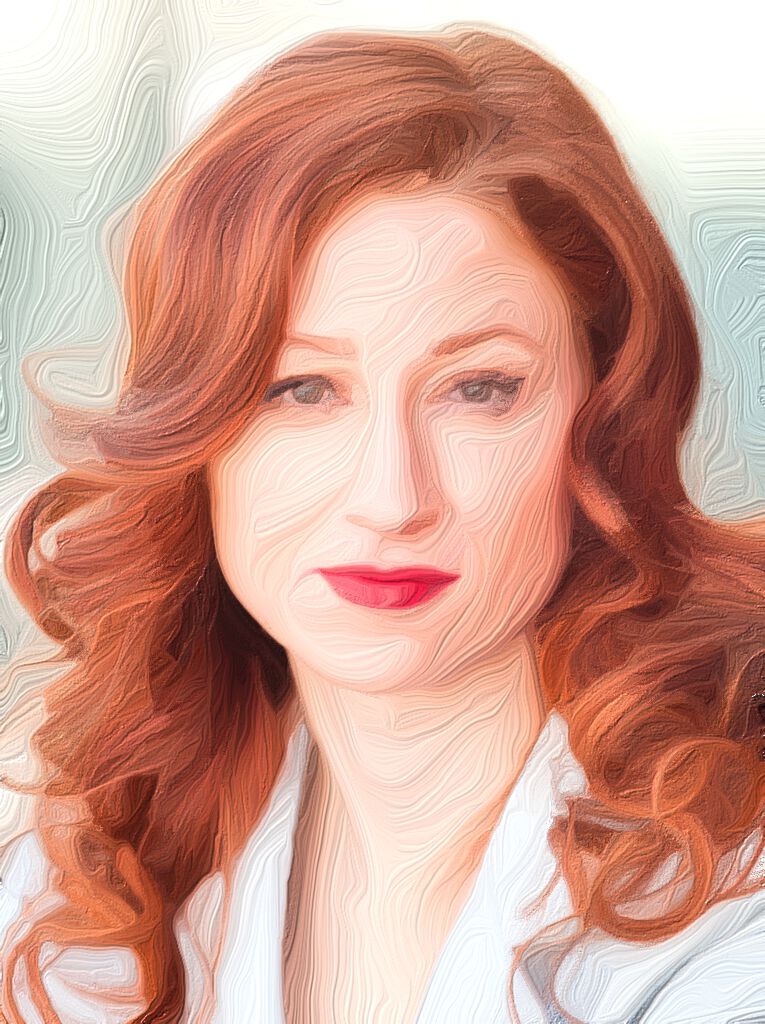}

\centerline{Level 1}
\medskip

\includegraphics[height=1.6cm]{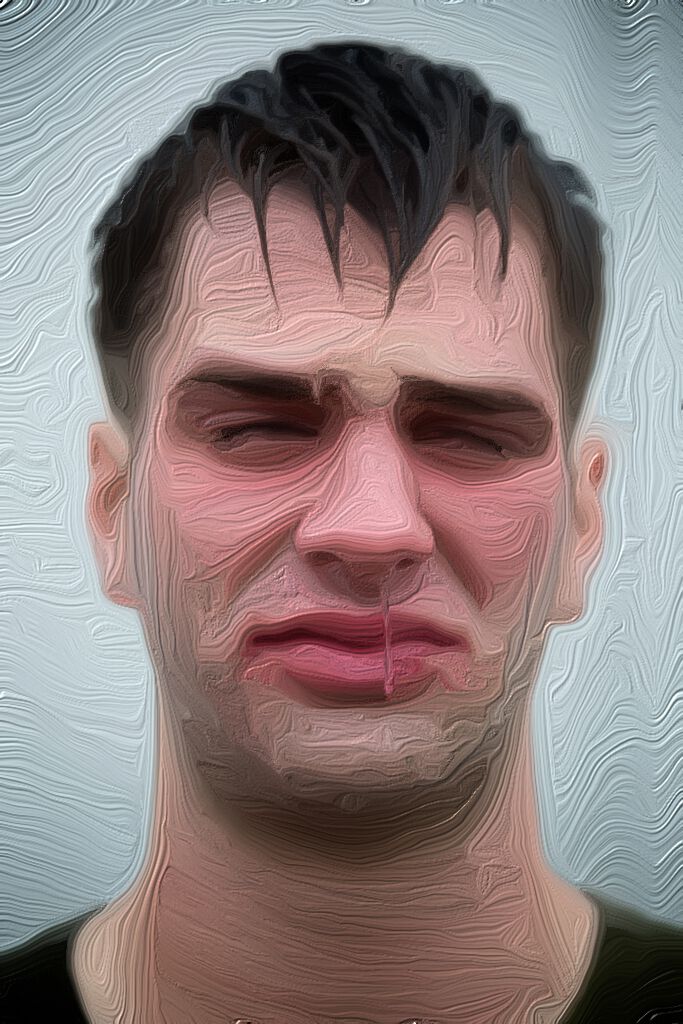}
\includegraphics[height=1.6cm]{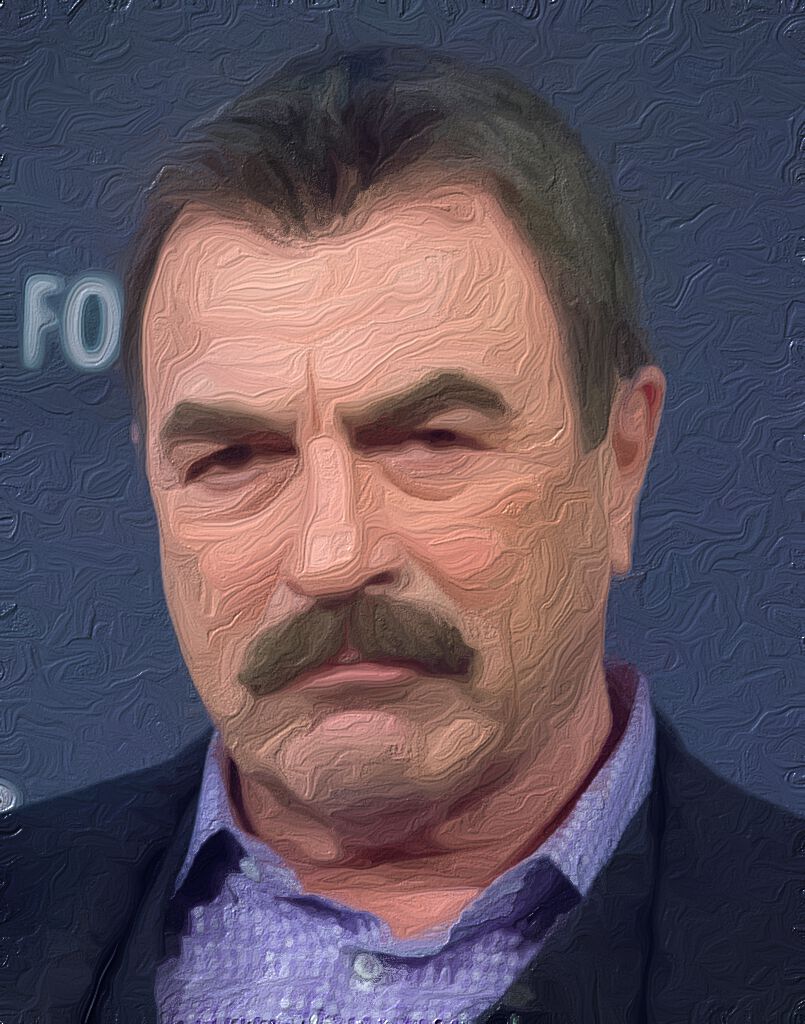}
\includegraphics[height=1.6cm]{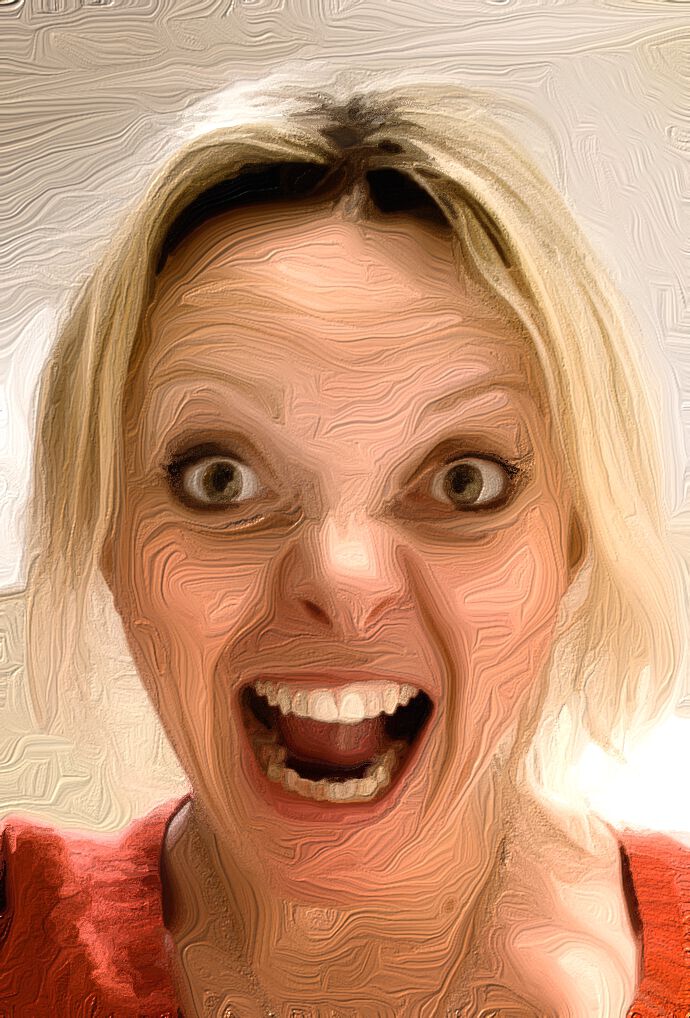}
\includegraphics[height=1.6cm]{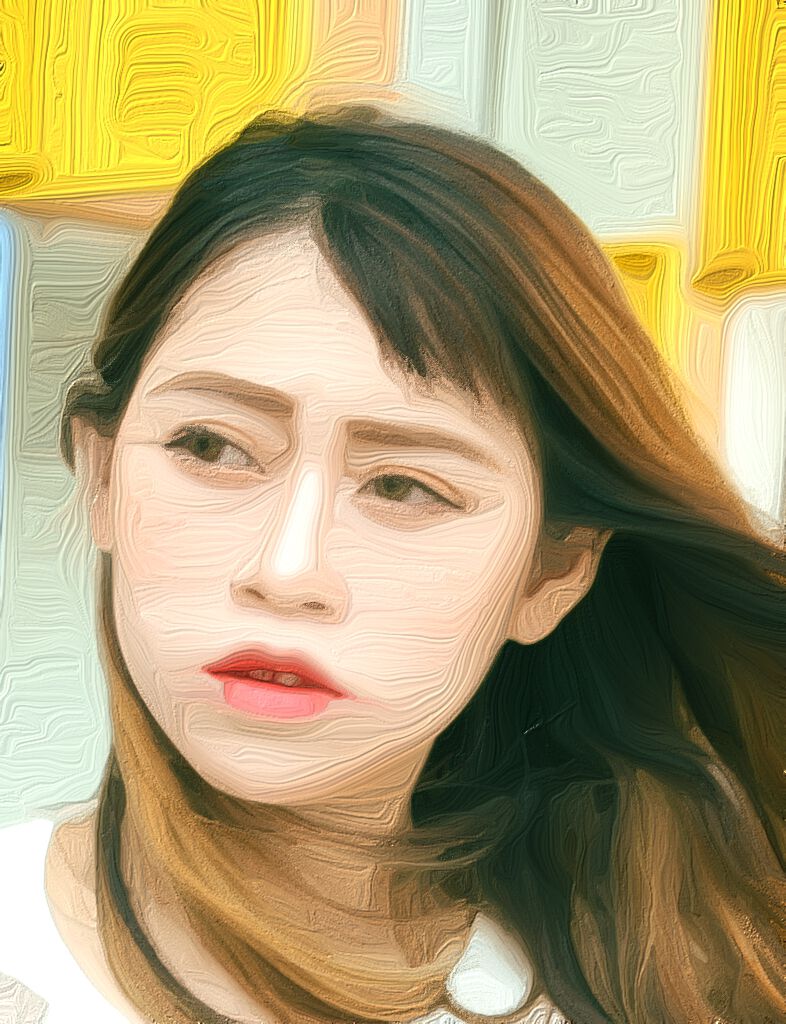}
\includegraphics[height=1.6cm]{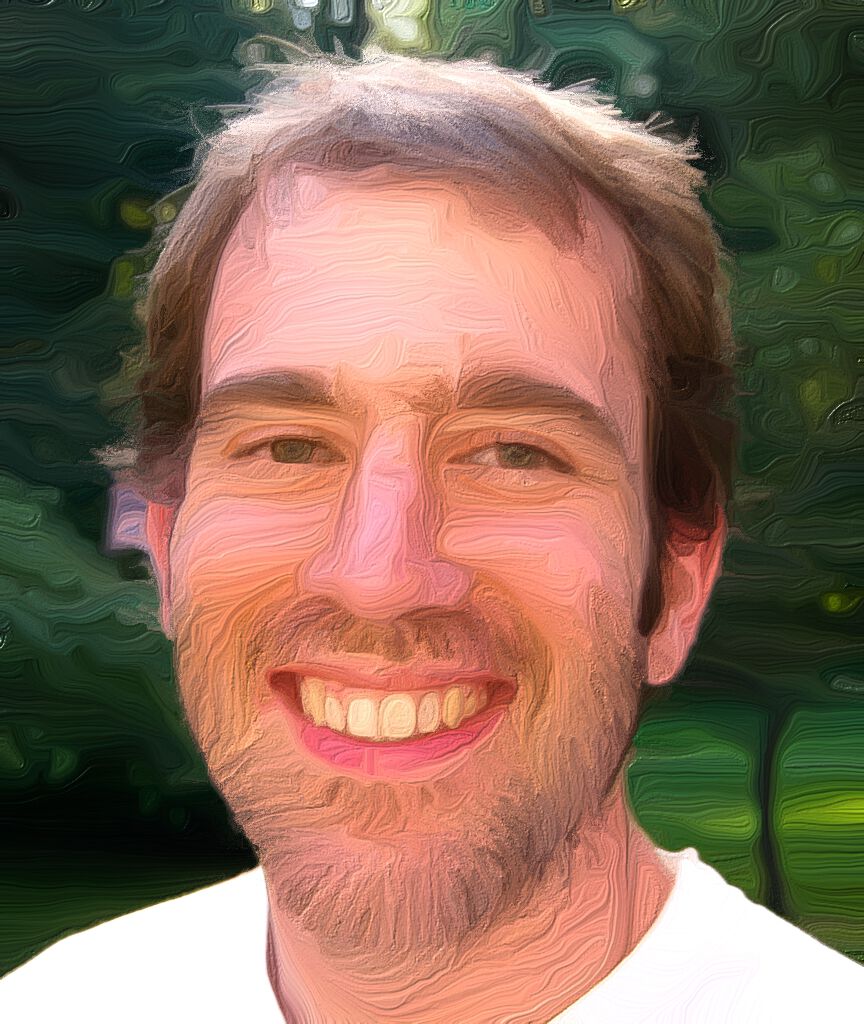}

\centerline{Level 2}
\medskip

\includegraphics[height=1.6cm]{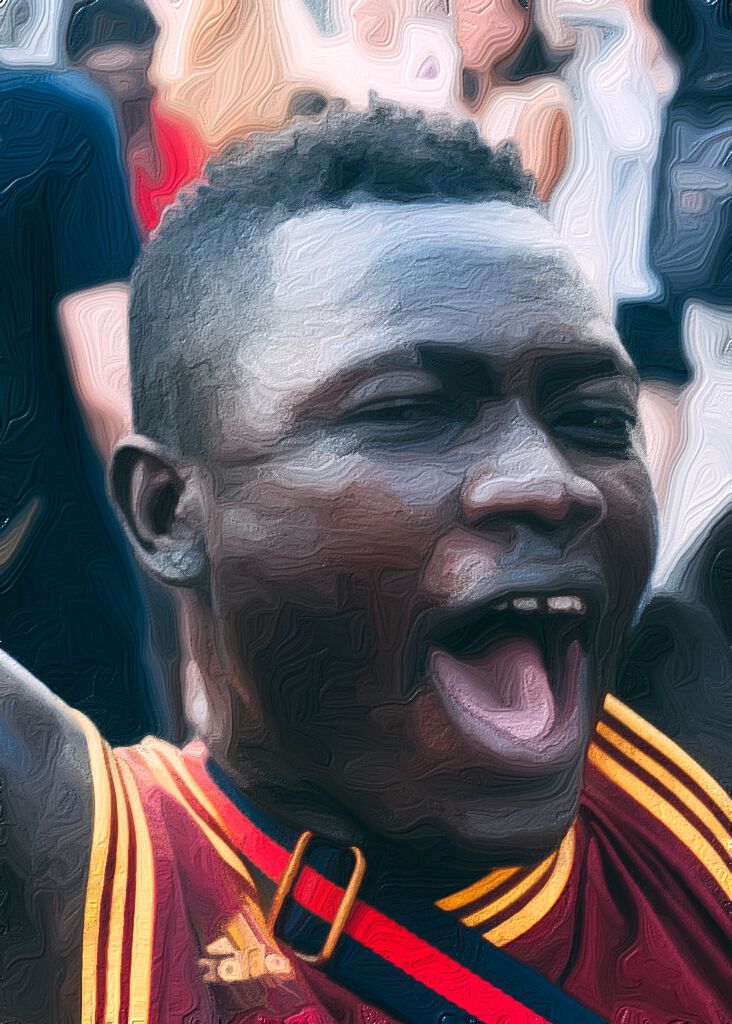}
\includegraphics[height=1.6cm]{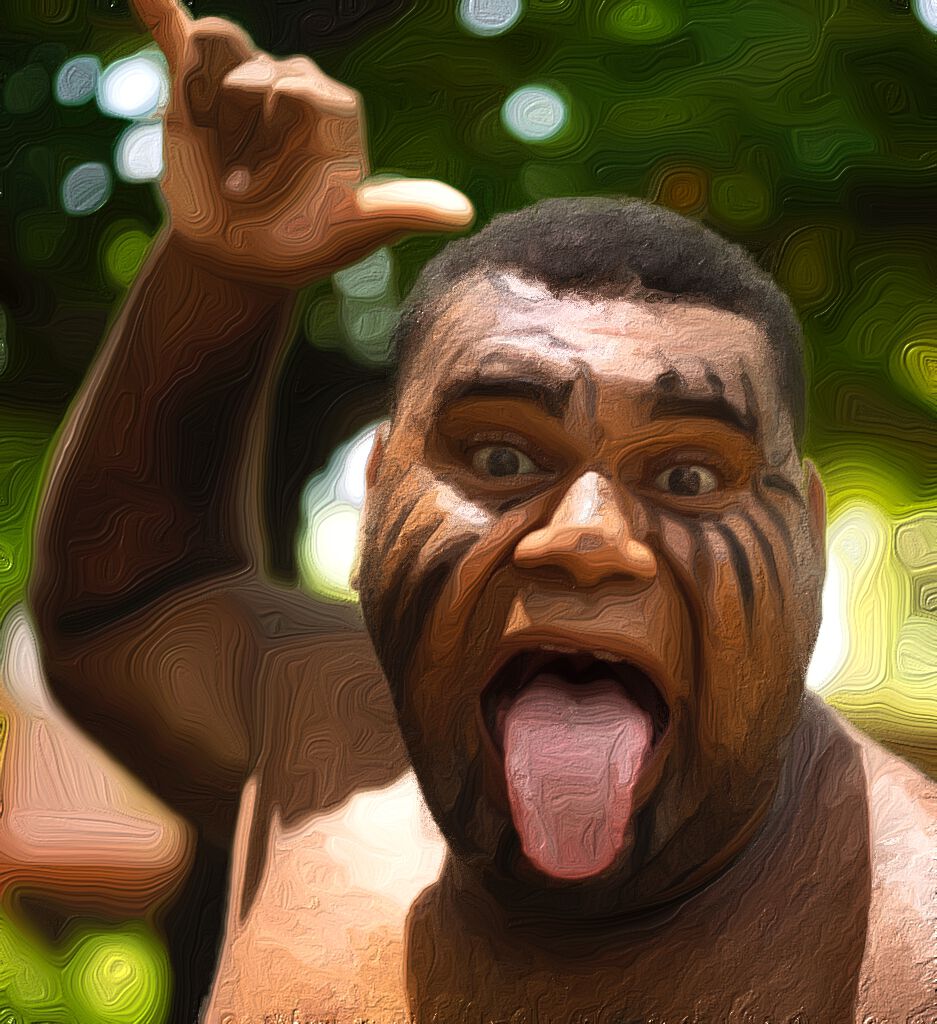}
\includegraphics[height=1.6cm]{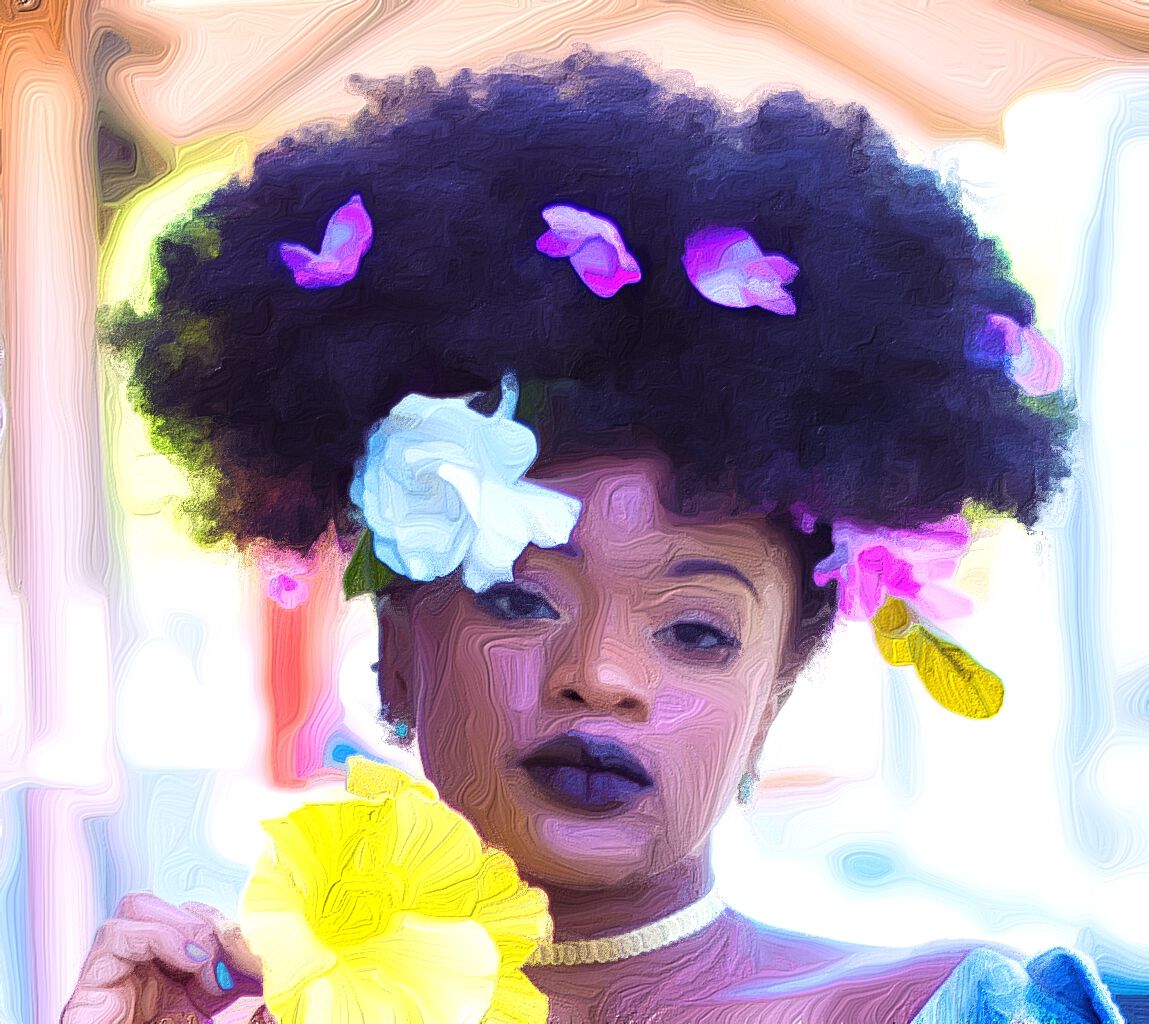}
\includegraphics[height=1.6cm]{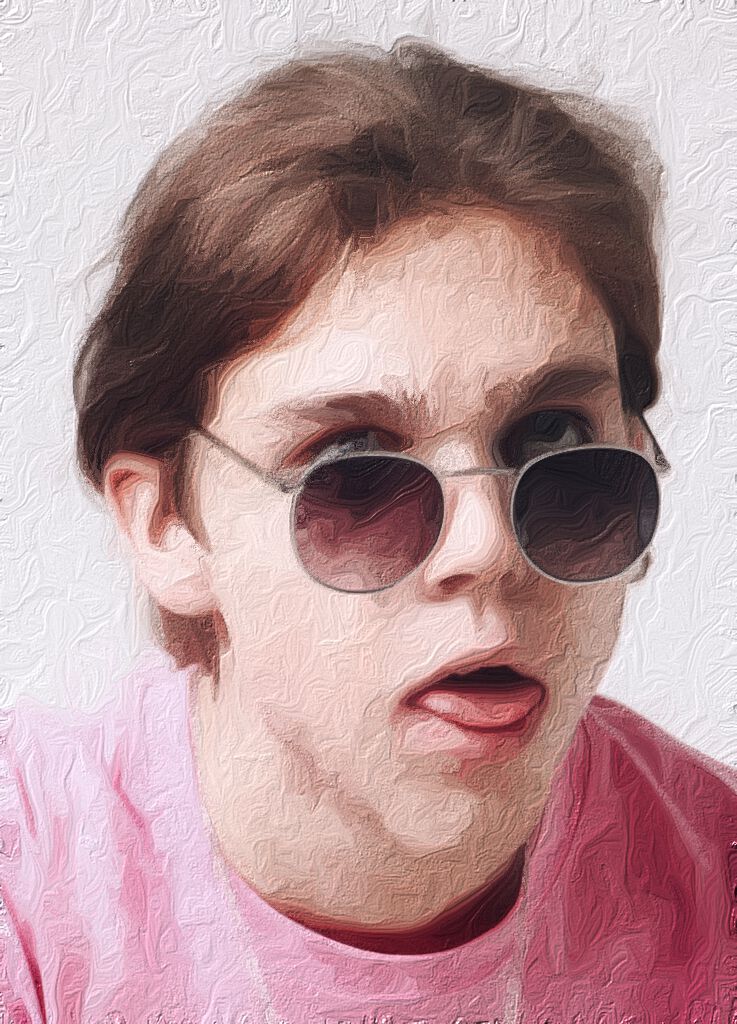}
\includegraphics[height=1.6cm]{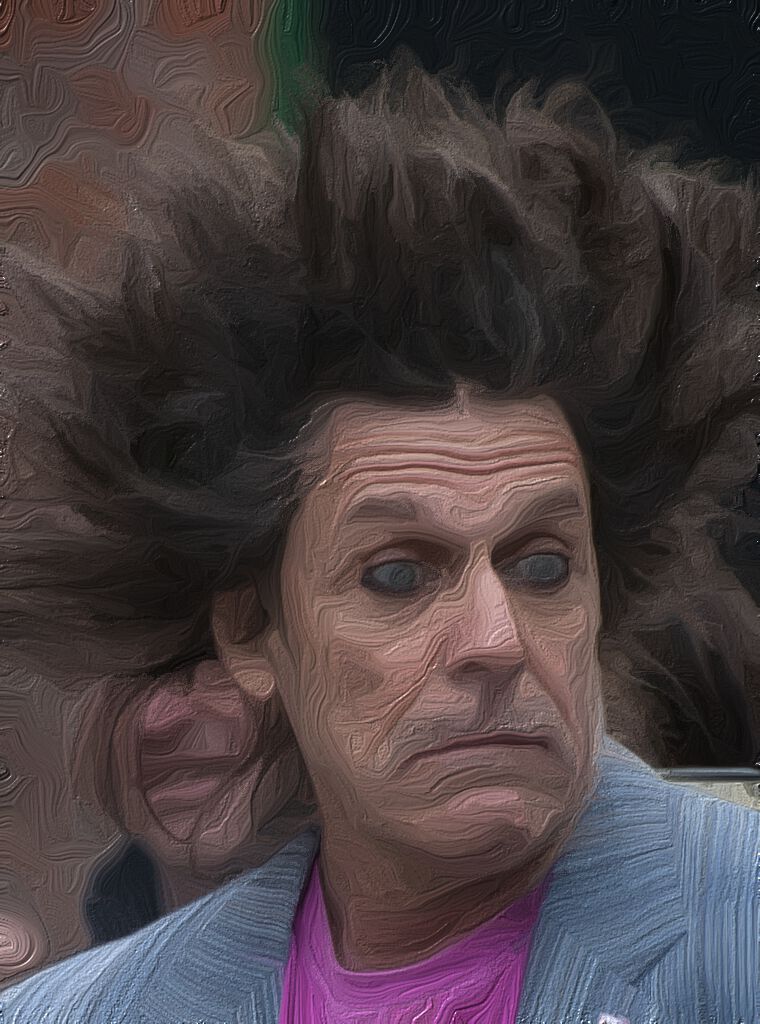}
\includegraphics[height=1.6cm]{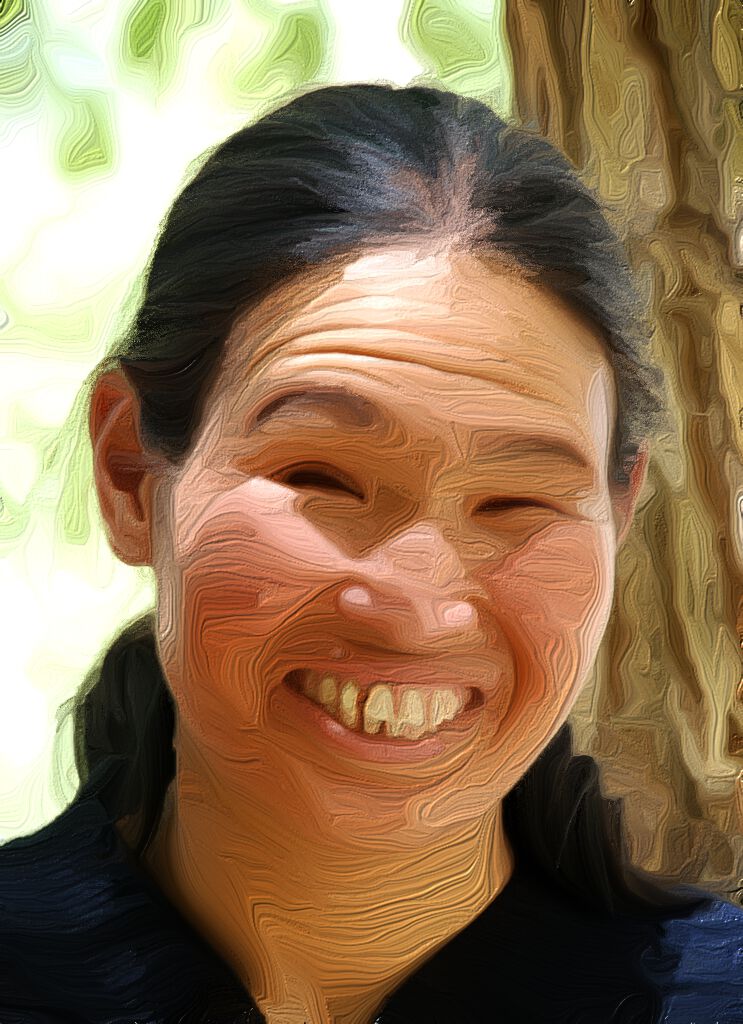}

\centerline{Level 3}
\medskip
\caption{Images from the \emph{NPRportrait1.0} benchmark stylised as oil paintings: Semmo \emph{et al.}~\cite{semmo2016image}}
\label{resultsoilpaint}
\end{figure}


\begin{figure}[!t]
\centering
\includegraphics[height=1.6cm]{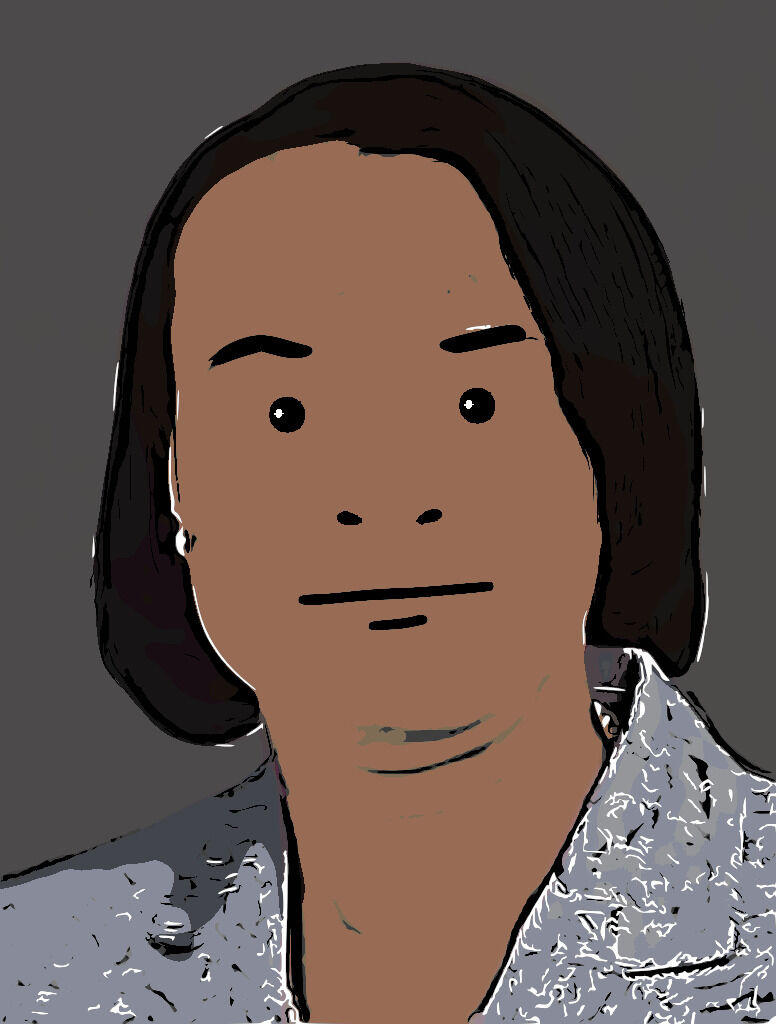}
\includegraphics[height=1.6cm]{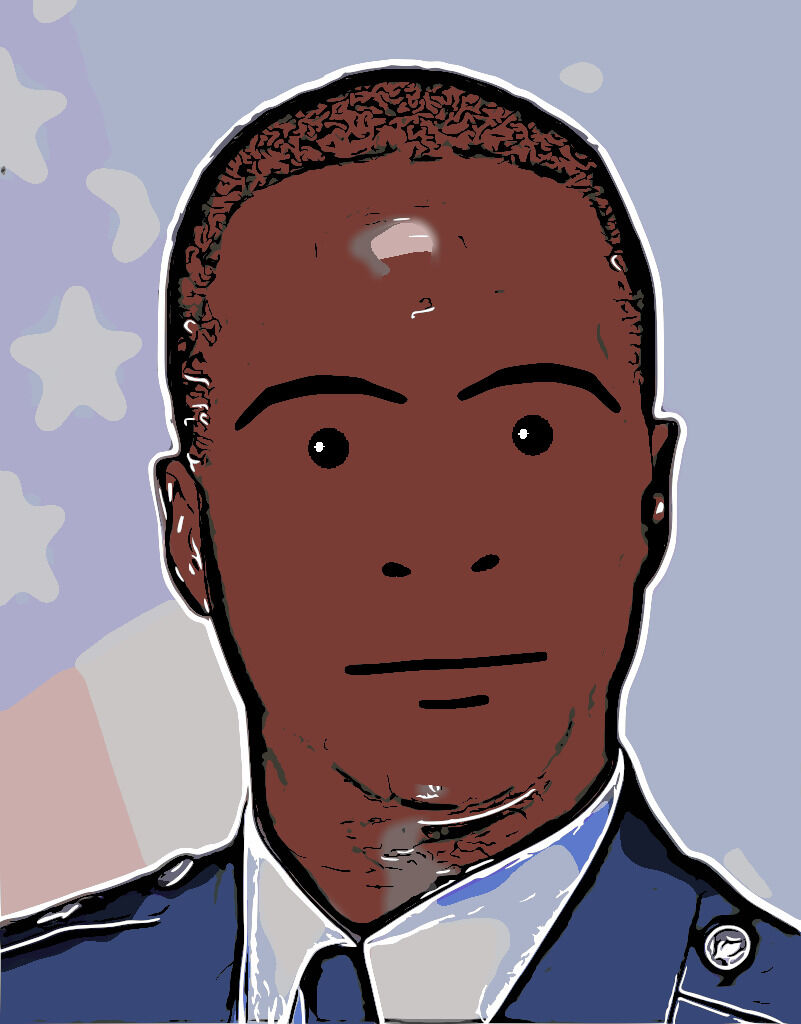}
\includegraphics[height=1.6cm]{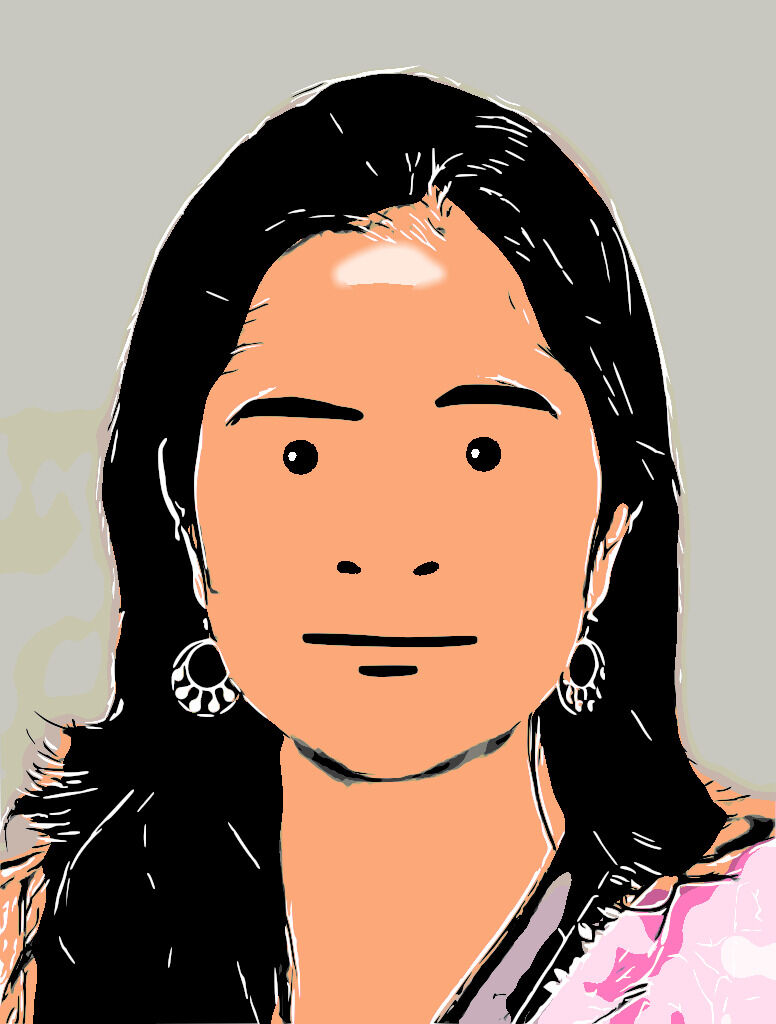}
\includegraphics[height=1.6cm]{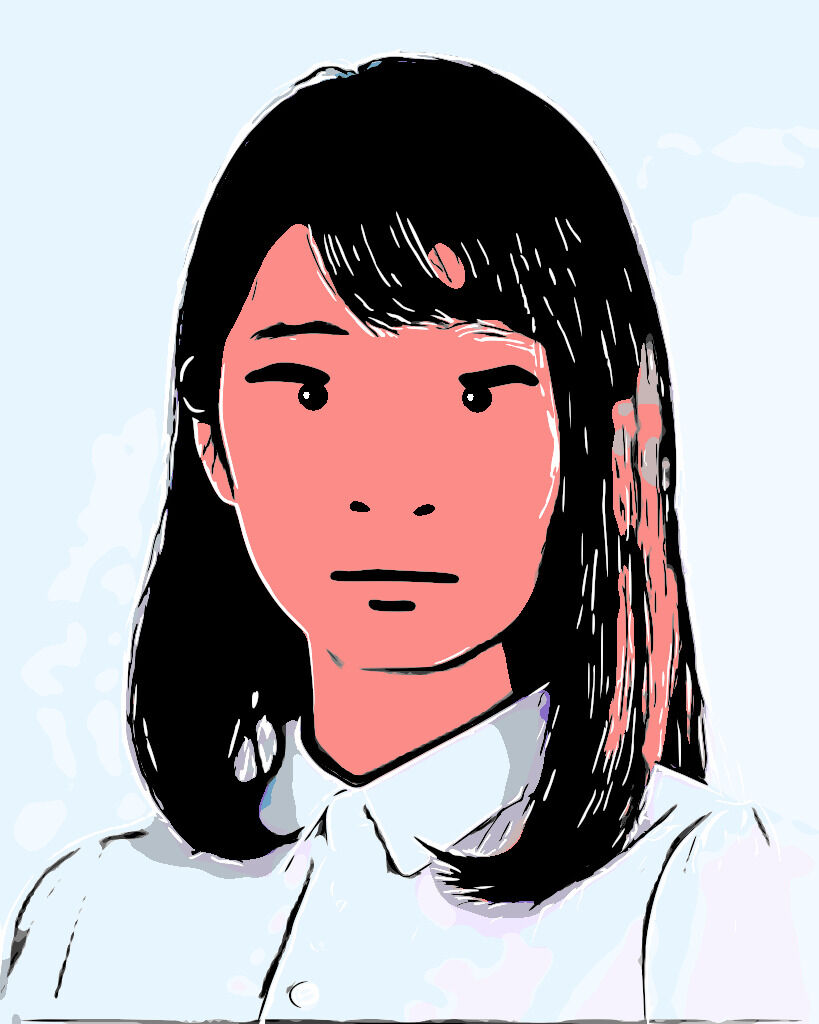}
\includegraphics[height=1.6cm]{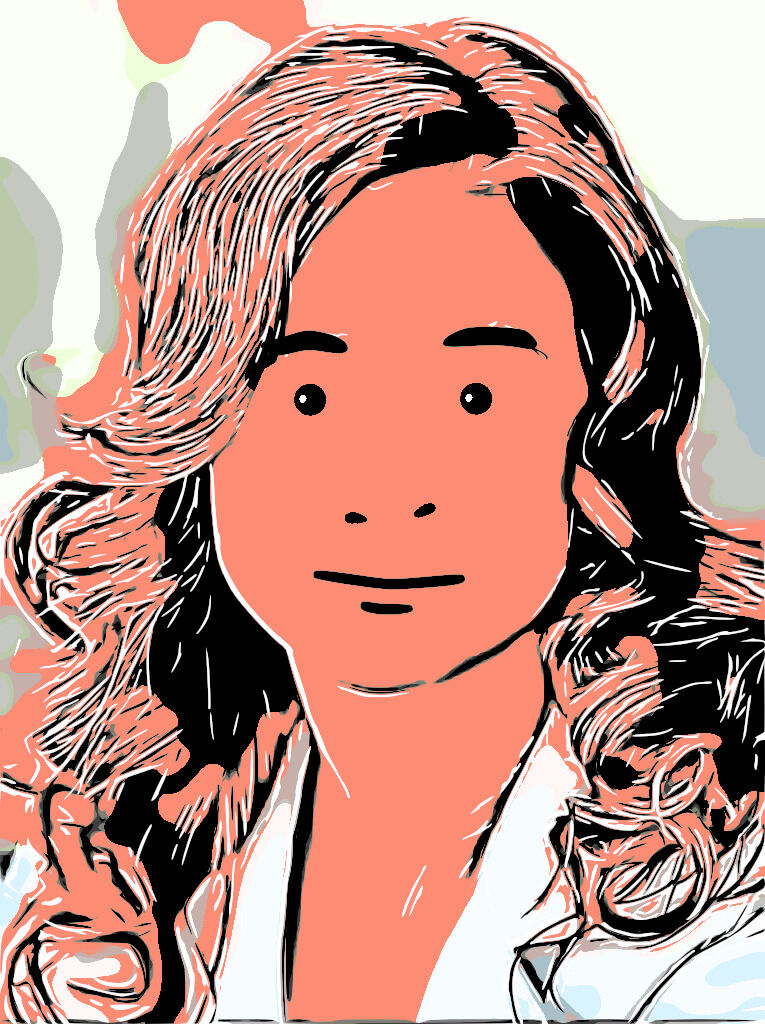}

\centerline{Level 1}
\medskip

\includegraphics[height=1.6cm]{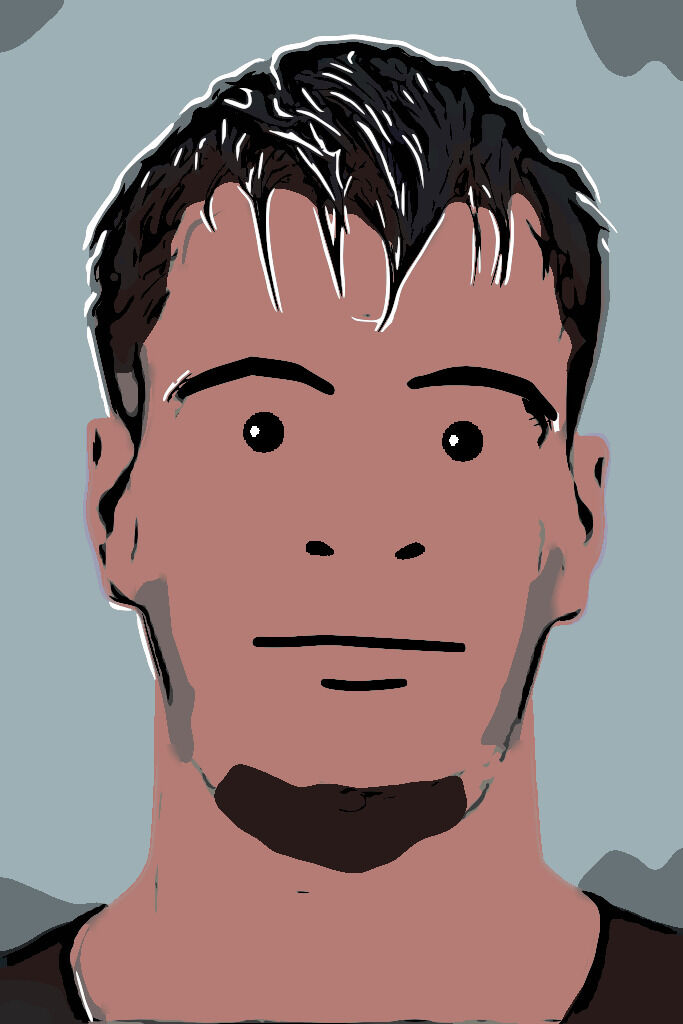}
\includegraphics[height=1.6cm]{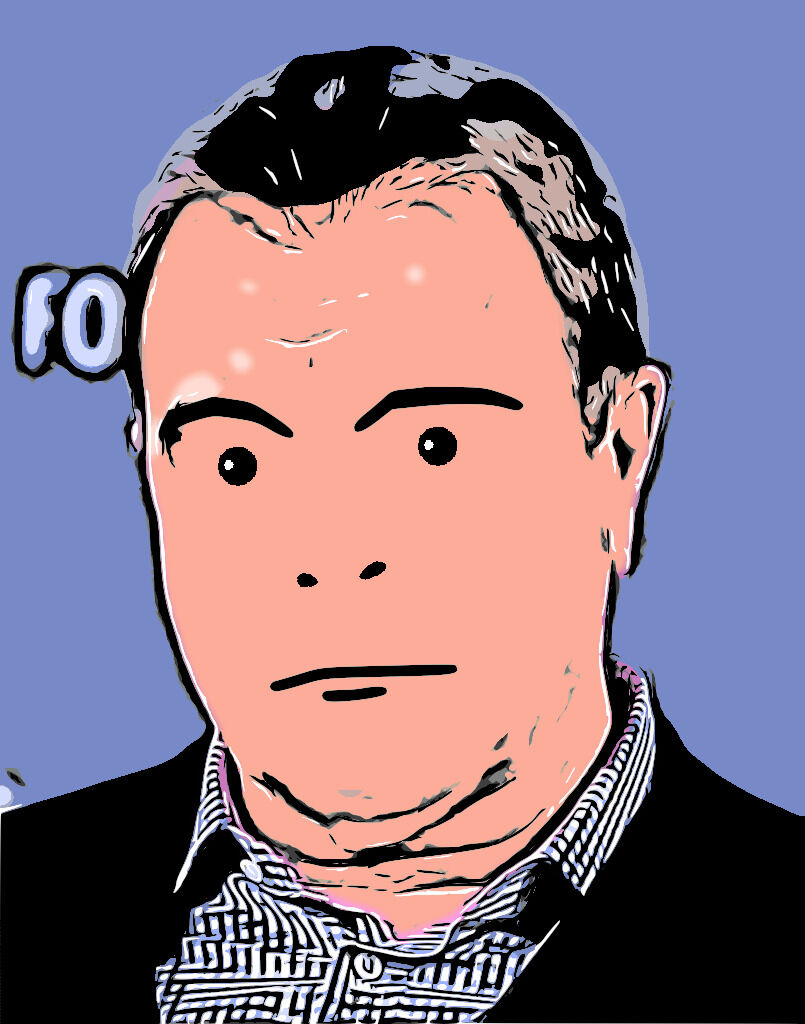}
\includegraphics[height=1.6cm]{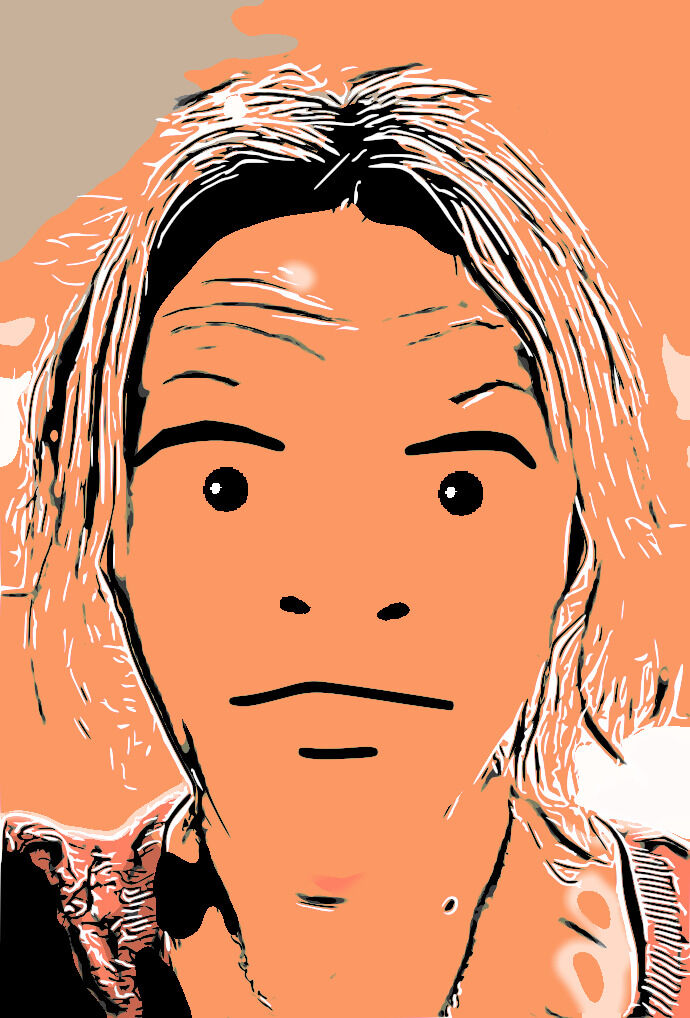}
\includegraphics[height=1.6cm]{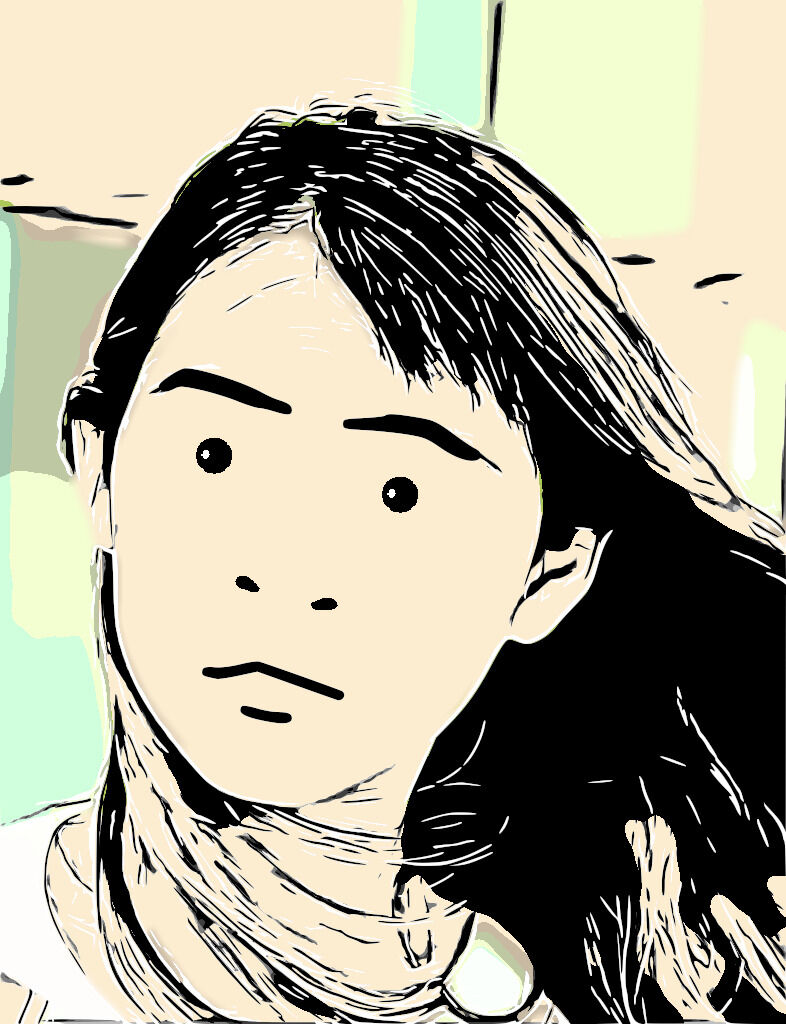}
\includegraphics[height=1.6cm]{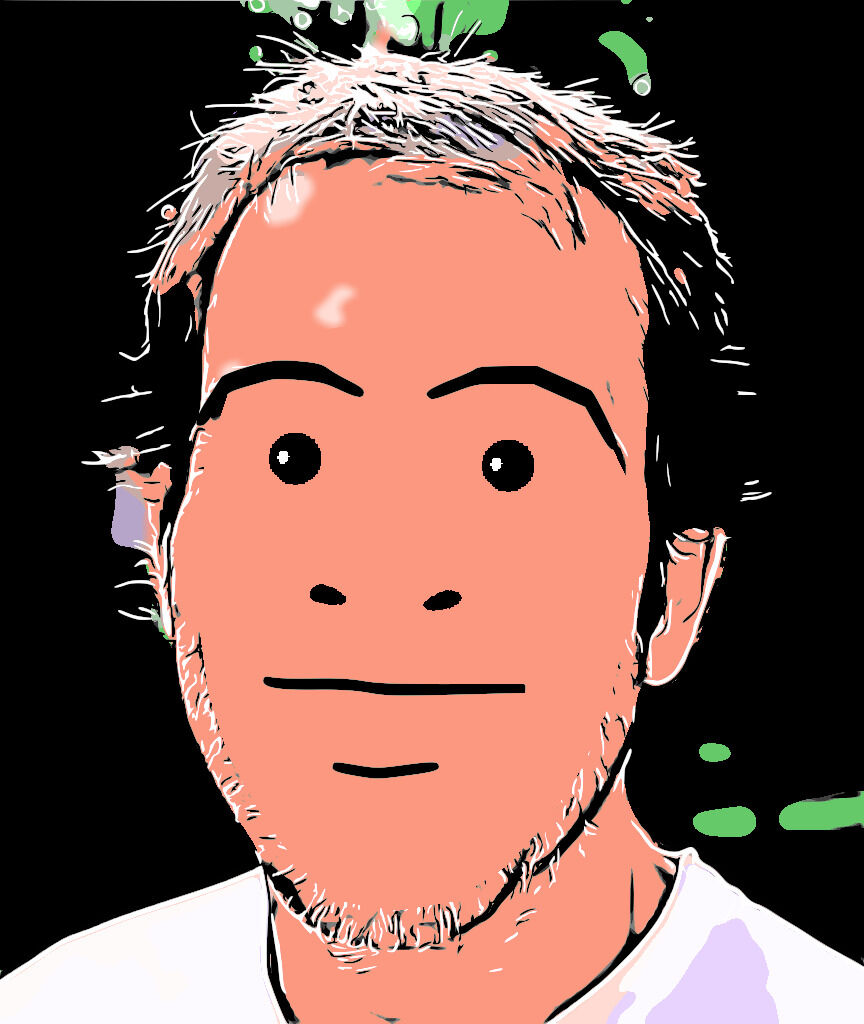}

\centerline{Level 2}
\medskip

\includegraphics[height=1.6cm]{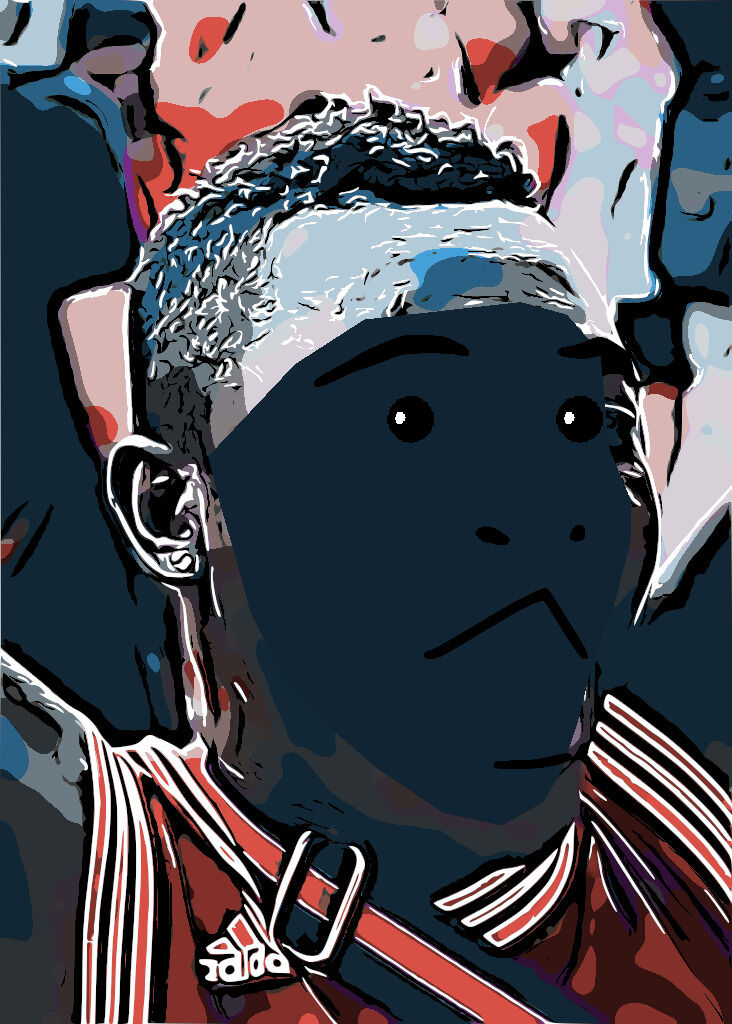}
\includegraphics[height=1.6cm]{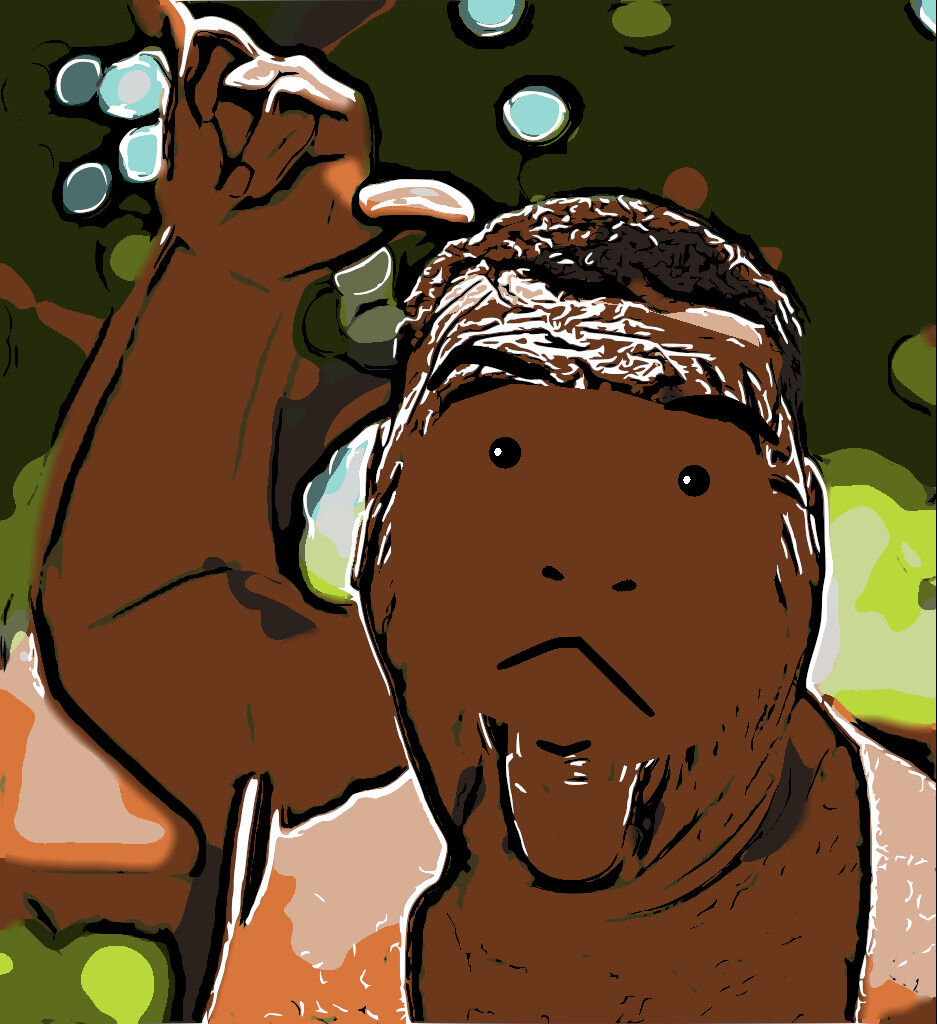}
\includegraphics[height=1.6cm]{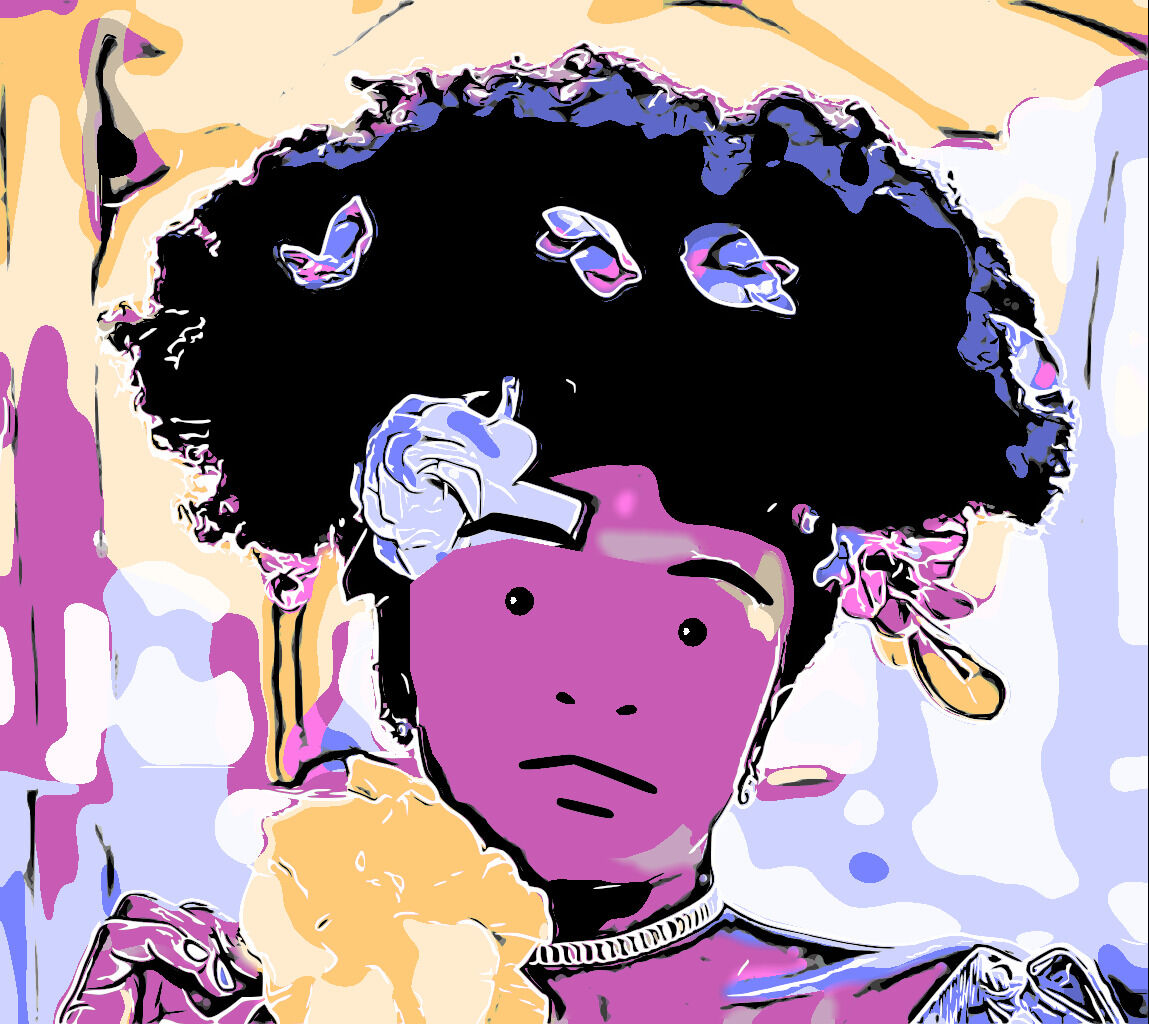}
\includegraphics[height=1.6cm]{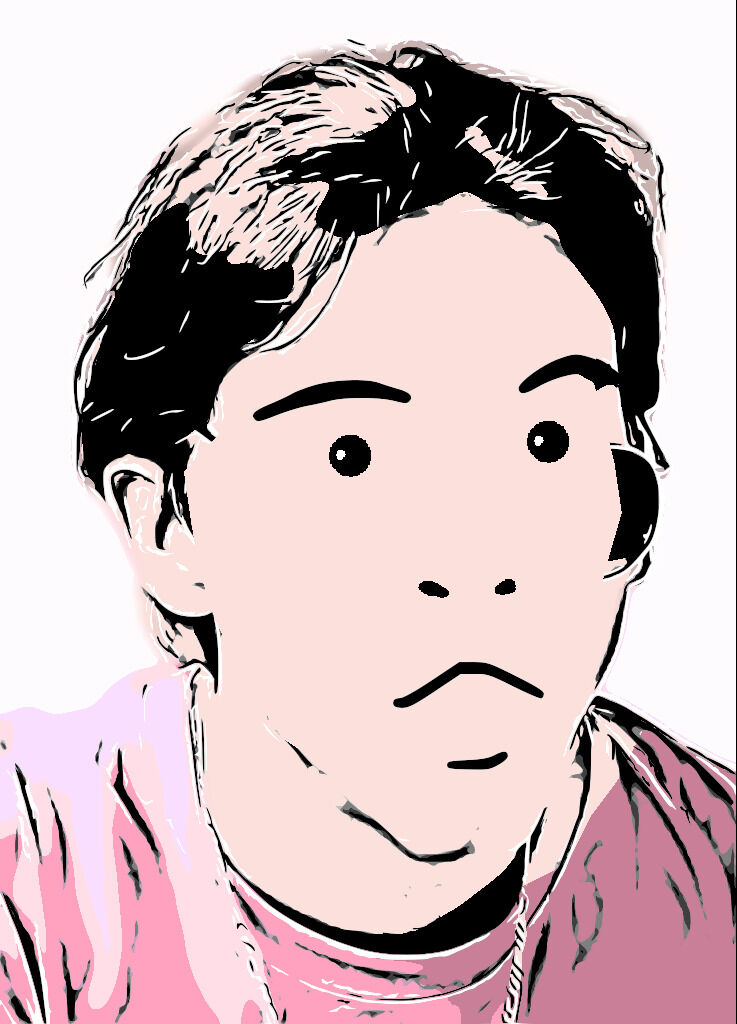}
\includegraphics[height=1.6cm]{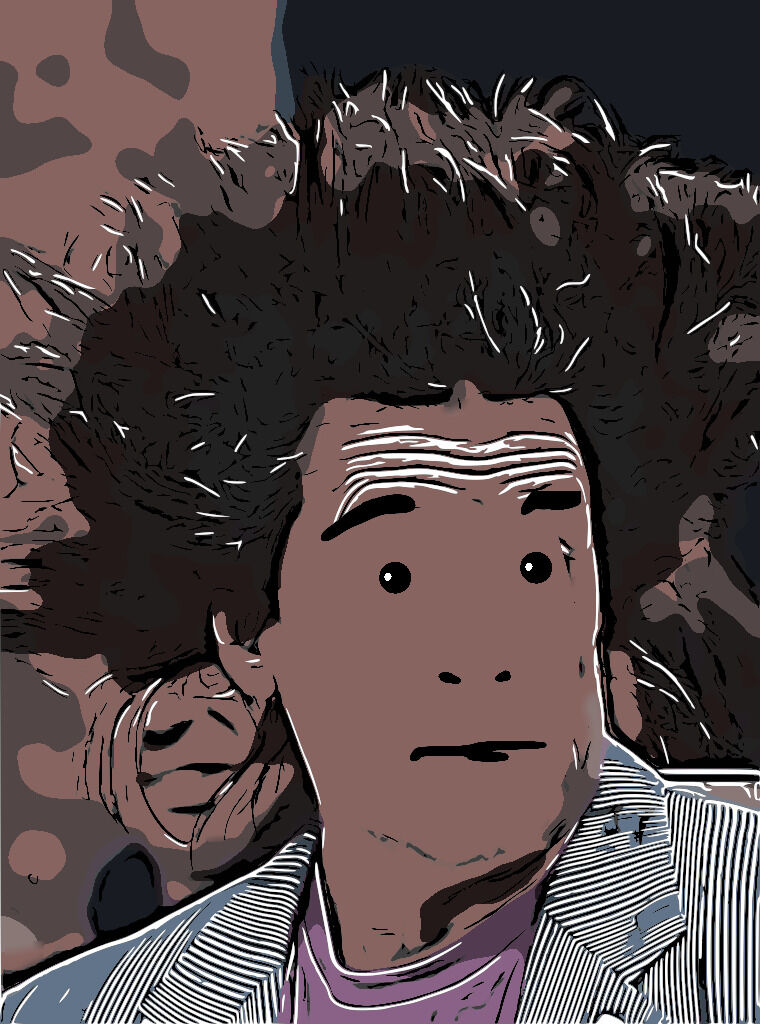}
\includegraphics[height=1.6cm]{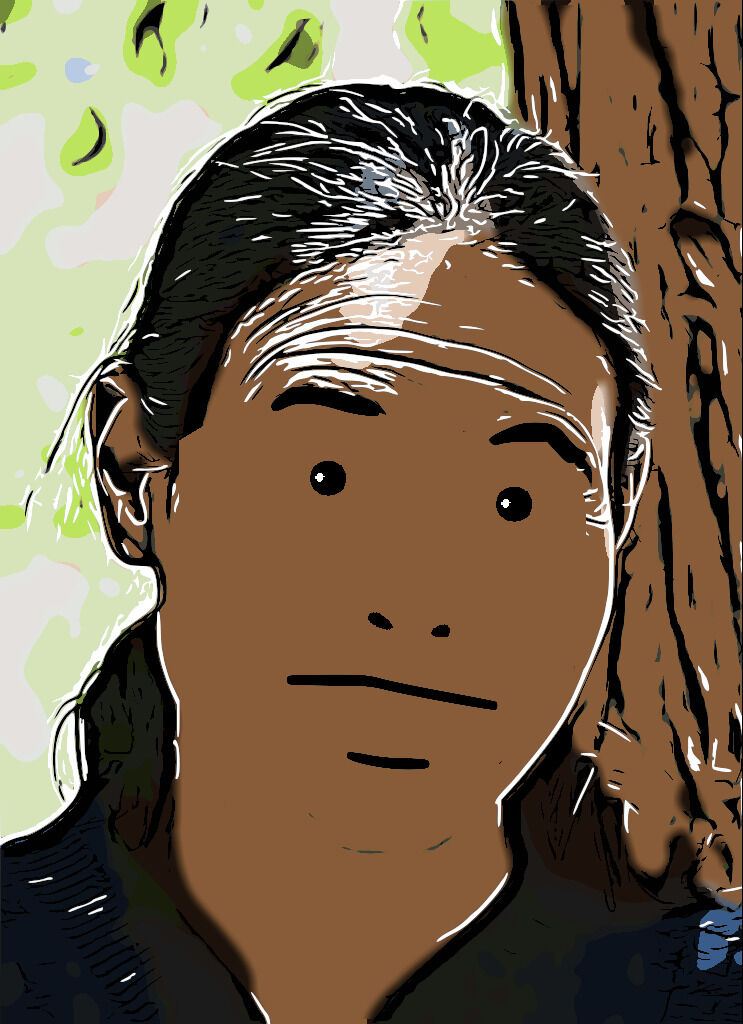}

\centerline{Level 3}
\medskip
\caption{Images from the \emph{NPRportrait1.0} benchmark stylised in the Julian Opie style: Rosin and Lai~\cite{rosin-portrait}}
\label{resultsopie}
\end{figure}


\begin{figure}[!t]
\centering
\includegraphics[height=1.6cm]{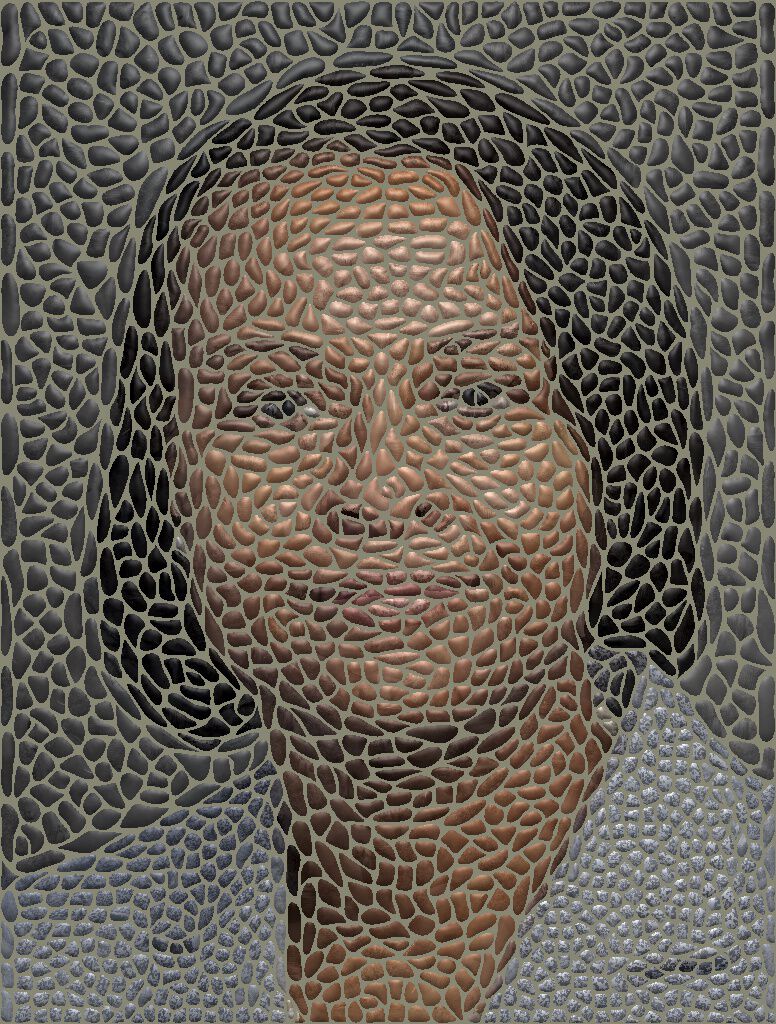}
\includegraphics[height=1.6cm]{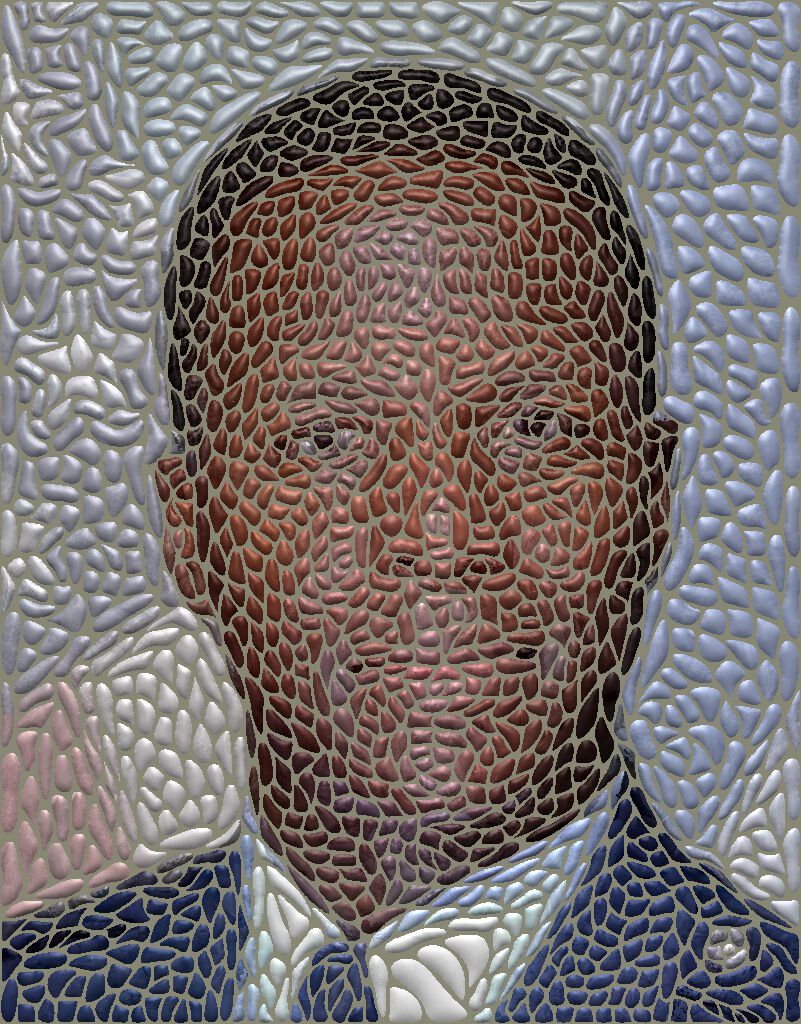}
\includegraphics[height=1.6cm]{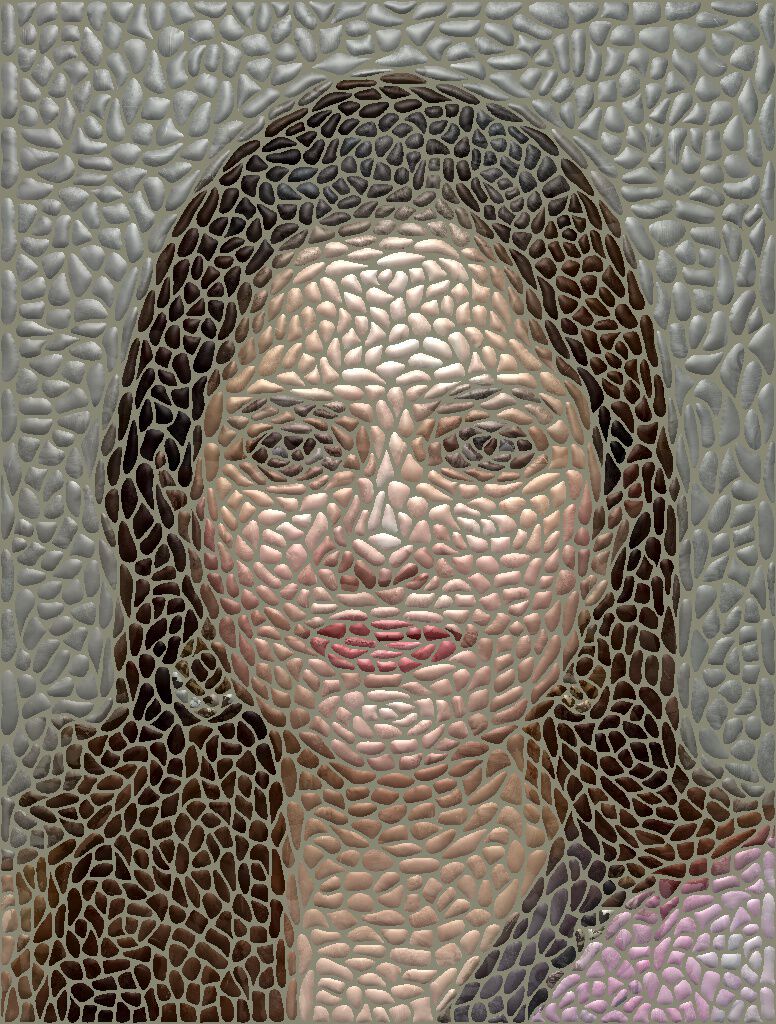}
\includegraphics[height=1.6cm]{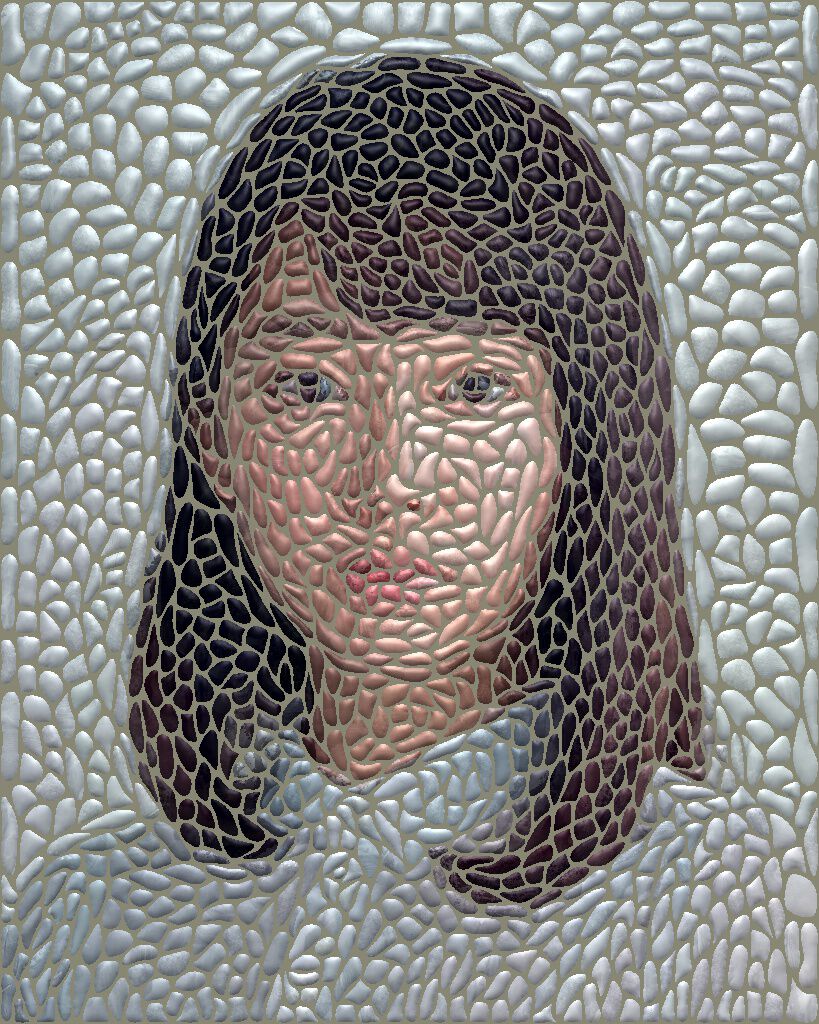}
\includegraphics[height=1.6cm]{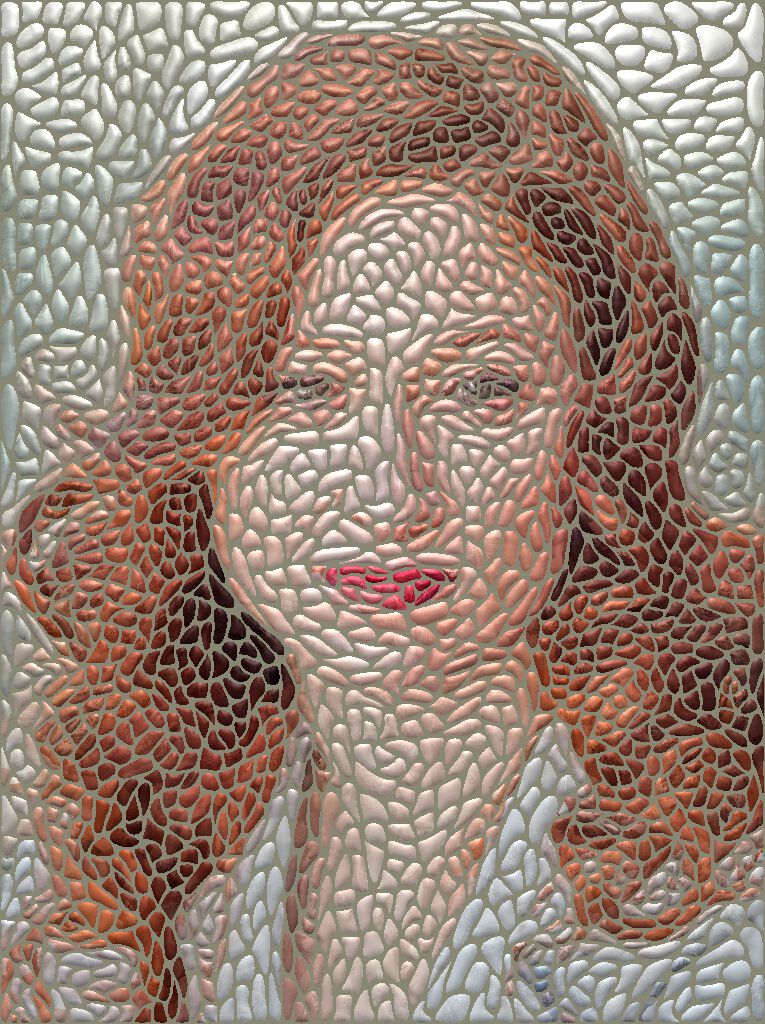}

\centerline{Level 1}
\medskip

\includegraphics[height=1.6cm]{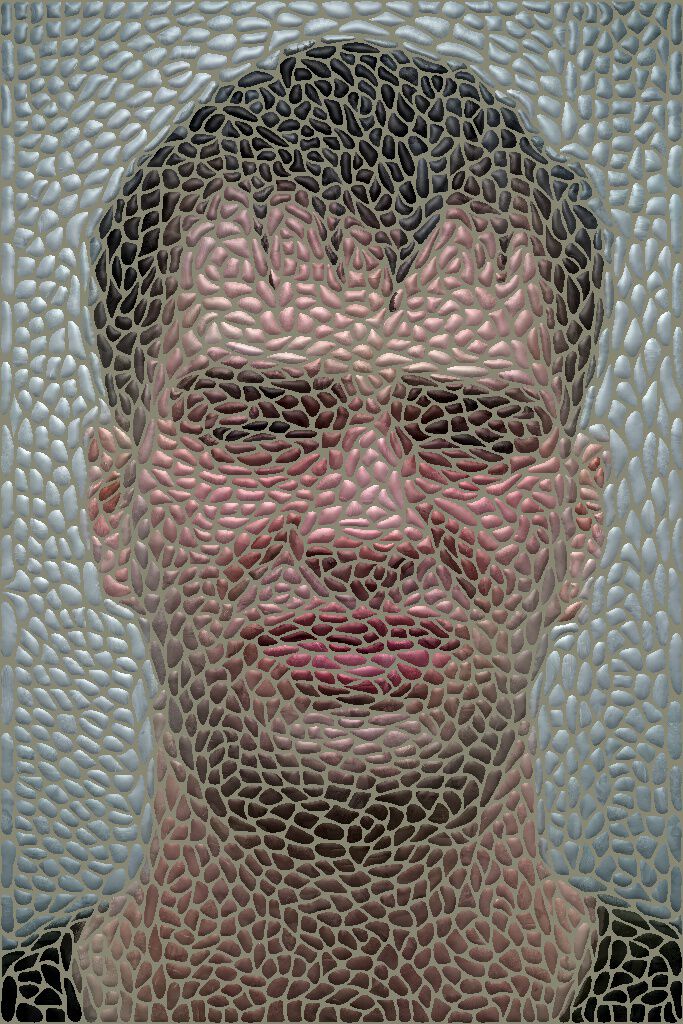}
\includegraphics[height=1.6cm]{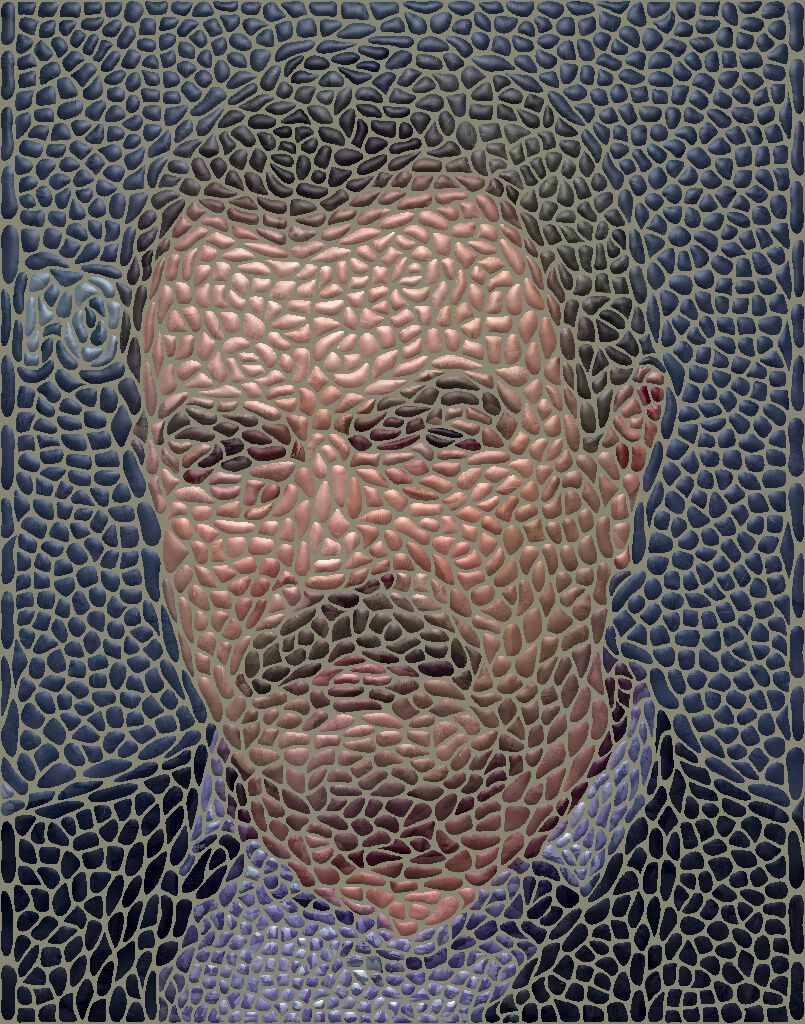}
\includegraphics[height=1.6cm]{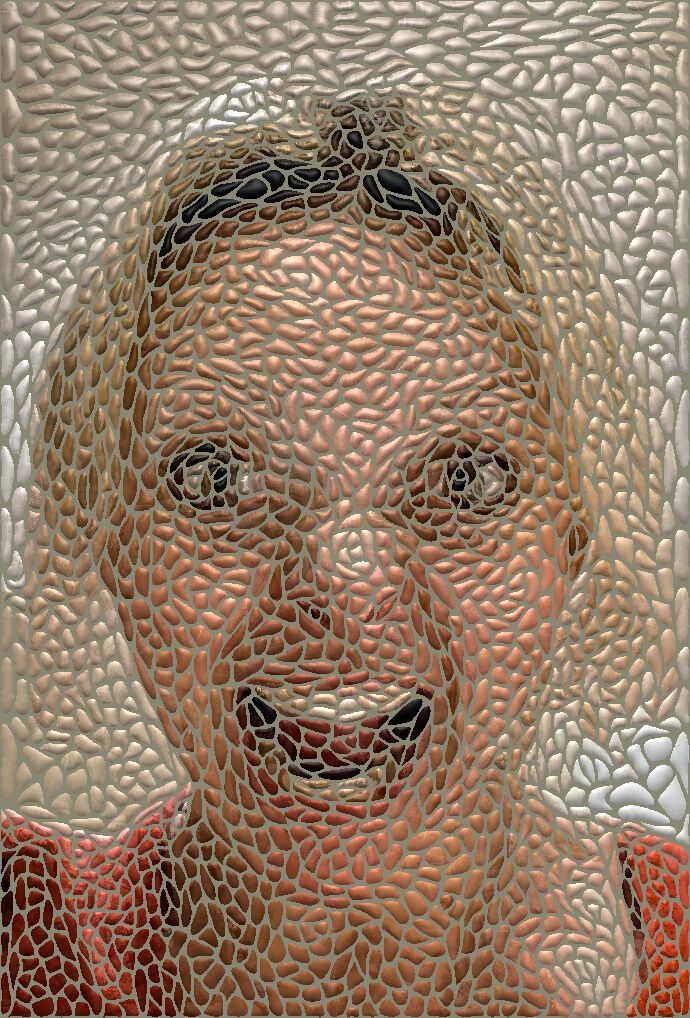}
\includegraphics[height=1.6cm]{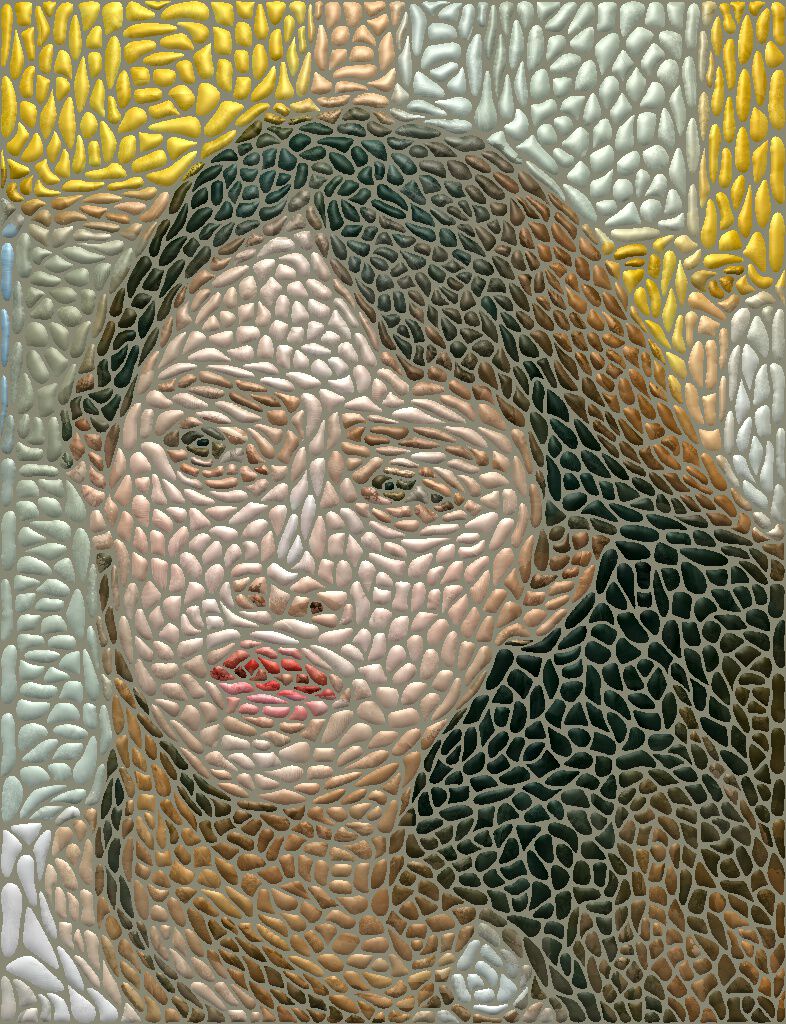}
\includegraphics[height=1.6cm]{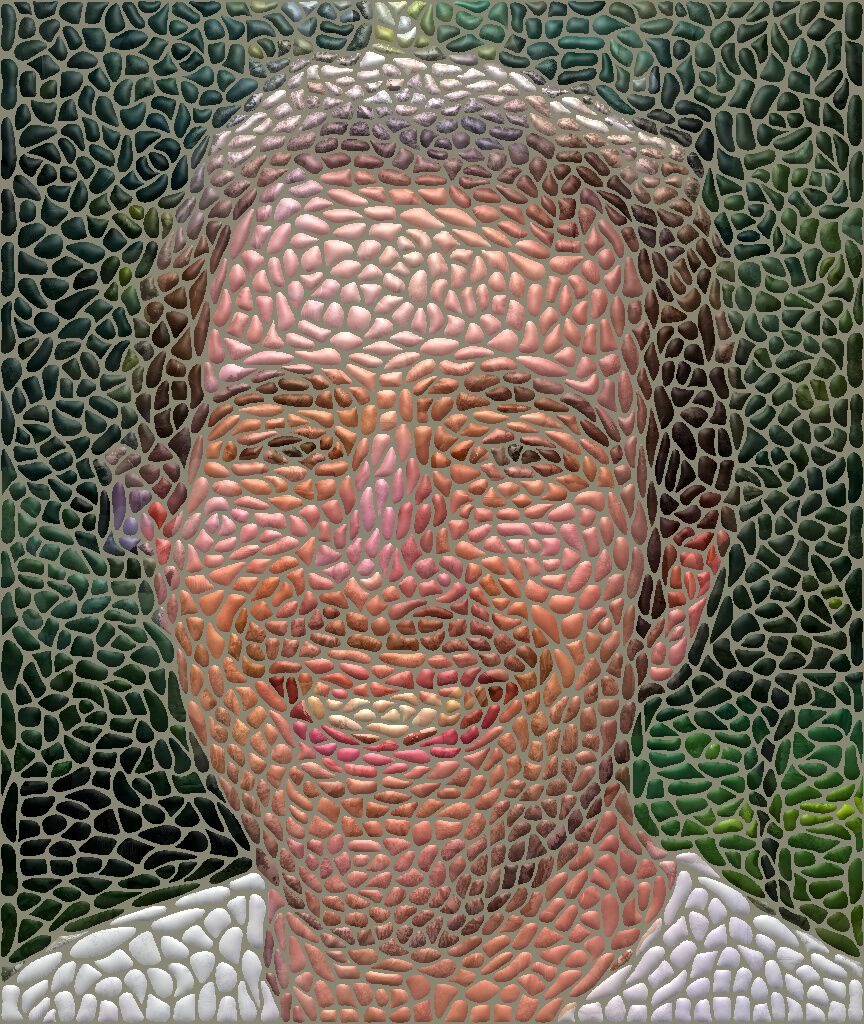}

\centerline{Level 2}
\medskip

\includegraphics[height=1.6cm]{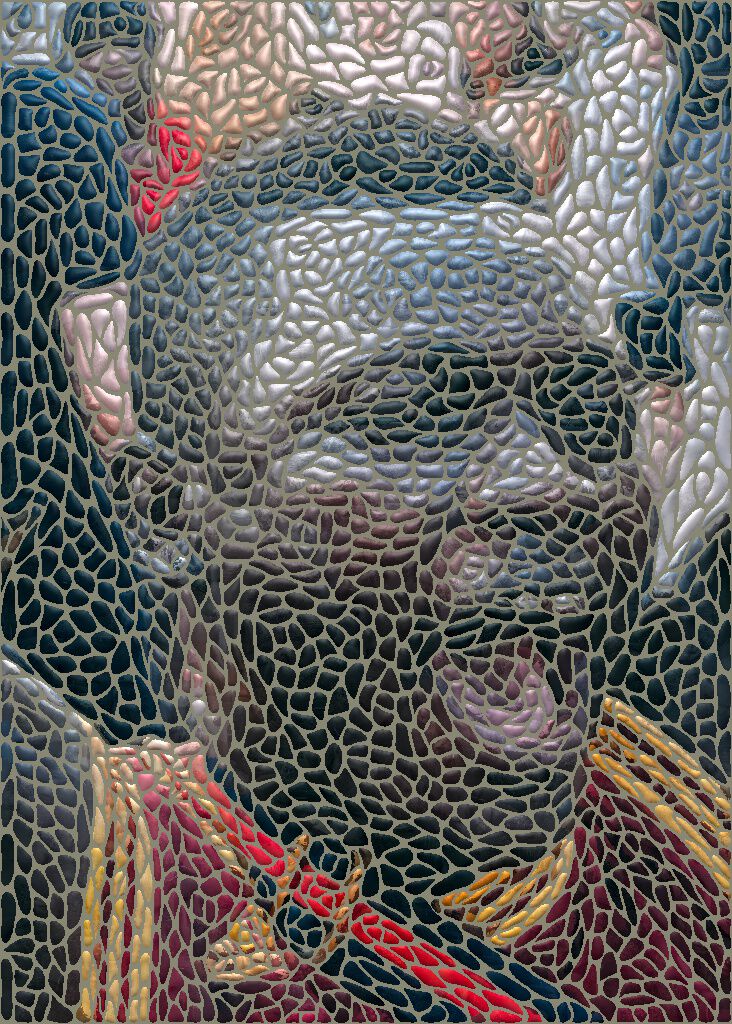}
\includegraphics[height=1.6cm]{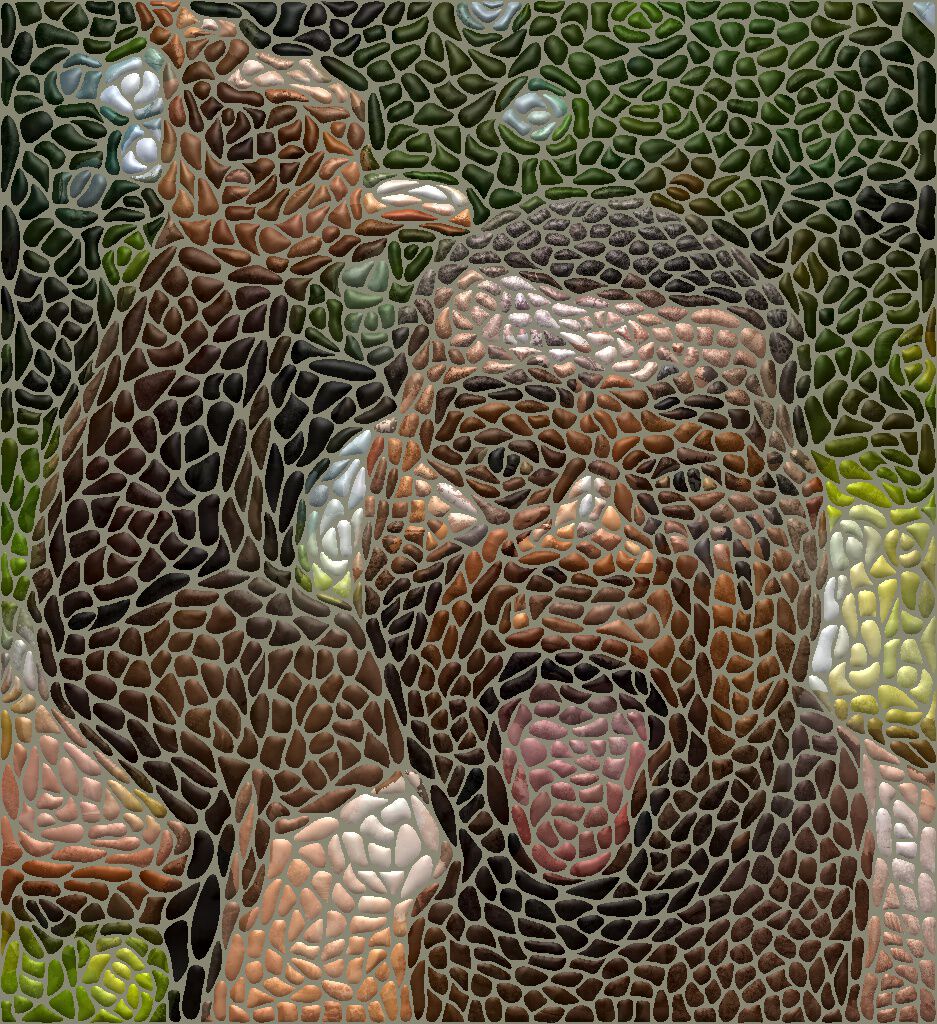}
\includegraphics[height=1.6cm]{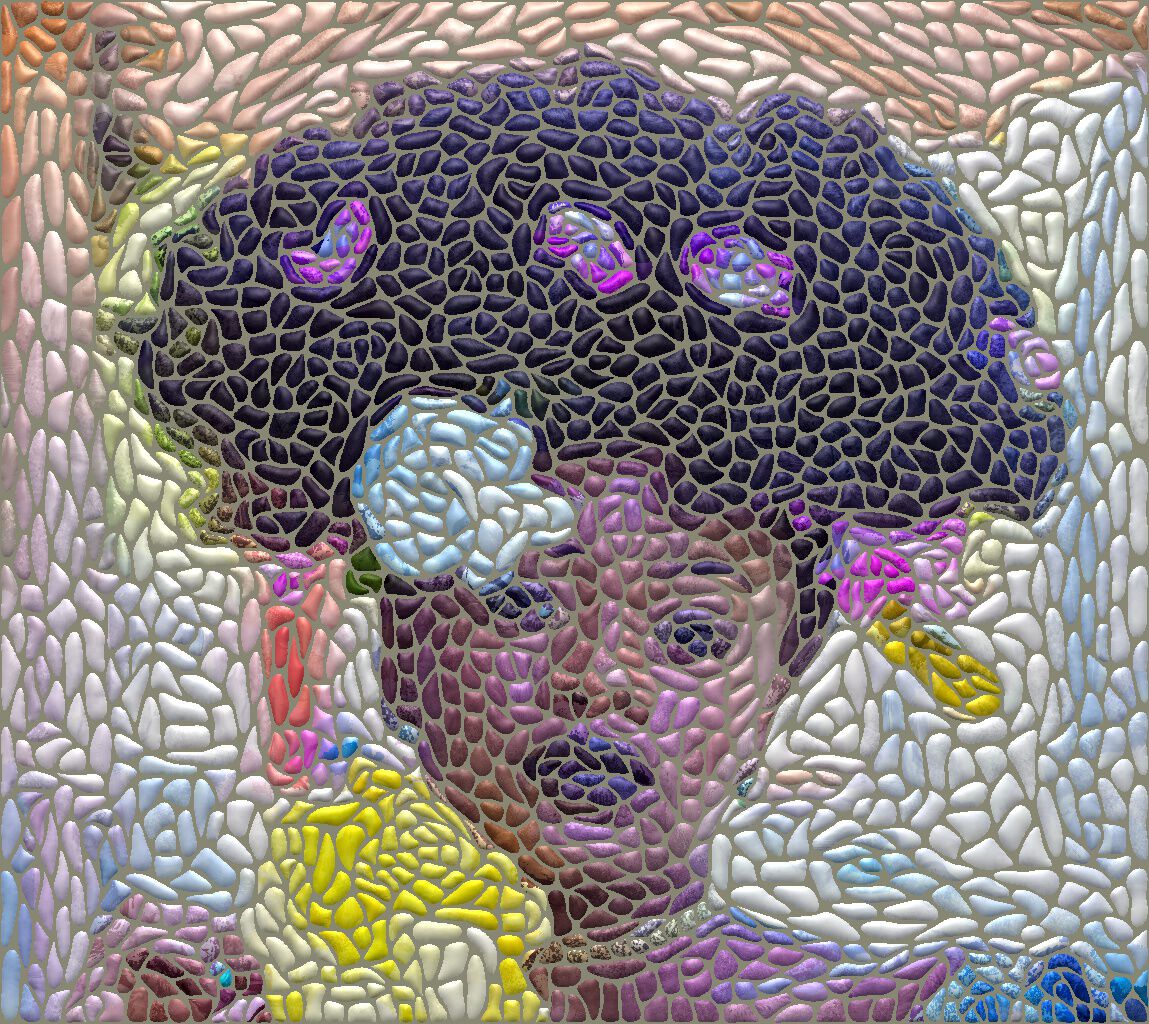}
\includegraphics[height=1.6cm]{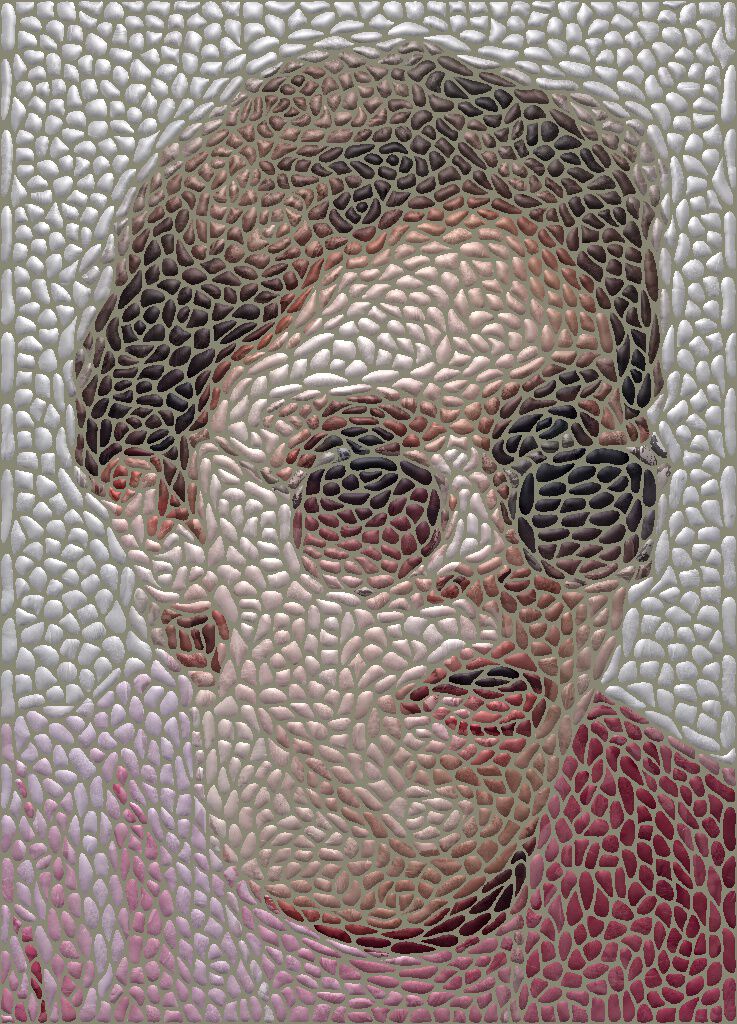}
\includegraphics[height=1.6cm]{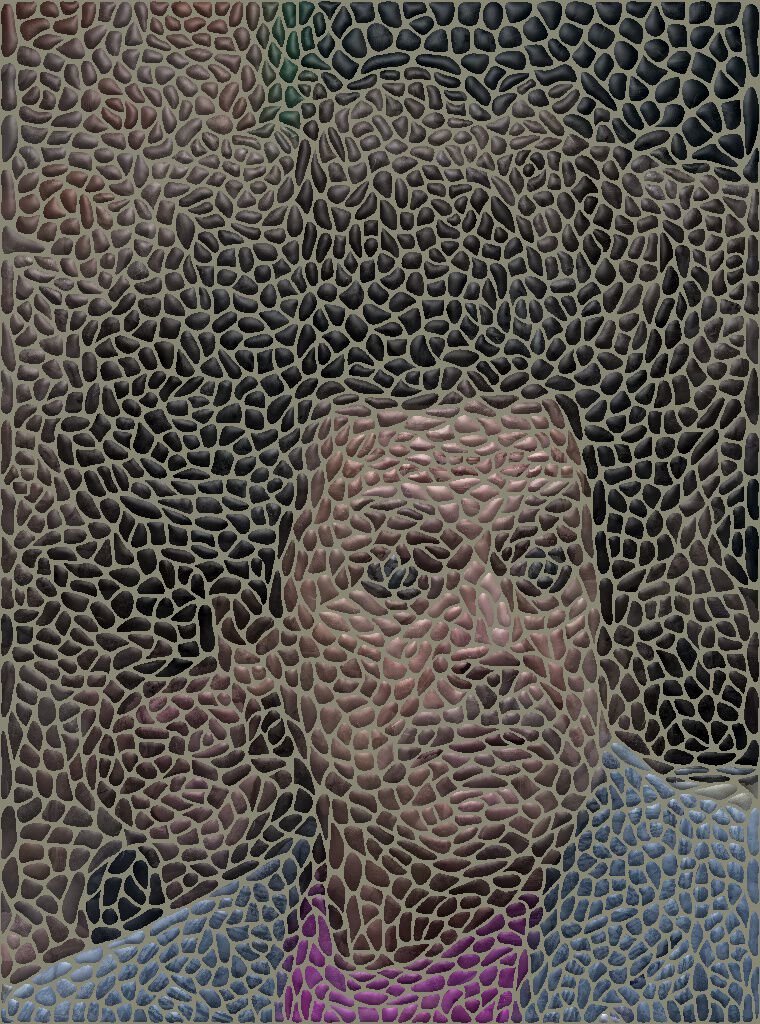}
\includegraphics[height=1.6cm]{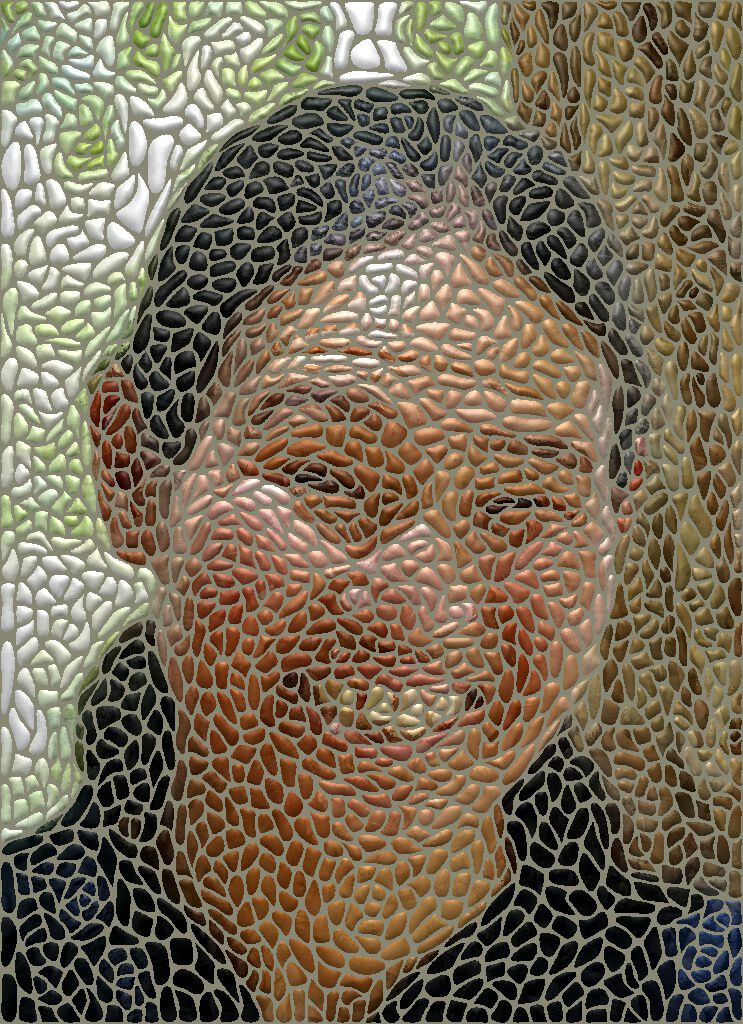}

\centerline{Level 3}
\medskip
\caption{Images from the \emph{NPRportrait1.0} benchmark stylised as pebble mosaics: Doyle \emph{et al.}~\cite{doyle2019automated}}
\label{resultspebble}
\end{figure}


\begin{figure}[!t]
\centering
\includegraphics[height=1.6cm]{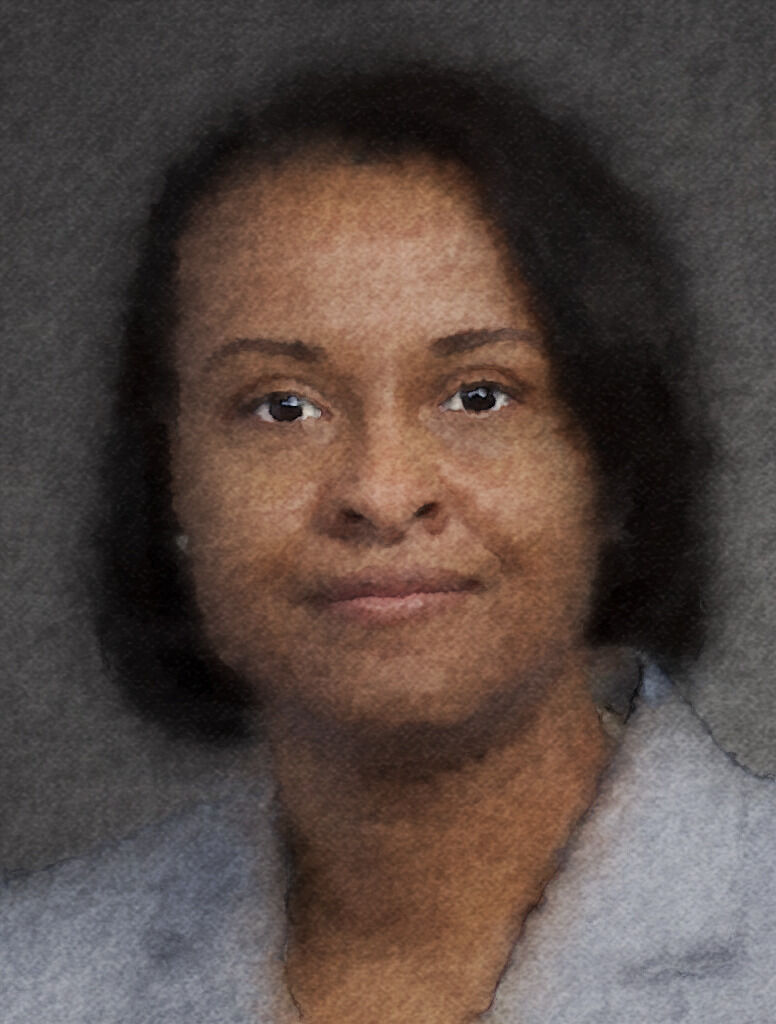}
\includegraphics[height=1.6cm]{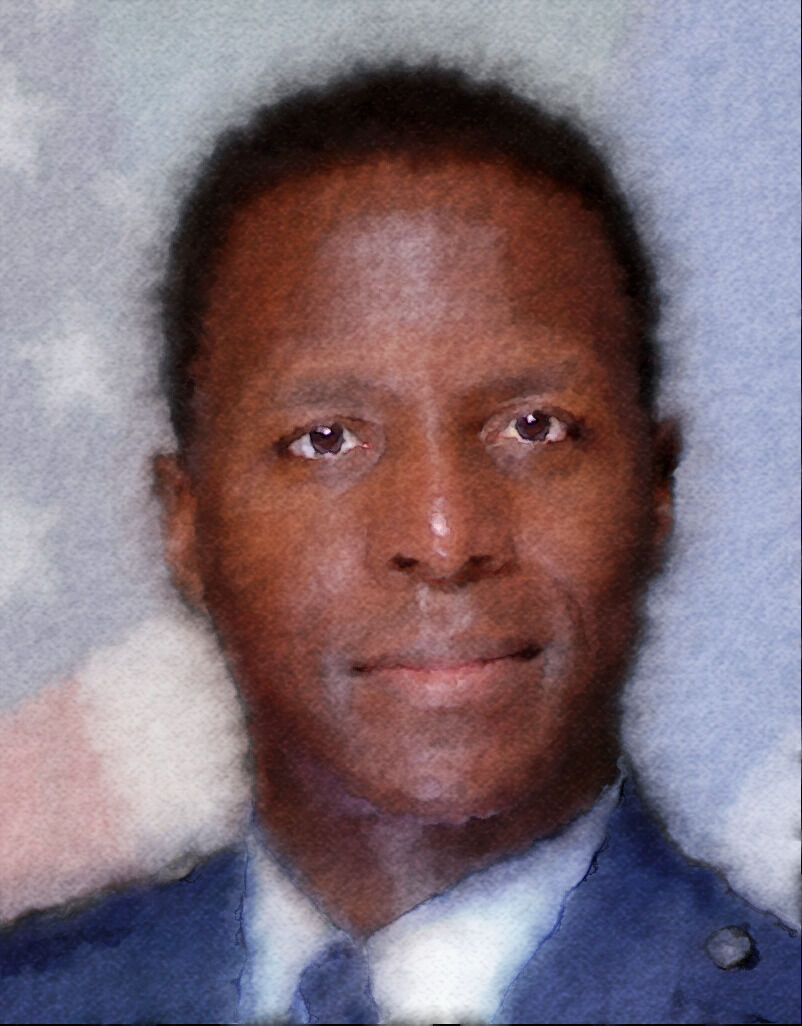}
\includegraphics[height=1.6cm]{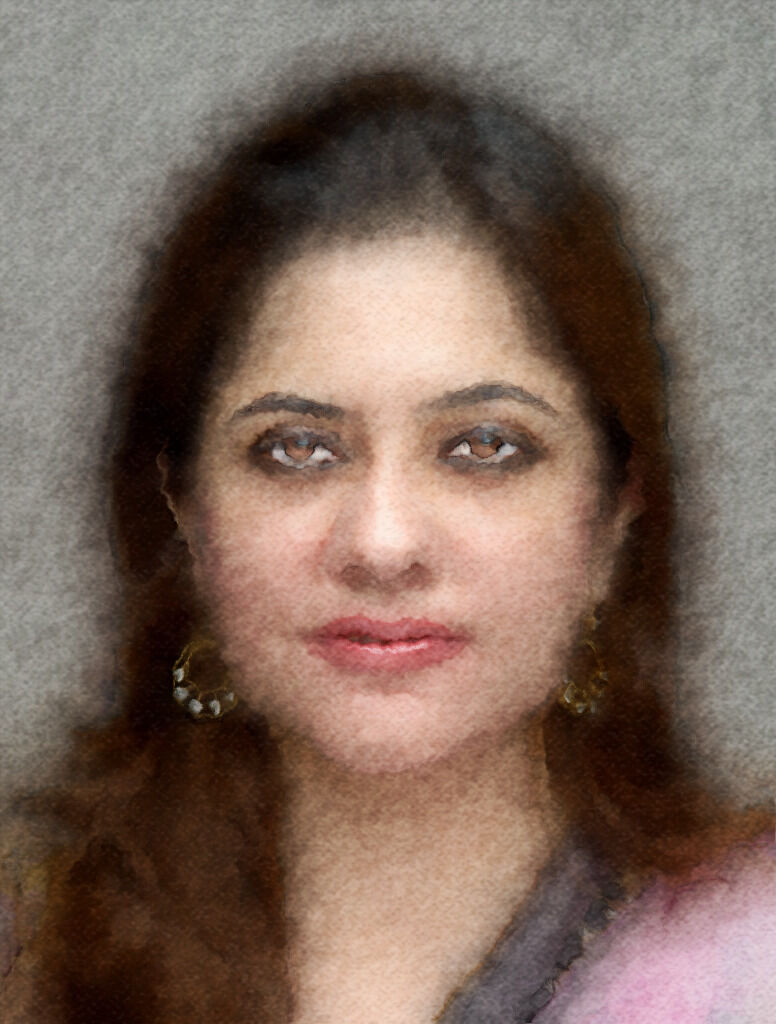}
\includegraphics[height=1.6cm]{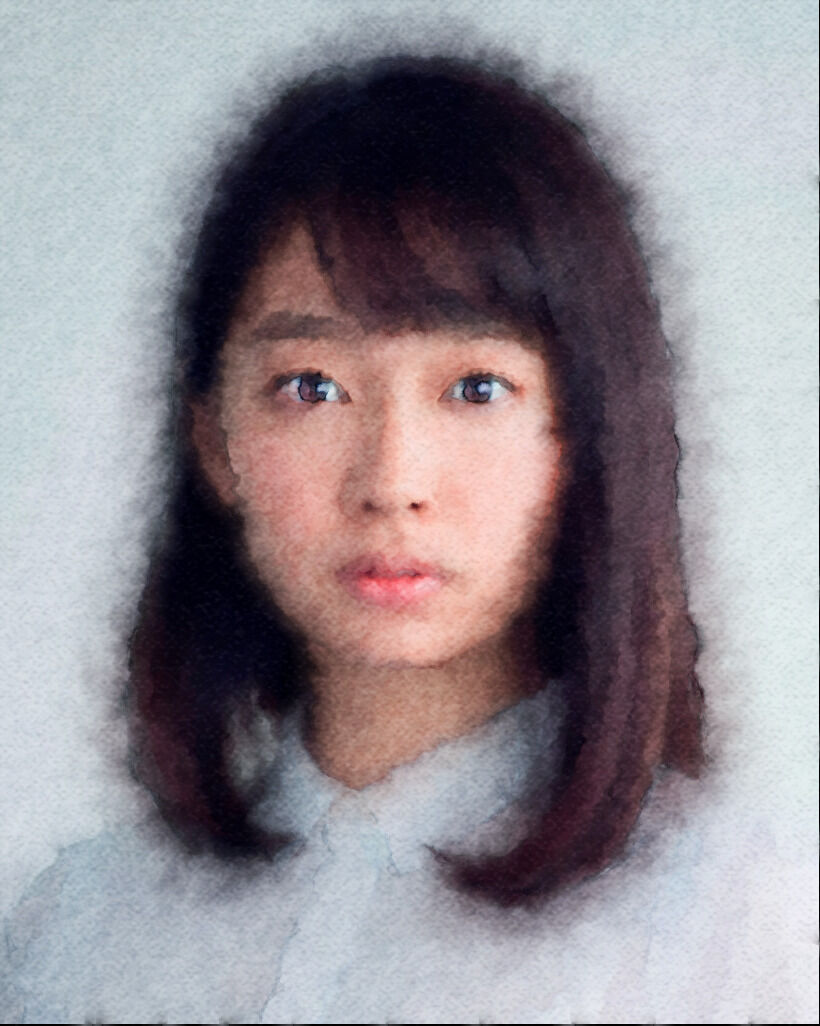}
\includegraphics[height=1.6cm]{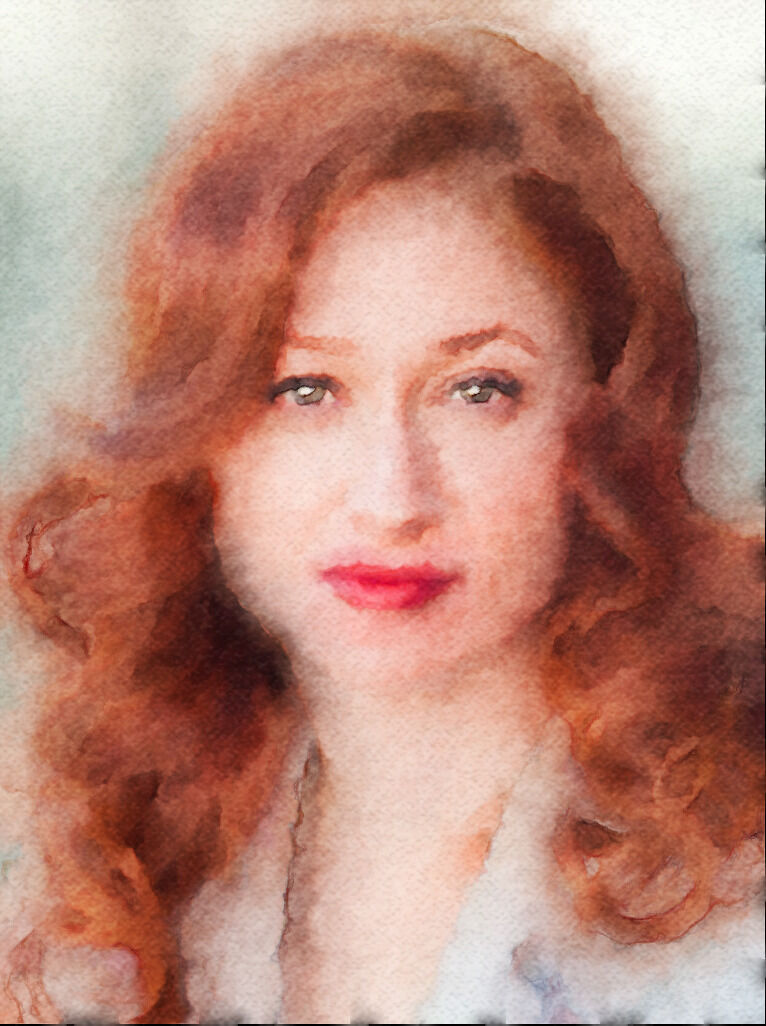}

\centerline{Level 1}
\medskip

\includegraphics[height=1.6cm]{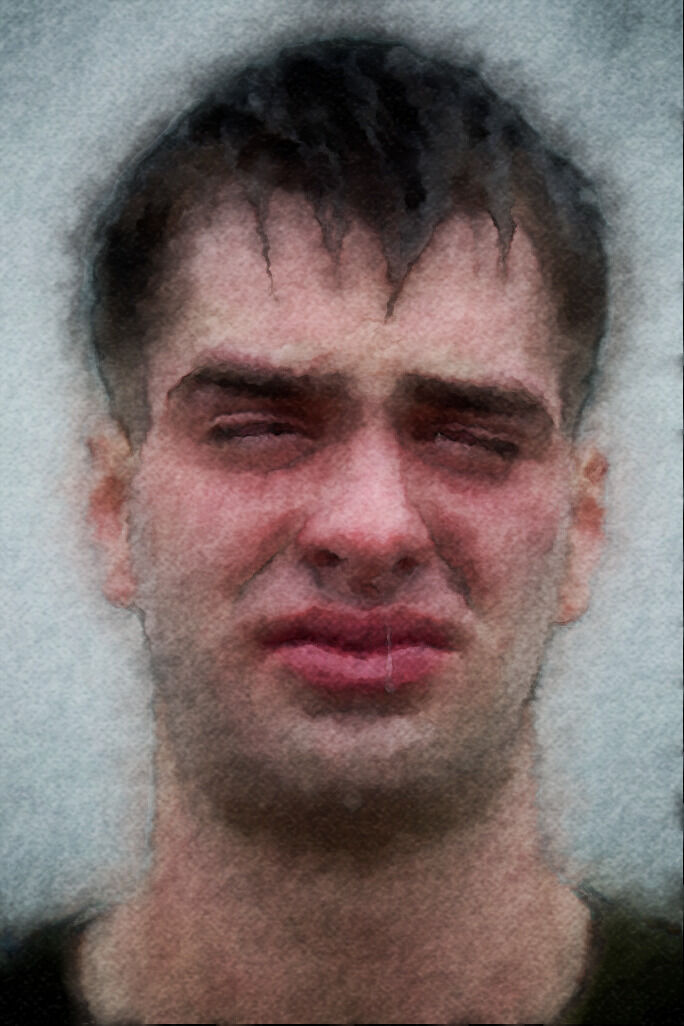}
\includegraphics[height=1.6cm]{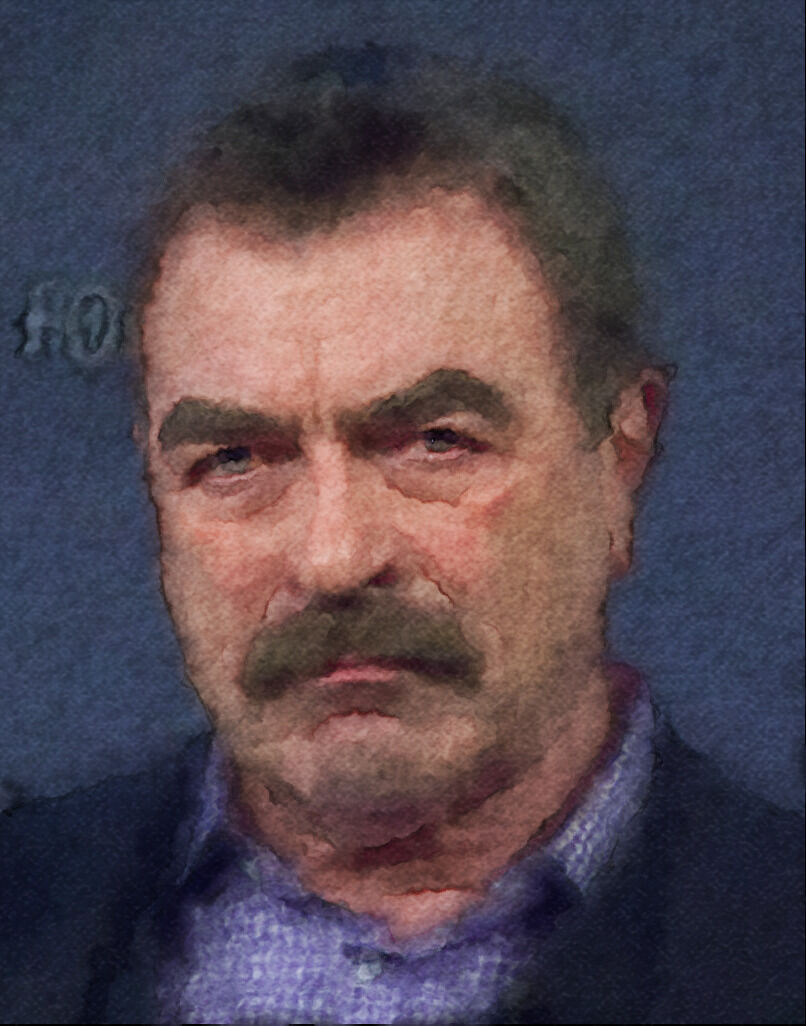}
\includegraphics[height=1.6cm]{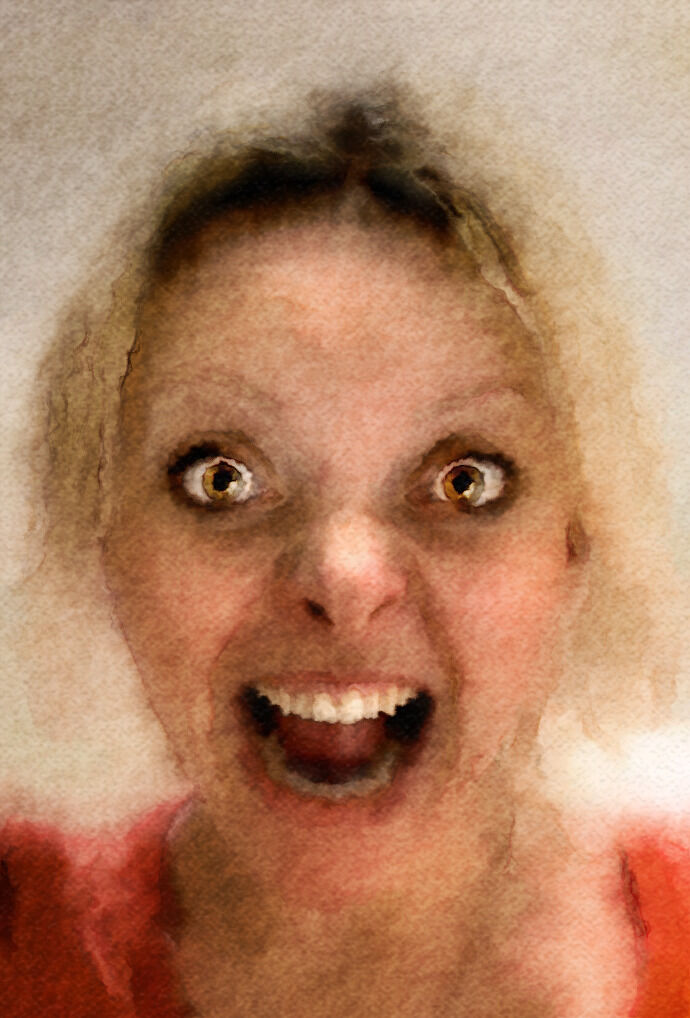}
\includegraphics[height=1.6cm]{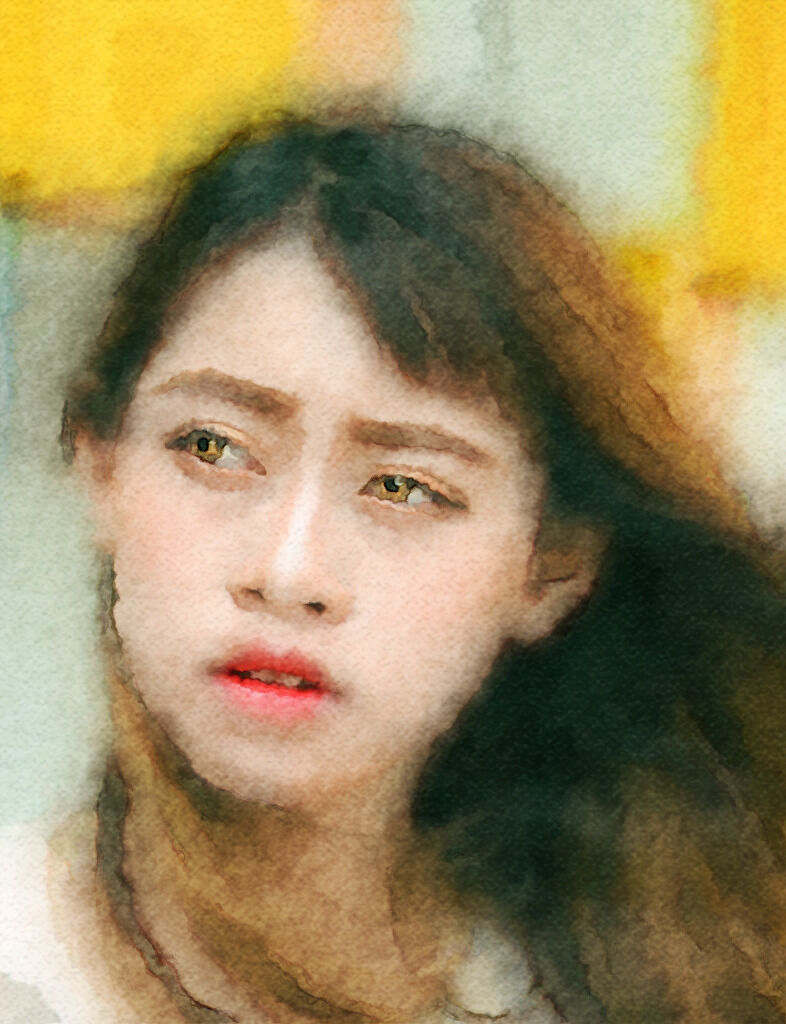}
\includegraphics[height=1.6cm]{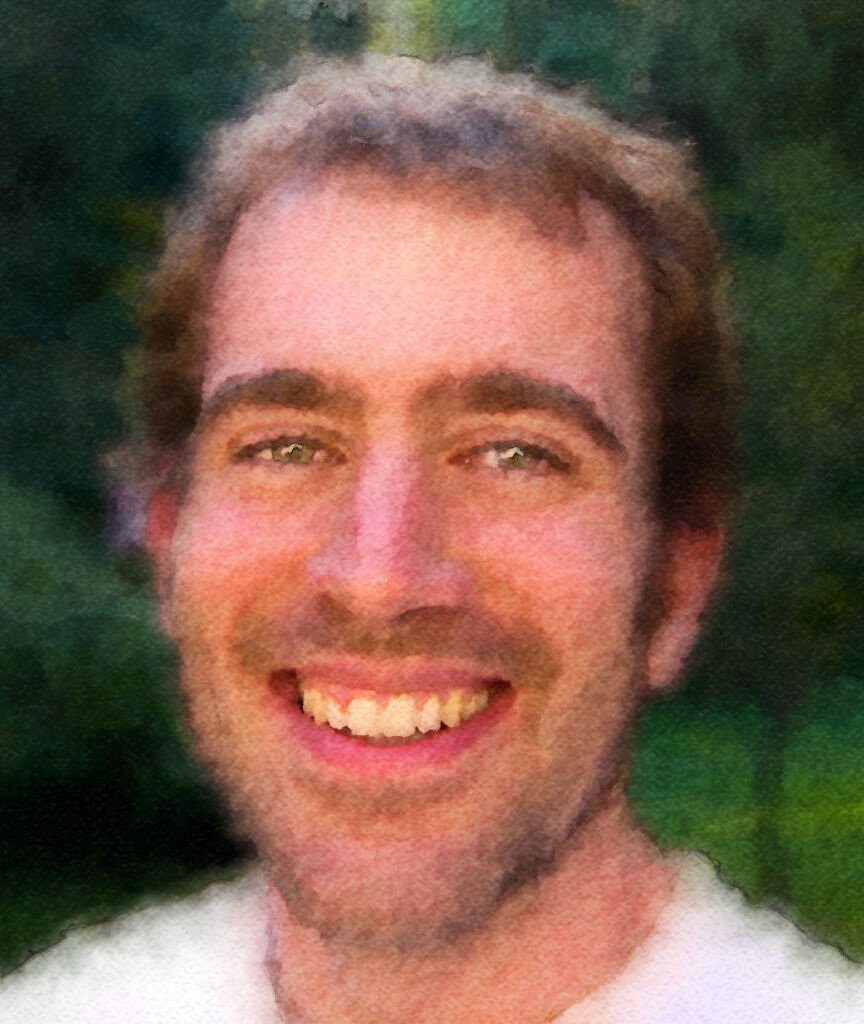}

\centerline{Level 2}
\medskip

\includegraphics[height=1.6cm]{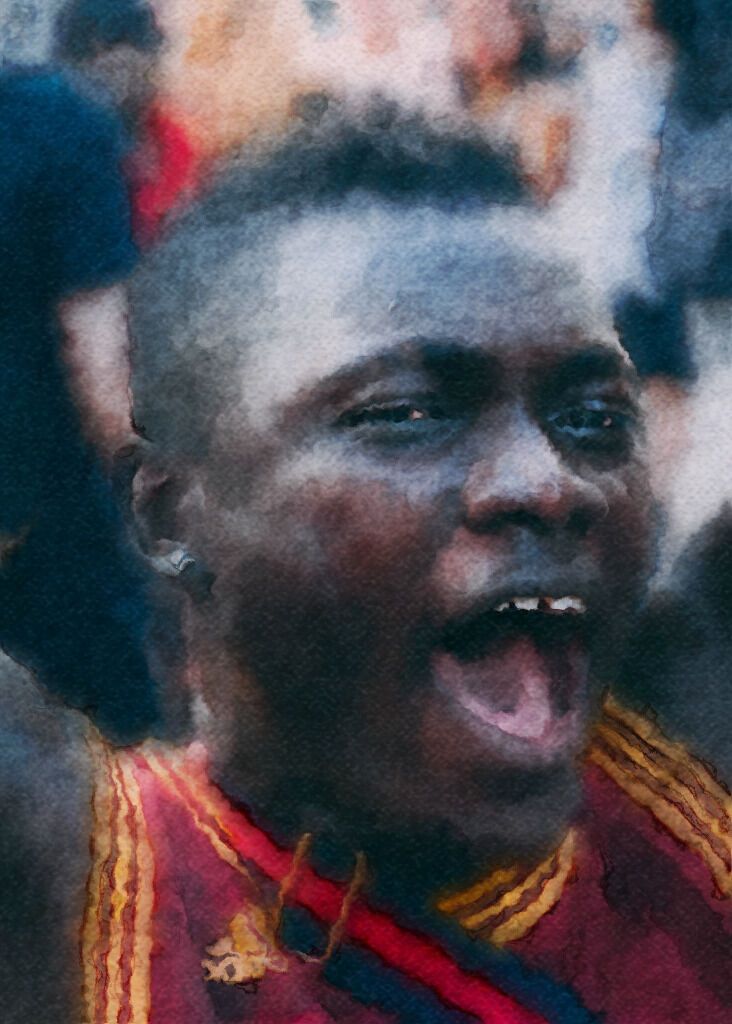}
\includegraphics[height=1.6cm]{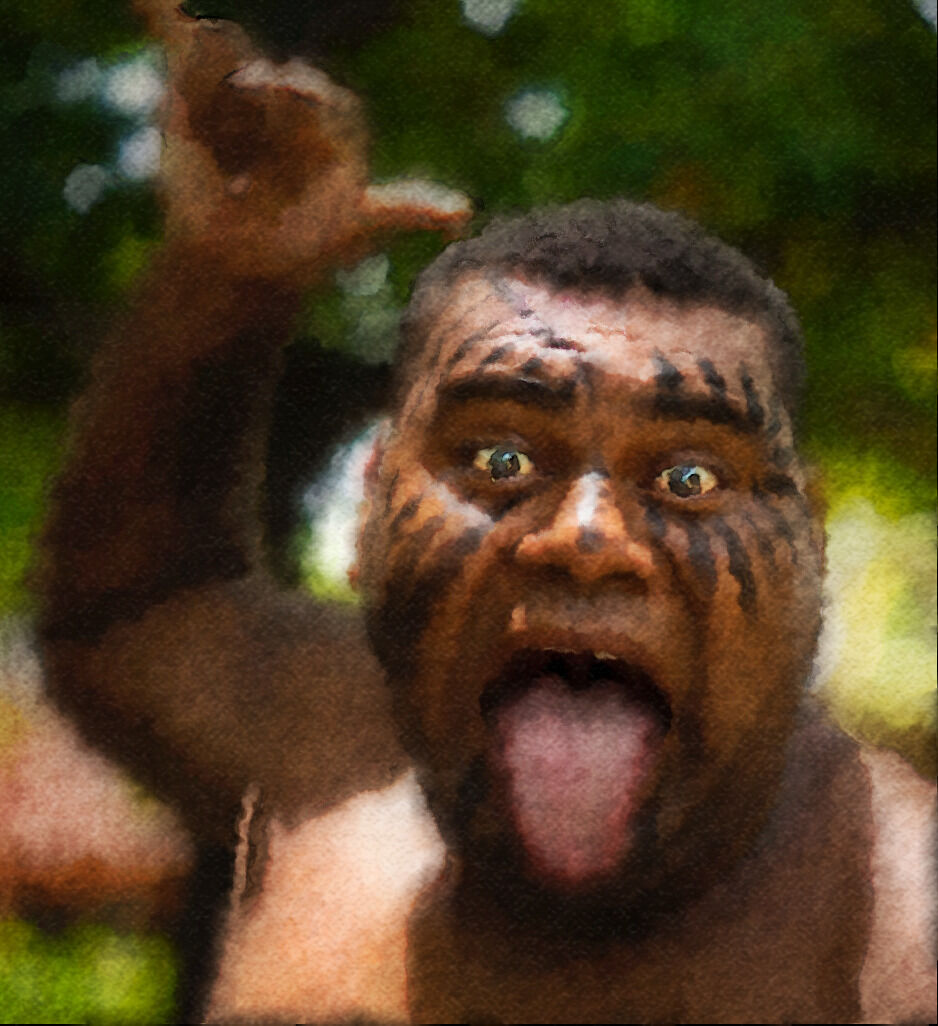}
\includegraphics[height=1.6cm]{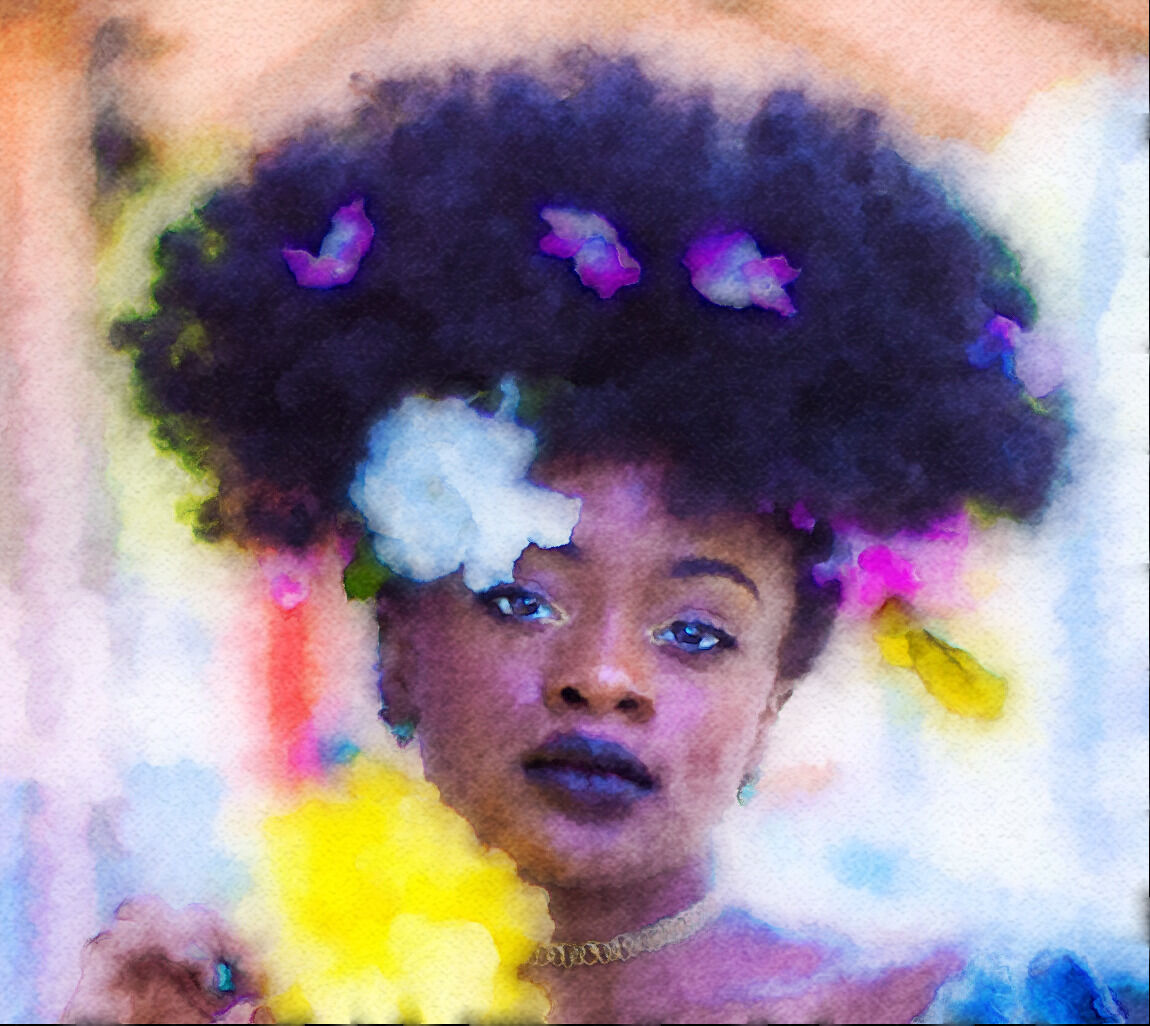}
\includegraphics[height=1.6cm]{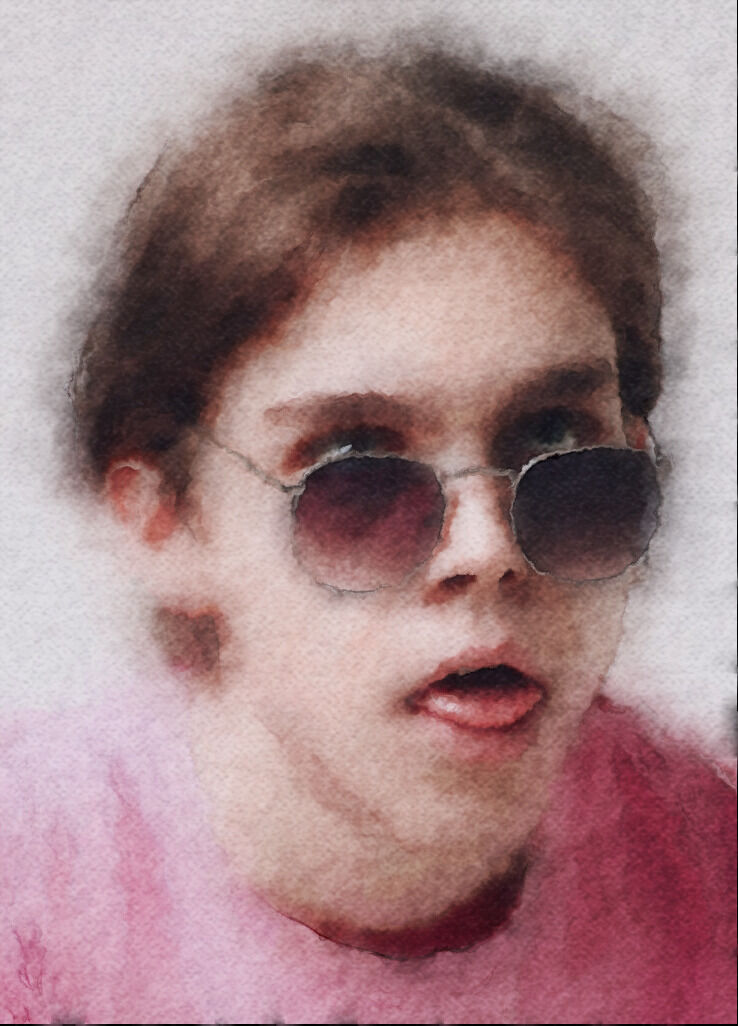}
\includegraphics[height=1.6cm]{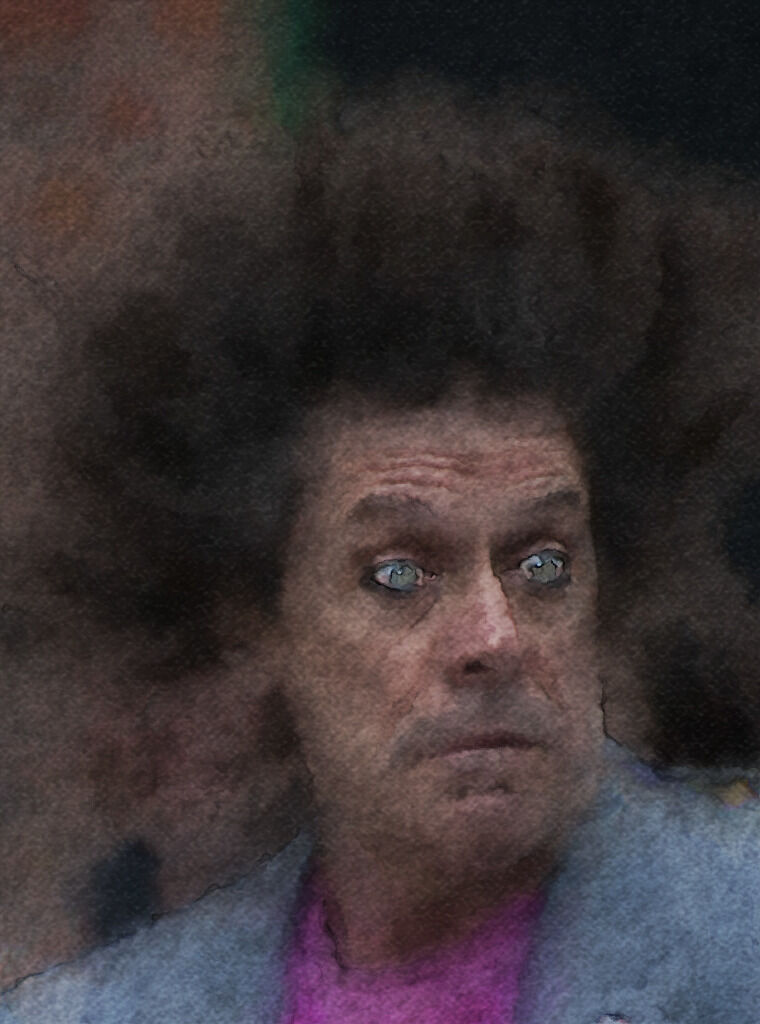}
\includegraphics[height=1.6cm]{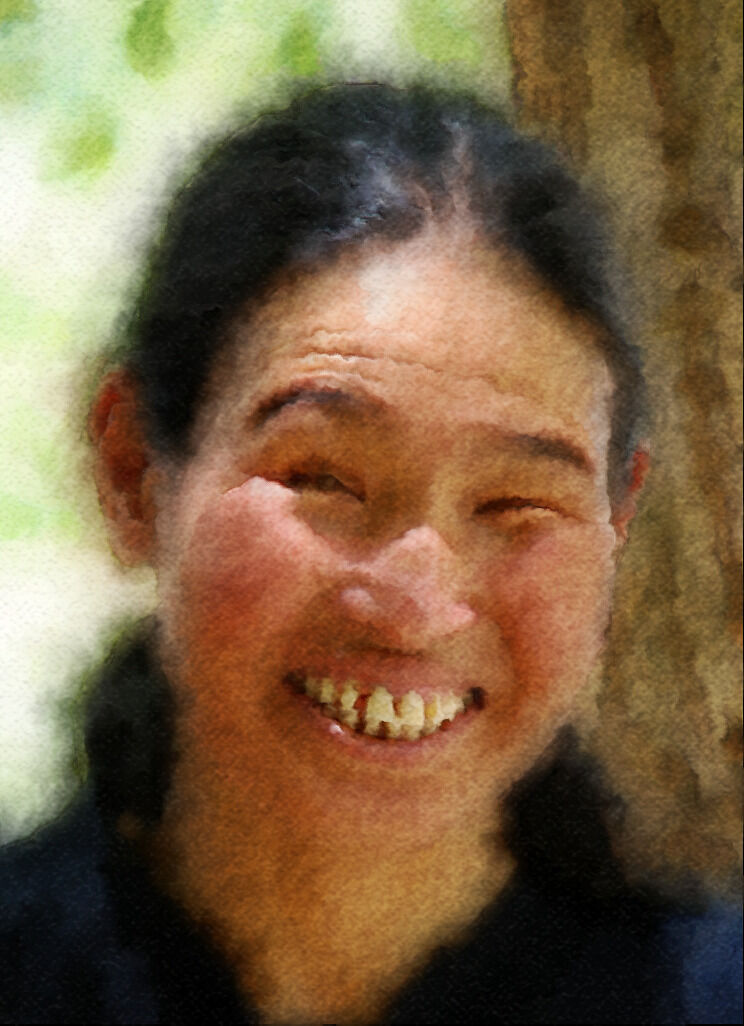}

\centerline{Level 3}
\medskip
\caption{Images from the \emph{NPRportrait1.0} benchmark stylised as watercolours: Rosin and Lai~\cite{rosin2017watercolour}}
\label{resultswatercolour}
\end{figure}

Experiment~2 described in section~\ref{experiments} is applied to the same 11 NPR algorithms as Experiment~1.
There are therefore $11 \times 20 \times 20 \times 20=88000$ possible stylised triplets.
The user study had 213 participants who saw 30 triples of images which are randomly generated with replacement,
leading to 6390 triples, of which 6171 triplets were unique.
Table~\ref{correlation} shows the Pearson and Kendall correlation coefficients;
the values confirm that general-purpose
filtering approaches such as XDoG~\cite{xdog} and oil painting~\cite{semmo2016image}
are not affected by the increasing complexity across the benchmark levels.
Although they are face-specific, watercolour~\cite{rosin2017watercolour} and engraving are also fairly robust
since their renderings are not highly dependent on the face model, and their results are reasonable despite inaccurate face detection.
The techniques with highest correlation to the levels are
neural style transfer~\cite{li2016combining},
which has a tendency to create
more spurious facial features (e.g. misplaced eyes)
as the images become more cluttered;
and both the line drawing methods (artistic sketch method~\cite{berger-shamir} and APDrawingGAN~\cite{YiLLR19})
which often produce fragmented or spurious lines
when there are variations in lighting.

\begin{figure*}[!tbp]
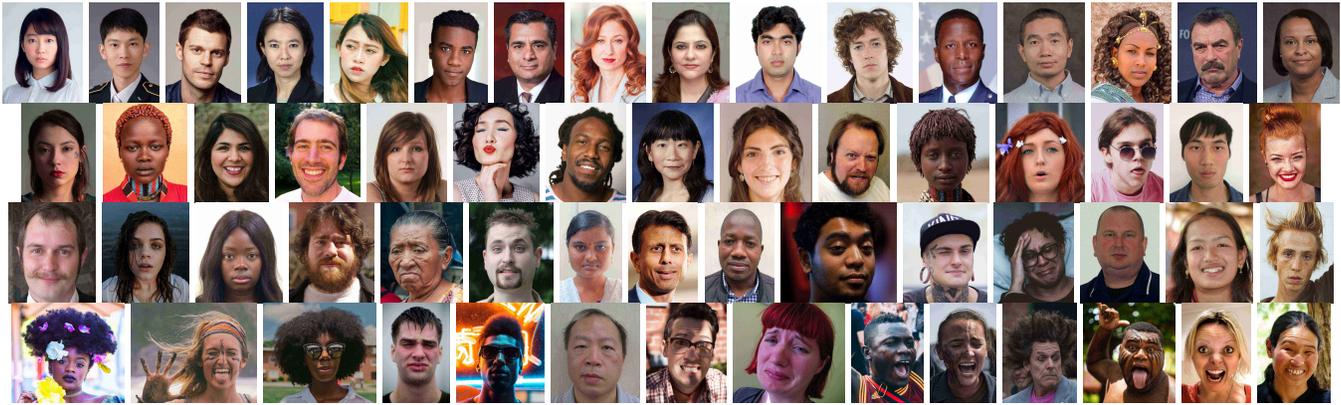

\centering
\includegraphics[height=1.33cm]{NPRportrait-level1new/15-Shimizu-kurumi}
\includegraphics[height=1.33cm]{NPRportrait-level1new/16-Yoon-Yeoil}
\includegraphics[height=1.33cm]{NPRportrait-level2new/18-Bjornar_Moxnes.jpg}
\includegraphics[height=1.33cm]{NPRportrait-level1new/11-Michelle-Chan}
\includegraphics[height=1.33cm]{NPRportrait-level2new/15-pexels-355020.jpg}
\includegraphics[height=1.33cm]{NPRportrait-level1new/06-Team-Karim}
\includegraphics[height=1.33cm]{NPRportrait-level1new/07-24930764475-3bfbc008f8-o}
\includegraphics[height=1.33cm]{NPRportrait-level1new/20-8552893573-1473209795-o}
\includegraphics[height=1.33cm]{NPRportrait-level1new/10-Saira-Shah-Halim}
\includegraphics[height=1.33cm]{NPRportrait-level1new/08-Moid-Rasheedi}
\includegraphics[height=1.33cm]{NPRportrait-level1new/18-Huw-Evans}
\includegraphics[height=1.33cm]{NPRportrait-level1new/05-Major-General-Edward-Rice}
\includegraphics[height=1.33cm]{NPRportrait-level1new/12-38753438484-f65be5b961-o}
\includegraphics[height=1.33cm]{NPRportrait-level2new/03-34694233905_6f774ba164_o.jpg}
\includegraphics[height=1.33cm]{NPRportrait-level2new/05-Tom_Selleck_at_PaleyFest_2014.jpg}
\includegraphics[height=1.33cm]{NPRportrait-level1new/01-Kimberly-Howell}
\includegraphics[height=1.33cm]{NPRportrait-level3new/03-felipe-sagn-2xd0_6wEj6k-unsplash}
\includegraphics[height=1.33cm]{NPRportrait-level1new/02-22636558997-80ee36b602-o}
\includegraphics[height=1.33cm]{NPRportrait-level2new/04-33916867564_c6af07fd52_o.jpg}
\includegraphics[height=1.33cm]{NPRportrait-level2new/20-3900823607_9f9d376bb8_o.jpg}
\includegraphics[height=1.33cm]{NPRportrait-level1new/17-4891358118-32c20e9d8e-o}
\includegraphics[height=1.33cm]{NPRportrait-level3new/07-azamat-zhanisov-h9Oo45soK_0-unsplash}
\includegraphics[height=1.33cm]{NPRportrait-level2new/16-14610754861_8fd4939744_k.jpg}
\includegraphics[height=1.33cm]{NPRportrait-level1new/13-6884042760-1ee2b00829-o}
\includegraphics[height=1.33cm]{NPRportrait-level2new/08-16022313531_5aa50d89d3_o.jpg}
\includegraphics[height=1.33cm]{NPRportrait-level2new/09-306039906_24c5e9600c_o.jpg}
\includegraphics[height=1.33cm]{NPRportrait-level2new/06-15158628486_c93546d21d_o.jpg}
\includegraphics[height=1.33cm]{NPRportrait-level2new/19-7741742246_502b485404_h.jpg}
\includegraphics[height=1.33cm]{NPRportrait-level3new/11-artyom-kim-zEa75CRX88M-unsplash}
\includegraphics[height=1.33cm]{NPRportrait-level1new/14-Rosin}
\includegraphics[height=1.33cm]{NPRportrait-level3new/16-gabriel-silverio-u3WmDyKGsrY-unsplash}
\includegraphics[height=1.33cm]{NPRportrait-level2new/07-9467963321_63d0375465_o.jpg}
\includegraphics[height=1.33cm]{NPRportrait-level2new/13-noah-buscher-_E-ogRrpM0s-unsplash.jpg}
\includegraphics[height=1.33cm]{NPRportrait-level1new/04-Rosin}
\includegraphics[height=1.33cm]{NPRportrait-level2new/11-2389874253_759d8164ab_o.jpg}
\includegraphics[height=1.33cm]{NPRportrait-level2new/17-wikimedia.jpg}
\includegraphics[height=1.33cm]{NPRportrait-level2new/14-230501453_85963e7ee3_o.jpg}
\includegraphics[height=1.33cm]{NPRportrait-level1new/09-Aswini-Phy-ALC}
\includegraphics[height=1.33cm]{NPRportrait-level3new/18-16585637733_67b0e18bcf_o}
\includegraphics[height=1.33cm]{NPRportrait-level1new/03-Rosin}
\includegraphics[height=1.33cm]{NPRportrait-level3new/17-6262243021_47792d9ca0_o}
\includegraphics[height=1.33cm]{NPRportrait-level3new/06-olesya-yemets-AjilVpkggN8-unsplash}
\includegraphics[height=1.33cm]{NPRportrait-level3new/08-claudia-owBcefxgrIE-unsplash}
\includegraphics[height=1.33cm]{NPRportrait-level1new/19-Zboralski-waldemar-2012}
\includegraphics[height=1.33cm]{NPRportrait-level2new/12-14452459445_1bb06c6302_o.jpg}
\includegraphics[height=1.33cm]{NPRportrait-level3new/04-nathan-dumlao-cibBnDQ9hcQ-unsplash}
\includegraphics[height=1.33cm]{NPRportrait-level3new/10-calvin-lupiya-Mx4auh5zO4w-unsplash}
\includegraphics[height=1.33cm]{NPRportrait-level3new/12-jordan-bauer-Is3VRzUaXVk-unsplash}
\includegraphics[height=1.33cm]{NPRportrait-level3new/14-andrew-robinson-4ar-CSxLcMg-unsplash}
\includegraphics[height=1.33cm]{NPRportrait-level2new/01-wikimedia.jpg}
\includegraphics[height=1.33cm]{NPRportrait-level3new/13-alex-iby-470eBDOc8bk-unsplash}
\includegraphics[height=1.33cm]{NPRportrait-level2new/02-8115174843_7abf278359_o.jpg}
\includegraphics[height=1.33cm]{NPRportrait-level3new/09-8079036040_6e5d7798f5_o}
\includegraphics[height=1.33cm]{NPRportrait-level3new/19-1824233430_59f1a20f0d_o}
\includegraphics[height=1.33cm]{NPRportrait-level3new/01-johanna-buguet-9GOAzu0G4oM-unsplash}
\includegraphics[height=1.33cm]{NPRportrait-level3new/02-29881657997_24106d0716_o}
\includegraphics[height=1.33cm]{NPRportrait-level3new/15-8717570008_edc9120e59_o}
\includegraphics[height=1.33cm]{NPRportrait-level3new/05-old-youth-tAJog0uJkT0-unsplash}
\includegraphics[height=1.33cm]{NPRportrait-level2new/10-3761108471_08e3f9f80d_o.jpg}
\includegraphics[height=1.33cm]{NPRportrait-level3new/20-3683799501_052eb48752_o}
\caption{\emph{NPRportrait1.0} benchmark ranked according to Experiment~2 aggregated over all 11 NPR styles.}
\label{rankingAllStyles}
\end{figure*}

\begin{figure}[tbp]
\centering
\includegraphics[height=1.6cm]{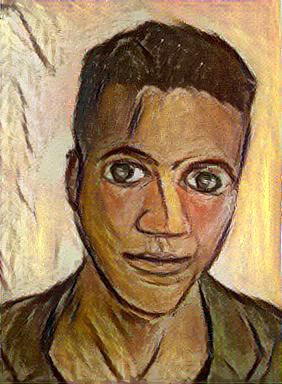}
\includegraphics[height=1.6cm]{level1-Chuan/05-Major-General-Edward-Rice.jpg}
\includegraphics[height=1.6cm]{level1-Chuan/01-Kimberly-Howell.jpg}
\hfill
\includegraphics[height=1.6cm]{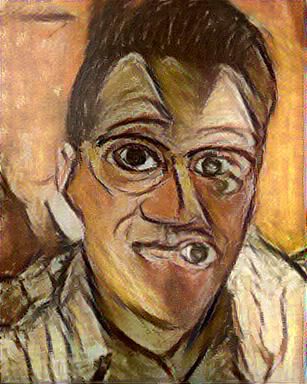}
\includegraphics[height=1.6cm]{level3-Chuan/10-calvin-lupiya-Mx4auh5zO4w-unsplash.jpg}
\includegraphics[height=1.6cm]{level3-Chuan/20-3683799501_052eb48752_o.jpg}

\smallskip
\includegraphics[height=1.6cm]{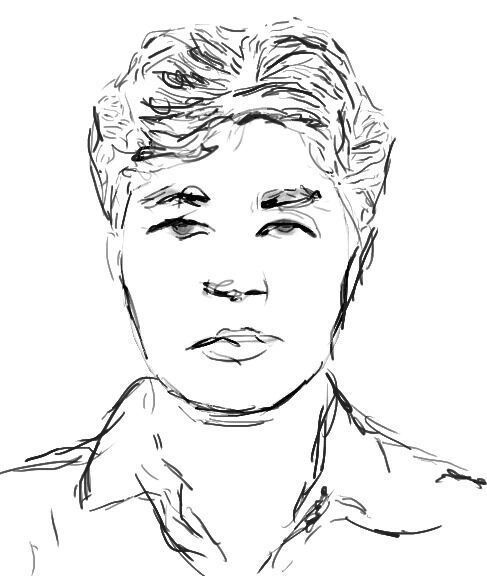}
\includegraphics[height=1.6cm]{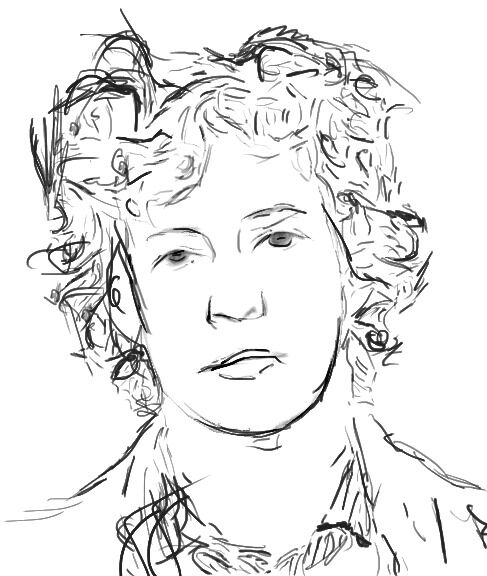}
\includegraphics[height=1.6cm]{level1-Itamar/10-Saira-Shah-Halim}
\hfill
\includegraphics[height=1.6cm]{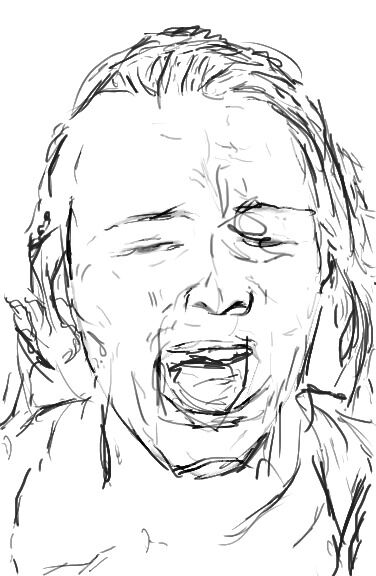}
\includegraphics[height=1.6cm]{level3-Itamar/05-old-youth-tAJog0uJkT0-unsplash}
\includegraphics[height=1.6cm]{level3-Itamar/20-3683799501_052eb48752_o}

\smallskip
\includegraphics[height=1.45cm]{level1-Ran/15-Shimizu-kurumi}
\includegraphics[height=1.45cm]{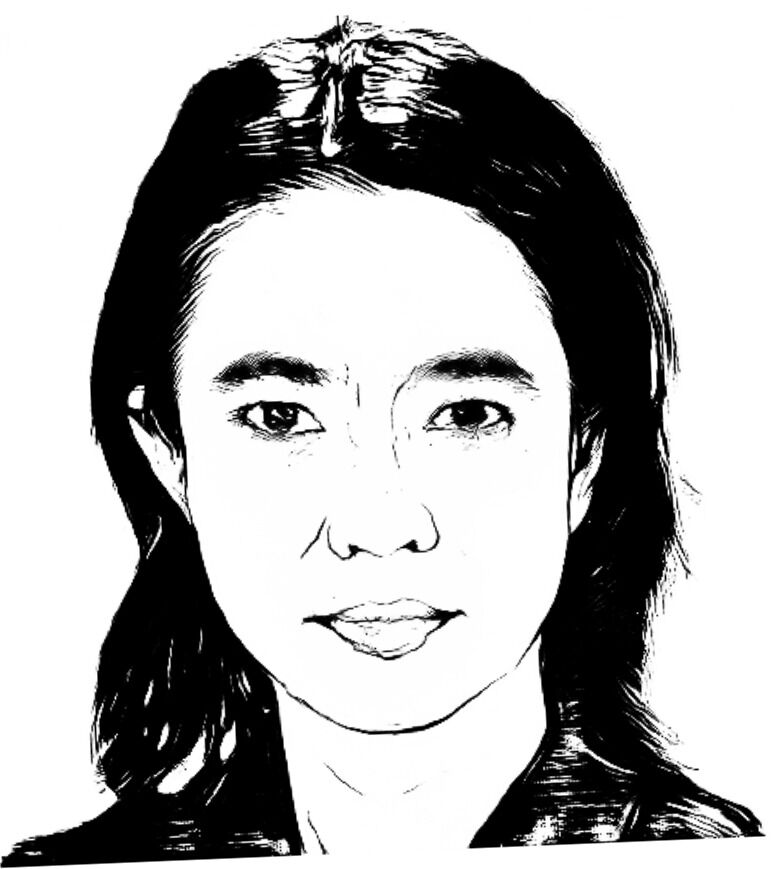}
\includegraphics[height=1.45cm]{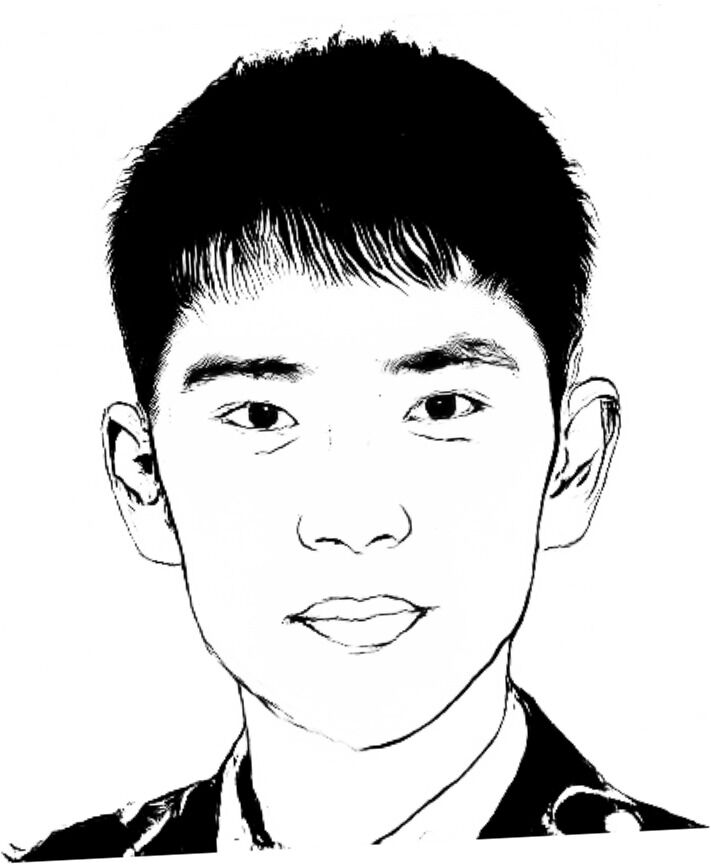}
\hfill
\includegraphics[height=1.45cm]{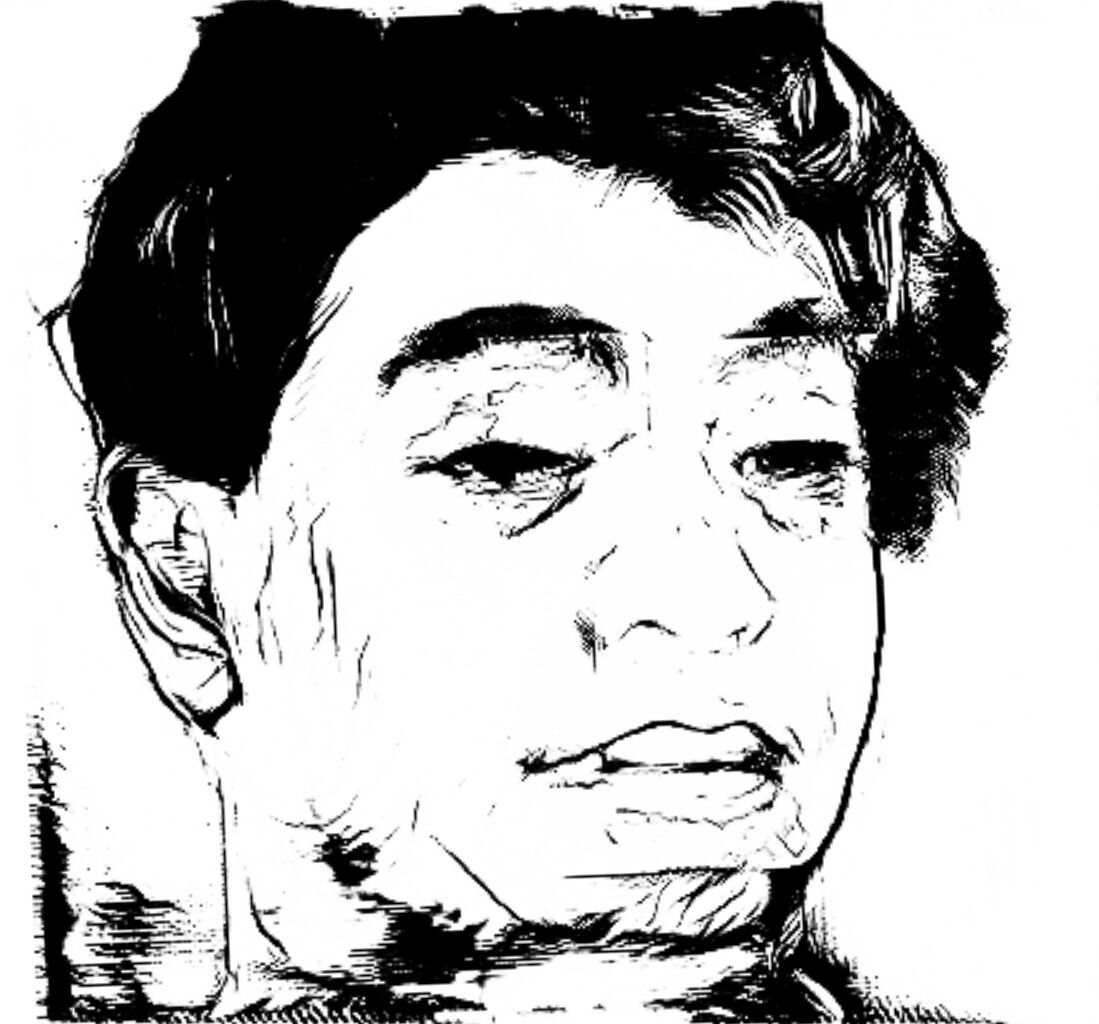}
\includegraphics[height=1.45cm]{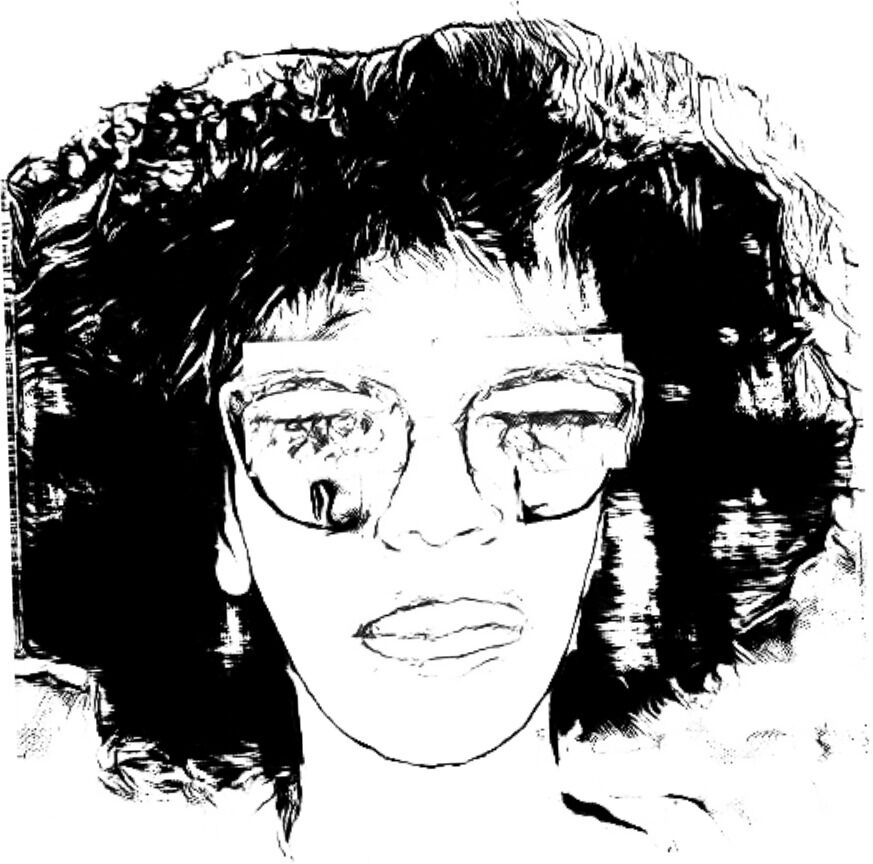}
\includegraphics[height=1.45cm]{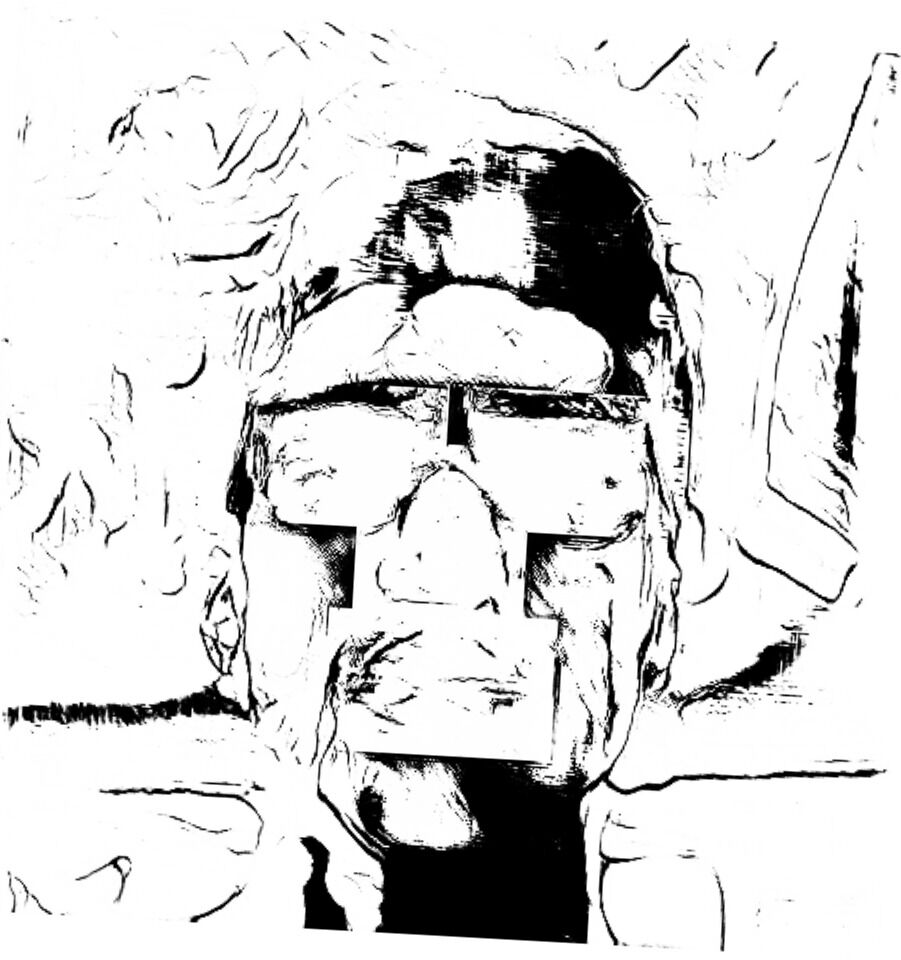}

\smallskip
\includegraphics[height=1.6cm]{level2-RosinLai/15-pexels-355020}
\includegraphics[height=1.6cm]{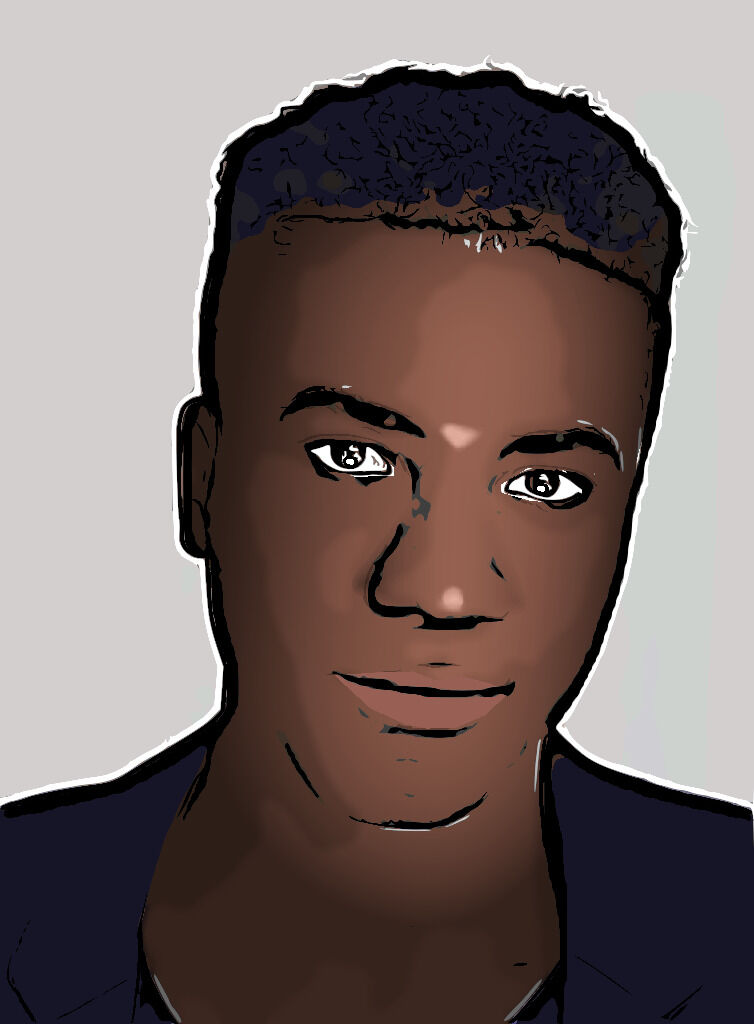}
\includegraphics[height=1.6cm]{level1-RosinLai/15-Shimizu-kurumi}
\hfill
\includegraphics[height=1.6cm]{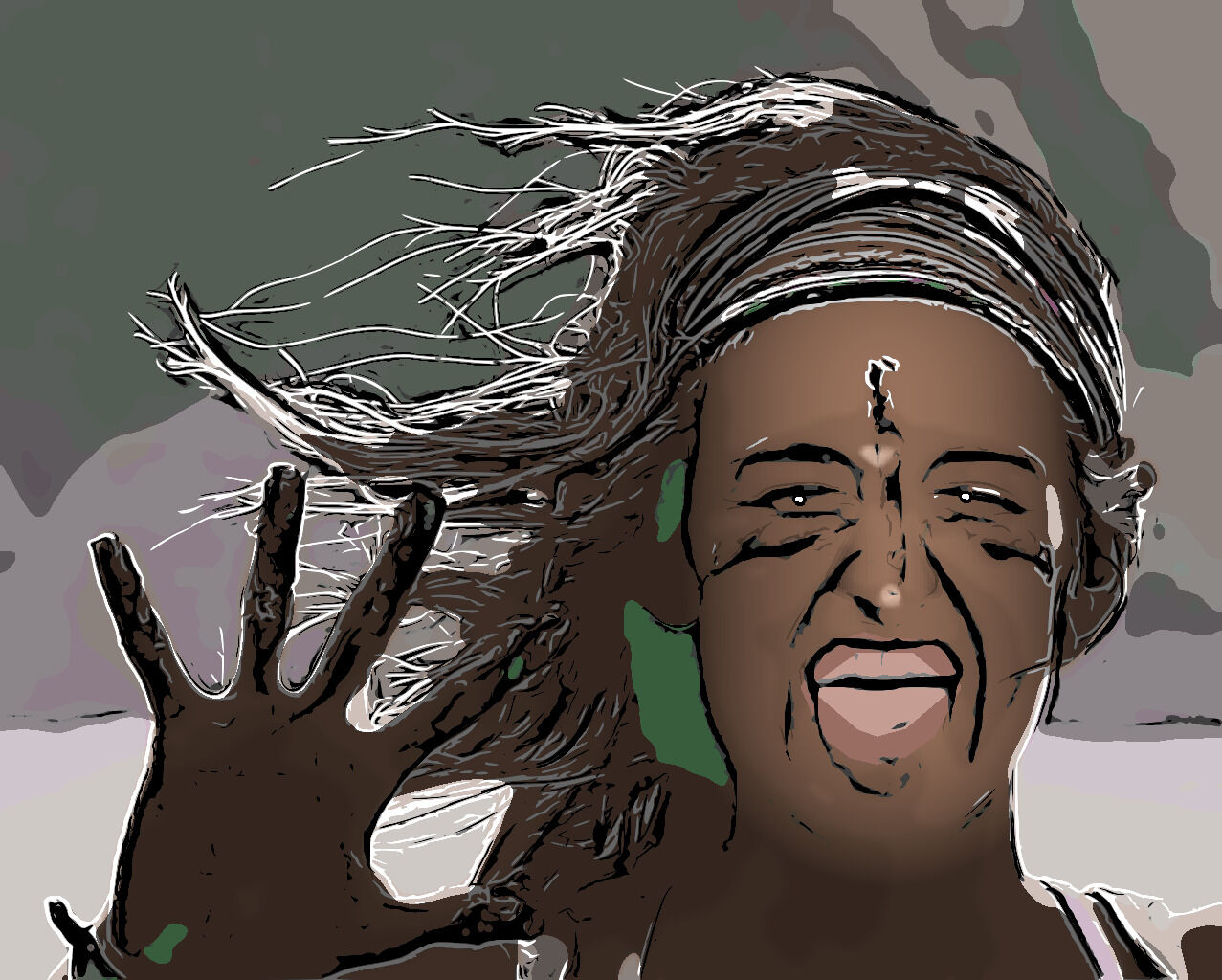}
\includegraphics[height=1.6cm]{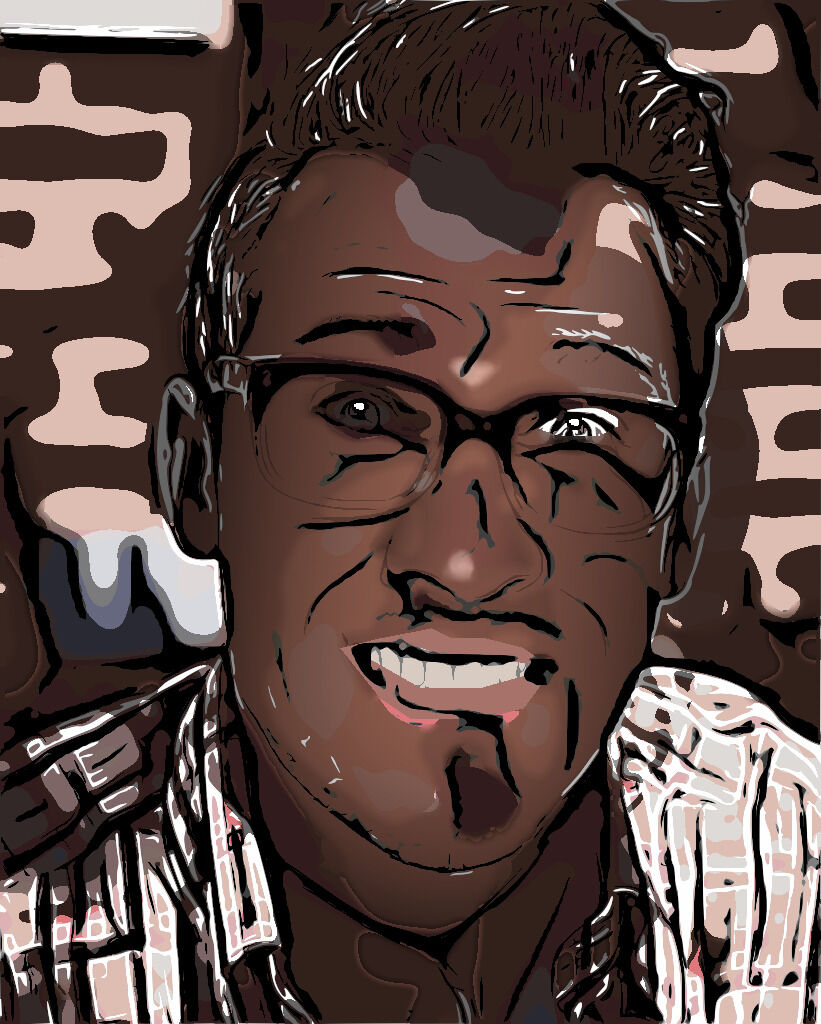}
\includegraphics[height=1.6cm]{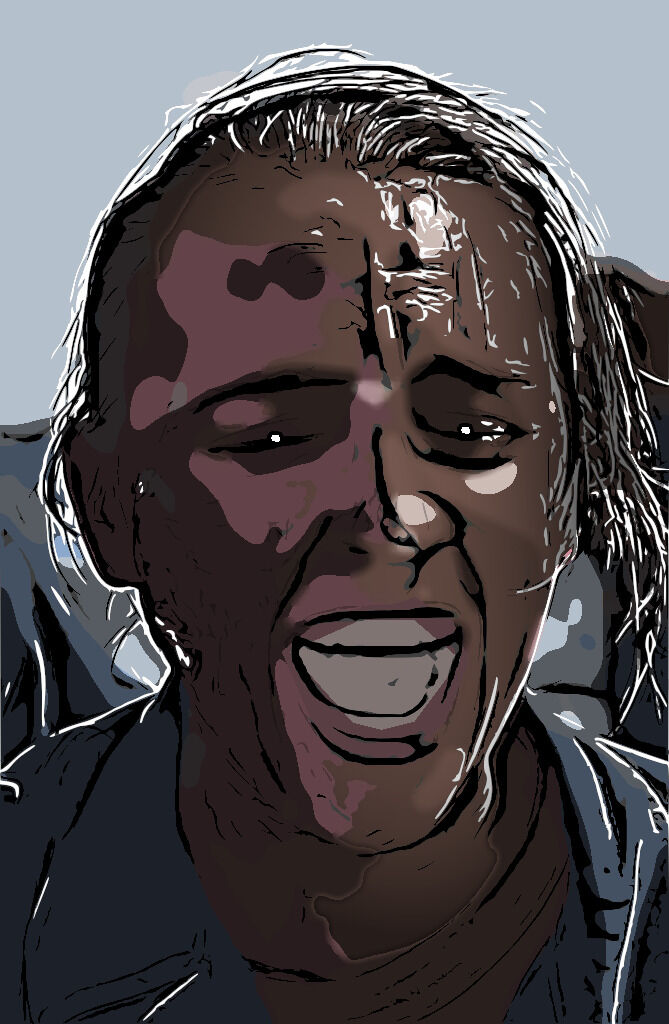}

\smallskip
\includegraphics[height=1.6cm]{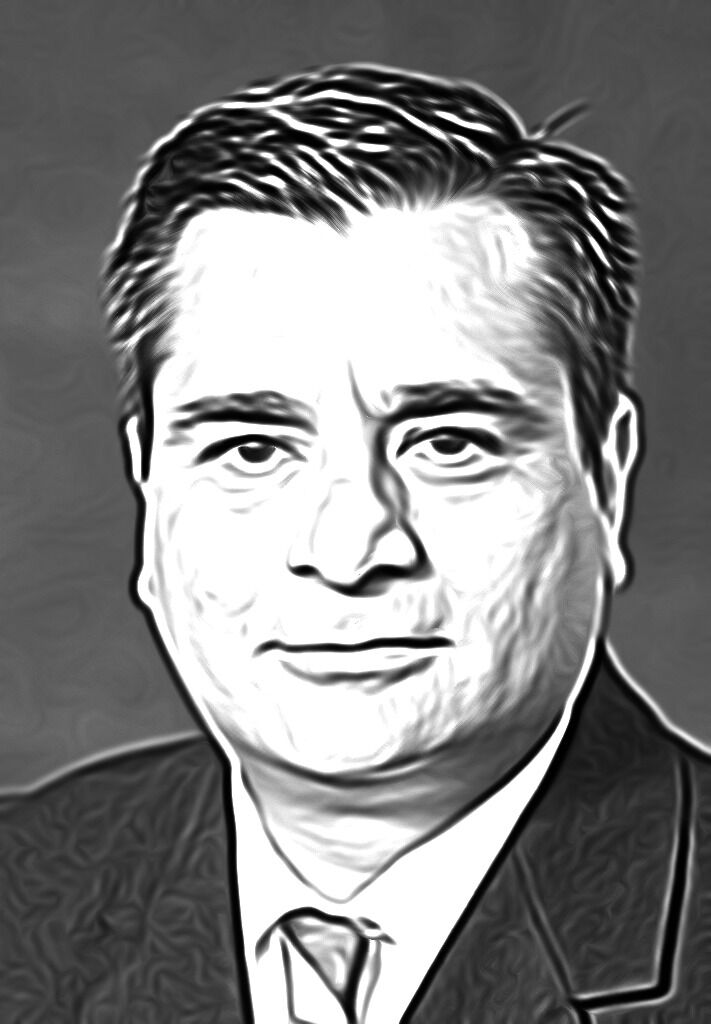}
\includegraphics[height=1.6cm]{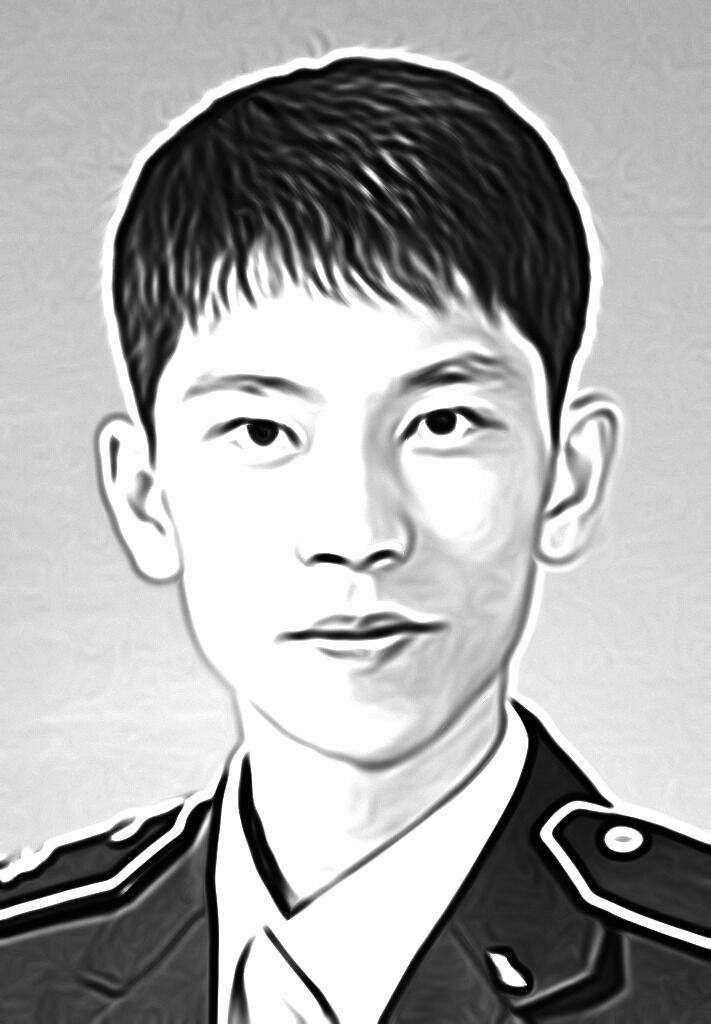}
\includegraphics[height=1.6cm]{level1-XDoG/15-Shimizu-kurumi}
\hfill
\includegraphics[height=1.6cm]{level2-XDoG/01-wikimedia}
\includegraphics[height=1.6cm]{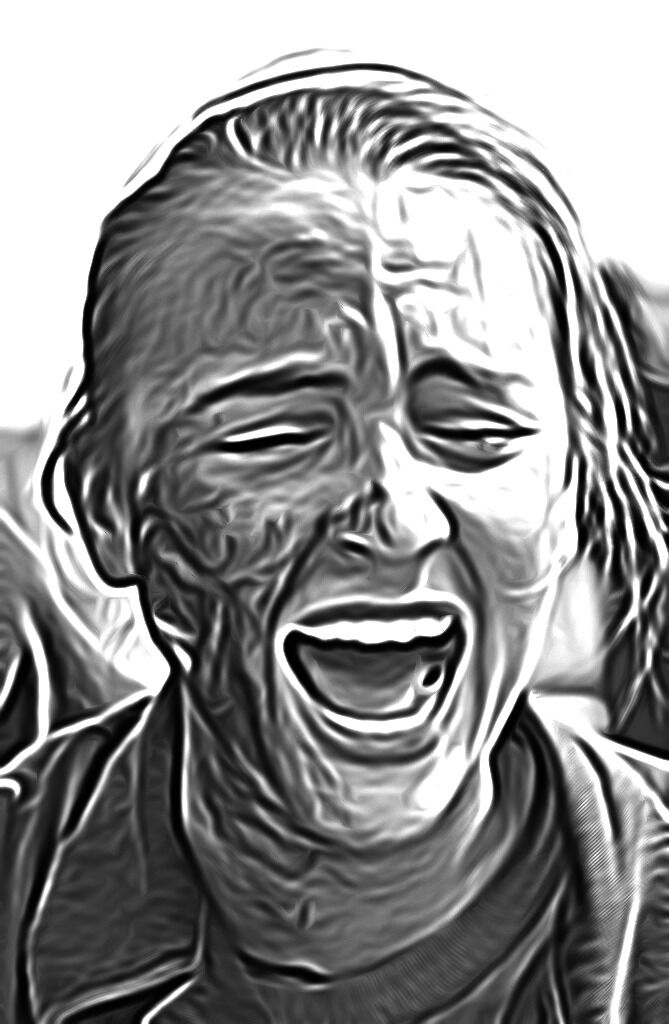}
\includegraphics[height=1.6cm]{level3-XDoG/20-3683799501_052eb48752_o}

\includegraphics[height=1.6cm]{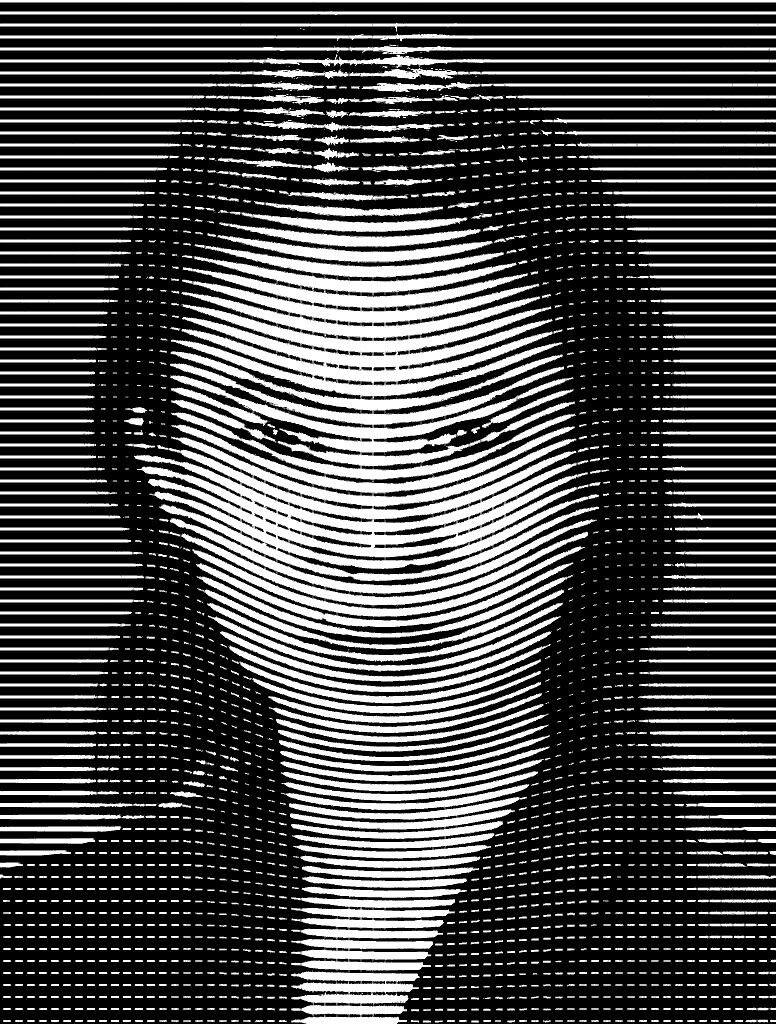}
\includegraphics[height=1.6cm]{level1-engraving/20-8552893573-1473209795-o}
\includegraphics[height=1.6cm]{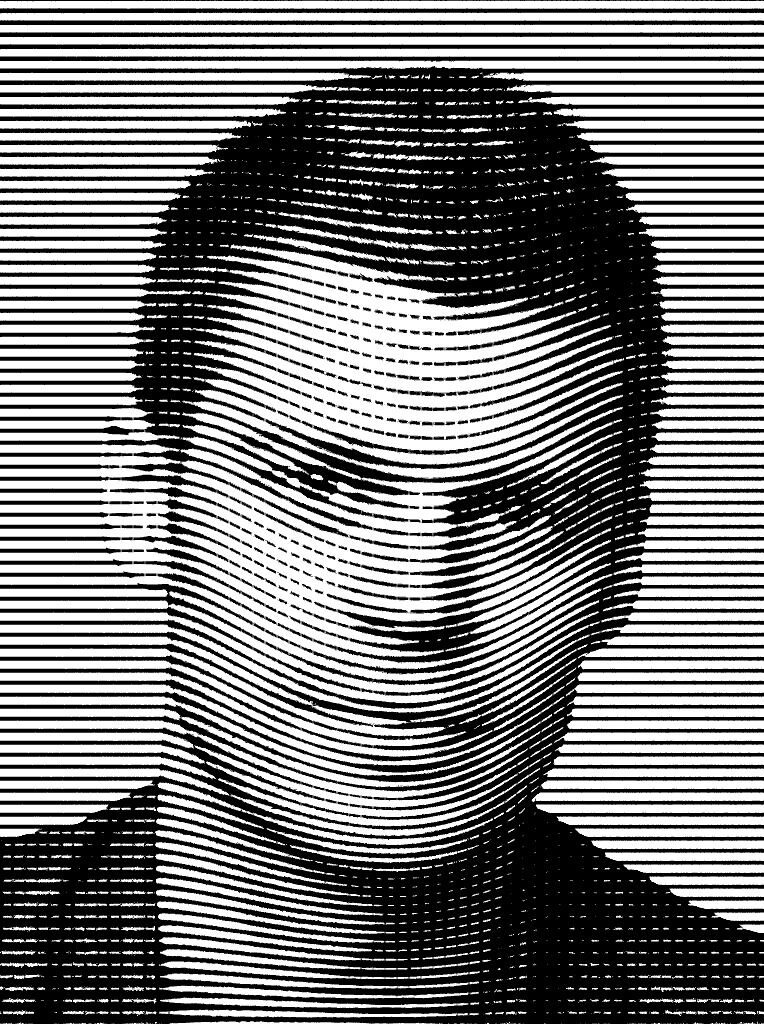}
\hfill
\includegraphics[height=1.6cm]{level3-engraving/01-johanna-buguet-9GOAzu0G4oM-unsplash}
\includegraphics[height=1.6cm]{level3-engraving/20-3683799501_052eb48752_o}
\includegraphics[height=1.6cm]{level3-engraving/15-8717570008_edc9120e59_o}

\includegraphics[height=1.6cm]{level1-hedcut/20-8552893573-1473209795-o}
\includegraphics[height=1.6cm]{level1-hedcut/15-Shimizu-kurumi}
\includegraphics[height=1.6cm]{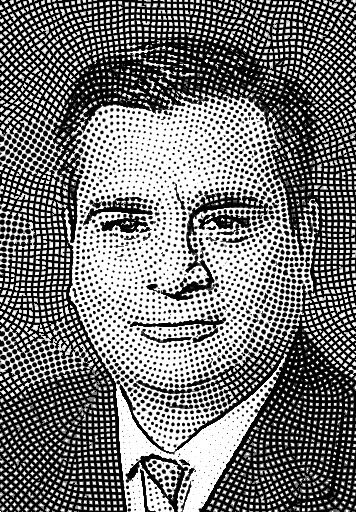}
\hfill
\includegraphics[height=1.6cm]{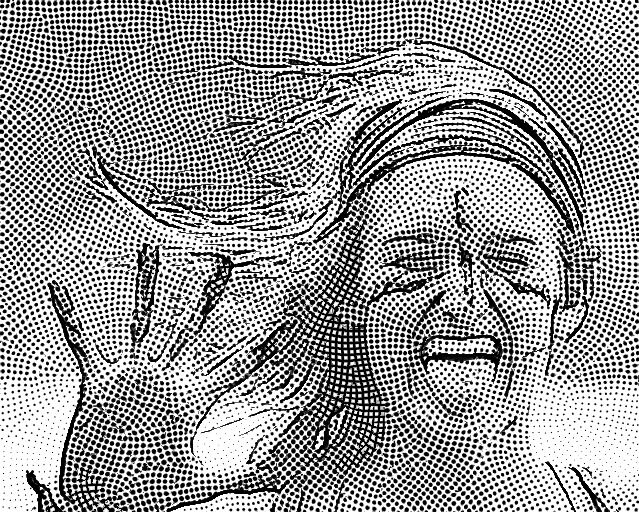}
\includegraphics[height=1.6cm]{level3-hedcut/20-3683799501_052eb48752_o}
\includegraphics[height=1.6cm]{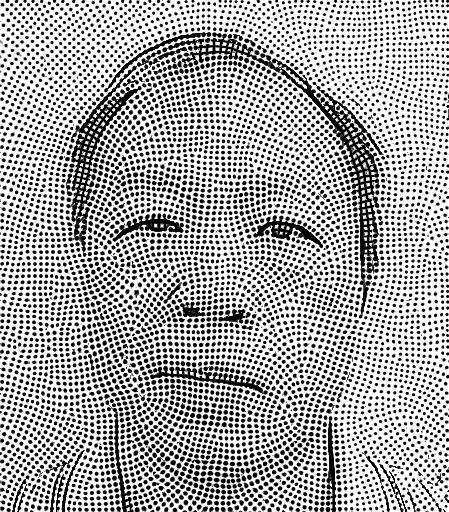}

\includegraphics[height=1.6cm]{level1-oilpaint/15-Shimizu-kurumi.jpg}
\includegraphics[height=1.6cm]{level3-oilpaint/11-artyom-kim-zEa75CRX88M-unsplash.jpg}
\includegraphics[height=1.6cm]{level2-oilpaint/15-pexels-355020.jpg}
\hfill
\includegraphics[height=1.6cm]{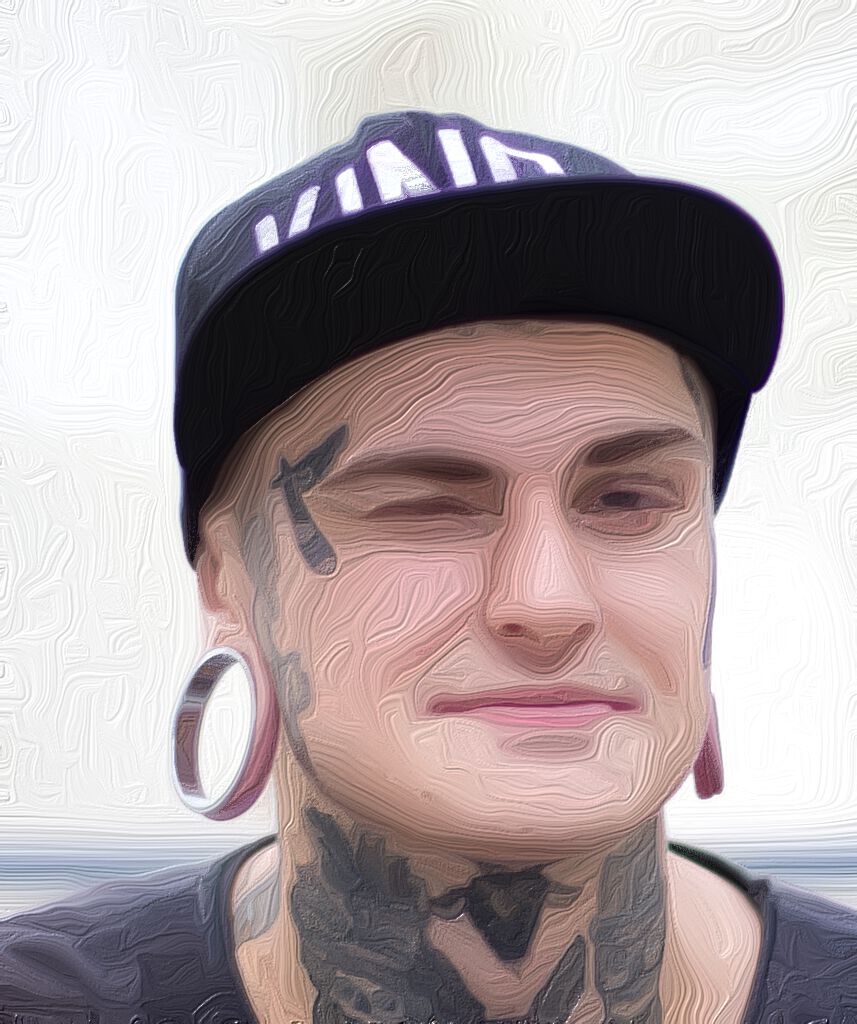}
\includegraphics[height=1.6cm]{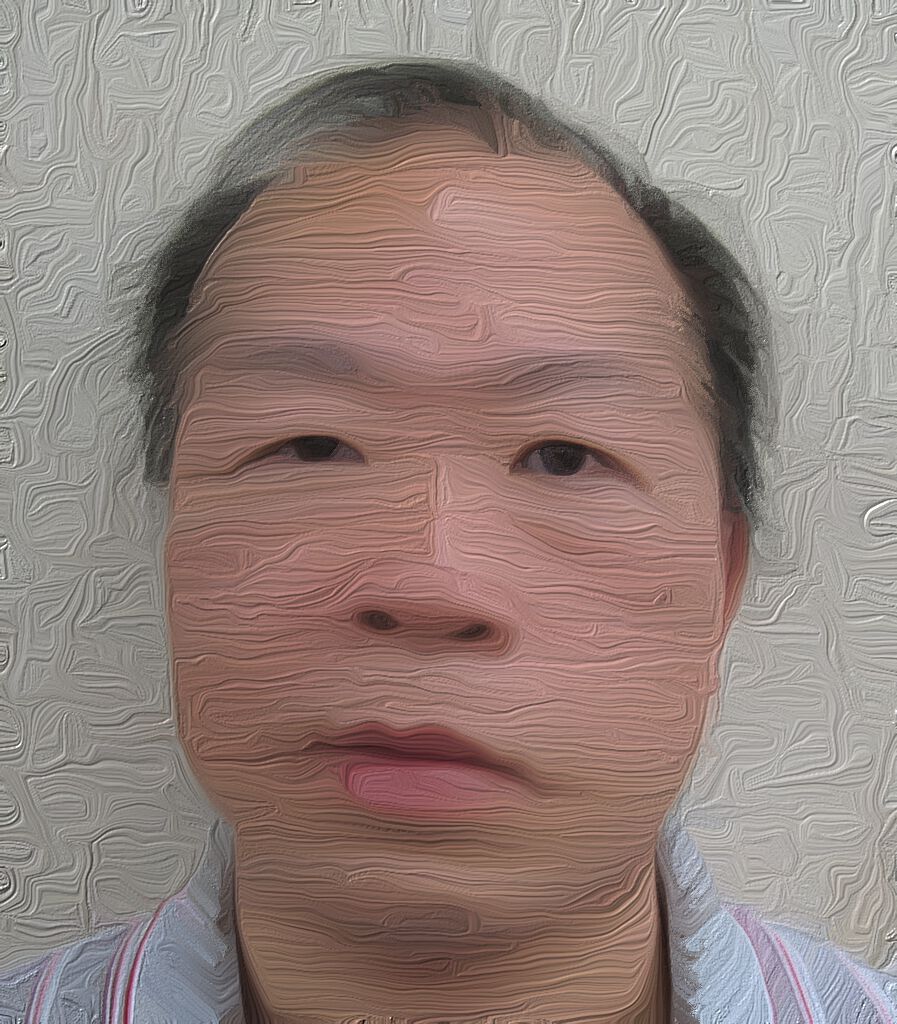}
\includegraphics[height=1.6cm]{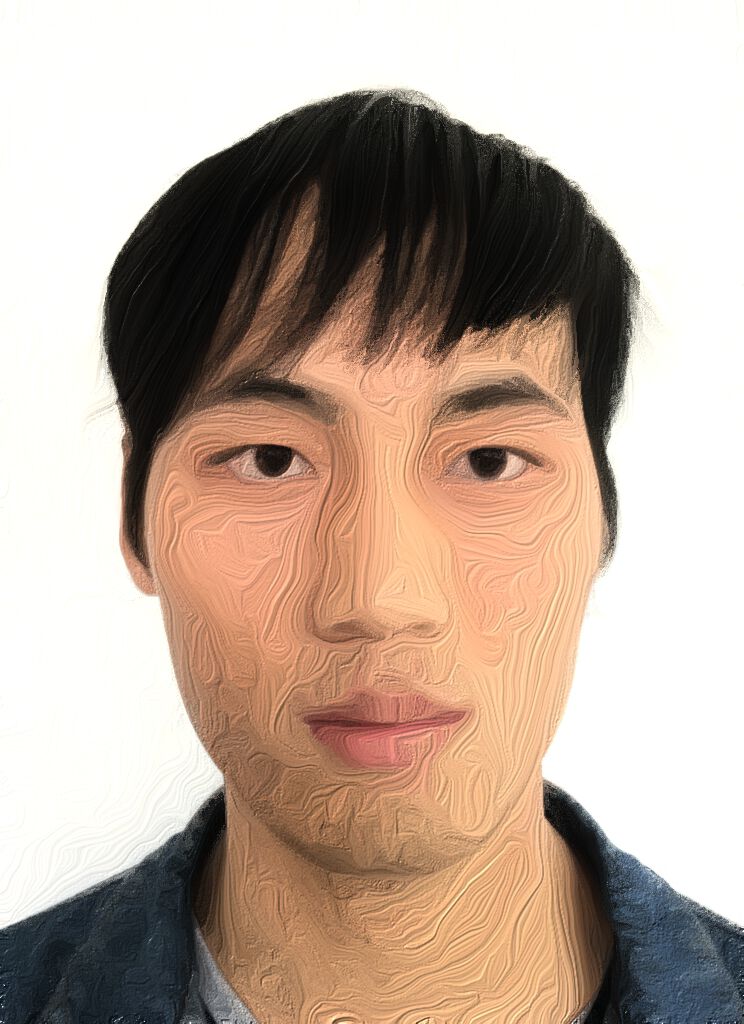}

\includegraphics[height=1.6cm]{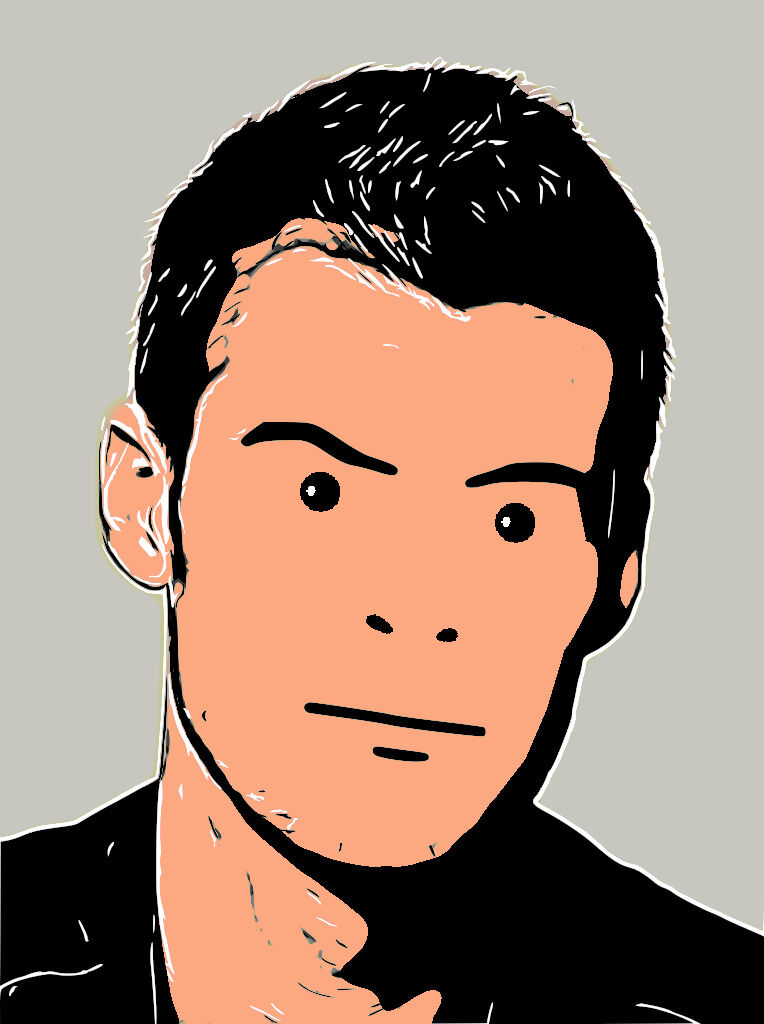}
\includegraphics[height=1.6cm]{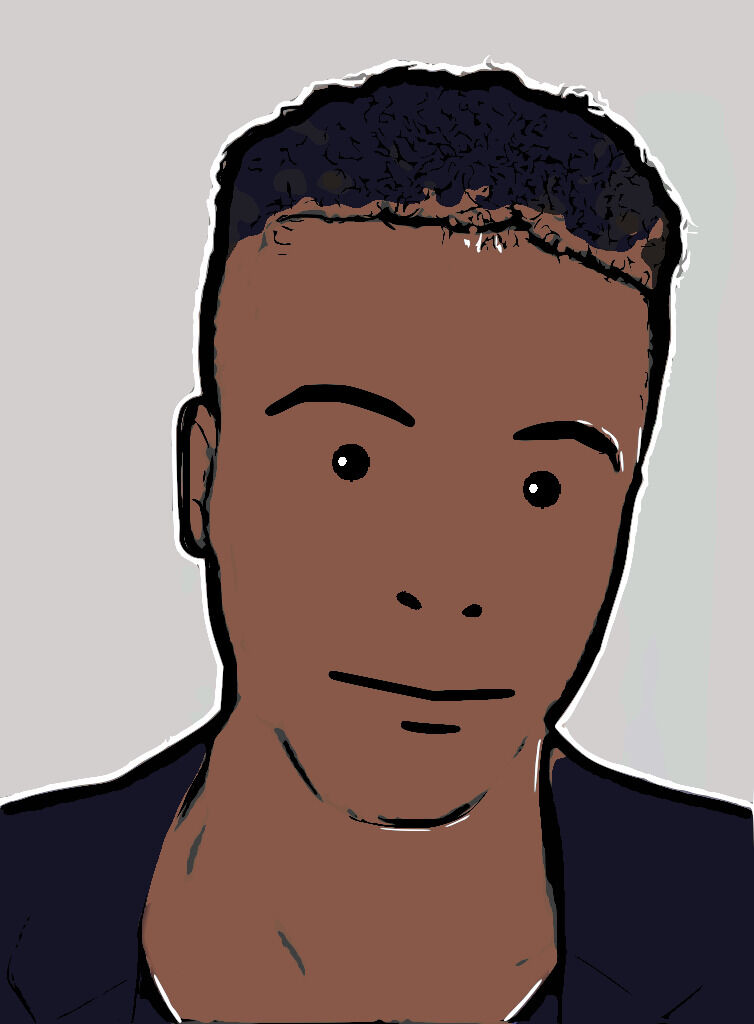}
\includegraphics[height=1.6cm]{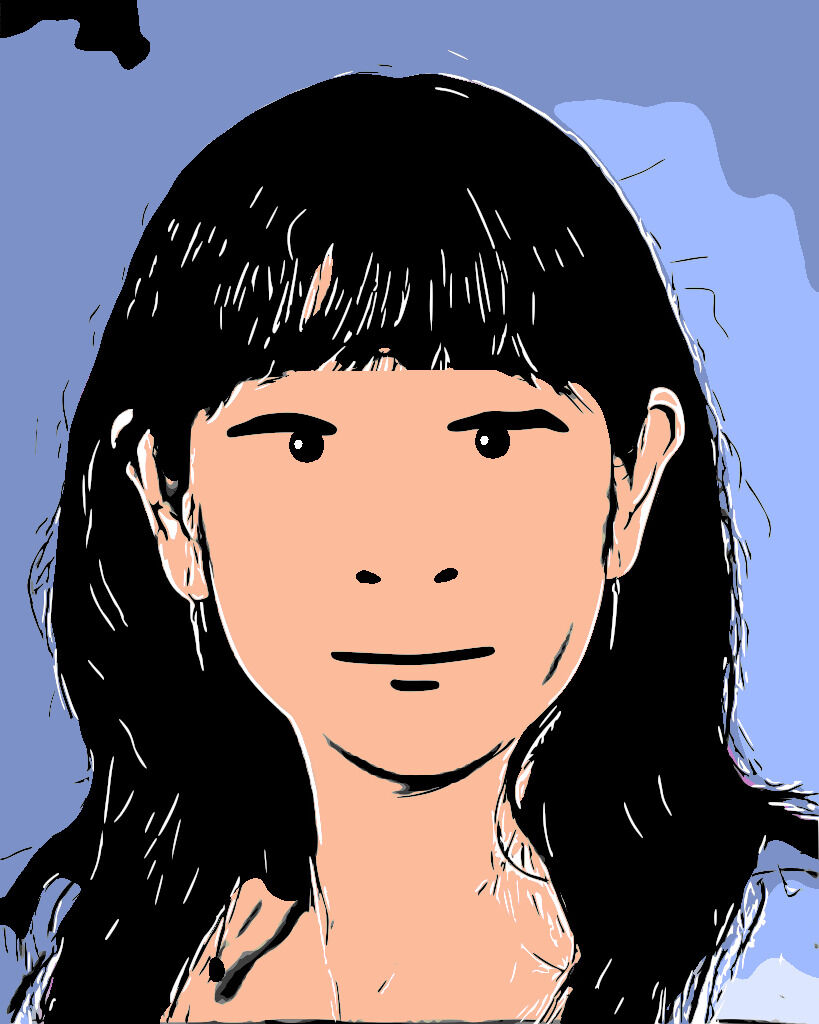}
\hfill
\includegraphics[height=1.6cm]{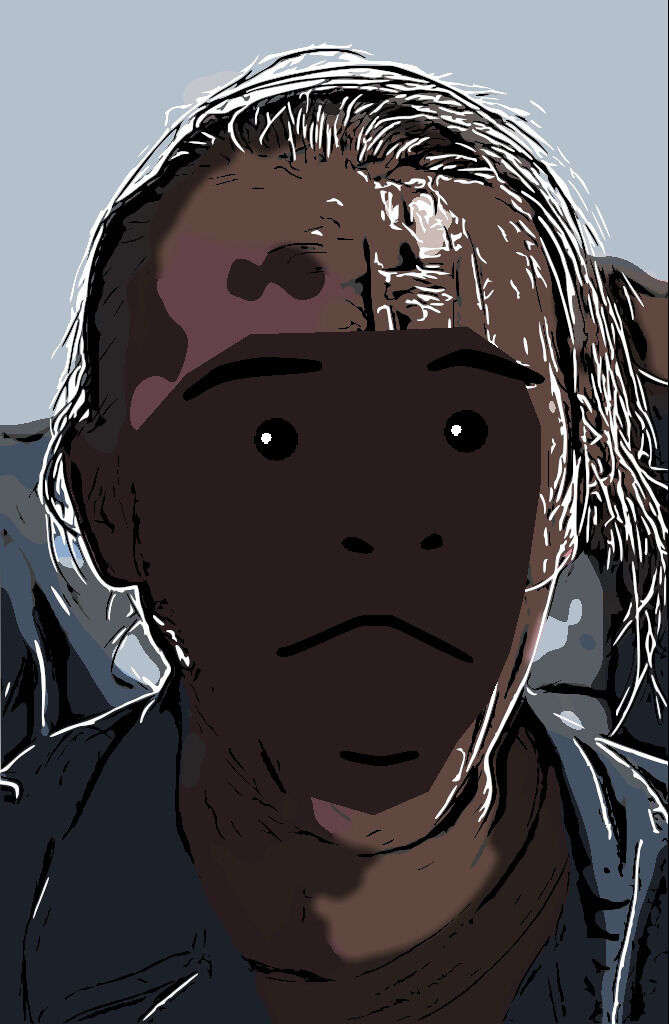}
\includegraphics[height=1.6cm]{level3-opie/05-old-youth-tAJog0uJkT0-unsplash}
\includegraphics[height=1.6cm]{level3-opie/01-johanna-buguet-9GOAzu0G4oM-unsplash}

\includegraphics[height=1.6cm]{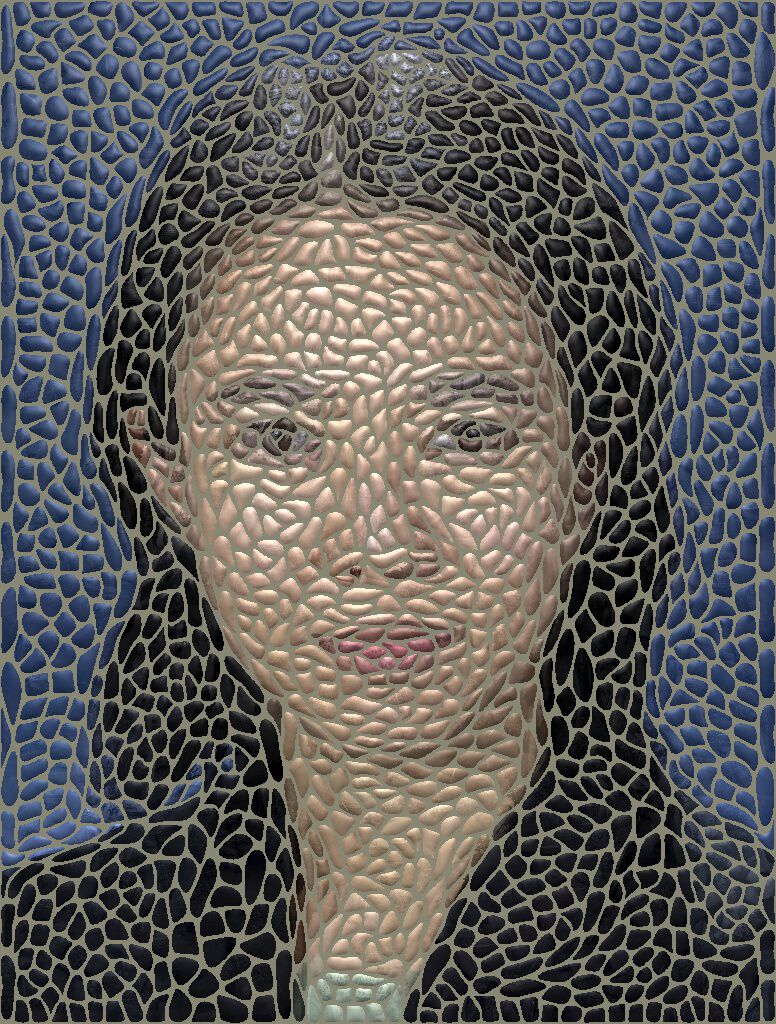}
\includegraphics[height=1.6cm]{level1-pebble/15-Shimizu-kurumi.jpg}
\includegraphics[height=1.6cm]{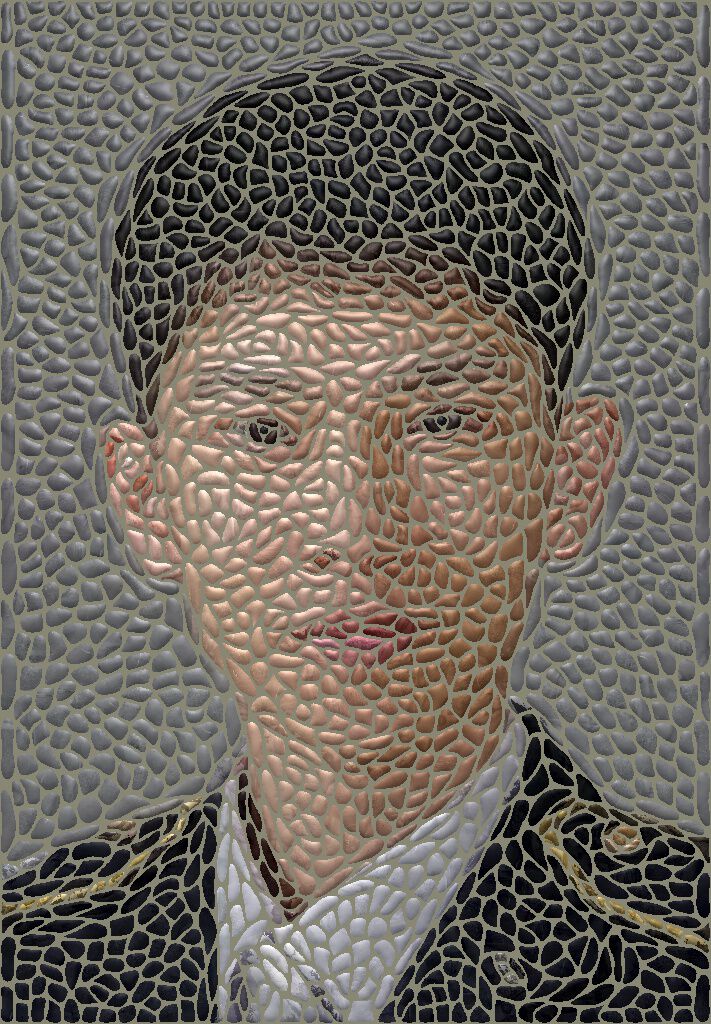}
\hfill
\includegraphics[height=1.6cm]{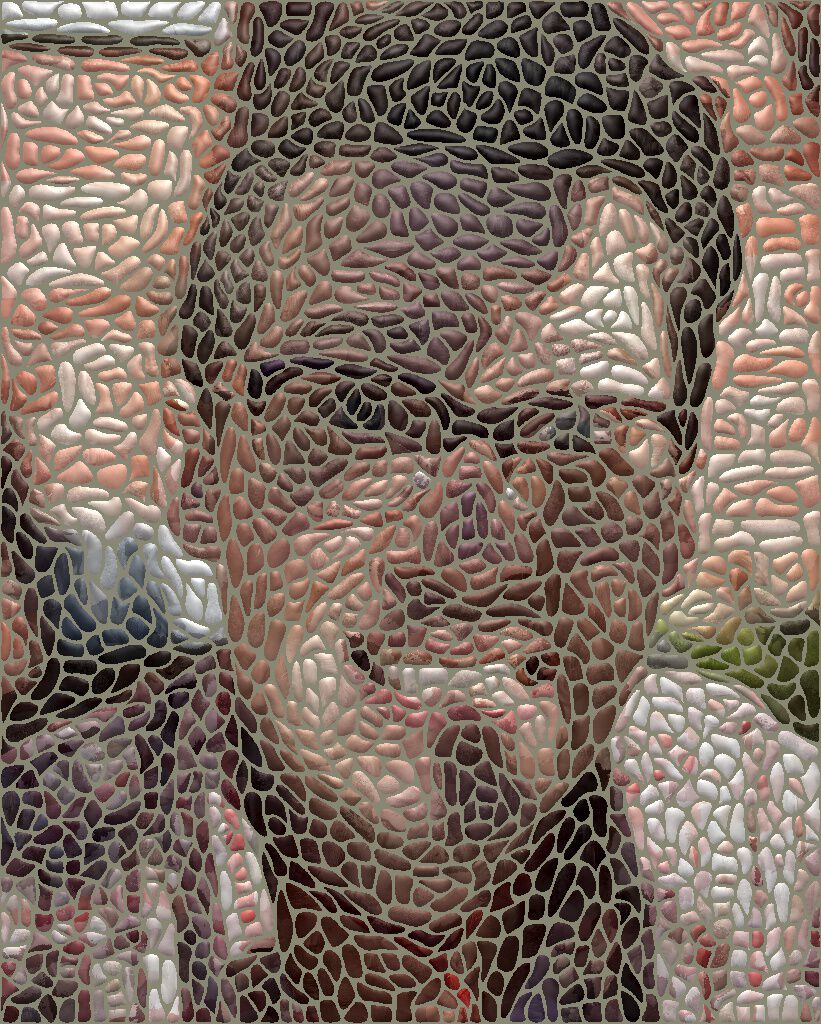}
\includegraphics[height=1.6cm]{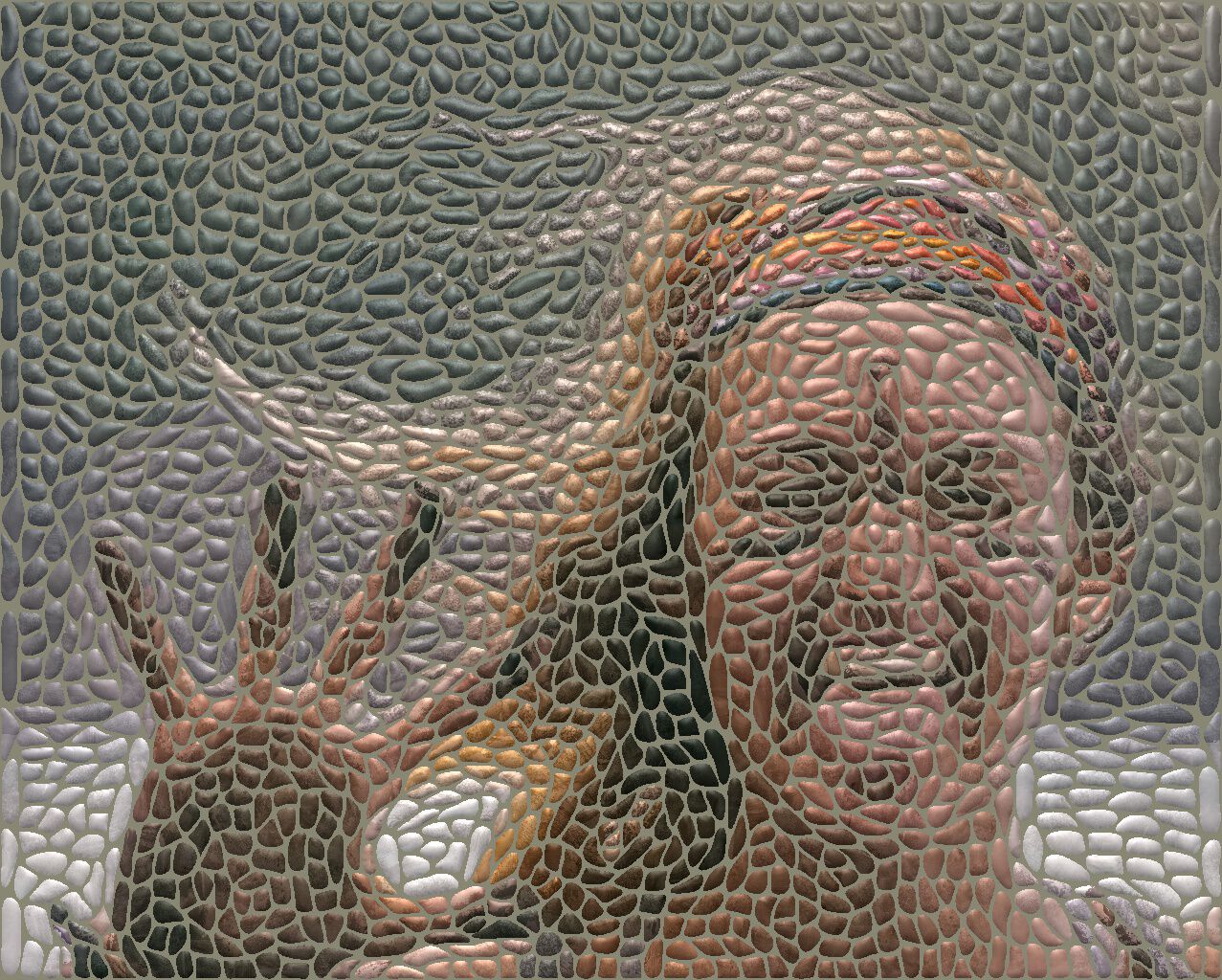}
\includegraphics[height=1.6cm]{level3-pebble/15-8717570008_edc9120e59_o.jpg}

\includegraphics[height=1.6cm]{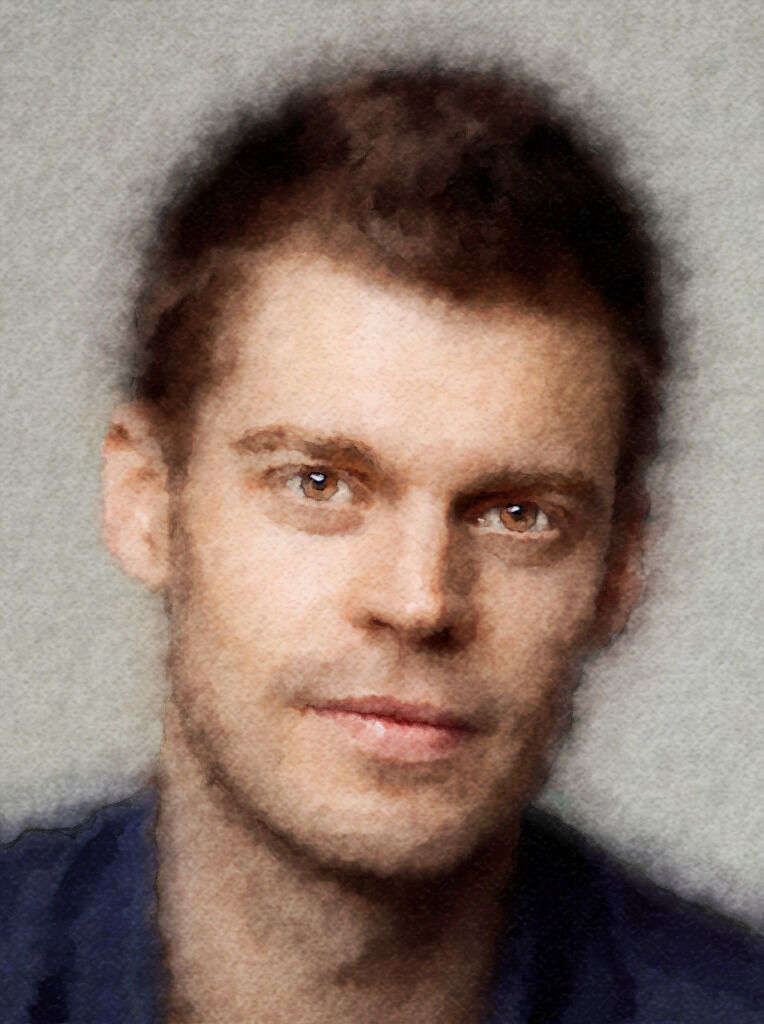}
\includegraphics[height=1.6cm]{level1-watercolour/15-Shimizu-kurumi.jpg}
\includegraphics[height=1.6cm]{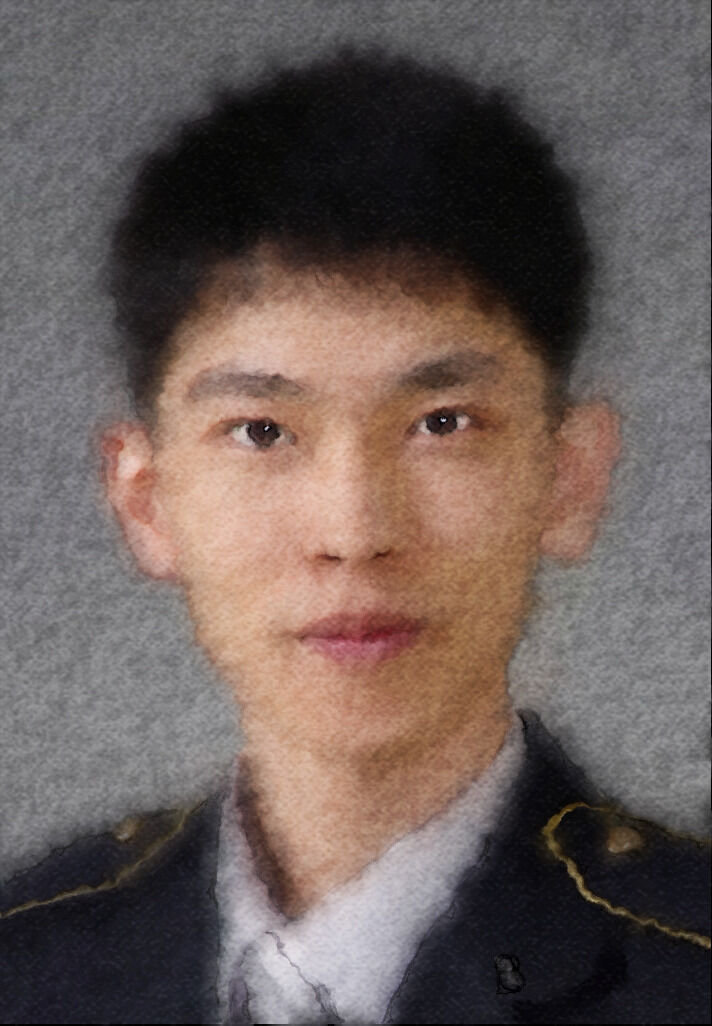}
\hfill
\includegraphics[height=1.6cm]{level2-watercolour/10-3761108471_08e3f9f80d_o.jpg}
\includegraphics[height=1.6cm]{level3-watercolour/20-3683799501_052eb48752_o.jpg}
\includegraphics[height=1.6cm]{level3-watercolour/15-8717570008_edc9120e59_o.jpg}
\caption{Images from \emph{NPRportrait1.0} benchmark stylised by the 11 NPR algorithms;
the rows show (in order):
neural style transfer~\cite{li2016combining},
artistic sketch method~\cite{berger-shamir},
APDrawingGAN~\cite{YiLLR19},
puppet style~\cite{rosin-portrait}
XDoG~\cite{xdog},
engraving~\cite{engraving},
hedcut~\cite{Son-hedcut},
oil painting~\cite{semmo2016image},
Julian Opie style~\cite{rosin-portrait},
pebble mosaic~\cite{doyle2019automated},
watercolour~\cite{rosin2017watercolour}.
The stylisations are ranked according to the outcomes of Experiment~2;
for each method we show the top three results on the left and the bottom three on the right.}
\label{ranking3x2}
\end{figure}

This user study can be further used to analyse both the benchmark and
the NPR algorithms.  The triplets were converted to a global ranking
using Wauthier \emph{et al.}'s~\cite{wauthier2013efficient} Balanced
Rank Estimation method, applied both (1) separately for each NPR
algorithm, and also (2) across all the NPR algorithms, by aggregating
the local scores for each benchmark image across the stylisations.
Ranking the images in this way enables us to see which aspects of
images lead to good stylisations either for a specific algorithm, or
more generally across a range of algorithms.
Figure~\ref{rankingAllStyles} reveals that images which are the top
ranked, and therefore more amenable to current stylisation algorithms,
tend to be portraits with frontal views, fairly neutral expressions,
good lighting, and plain backgrounds.  At the other end of the scale,
the bottom ranked images tend to have one or more of the following
characteristics: non-frontal views, strong expressions, patterns on
the face, strong lighting effects, and cluttered backgrounds.

The top and bottom three ranked results for each of the 11 NPR methods
are shown in figure~\ref{ranking3x2}.  The bottom ranked results
reveal a variety of artifacts, including inappropriate rendering of
facial features, messy rendering, segmentation
errors, and rendering that does not clearly delineate facial
components and structure.  We note that the top and bottom ranked
images in figure~\ref{rankingAllStyles} appear in many of the top and
bottom three rankings in figure~\ref{ranking3x2} (i.e. 7 and 6 out of
11 respectively).  However, it is possible that, despite their
instructions to score according to stylisation quality, the users'
responses in Experiment~2 were biased by other factors.  The images
that were ranked top and bottom ranked in
figure~\ref{rankingAllStyles} according to the overall quality of
their stylisations also have the most and least attractiveness ratings
for source images in the \emph{NPRportrait1.0} dataset.  There is a
moderate degree of correlation between the overall stylisation
rankings and the source image attractiveness ratings: 0.6732 (Pearson)
and 0.4666 (Kendall).

\begin{figure}[htbp]
\centering
\subfloat[]{\includegraphics[height=4cm]{NPRportrait-level3new/16-gabriel-silverio-u3WmDyKGsrY-unsplash}}
\qquad
\subfloat[]{\includegraphics[height=4cm]{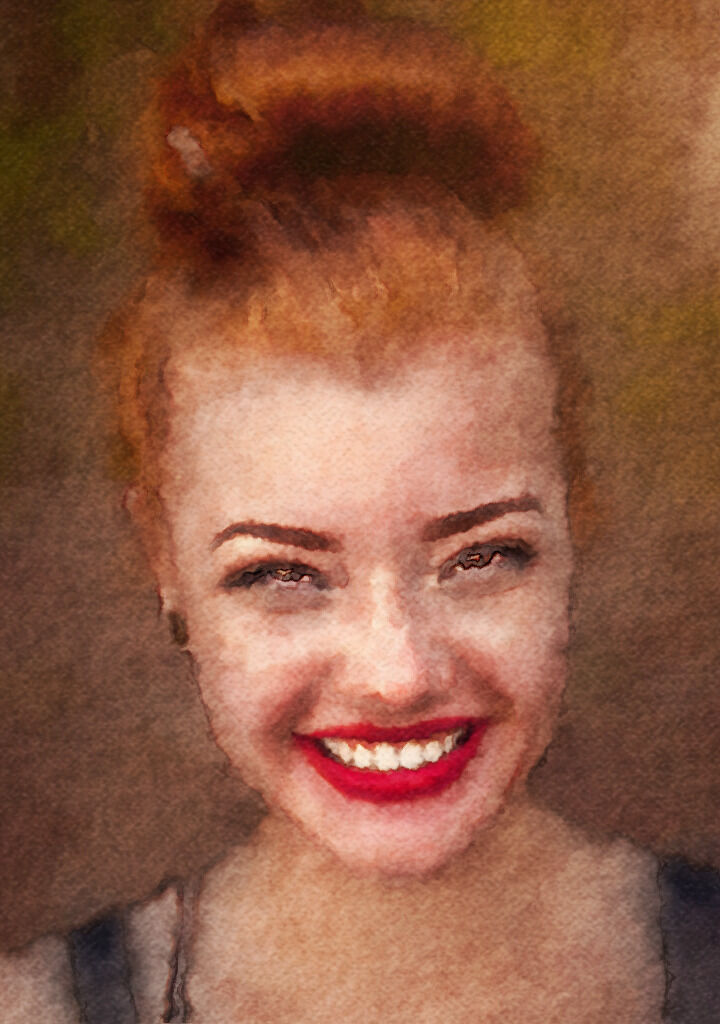}}\\
\subfloat[]{\includegraphics[height=5cm]{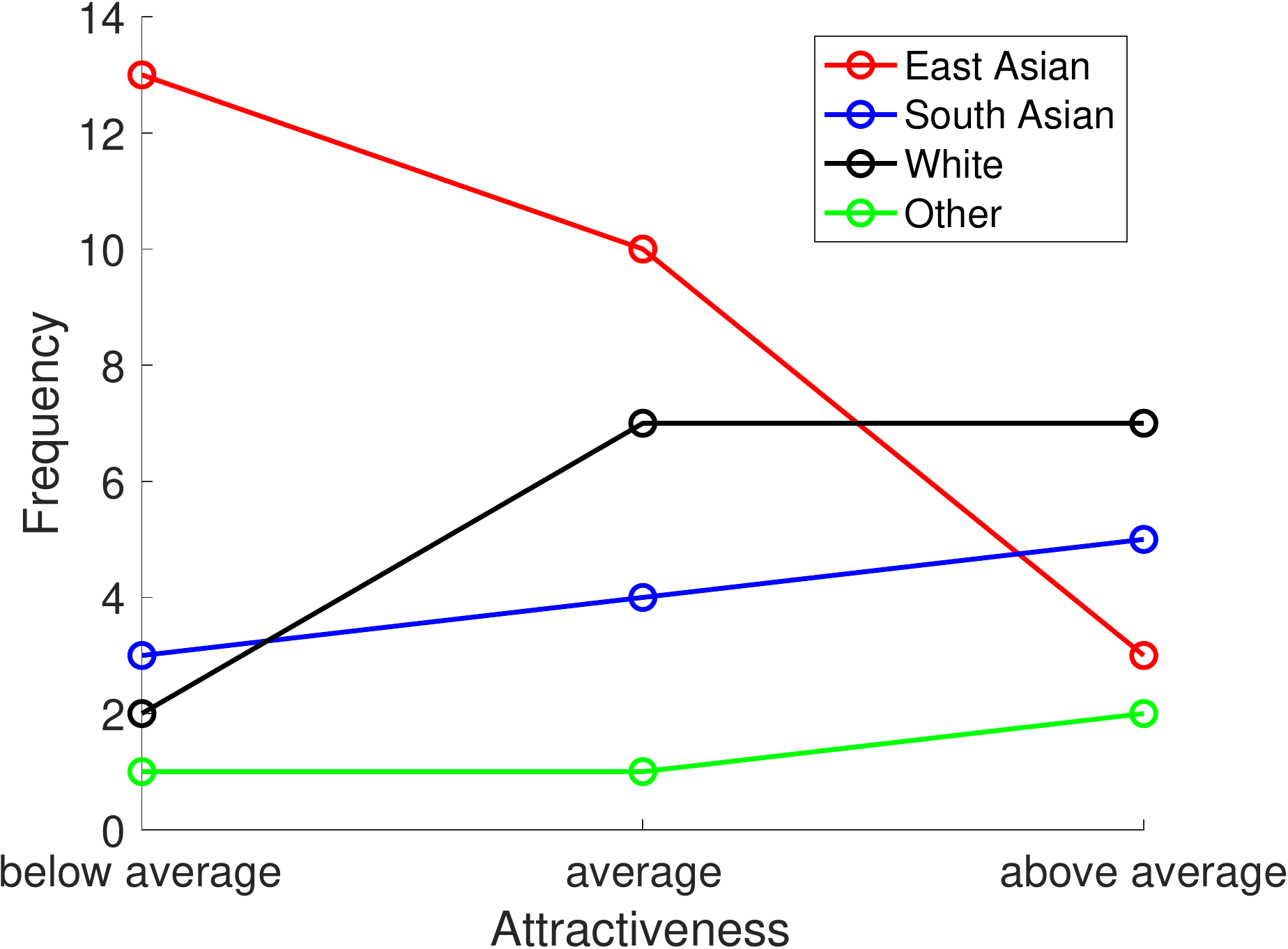}}
\caption{Cultural effects on perceptual judgments; (a) source image, (b) its watercolour stylisation,
and the distribution of ethnic backgrounds of the participants who saw this \emph{source} image during the user study.}
\label{freckles}
\end{figure}

The cultural
backgrounds of the dtudy participants
can also affect their perceptual judgments of the images
and their stylisations.
  This was highlighted in the image from
level~3 shown in figure~\ref{freckles}.  Applying watercolour
stylisation produced a surprisingly large improvement in
attractiveness score, from -0.385 to +0.800. We learned that
this was because the stylisation removed freckles from
the original portrait, a feature that is considered unsightly in some Asian
cultures.  Figure~\ref{freckles}c shows that a large proportion of the
participants who rated the source image were East
Asian, and that they considered the
source image (with freckles) to be unattractive.\footnote{ The user
study for rating the stylised images involved different users, and for
this image also contained a large proportion of East Asian evaluators
(50\%).  However, this was not critical in determing the
attractiveness rating for this stylisation.  }

\section{Conclusions and Future Work}

Currently the field of non-photorealistic rendering and neural style
transfer is hampered by a lack of benchmark datasets and objective
measures, leading to most papers providing limited and rudimentary
performance evaluation.  In the specific area of portrait stylisation,
this paper has presented a benchmark dataset that is structured into
three levels to provide clearly specified degrees of difficulty.  The
criteria for selecting images for each level were clearly specified,
and used to construct a design matrix.  User studies were used to
validate the suitability of each image with respect to the design
matrix.

Alongside the new dataset a new methodology has been proposed for
evaluating portrait stylisation algorithms.  Rather than rely on users
articulating aesthetic judgments, a challenging and ill-defined task, the user
studies also incorporate more straightforward judgments, such as
identification of gender or age.

The new benchmark and methodology enabled us to evaluate 11 NPR
algorithms, both portrait-specific and general-purpose, and
quantitatively compare them in terms of their preservation of the
portraits' characteristics, and their robustness to increasing levels
of image complexity.  By applying Balanced Rank Estimation it was
possible to determine a global ranking of the stylised benchmark
images so that the problematic images for each NPR algorithm could be
identified.  The bottom ranked results reveal that typical defects are
inappropriate rendering of facial features, messy
rendering, segmentation errors, and rendering that does not clearly
delineate facial components and structure.  Likewise, the global
ranking computed across all the algorithms highlighted image types
that are problematic for many state of the art algorithms.  Typically
they contained non-frontal views, strong expressions, patterns on the
face, strong lighting effects, and cluttered backgrounds.

The identification of challenging cases will help direct future
research to in useful directions.  Of course, there is scope for
increasing the benchmark by adding addition levels, covering more
complicated scenes as well as broader coverage of portrait subjects.
Possible complications include images with
multiple people, full bodies, substantial
occlusion, heavily cluttered background, extreme poses and
expressions, and extreme perspective and other photographic
distortions. Additional portrait subjects could include
children, the elderly, and more ethnicities.  In addition, more NPR
benchmarks should be developed for different kinds of content.
For example, landscapes, cityscapes, and animal portraiture have different
requirements, and have evolved traditionally distinctive depiction
styles.  Whereas curating images is a relatively tractable task, truly
capturing the perceptual and artistic aspects of stylisations in an
evaluation measure is challenging.  For instance, one limitation of
the methodology of Experiment~1 is that measuring the degree of
preservation of facial characteristics may unfairly penalise methods
that involve geometric distortions or extreme stylisations.  Thus,
future research should investigate further novel measures that can achieve this whilst reducing the dependence on user studies.

\ifCLASSOPTIONcaptionsoff
  \newpage
\fi

\bibliographystyle{IEEEtran}
\bibliography{benchmarkPortrait2}

\vspace{-30pt}
\begin{IEEEbiographynophoto}{Paul L. Rosin}
is a Professor at the School of Computer Science and Informatics, Cardiff University, UK.
He received his PhD from City University, London in 1988.
Previous posts were at Brunel University, UK; the Institute for Remote Sensing Applications, Joint Research Centre, Italy; and Curtin University of Technology, Australia.
His research interests include
low level image processing,
performance evaluation,
shape analysis,
facial analysis,
medical image analysis,
3D mesh processing,
cellular automata,
non-photorealistic rendering and cultural heritage.
\end{IEEEbiographynophoto}
\vspace{-30pt}
\begin{IEEEbiographynophoto}{Yu-Kun Lai}
is a Professor at School of Computer Science and Informatics,
Cardiff University, UK. He received his B.S and PhD degrees in Computer Science from Tsinghua University, in 2003 and 2008 respectively.
His research interests include computer graphics, computer vision, geometric modeling and image processing.
For more information, visit \url{https://users.cs.cf.ac.uk/Yukun.Lai/}
\end{IEEEbiographynophoto}
\vspace{-30pt}
\begin{IEEEbiographynophoto}{Dr. Mould}
received his PhD from the University of Toronto in 2002.
Following a faculty appointment at the University of Saskatchewan, he
became a professor at Carleton University, where he founded the
Graphics, Imaging, and Games Lab. Dr. Mould is broadly interested in
algorithmic creation of aesthetic objects, including images, music, 3D
models, and computer-mediated experiences. His research centres on
computer graphics and interactive systems, with particular emphasis on
image stylisation, computer games, and procedural modeling.
\end{IEEEbiographynophoto}
\vspace{-30pt}
\begin{IEEEbiographynophoto}{Ran Yi}
is a Ph.D student with Department of Computer Science and Technology, Tsinghua University.
She received her B.Eng. degree from Tsinghua University, China, in 2016.
Her research interests include computational geometry, computer vision and computer graphics.
\end{IEEEbiographynophoto}
\vspace{-30pt}
\begin{IEEEbiographynophoto}{Itamar Berger}
received his Msc in Computer Science in 2012 from the Efi Arazi school of Computer Science at the Interdisciplinary Center in Israel, specializing in Computer Graphics, Deep Learning and Augmented Reality.
\end{IEEEbiographynophoto}
\vspace{-30pt}
\begin{IEEEbiographynophoto}{Lars Doyle}
is a Ph.D. student in the School of Computer Science at Carleton University where he works in the Graphics, Imaging, and Games Lab. His research interests focus on image processing, image stylization, and super-resolution. He received his master and bachelor degrees in computer science from Carleton University. Previously, he worked as a graphic designer.
\end{IEEEbiographynophoto}
\vspace{-30pt}
\begin{IEEEbiographynophoto}{Seungyong Lee}
is a professor of computer science and engineering at Pohang University of Science and Technology (POSTECH), Korea. He received a PhD degree in computer science from Korea Advanced Institute of Science and Technology (KAIST) in 1995. His current research interests include image and video processing, deep learning based computational photography, and 3D scene reconstruction.
\end{IEEEbiographynophoto}
\vspace{-30pt}
\begin{IEEEbiographynophoto}{Chuan Li}
is a research scientist at Lambda Labs. His work focuses specifically on the convergent field of computer graphics, computer vision, and machine learning. He completed his Ph.D. in image-based modeling at the University of Bath. Before joining Lambda Labs, he was a Postdoc researcher at Max Planck Institute of Informatics and a research associate at Utrecht University and Mainz University. His research in visual data analysis and synthesis was published at CVPR, ICCV, ECCV, NIPS, Siggraph.
\end{IEEEbiographynophoto}
\vspace{-30pt}
\begin{IEEEbiographynophoto}{Yong-Jin Liu}
(SM'16) is a tenured full professor with Department of Computer Science and Technology, Tsinghua University, China. He received his B.Eng degree from Tianjin University, China, in 1998, and the PhD degree from the Hong Kong University of Science and Technology, Hong Kong, China, in 2004. His research interests include cognition computation, computational geometry, computer graphics and computer vision.
\end{IEEEbiographynophoto}
\vspace{-30pt}
\begin{IEEEbiographynophoto}{Amir Semmo}
is a post-doctoral researcher with the Visual Computing \& Visual Analytics group of the Hasso Plattner Institute, Germany, and is the Head of R\&D at Digital Masterpieces. In 2016, he received a doctoral degree on non-photorealistic rendering for 3D geospatial data. His main research topics include image and video processing, computer vision and GPU computing. He is particularly interested in expressive rendering on mobile devices, image stylisation, and the processing of multi-dimensional video data.
\end{IEEEbiographynophoto}
\vspace{-30pt}
\begin{IEEEbiographynophoto}{Prof. Ariel Shamir}
is the Dean of the Efi Arazi school of Computer Science at the
Interdisciplinary Center in Israel. He received his Ph.D. in computer science in 2000 from the Hebrew University in Jerusalem, and spent two years as PostDoc at the University of Texas in Austin.
He is currently an associate editor for ACM TOG and CVM.
Prof. Shamir was named one of the most highly cited researchers on the Thomson Reuters list in 2015. He has a broad commercial experience consulting for various companies.
Prof. Shamir specializes in geometric modeling, computer graphics, image processing and machine learning.
\end{IEEEbiographynophoto}
\vspace{-30pt}
\begin{IEEEbiographynophoto}{Minjung Son}
received the B.S., M.S., and Ph.D. degrees from the Pohang University of Science and Technology (POSTECH), South Korea, in 2005, 2007, and 2014, respectively, all in computer science and engineering. Since 2014, she has been with the Samsung Advanced Institute of Technology, Suwon, South Korea, as a Senior Researcher.
\end{IEEEbiographynophoto}
\vspace{-30pt}
\begin{IEEEbiographynophoto}{Holger Winnem{\"oller}}
received the BSc, BSc (Hons), and MSc degrees in computer science from Rhodes University, South Africa, between 1998 and 2002. He then moved to the US, where in 2006 he received his PhD from Northwestern University. Since 2007, he has been with Adobe Research in Seattle, Washington, where he is currently a principal scientist. His research domains include nonphotorealistic rendering and novel digital media, while his current research focuses on creative tools for aspiring (nonprofessional) artists and casual creativity.
\end{IEEEbiographynophoto}

\end{document}